\documentclass[dvipsnames,format=sigconf,authorversion=true]{acmart}

\usepackage{xcolor}
\usepackage{adjustbox}
\usepackage{colortbl}
\usepackage{tabularx}
\usepackage{booktabs} 
\usepackage{graphicx}
\usepackage{subfig}
\usepackage[linesnumbered, vlined, ruled]{algorithm2e}
\SetArgSty{textup}
\usepackage{enumerate}
\usepackage{prettyref}
\usepackage{color, colortbl}
\usepackage{braket}
\definecolor{MyHiLiRow}{gray}{0.9}
\usepackage{tcolorbox}
\usepackage{caption}
\usepackage{seqsplit}

\usepackage{xr}
\newcommand{\argmax}{\mathop{\rm argmax}\limits}
\newcommand{\argmin}{\mathop{\rm argmin}\limits}

\newcommand\tabblue[1]{{\color[HTML]{1f77b4}{#1}}}
\newcommand\taborange[1]{{\color[HTML]{ff7f0e}{#1}}}
\newcommand\tabgreen[1]{{\color[HTML]{2ca02c}{#1}}}

\newcommand{\pref}{\prettyref}

\newrefformat{fig}{Figure~\ref{#1}}
\newrefformat{tab}{Table~\ref{#1}}
\newrefformat{sec}{Section~\ref{#1}}
\newrefformat{app}{Appendix~\ref{#1}}
\newrefformat{alg}{Algorithm~\ref{#1}}
\newrefformat{property}{Property~\ref{#1}}
\newrefformat{theorem}{Theorem~\ref{#1}}
\newrefformat{corollary}{Corollary~\ref{#1}}
\newrefformat{proposition}{Proposition~\ref{#1}}
\newrefformat{def}{Definition~\ref{#1}}
\newrefformat{eq}{equation~(\ref{#1})}

\AtBeginDocument{%
  \providecommand\BibTeX{{%
    \normalfont B\kern-0.5em{\scshape i\kern-0.25em b}\kern-0.8em\TeX}}}

\setcopyright{acmcopyright}

\copyrightyear{2023}
\acmYear{2023}
\setcopyright{acmlicensed}\acmConference[GECCO '23]{Genetic and
Evolutionary Computation Conference}{July 15--19, 2023}{Lisbon, Portugal}
\acmBooktitle{Genetic and Evolutionary Computation Conference (GECCO '23),
July 15--19, 2023, Lisbon, Portugal}
\acmPrice{15.00}
\acmDOI{10.1145/3583131.3590511}
\acmISBN{979-8-4007-0119-1/23/07}

\begin{document}

\title[On the UA and Population Size in PBEMO Using a Reference Point]{On the Unbounded External Archive and Population Size in Preference-based Evolutionary Multi-objective Optimization Using a Reference Point}





\author{Ryoji Tanabe}
\affiliation{%
  \institution{Yokohama National University}
  \city{Yokohama}
  \state{Kanagawa}
  \country{Japan}}
  \email{rt.ryoji.tanabe@gmail.com}







\begin{abstract}

Although the population size is an important parameter in evolutionary multi-objective optimization (EMO), little is known about its influence on preference-based EMO (PBEMO).
The effectiveness of an unbounded external archive (UA) in PBEMO is also poorly understood, where the UA maintains all non-dominated solutions found so far.
In addition, existing methods for postprocessing the UA cannot handle the decision maker's preference information.
In this context, first, this paper proposes a preference-based postprocessing method for selecting representative solutions from the UA.
Then, we investigate the influence of the UA and population size on the performance of PBEMO algorithms.
Our results show that the performance of PBEMO algorithms (e.g., R-NSGA-II) can be significantly improved by using the UA and the proposed method.
We demonstrate that a smaller population size than commonly used is effective in most PBEMO algorithms for a small budget of function evaluations, even for many objectives.
We found that the size of the region of interest is a less important factor in selecting the population size of the PBEMO algorithms on real-world problems.

\end{abstract}





\maketitle

\section{Introduction}
\label{sec:introduction}

\smallskip
\noindent \textit{\textbf{General context.}}
The ultimate goal of multi-objective optimization involves finding a Pareto optimal solution preferred by a decision maker (DM) \cite{Miettinen98}.
Evolutionary multi-objective optimization (EMO) is helpful for finding a solution set that approximates the Pareto front (PF) in the objective space \cite{Deb01}.
Representative EMO algorithms include NSGA-II \cite{DebAPM02}, IBEA \cite{ZitzlerK04}, and MOEA/D \cite{ZhangL07}.
The solution set found by an EMO algorithm is generally used for an \textit{a posteriori} decision-making, where the DM selects a single solution from the solution set.
The \textit{a posteriori} decision-making is practical when any preference information from the DM is unavailable. 

When the DM's preference information is available \textit{a priori}, it can be incorporated into EMO algorithms \cite{PurshouseDMMW14}.
A preference-based EMO (PBEMO) algorithm \cite{CoelloCoello00,BechikhKSG15} aims to approximate the region of interest (ROI), which is a subregion of the PF defined by the DM's preference information \cite{RuizSL15}.
From the perspective of optimization algorithms, approximating a subregion of the PF is generally easier than approximating the whole one, especially for many objectives.
From the perspective of the DM, she/he can examine only a set of her/his potentially preferred solutions, rather than a set of irrelevant solutions.
That is, using the DM's preference information can reduce her/his workload. 


Although a number of ways of expressing the preference have been proposed in the literature \cite{BechikhKSG15}, using a reference point \cite{Wierzbicki80} is one of the most popular approaches \cite{LiLDMY20,AfsarMR21}.
The reference point consists of objective values desired by the DM.
Compared to other approaches, the reference point-based approach is intuitive and thus relatively easy for the DM to express her/his preference information.
Throughout this paper, we focus only on PBEMO using the reference point.
Representative PBEMO algorithms include R-NSGA-II \cite{DebSBC06}, PBEA \cite{ThieleMKL09}, and MOEA/D-NUMS \cite{LiCMY18}.
These PBEMO algorithms were designed to approximate the ROI.

The population size $\mu$ is one of the most important parameters in evolutionary algorithms.
Some previous studies investigated the influence of the population size on the performance of conventional EMO algorithms (e.g., NSGA-II, not R-NSGA-II).
For example,  Dymond et al. \cite{DymondKH13} showed that the best population size of NSGA-II and SPEA2 \cite{ZitzlerLT01} depends on a budget of function evaluations.
Brockhoff et al. \cite{BrockhoffTH15} showed that NSGA-II with a large population size performs poorly at an early stage of the search but performs well at a later stage. 
Ishibuchi et al. \cite{IshibuchiSMN16a} showed that the best population size depends on the type of EMO algorithm.
Some previous studies (e.g., \cite{RadulescuLS13,Bezerra0S19}) analyzed the influence of the population size on EMO algorithms by using an automatic algorithm configurator. 

With some exceptions (e.g., PESA \cite{CorneKO00} and $\epsilon$-MOEA \cite{DebMM05}), EMO algorithms maintain only $\mu$ solutions in the population.
Thus, as pointed out in \cite{BringmannFK14}, most EMO algorithms are likely to discard solutions in which the DM may be interested.
As demonstrated in previous studies (e.g., \cite{Lopez-IbanezKL11,BringmannFK14}), this issue can be easily addressed by using an unbounded external archive (UA), which maintains all non-dominated solutions found so far.
The UA can be incorporated into any EMO algorithm in a plug-in manner.
Since the UA works independently of EMO algorithms, the UA does not influence their search behavior.
An EMO algorithm with the UA generally outperforms the one without the UA \cite{Lopez-IbanezKL11,TanabeI18,Bezerra0S19}.

One disadvantage of using the UA is that its size can be very large, especially for many-objective optimization. 
In practice, the DM does not want to examine so many solutions \cite{ZhangL07}.
To address this issue, some methods for postprocessing the UA have been proposed (e.g., \cite{BringmannFK14,ShangIN21}).
One of the simplest methods is the distance-based subset selection method \cite{ShangIN21}, which aims to select a subset of uniformly distributed solutions in the objective space from the UA.


\smallskip
\noindent \textit{\textbf{Motivation.}}
Unlike the situation in EMO, little is known about the influence of the population size $\mu$ on the performance of PBEMO algorithms.
The population size $\mu$ in PBEMO has also not been standardized in the literature.
\pref{tab:popsize_literature} shows the number of objectives $m$, population size $\mu$, and maximum number of function evaluations \texttt{max\_evals} used in previous studies.
For the sake of reference, \pref{tab:popsize_literature} shows the setup in the previous study \cite{DebAPM02} that proposed NSGA-II.
As shown in \pref{tab:popsize_literature}, the four previous studies set $\mu$ and \texttt{max\_evals} to different values.
PBEMO algorithms approximate the ROI rather than the whole PF.
Thus, it is \textit{intuitively} expected that PBEMO requires only small values of $\mu$ and \texttt{max\_evals} compared to EMO.
However, except for \cite{ThieleMKL09}, the previous studies used relatively larger values of $\mu$ and \texttt{max\_evals} compared to \cite{DebAPM02}.
%

In addition, the UA has received little attention in the field of PBEMO.
Except for the earliest study \cite{FonsecaF93}, most previous studies discussed the performance of PBEMO algorithms based only on the quality of the final population.
The reason for not using the UA may be that existing methods for postprocessing the UA are unlikely to work well for PBEMO.
This is because existing postprocessing methods cannot handle the DM's preference information.

\smallskip
\noindent \textit{\textbf{Contributions.}}
Motivated by the above discussion, first, we propose a preference-based postprocessing method for selecting representative solutions from the UA.
Then, we analyze the behavior of PBEMO algorithms with different population sizes.
Through an analysis, we address the following three research questions:

\begin{enumerate}[RQ1:]
\item Can the performance of PBEMO algorithms be improved by using the UA? Should a postprocessing method of the UA handle the DM's preference information?
\item How do the number of objectives and a budget of function evaluations influence the best population size? Is the population size used in previous studies reasonable? 
\item Is the size of the ROI a key factor in selecting the population size of PBEMO algorithms?
\end{enumerate}

\noindent \textit{\textbf{Outline.}}
Section \ref{sec:preliminaries} gives some preliminaries.
Section \ref{sec:proposed_method} introduces the proposed postprocessing method.
Section \ref{sec:setting} describes our experimental setup.
Section \ref{sec:results} shows the analysis results to answer the three research questions.
Section \ref{sec:conclusion} concludes this paper.

\smallskip
\noindent \textit{\textbf{Supplementary file.}}
Figure S.$*$ and Table S.$*$ indicate a figure and a table in the supplementary file, respectively.

\smallskip
\noindent \textit{\textbf{Code availability.}}
The implementation of the proposed postprocessing method is available at \url{https://github.com/ryojitanabe/prefpp}.


\begin{table}[t]  
\setlength{\tabcolsep}{4.5pt} 
\renewcommand{\arraystretch}{0.65}
\centering
\caption{$m$, $\mu$, and \texttt{max\_evals} used in previous studies.}
{\footnotesize
  \label{tab:popsize_literature}
\scalebox{1}[1]{ 
\begin{tabular}{lllllcccc}
\toprule 
Ref. & $m$ & $\mu$ & \texttt{max\_evals}\\
\midrule
NSGA-II \cite{DebAPM02} & $2, 5$ & 100 &  $5 \times 10^4$\\
\midrule
R-NSGA-II \cite{DebSBC06} & $2, 3, 5,10$ & 100--500 &  $5 \times 10^4$ -- $ 2.5 \times 10^5$\\
PBEA \cite{ThieleMKL09} & $2, 5$ & 20--200 & $2 \times 10^3$, $2 \times 10^4$\\
R-MEAD2 \cite{MohammadiOLD14} & $4, \ldots, 10$ & 200--350 & $6\times 10^4, 9\times 10^4, 1.05\times 10^5$\\
NUMS \cite{LiCMY18} & 2, 3, 5, 8, 10 & 100--660 & $4 \times 10^4$ -- $1.188 \times 10^6$\\
\bottomrule 
\end{tabular}
}
}
\end{table}

\section{Preliminaries}
\label{sec:preliminaries}

\subsection{Multi-objective optimization}
\label{sec:def_MOPs}

This paper considers minimization problems.
Multi-objective optimization aims to find a solution $\mathbf{x} \in \mathbb{X}$ that minimizes an objective function vector $\mathbf{f}: \mathbb{X} \rightarrow \mathbb{R}^m$, where $\mathbb{X} \subseteq \mathbb{R}^n$ is the feasible solution space, and $\mathbb{R}^m$ is the objective space.
Thus, $n$ is the dimension of the solution space, and $m$ is the dimension of the objective space.

A solution $\mathbf{x}_1$ is said to dominate $\mathbf{x}_2$ if $f_i (\mathbf{x}_1) \leq f_i (\mathbf{x}_2)$ for all $i \in \{1, \ldots, m\}$ and $f_i (\mathbf{x}_1) < f_i (\mathbf{x}_2)$ for at least one index $i$.
We denote $\mathbf{x}_1 \prec \mathbf{x}_2$ when $\mathbf{x}_1$ dominates $\mathbf{x}_2$.
In addition, $\mathbf{x}_1$ is said to weakly dominate $\mathbf{x}_2$ if $f_i (\mathbf{x}_1) \leq f_i (\mathbf{x}_2)$ for all $i \in \{1, \ldots, m\}$.
A solution $\mathbf{x}^\ast$ is a Pareto optimal solution if $\mathbf{x}^\ast$ is not dominated by any solution in $\mathbb{X}$.
The set of all Pareto optimal solutions in $\mathbb{X}$ is called the Pareto optimal solution set $\mathcal{X}^{*} = \{\mathbf{x}^* \in  \mathbb{X} \,|\, \nexists \mathbf{x} \in  \mathbb{X} \: \text{s.t.} \: \mathbf{x} \prec \mathbf{x}^* \}$.
The image of the Pareto optimal solution set in $\mathbb{R}^m$ is also called the PF $\mathbf{f}(\mathcal{X}^{*})$. 
The ideal and nadir points consist of the minimum and maximum values of the PF for $m$ objective functions, respectively.

\subsection{Preference-based multi-objective optimization using the reference point}
\label{sec:def_PBMOPs}

Preference-based multi-objective optimization using the reference point $\mathbf{z}$ involves finding a set of solutions in the ROI in the objective space.
Here, the ROI is a subregion of the PF defined by $\mathbf{z}$.
While EMO aims to approximate the PF, PBEMO aims to approximate the ROI.
It is expected that approximating the ROI is relatively easier than approximating the PF.
In addition, most solutions in the ROI are potentially preferred by the DM.
Thus, the DM's workload is relatively low in this case.

This paper considers the ROI based on the closest Pareto optimal objective vector $\mathbf{f}(\mathbf{x}^{\mathrm{c}*})$ to the reference point $\mathbf{z}$ in terms of the Euclidean distance.
This ROI is called the ROI-C in \cite{TanabeL23}.
Although three ROIs were reviewed in \cite{TanabeL23}, this ROI is the most intuitive ROI.
Mathematically, this ROI is defined as follows:
\begin{align}
  \label{eq:roi}
  \text{ROI}&=\left\{\mathbf{f}(\mathbf{x}^*) \in \mathbf{f}(\mathcal{X}^*) \, | \, \texttt{distance}\bigl(\mathbf{f}(\mathbf{x}^*), \mathbf{f}(\mathbf{x}^{\mathrm{c}*})\bigr)<r\right\},\\
  \mathbf{x}^{\mathrm{c}*}&=\argmin_{\mathbf{x}^* \in \mathcal{X}^*} \left\{\texttt{distance}\left(\mathbf{f}(\mathbf{x}^*),\mathbf{z}\right) \right\}, \notag
\end{align}
where $\texttt{distance}(\cdot,\cdot)$ returns the Euclidean distance between two inputs.
In \pref{eq:roi}, $r$ determines the radius of the ROI.

\begin{figure}[t]
  \centering
  \includegraphics[width=0.2\textwidth]{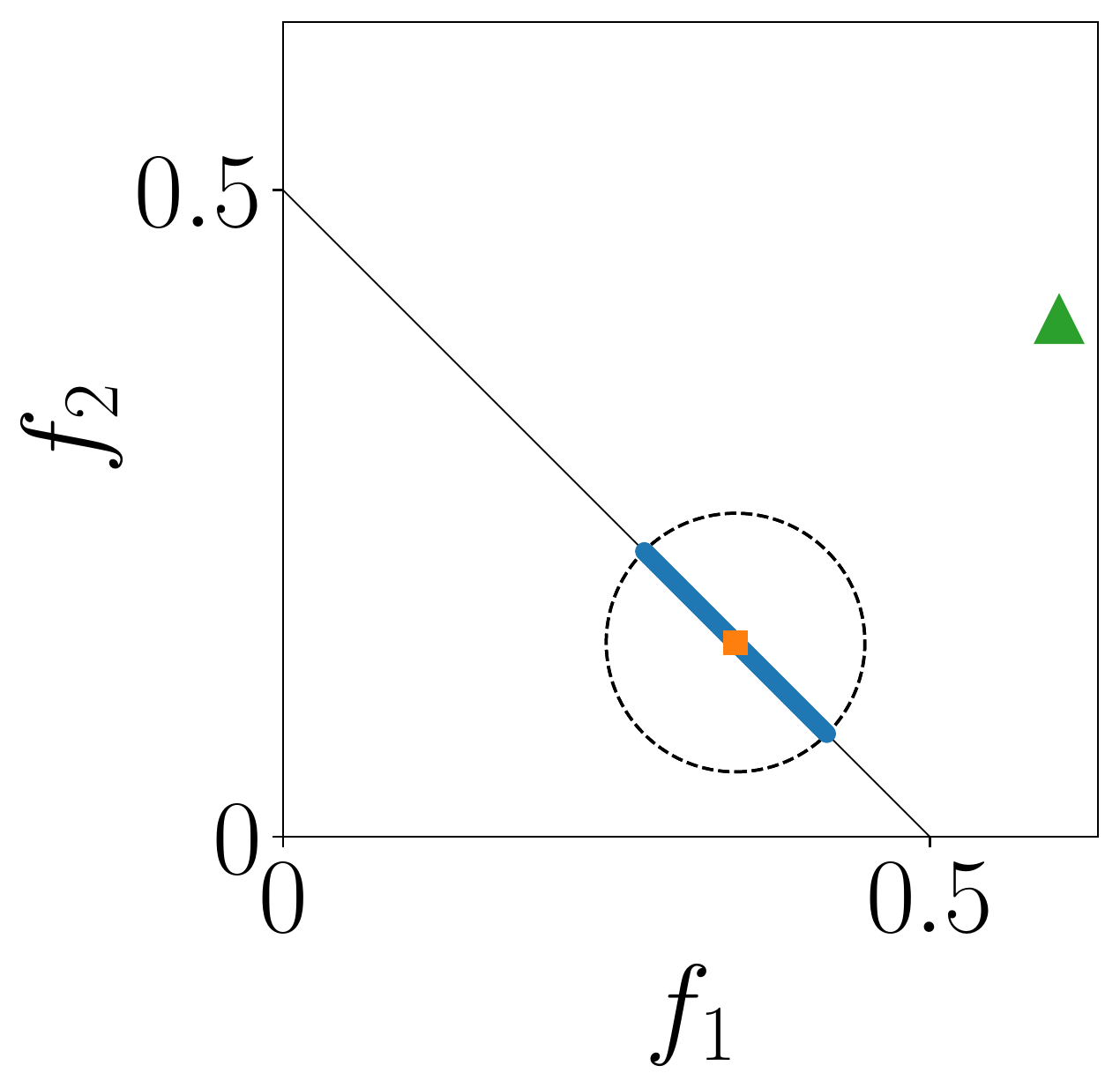}
  \caption{Example of the ROI on DTLZ1 with $m=2$, where \tabgreen{$\blacktriangle$} is the reference point $\mathbf{z}=(0.6,0.4)^\top$, \taborange{$\blacksquare$} is the objective vector of the closet Pareto optimal solution $\mathbf{f}(\mathbf{x}^{\mathrm{c}*})$ to $\mathbf{z}$, and $r=0.1$. A set of the objective vectors of Pareto optimal solutions \tabblue{$\bullet$} in the dotted circle is the ROI.
  }
  \label{fig:roi}
\end{figure}

\pref{fig:roi} shows an example of the ROI on the bi-objective DTLZ1 problem, where the shape of the PF is linear.
As shown in \pref{fig:roi}, the ROI is a set of objective vectors of Pareto optimal solutions in a hypersphere of a radius $r$ centered at $\mathbf{f}(\mathbf{x}^{\mathrm{c}*})$.
As discussed in \cite{TanabeL23}, some PBEMO algorithms (e.g., R-NSGA-II \cite{DebSBC06} and R-MEAD2 \cite{MohammadiOLD14}) were implicitly designed to approximate this ROI.


\subsection{Quality indicators for PBEMO}
\label{sec:q_indicators}

Quality indicators evaluate the quality of a set of objective vectors \cite{ZitzlerTLFF03,LiY19}.
Quality indicators play a crucial role in benchmarking EMO algorithms.
Representative quality indicators include the hypervolume (HV) \cite{ZitzlerT98}, additive $\epsilon$-indicator ($I_{\epsilon+}$) \cite{ZitzlerTLFF03}, and inverted generational distance (IGD) \cite{CoelloS04}.
For example, the IGD value of a solution set $\mathcal{X}$ is calculated as follows:
%
\begin{align}
\label{eq:igd}
{\rm IGD} (\mathcal{X}) = \frac{1}{|\mathcal{S}|} \left(\sum_{\mathbf{s} \in \mathcal{S}} \min_{\mathbf{x} \in \mathcal{X}} \Bigl\{ \mathtt{distance}\bigl(\mathbf{f}(\mathbf{x}), \mathbf{s}\bigr) \Bigr\} \right),
\end{align}
where $\mathcal{S}$ is a set of IGD-reference points $\mathbf{s} \in \mathbb{R}^m$.
It is desirable that the IGD-reference points in $\mathcal{S}$ are uniformly distributed on the PF.
In \pref{eq:igd}, IGD measures the average distance from each $\mathbf{s} \in \mathcal{S}$ to its nearest objective vector $\mathbf{f}(\mathbf{x})$.
A small IGD value indicates a good quality of $\mathcal{X}$ in terms of both convergence and diversity.

Since IGD is not Pareto-compliant, IGD can lead to the wrong conclusion \cite{IshibuchiMTN15,SchutzeELC12}.
To address this issue, a weakly Pareto-compliant version of IGD (IGD$^+$) was proposed in \cite{IshibuchiMTN15}.
IGD$^+$ uses the following distance measure instead the $\mathtt{distance}$ function in \pref{eq:igd}: $\mathtt{distance}_{\mathrm{IGD}^+}(\mathbf{f}(\mathbf{x}), \mathbf{s}) = \sqrt{\sum^m_{i=1} \bigl(\max\{f_i (\mathbf{x})- s_i, 0\}\bigr)^2}$.

Conventional quality indicators  (e.g., IGD) do not take into account any preference information defined by the reference point $\mathbf{z}$.
Thus, conventional quality indicators are not suitable for benchmarking PBEMO algorithms.
As discussed in \cite{TanabeL23}, quality indicators for PBEMO should be able to assess $\mathbf{f}(\mathcal{X})$ from the following three aspects: $i$) the convergence to the PF, $ii$) the convergence to the reference point $\mathbf{z}$, and $iii$) the diversity in the ROI.
Quality indicators for PBEMO should also be able to evaluate the quality of $\mathbf{f}(\mathcal{X})$ even when it is outside the ROI.

Although some quality indicators for benchmarking PBEMO algorithms (e.g., \cite{MohammadiOL13,LiDY18,BandaruS19}) have been proposed, IGD-C \cite{MohammadiOLD14} was recommended in \cite{TanabeL23}.
IGD-C was implicitly designed for the ROI in \pref{eq:roi} and can address the three aspects $i)$ -- $iii)$.
%
The IGD-C value of $\mathcal{X}$ is calculated by \pref{eq:igd} using $\mathcal{S}^{\prime} \subseteq \mathcal{S}$ instead of $\mathcal{S}$.
While the IGD-reference points in $\mathcal{S}$ are distributed on the PF, those in $\mathcal{S}^{\prime}$ are distributed only on the ROI.
In the example in \pref{fig:roi}, $\mathcal{S}^{\prime}$ can be only in the dotted circle.
%
To select $\mathcal{S}^{\prime}$ from $ \mathcal{S}$, first, the closest point $\mathbf{s}^{\mathrm{c}}$ to $\mathbf{z}$ is selected from $\mathcal{S}$, i.e., $\mathbf{s}^{\mathrm{c}}=\argmin_{\mathbf{s} \in \mathcal{S}} \left\{\texttt{distance}\left(\mathbf{s},\mathbf{z}\right) \right\}$.
Then, $\mathcal{S}^{\prime}$ is a set of all points in the region of a hypersphere of radius $r$ centered at $\mathbf{s}^{\mathrm{c}}$, i.e., $\mathcal{S}^{\prime} = \{\mathbf{s} \in \mathcal{S} \, | \, \texttt{distance}(\mathbf{s}, \mathbf{s}^{\mathrm{c}}) < r \}$.

\subsection{Preference-based EMO algorithms}
\label{sec:pemo_ref}


As reviewed in \cite{BechikhKSG15}, some PBEMO algorithms have been proposed in the literature.
As in \cite{LiLDMY20}, this paper focuses on the following six PBEMO algorithms: R-NSGA-II \cite{DebSBC06}, r-NSGA-II \cite{SaidBG10}, g-NSGA-II \cite{MolinaSHCC09}, PBEA \cite{ThieleMKL09}, R-MEAD2 \cite{MohammadiOLD14}, and MOEA/D-NUMS \cite{LiCMY18}.
Note that none of the six PBEMO algorithms use the UA.

R-NSGA-II, r-NSGA-II, and g-NSGA-II are extended versions of NSGA-II \cite{DebAPM02} for preference-based multi-objective optimization.
R-NSGA-II measures the distance to the reference point $\mathbf{z}$, instead of the crowding distance.
Thus, R-NSGA-II prefers non-dominated individuals close to $\mathbf{z}$.
R-NSGA-II also performs $\epsilon$-clearing to maintain the diversity of the population.
r-NSGA-II uses the r-dominance relation instead of the Pareto dominance relation.
The r-dominance relation prefers individuals close to $\mathbf{z}$, and the size of its preferred region is determined by a parameter $\delta \in [0, 1]$.
Similarly, g-NSGA-II uses the g-dominance relation instead of the Pareto dominance relation.
The g-dominance relation prefers non-dominated individuals in a region, which is the set of all objective vectors in $\mathbb{R}^m$ that dominate $\mathbf{z}$ or are dominated by  $\mathbf{z}$.

PBEA \cite{ThieleMKL09} is an extended version of IBEA \cite{ZitzlerK04} by replacing the binary additive $\epsilon$-indicator ($I_{\epsilon +}$) with a preference-based one.
The indicator in PBEA handles the DM's preference information by using an augmented achievement scalarizing function \cite{MiettinenM02}.


R-MEAD2 and MOEA/D-NUMS are scalarizing function-based PBEMO algorithms using a set of $\mu$ weight vectors $\mathcal{W} = \{\mathbf{w}_i\}^{\mu}_{i=1}$.
Although R-MEAD2 is similar to MOEA/D \cite{ZhangL07}, R-MEAD2 adaptively adjusts the $\mu$ weight vectors so that the $\mu$ individuals move toward $\mathbf{z}$ in the objective space.
The previous study \cite{LiCMY18} proposed a nonuniform mapping scheme (NUMS) that shifts the $\mu$ weight vectors toward $\mathbf{z}$.
In contrast to R-MEAD2, NUMS adjusts the weight vectors in an offline manner.
MOEA/D-NUMS \cite{LiCMY18} is an extended MOEA/D using $\mathcal{W}$ generated by NUMS.
MOEA/D-NUMS also uses an augmented achievement scalarizing function \cite{MiettinenM02}.

\subsection{Subset selection method}
\label{sec:ss}

As mentioned in \pref{sec:introduction}, a postprocessing method is necessary to select $k$ representative solutions from the UA $\mathcal{A}$.
This is exactly the subset selection problem that aims to find a subset $\mathcal{X}$ of size $k$ from $\mathcal{A}$ \cite{BasseurDGL16}.
Since the subset selection problem is generally NP-hard \cite{BringmannCE17}, an inexact approach (e.g., greedy search and local search) is a reasonable first choice.
However, there is a trade-off between the computational cost and the quality of a subset.

Shang et al. \cite{ShangIN21} investigated the performance of some distance-based subset selection (DSS) methods that aim to find a subset of uniformly distributed solutions in the objective space. 
Their results showed that an iterative DSS (IDSS) method can obtain a better subset than other methods while keeping the computational cost low.
The IDSS method aims to find a subset $\mathcal{X}$ of size $k$ that maximizes the following uniformity level \cite{Sayin00}.
\begin{align}
\mathtt{uniformity}(\mathcal{X}) = \underset{\mathbf{x}_1 \in \mathcal{X}, \mathbf{x}_2 \in \mathcal{X}  \setminus \{\mathbf{x}_1\}}{\min} \mathtt{distance}(\mathbf{f}^{\prime}\left(\mathbf{x}_1), \mathbf{f}^{\prime}(\mathbf{x}_2)\right), \notag
\end{align}
where $\mathbf{f}^{\prime}(\mathbf{x})$ is the normalized objective vector of $\mathbf{x}$ based on the maximum and minimum values of $\mathbf{f}(\mathcal{X})$.

\pref{alg:idss} shows the IDSS method.
$\mathcal{A}$ is the UA that includes only non-dominated solutions.
If the size of $\mathcal{A}$ is equal to or smaller than $k$, \pref{alg:idss} returns $\mathcal{A}$ (lines 1--2).
Otherwise, $\mathcal{X}$ is set to $k$ solutions randomly selected from $\mathcal{A}$ (line 4).
Then, the following steps are performed $t^{\mathrm{max}}$ times.
First, the new solution $\mathbf{x}^{\mathrm{new}}$ is randomly selected from
$\mathcal{A} \setminus \mathcal{X}$ (line 6), and $\mathbf{x}^{\mathrm{new}}$ is added to $\mathcal{X}$ (line 7).
Then, the IDSS method removes the solution $\mathbf{x}^{\mathrm{w}}$ with the worst contribution to the uniformity level of $\mathcal{X}$ (lines 8--9).
If $t^{\mathrm{max}}$ is large enough, IDSS is likely to obtain $\mathbf{f}(\mathcal{X})$ with the uniform distribution.

\IncMargin{0.5em}
\begin{algorithm}[t]
\SetSideCommentRight
\SetKwInOut{Input}{input}
\SetKwInOut{Output}{output}
%
%
\uIf{$|\mathcal{A}| \leq k$}{
  \Return $\mathcal{A}$\;
  }
\Else{  
  $\mathcal{X} \leftarrow$ Randomly select $k$ solutions from $\mathcal{A}$\;
  \For{$t \in \{1, \ldots, t^{\mathrm{max}}\}$}{
    $\mathbf{x}^{\mathrm{new}} \leftarrow $ Randomly select a solution from $\mathcal{A} \setminus \mathcal{X}$\;
    $\mathcal{X} \leftarrow \mathcal{X} \cup \{\mathbf{x}^{\mathrm{new}}\}$ \;    
    $\mathbf{x}^{\mathrm{w}} \leftarrow \argmax_{\mathbf{x} \in \mathcal{X}} \bigl\{\mathtt{uniformity}\left(\mathcal{X} \setminus \{\mathbf{x}\}\bigr) \right\}$\;
    $\mathcal{X} \leftarrow \mathcal{X} \setminus \{\mathbf{x}^{\mathrm{w}}\}$\;    
  }
  \Return $\mathcal{X}$\;
}
\caption{IDSS \cite{ShangIN21}}
\label{alg:idss}
\end{algorithm}
\DecMargin{0.5em}

\section{Proposed postprocessing method}
\label{sec:proposed_method}

This section introduces the proposed preference-based postprocessing method for the UA in PBEMO.
Let $\mathcal{A}$ be the UA that includes all non-dominated solutions found so far by a PBEMO algorithm.
The goal here is to find a subset $\mathcal{X}$ of size $k$ with the uniform distribution on the ROI in the objective space.
We believe that this goal addresses the three aspects $i$) -- $iii$) that are required for quality indicators for PBEMO (see \pref{sec:q_indicators}).
However, existing postprocessing methods (e.g., the IDSS method \cite{ShangIN21}) are unlikely to achieve this goal.
This is because they were designed to find $\mathcal{X}$ with the uniform distribution on \textit{the whole PF} in the objective space.
To address this issue, we propose the preference-based postprocessing method that can handle the DM's preference information. 
Note that the proposed method is designed for the ROI defined in \pref{eq:roi}.
Note also that the proposed method can be incorporated even into any conventional EMO algorithm.

\IncMargin{0.5em}
\begin{algorithm}[t]
\SetSideCommentRight
\SetKwInOut{Input}{input}
\SetKwInOut{Output}{output}
%
%
\uIf{$|\mathcal{A}| \leq k$}{
  \Return $\mathcal{A}$\;
  }
\Else{
  $\mathbf{x}^{\mathrm{c}} \leftarrow \argmin_{\mathbf{x} \in \mathcal{A}} \left\{\mathtt{distance}(\mathbf{f}(\mathbf{x}), \mathbf{z}) \right\}$\;
  $\mathcal{X} \leftarrow \{\mathbf{x} \in \mathcal{A} \: | \: \mathtt{distance}(\mathbf{f}(\mathbf{x}), \mathbf{f}(\mathbf{x}^{\mathrm{c}})) \leq r \}$\;
  \While{$|\mathcal{X}| < k$}{
    $\mathbf{x}^{\mathrm{c^{\prime}}} \leftarrow \argmin_{\mathbf{x} \in \mathcal{A} \setminus \mathcal{X}} \left\{\mathtt{distance}\left(\mathbf{f}(\mathbf{x}), \mathbf{f}(\mathbf{x}^{\mathrm{c}})\right) \right\}$\;
    $\mathcal{X} \leftarrow \mathcal{X} \cup \{\mathbf{x}^{\mathrm{c^{\prime}}}\}$\;    
  }
  \If{$|\mathcal{X}| > k$}{
    $\mathcal{X} \leftarrow \mathtt{subset\_selection}(\mathcal{X}, k)$\;
    }
  \Return $\mathcal{X}$\;
}
\caption{The proposed preference-based postprocessing method for PBEMO}
\label{alg:ppp}
\end{algorithm}
\DecMargin{0.5em}

\begin{figure}[t]
  \centering
  \subfloat[Case (1): $|\mathcal{X}| < k$]{
    \includegraphics[width=0.2\textwidth]{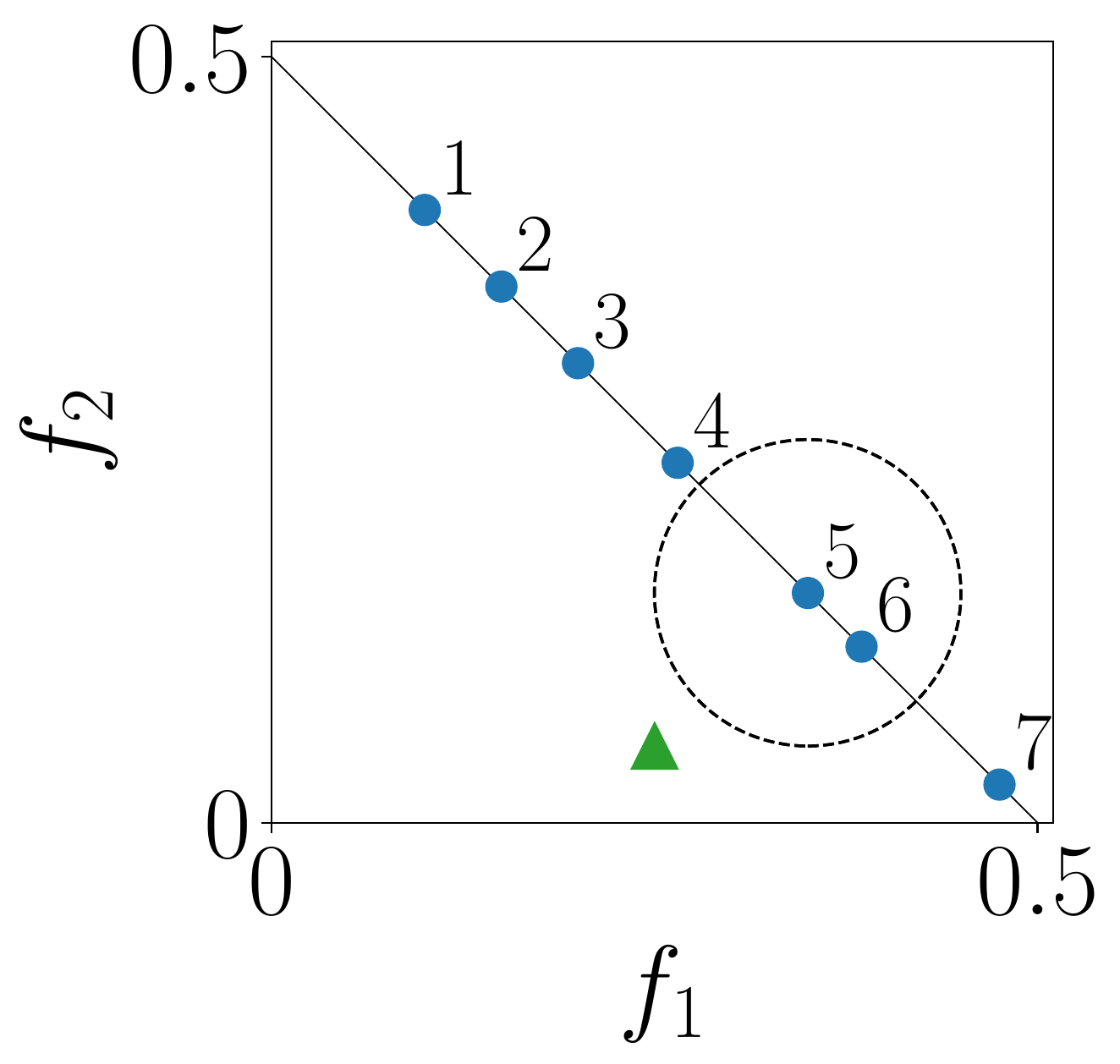}
  }
  \subfloat[Case (2):  $|\mathcal{X}| > k$]{
    \includegraphics[width=0.2\textwidth]{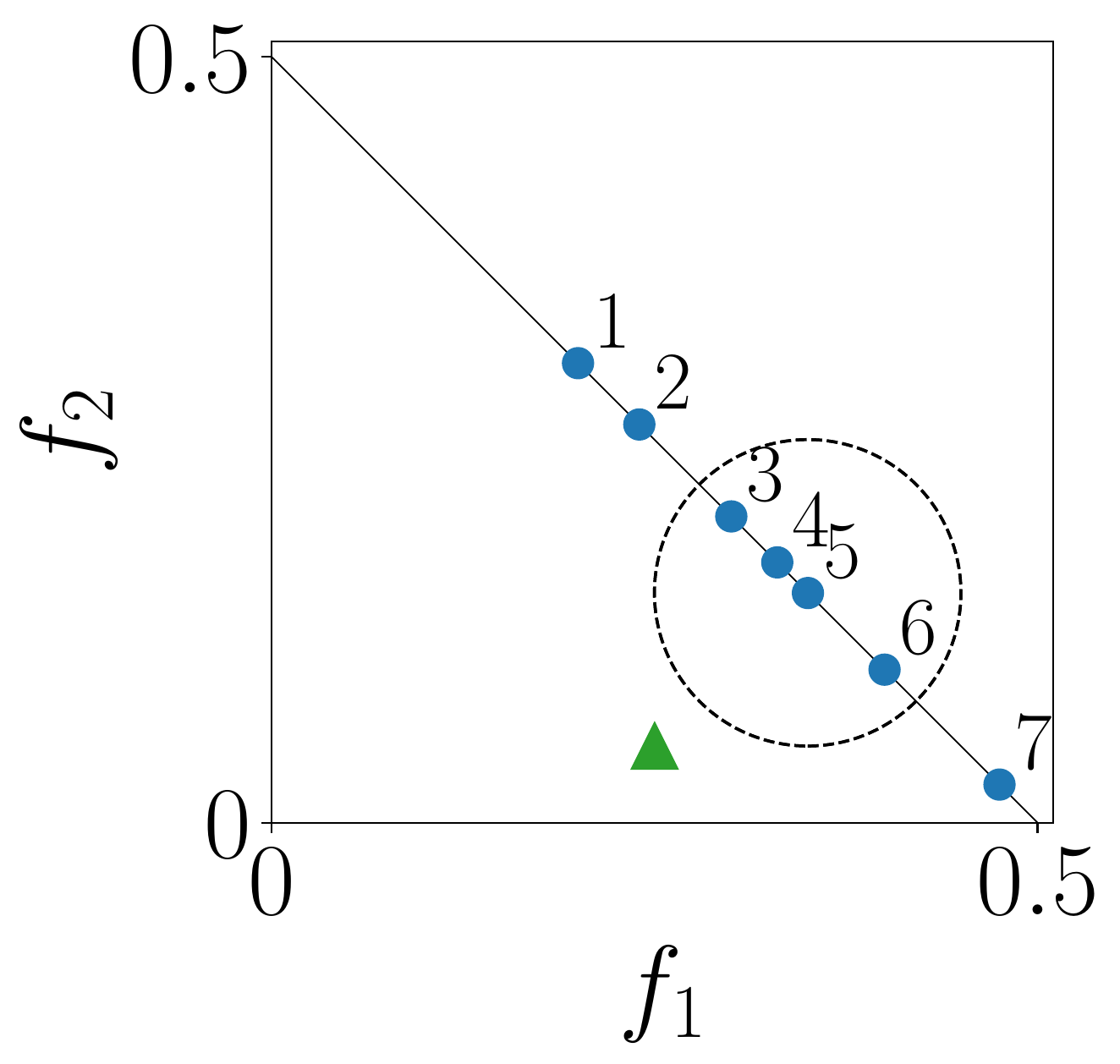}
  }
  \caption{Examples on DTLZ1 with $m=2$, where $|\mathcal{A}|=7$ and $k=3$. \tabgreen{$\blacktriangle$} is the reference point.} 
  \label{fig:example_p3}
\end{figure}

\pref{alg:ppp} shows the proposed method.
%
As in \pref{alg:idss}, if $|\mathcal{A}| \leq k$, \pref{alg:ppp} returns $\mathcal{A}$ (lines 1--2).
Otherwise, the following steps are performed.
First, the proposed method approximates the ROI, which is unknown in real-world problems.
The closest solution $\mathbf{x}^{\mathrm{c}}$ to the reference point $\mathbf{z}$ in the objective space is selected from $\mathcal{A}$ (line 4).
We denote the set of all non-dominated solutions in an $m$-dimensional hypersphere of radius $r$ centered at $\mathbf{x}^{\mathrm{c}}$ in the objective space as an \textit{approximated ROI}.
Here, the approximated ROI is equivalent to the true ROI in \pref{eq:roi} if $\mathcal{A} = \mathcal{X}^*$.
All non-dominated solutions in the approximated ROI are set to $\mathcal{X}$ (line 5).
If the size of $\mathcal{X}$ is exactly the same as $k$, \pref{alg:ppp} returns $\mathcal{X}$ (lines 11).
Otherwise, the following two cases are considered:
\begin{enumerate}
\item the size of $\mathcal{X}$ is less than $k$ (lines 6--8), and 
\item the size of $\mathcal{X}$ is greater than $k$ (line 9--10).
\end{enumerate}
\pref{fig:example_p3} shows examples of the two cases, where $|\mathcal{A}|=7$ and $k=3$.
In our preliminary results, we observed that case (1) occurs in an early stage of the search, while case (2) occurs in a later stage.

In case (1), we believe that the DM is interested in closer solutions to $\mathbf{x}^{\mathrm{c}}$ in the objective space even though they are out of the approximated ROI.
Thus, it is rationale to select $k - |\mathcal{X}|$ non-dominated solutions from $\mathcal{A} \setminus \mathcal{X}$.
While $|\mathcal{X}| < k$, the following steps are repeatedly performed.
First, the closest solution $\mathbf{x}^{\mathrm{c^{\prime}}}$ to $\mathbf{x}^{\mathrm{c}}$ in the objective space is selected from $\mathcal{A} \setminus \mathcal{X}$ (line 7).
Then, $\mathbf{x}^{\mathrm{c^{\prime}}}$ is added to $\mathcal{X}$ (line 8).
In the example of \pref{fig:example_p3}(a), $\mathcal{X}=\{\mathbf{x}_5, \mathbf{x}_6\}$, and $\mathbf{x}_5$ is $\mathbf{x}^{\mathrm{c}}$.
In this case, $\mathbf{x}_4$ is closer to $\mathbf{x}_5$ than $\mathbf{x}_1, \mathbf{x}_2, \mathbf{x}_3,$ and $\mathbf{x}_7$ in the objective space.
Thus, $\mathbf{x}_4$ is added to $\mathcal{X}$.

In case (2), $|\mathcal{X}| - k$ solutions should be removed from $\mathcal{X}$ while keeping the diversity of $\mathcal{X}$.
Fortunately, a conventional subset selection method can be used to select $k$ solutions from $\mathcal{X}$ (line 10).
Based on the promising results reported in \cite{ShangIC21}, this paper uses the IDSS method shown in \pref{alg:idss}.
Note that IDSS cannot handle the DM's preference information, but the objective vectors of the solutions in $\mathcal{X}$ are already in the approximated ROI.
After applying IDSS to $\mathcal{X}$, it is expected that the objective vectors of the $k$ solutions in $\mathcal{X}$ are uniformly distributed in the approximate ROI.
In the example of \pref{fig:example_p3}(b), $\mathcal{X}=\{\mathbf{x}_3, \mathbf{x}_4, \mathbf{x}_5, \mathbf{x}_6\}$, and $k=3$.
In this case, $\mathbf{x}_4$ is likely to be removed from $\mathcal{X}$ after IDSS is performed.

\section{Experimental setup}
\label{sec:setting}

Here, we describe the experimental setup for our analysis.
A non-dominated sorting method is required to maintain the UA. 
We used the ND-Tree-based update method \cite{JaszkiewiczL18}, which is one of the state-of-the-art methods.
We used the \texttt{C++} implementation of the ND-Tree-based update method provided by the authors of \cite{JaszkiewiczL18}.
As in \cite{TanabeIO17}, we periodically updated the archive at  $100, 200, \ldots$ $, 1\,000, 2\,000, \ldots,$ $ 50\,000$ function evaluations.
We also implemented the proposed preference-based postprocessing method in \texttt{C++}, including the IDSS method \cite{ShangIN21}.
Unless otherwise noted, we set the size of the subset $k$ and the radius of the ROI $r$ to 100 and 0.1, respectively.
As in \cite{ShangIN21}, we set the maximum number of iterations $t^{\mathrm{max}}$ in the IDSS to $10^4$.

We used the six PBEMO algorithms described in \pref{sec:pemo_ref}.
We used the source code of the PBEMO algorithms provided by the authors of \cite{LiLDMY20}. 
We performed $31$ independent runs for each PBEMO algorithm.
We set the population size $\mu$ to $8, 20, 40, 100, 200, 300, 400,$ and $500$. 
We set other parameters of the six PBEMO algorithms according to \cite{LiLDMY20}. 
We investigated the effect of the population size by fixing other parameters to default values.
Correlation analysis of all parameters (e.g., $\epsilon$ in R-NSGA-II) is beyond the scope of this paper.
We generated the weight vector set in MOEA/D-NUMS using the source code provided by the authors of~\cite{LiCMY18}. 
We also generated the original weight vector set using the method proposed in \cite{BlankDDBS21}, which can generate the weight vector set of any size.

As in \cite{BandaruS19}, we used the DTLZ1--DTLZ4 problems \cite{DebTLZ05}.
As observed in the results of EMO algorithms in \cite{TanabeI18}, we believe that PBEMO algorithms require a large population size on problems with irregular PFs (e.g., DTLZ5 and DTLZ7).
An investigation on problems with differently scaled objectives (e.g., SDTLZ1~\cite{DebJ14}) remains for future work.
We set the number of objectives $m$ as follows: $m \in \{2, 3, 4, 5, 6\}$.
We set the number of decision variables as in \cite{DebTLZ05}.
We set the reference point $\mathbf{z}$ as follows: $\mathbf{z} = (0.6, 0.4)^{\top}$, $\mathbf{z} = (0.5, 0.3, 0.2)^{\top}$,  $\mathbf{z} = (0.4, 0.3, 0.2, 0.1)^{\top}$, $\mathbf{z} = (0.3, 0.25, 0.2, 0.15, 0.1)^{\top}$, and $\mathbf{z} = (0.3, 0.2, 0.15, 0.13, 0.12, 0.1)^{\top}$ for $m=2, \ldots, 6$, respectively.
Most reference points are achievable, but the behavior of the PBEMO algorithms (except for g-NSGA-II) is not very sensitive to the position of $\mathbf{z}$ unless $\mathbf{z}$ is too far from the PF \cite{LiLDMY20,TanabeL23}.

%

Although the IGD-C indicator was recommended in \cite{TanabeL23}, IGD-C is not Pareto-compliant.
To address this issue, we replaced the IGD calculation in IGD-C with the IGD$^+$ calculation.
We denote this version of IGD-C as IGD$^+$-C.
IGD$^+$-C is more reliable than IGD-C in terms of Pareto-compliance.

\definecolor{c1}{RGB}{150,150,150}
\definecolor{c2}{RGB}{220,220,220}

\section{Results}
\label{sec:results}

This section describes our analysis results.
Through experiments, Sections \ref{sec:results_pp}--\ref{sec:results_roi_size} aim to address the three research questions (\textbf{RQ1}--\textbf{RQ3}) described in \pref{sec:introduction}, respectively.
%

\subsection{On the effectiveness of the proposed postprocessing method}
\label{sec:results_pp}

In this section, we consider the following three solution sets found by each PBEMO algorithm:
\setlength{\leftmargini}{1em}  
\begin{itemize}
\item \textbf{POP}: the final population of size $\mu$, 
\item \textbf{UA-IDDS}: a solution subset of size $k$ obtained by postprocessing the UA using the IDSS method, and 
\item \textbf{UA-PP}: a solution subset of size $k$ obtained by postprocessing the UA using the preference-based postprocessing (PP) method.
\end{itemize}
We focus on the results of the PBEMO algorithms with $\mu=100$, i.e., $\mu=k$.
Most previous studies evaluated the performance of a PBEMO algorithm based on its POP.
In contrast, this is the first study to evaluate the performance of a PBEMO algorithm based on its UA-IDDS and UA-PP.

\subsubsection{Distributions of objective vectors}

\begin{figure}[htp]
   \centering
   \subfloat[R-NSGA-II (POP)]{
     \includegraphics[width=0.15\textwidth]{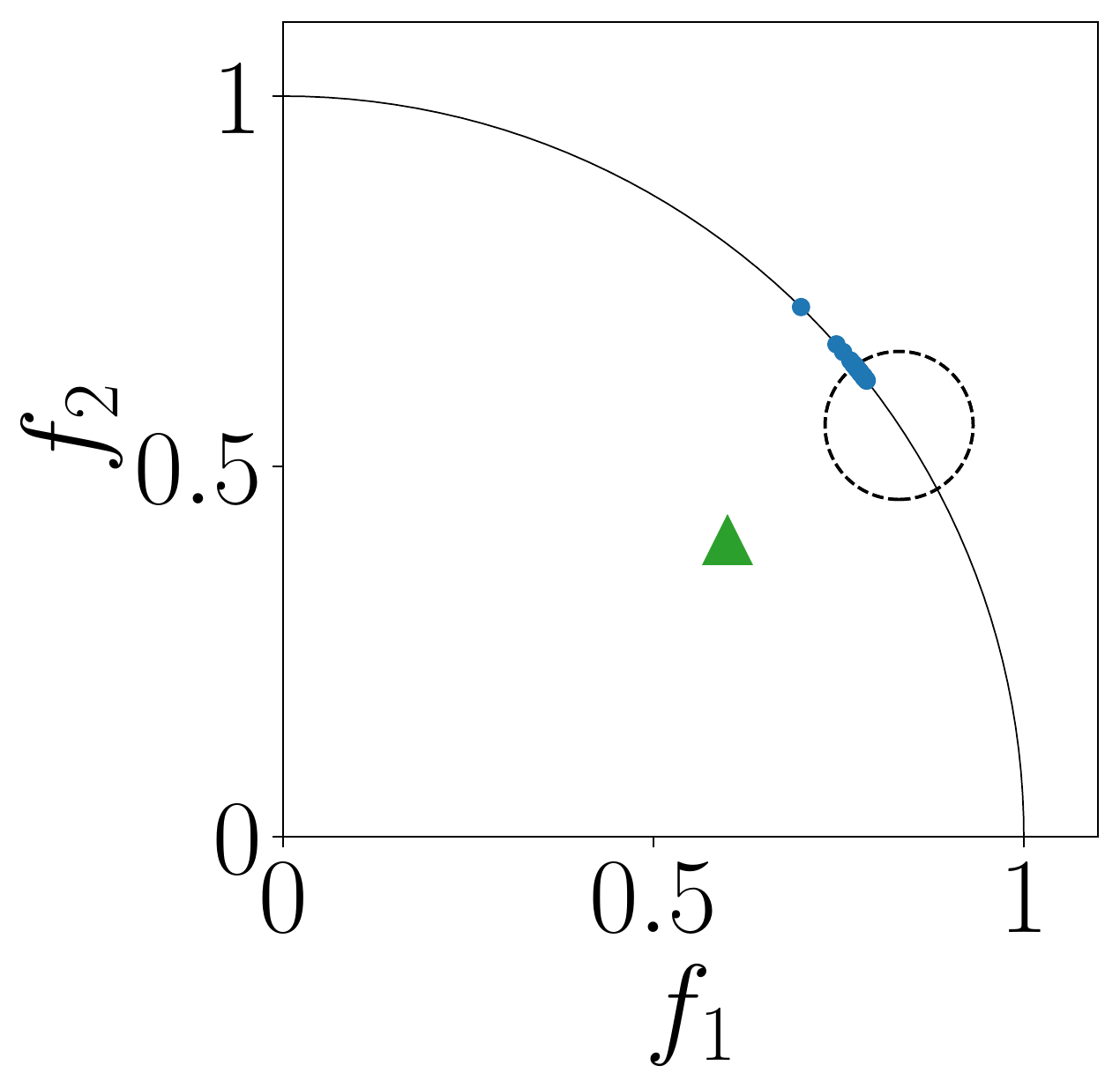}
   }
   \subfloat[R-NSGA-II (UA-IDDS)]{
  \includegraphics[width=0.15\textwidth]{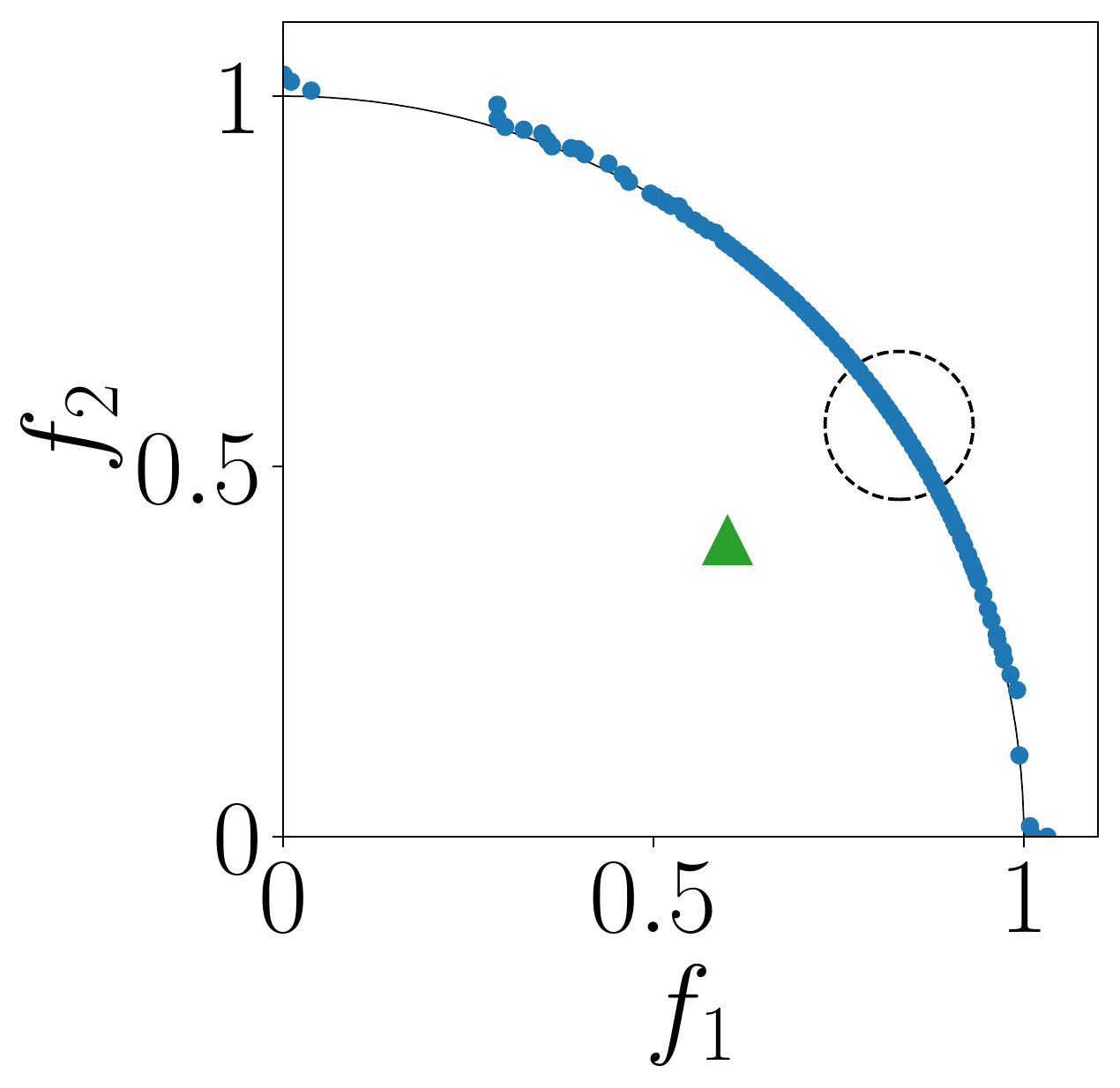}
}
\subfloat[R-NSGA-II (UA-PP)]{
  \includegraphics[width=0.15\textwidth]{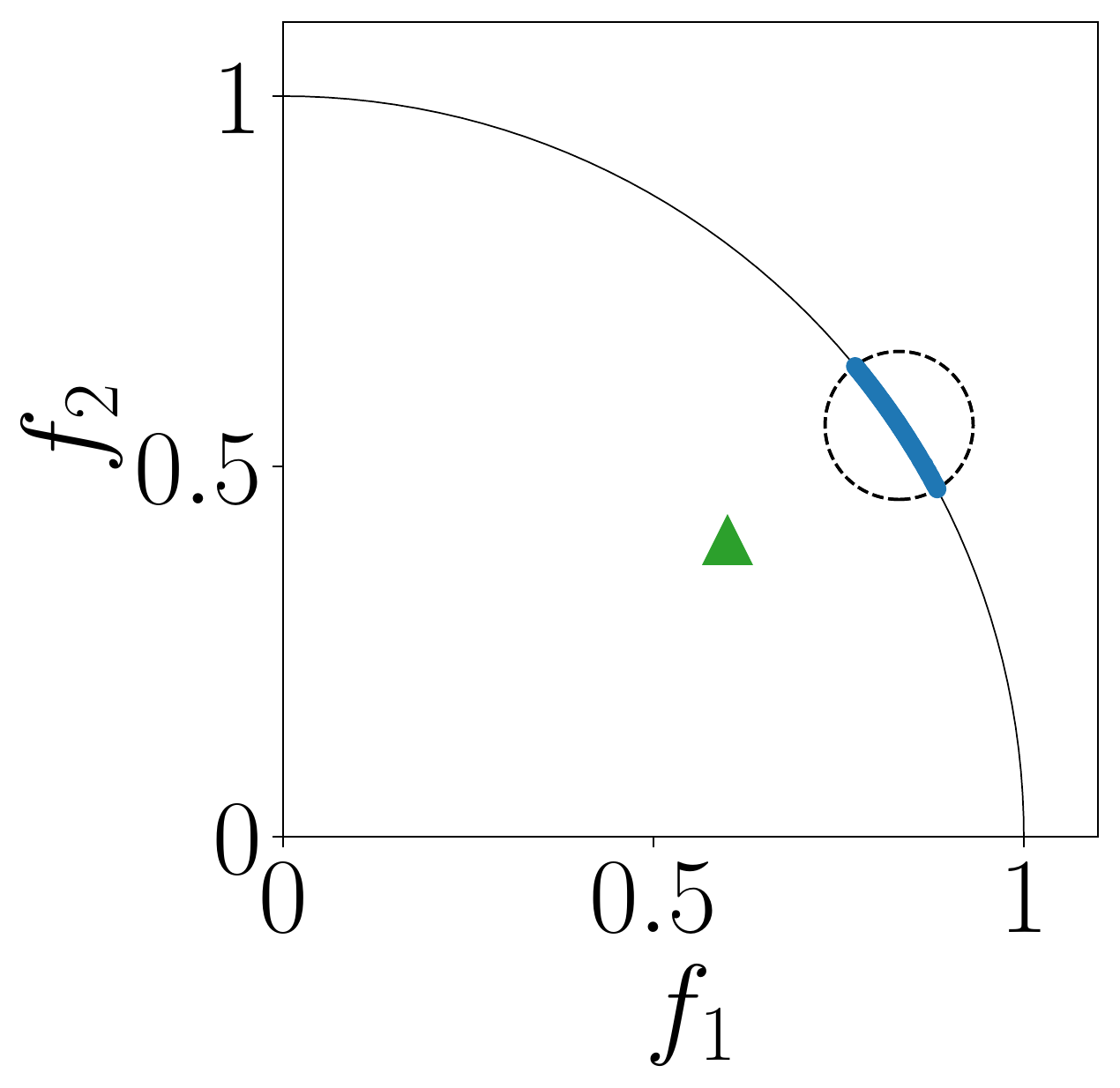}
}
\\
\subfloat[r-NSGA-II (POP)]{
     \includegraphics[width=0.15\textwidth]{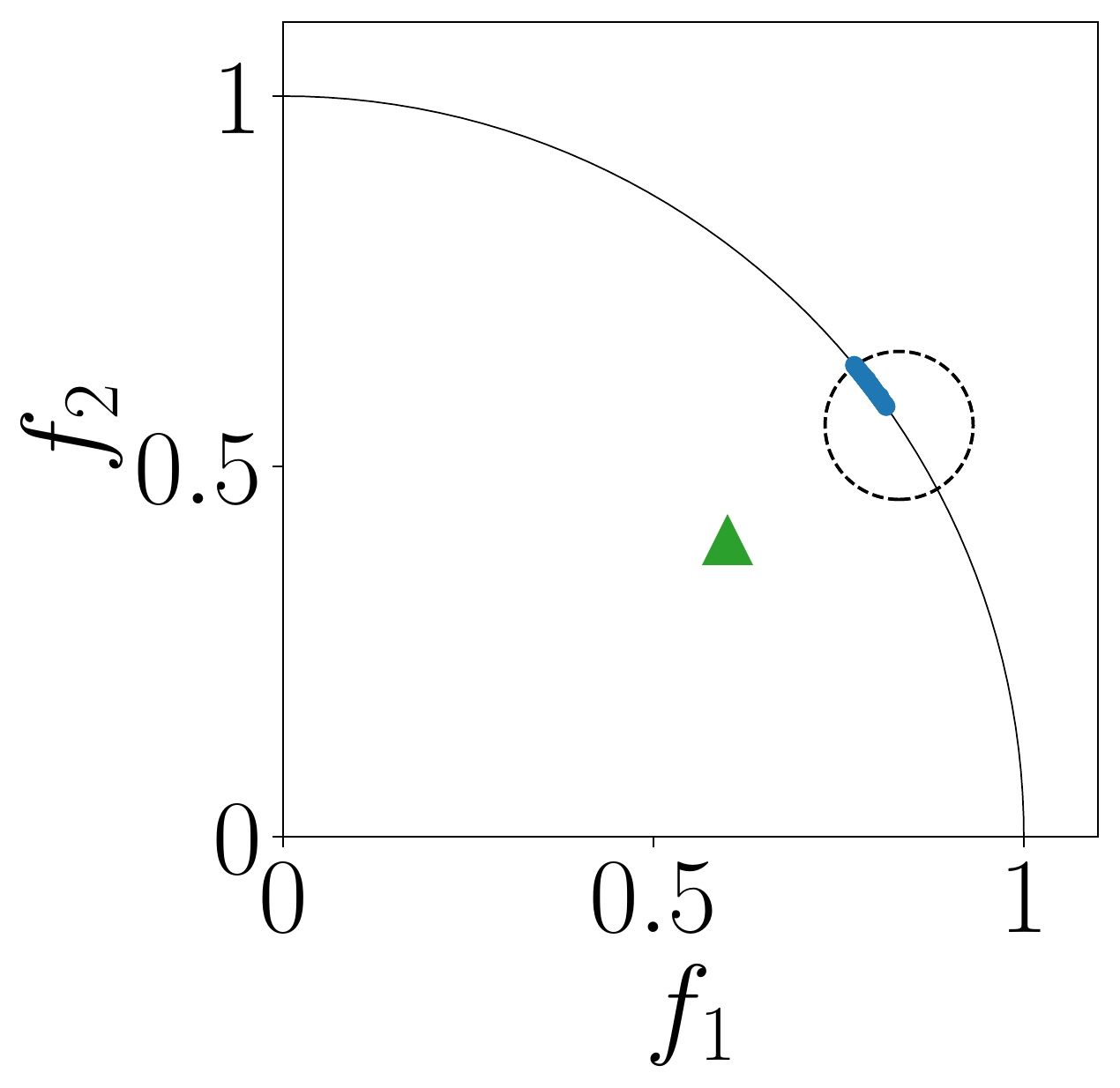}
   }
   \subfloat[r-NSGA-II (UA-IDDS)]{
  \includegraphics[width=0.15\textwidth]{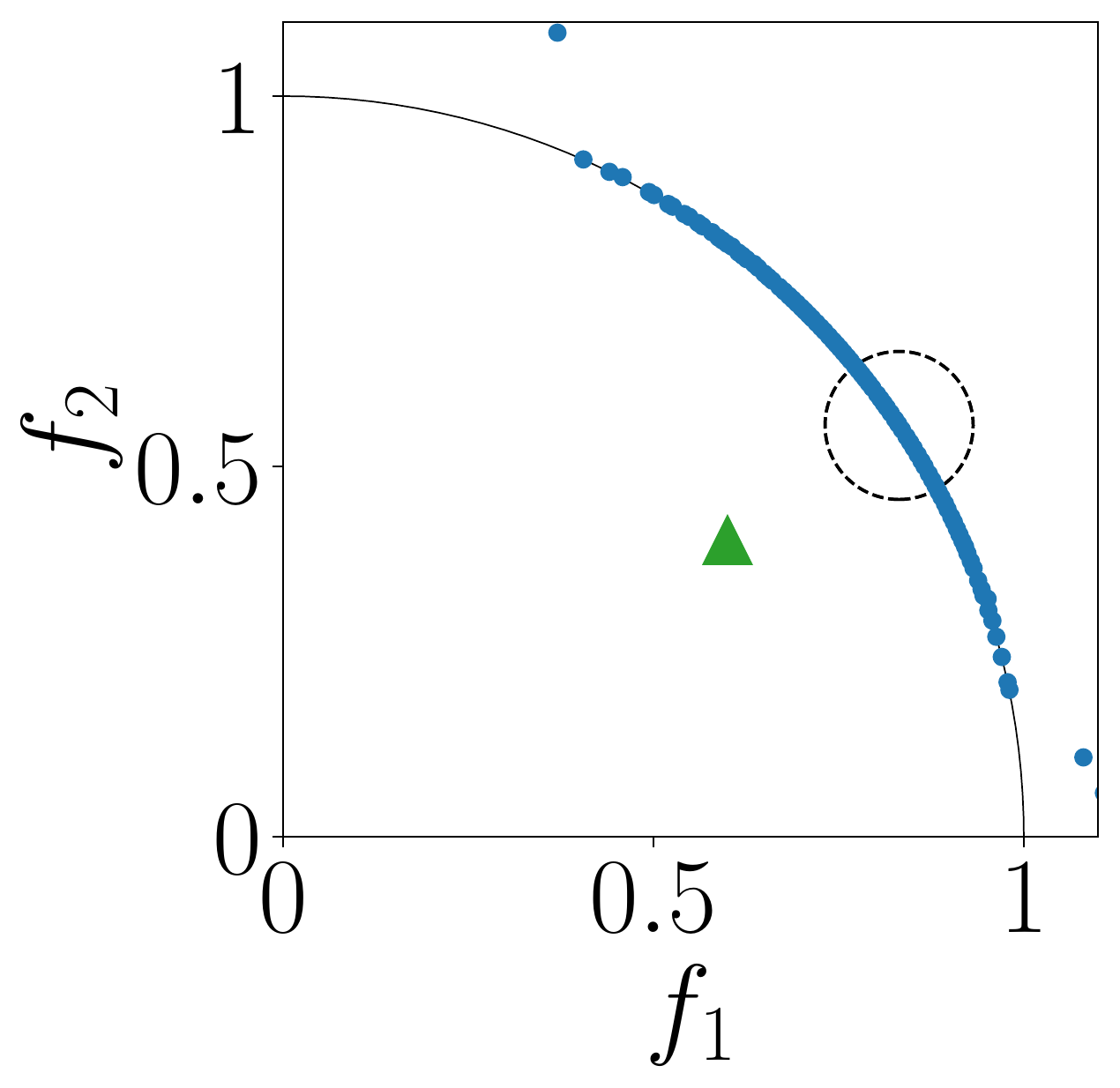}
}
\subfloat[r-NSGA-II (UA-PP)]{
  \includegraphics[width=0.15\textwidth]{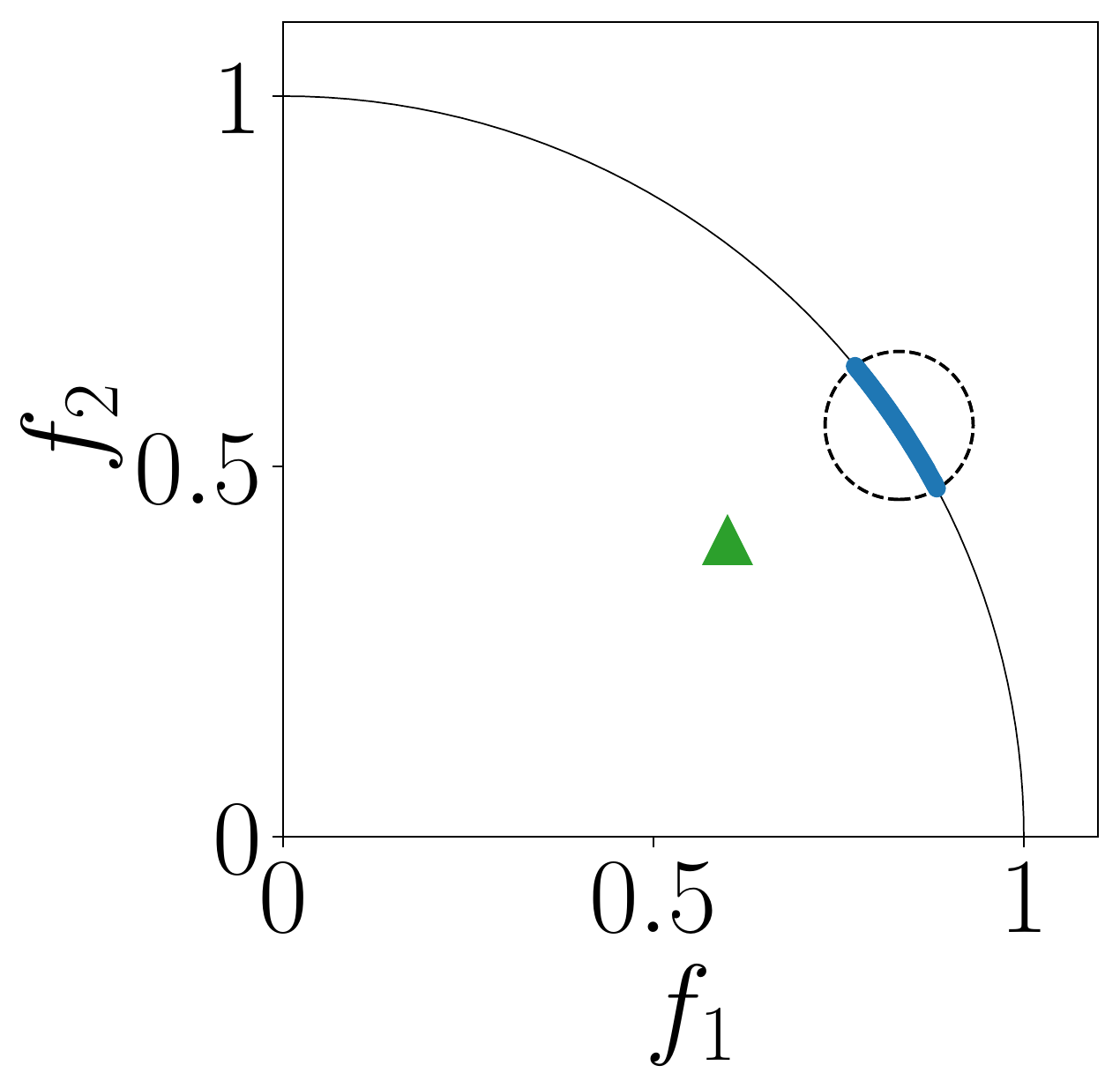}
}
\\
\subfloat[g-NSGA-II (POP)]{
     \includegraphics[width=0.15\textwidth]{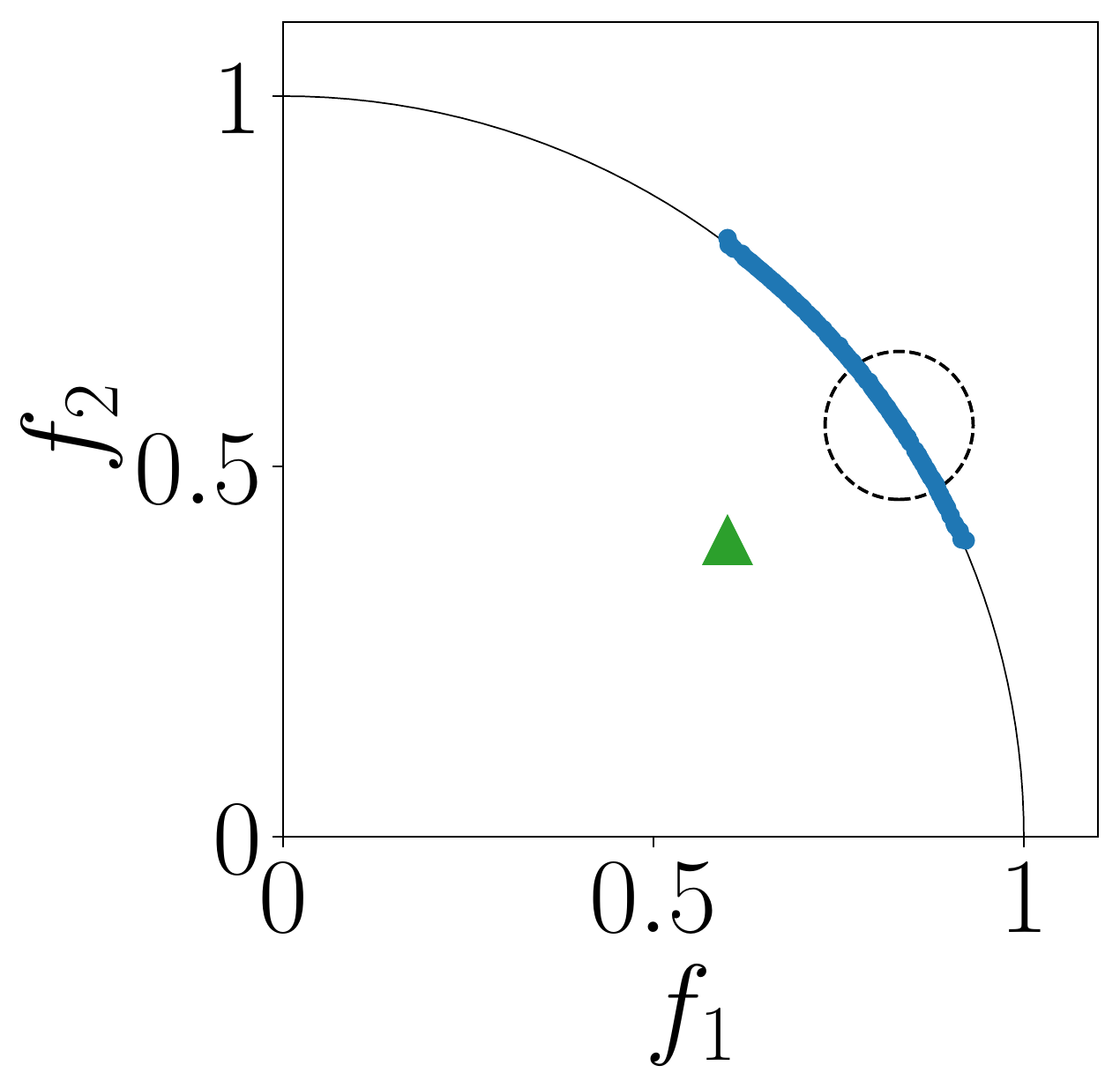}
   }
   \subfloat[g-NSGA-II (UA-IDDS)]{
  \includegraphics[width=0.15\textwidth]{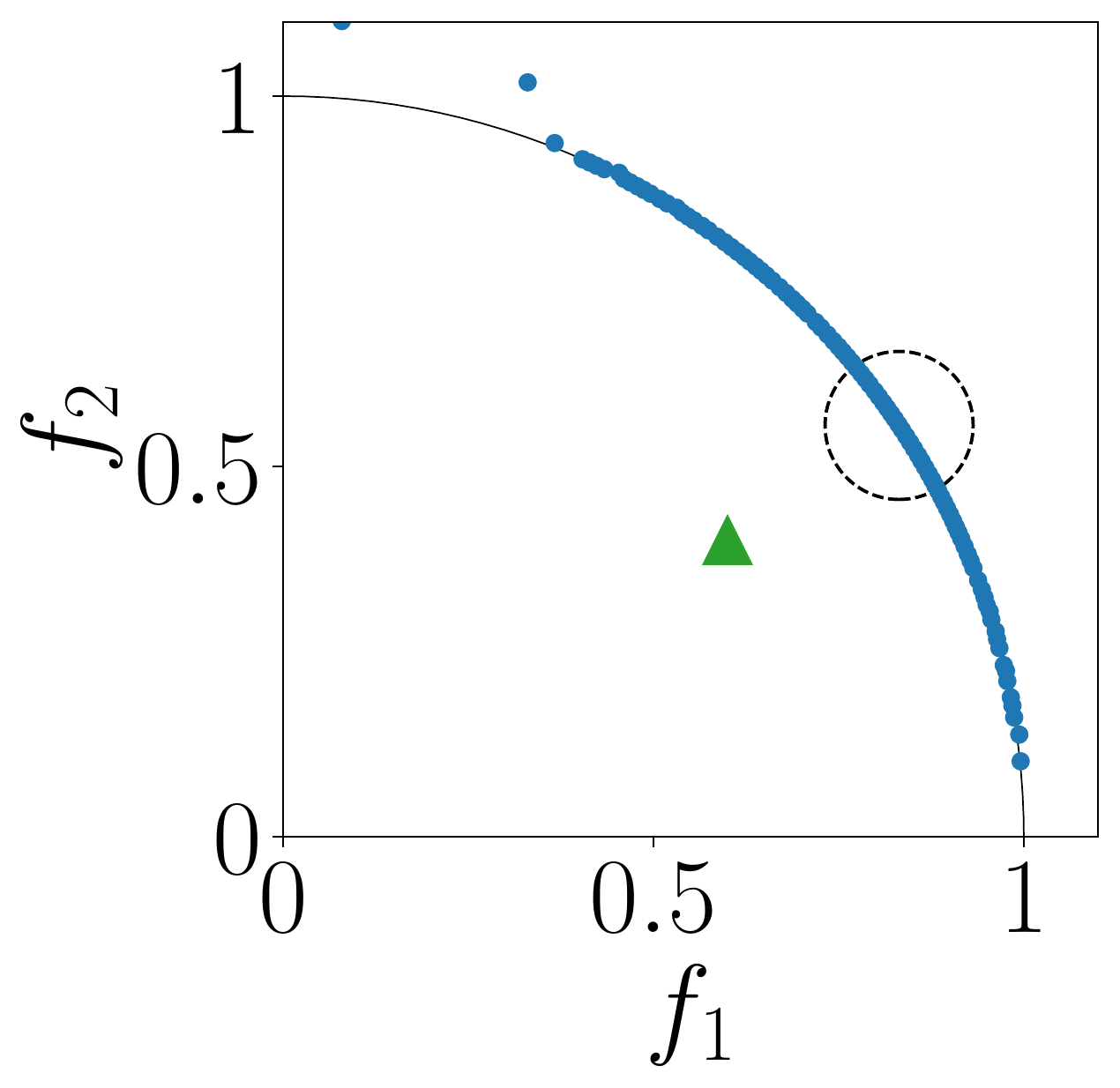}
}
\subfloat[g-NSGA-II (UA-PP)]{
  \includegraphics[width=0.15\textwidth]{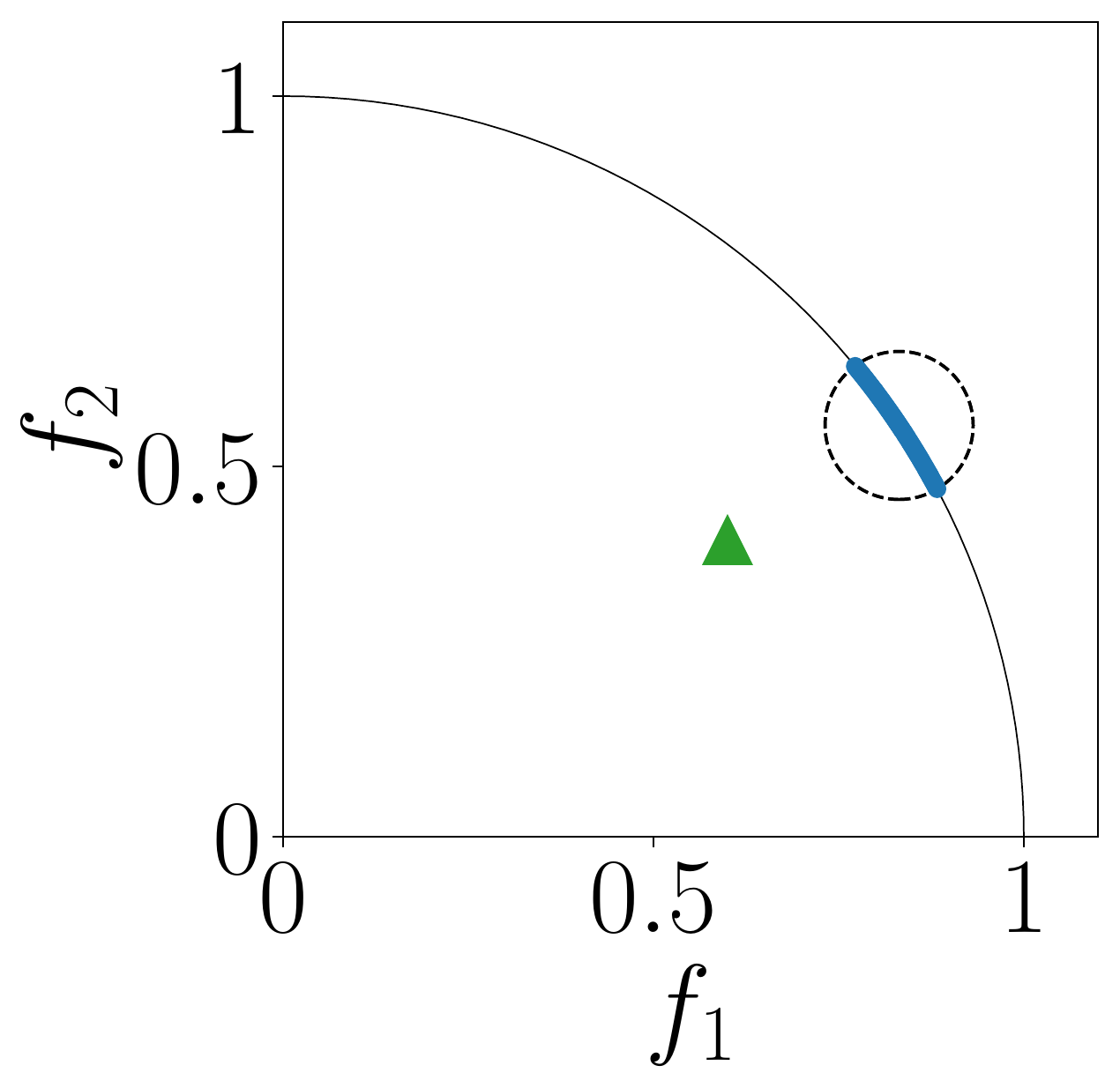}
}
\\
\subfloat[PBEA (POP)]{
     \includegraphics[width=0.15\textwidth]{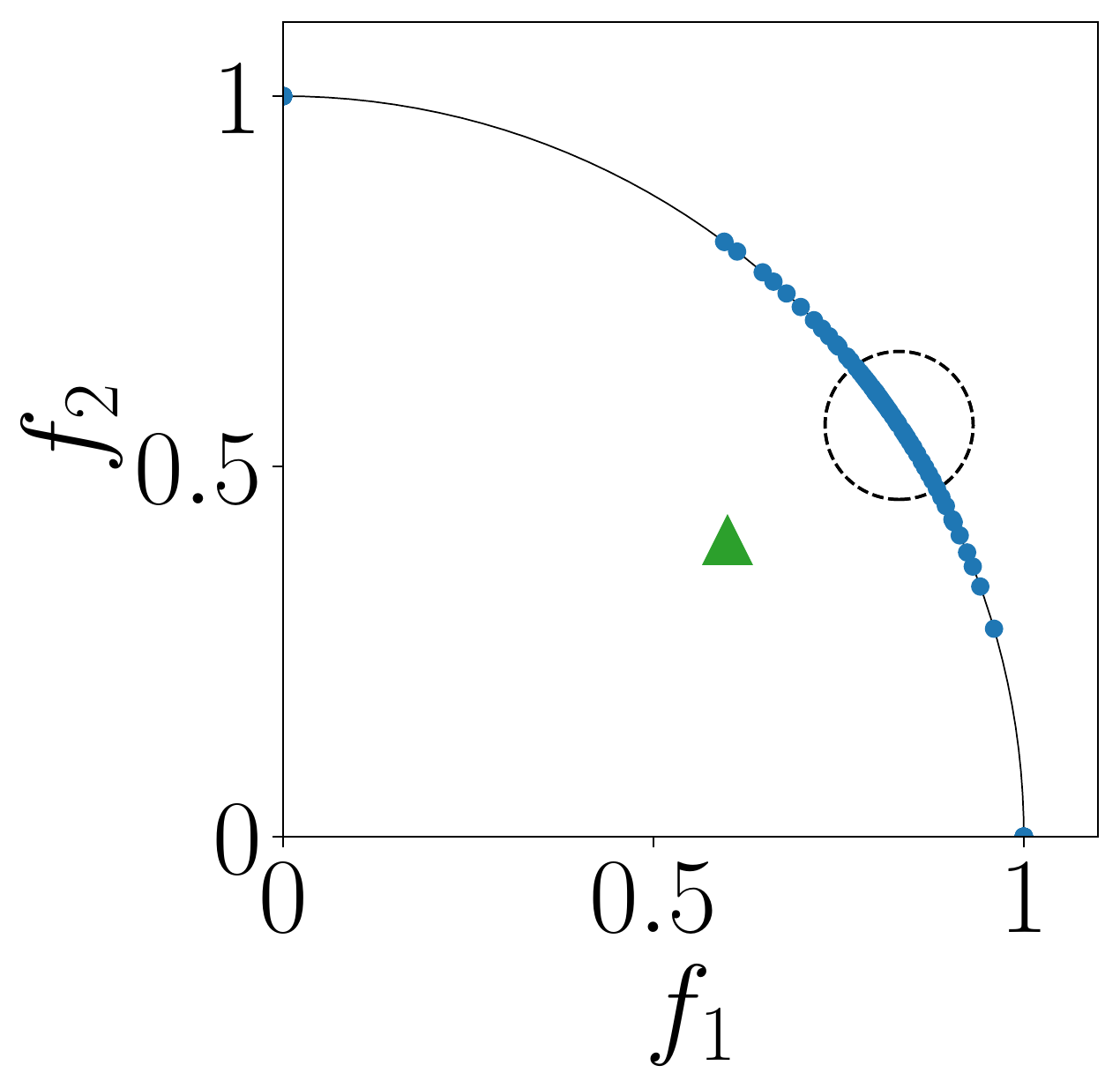}
   }
   \subfloat[PBEA (UA-IDDS)]{
  \includegraphics[width=0.15\textwidth]{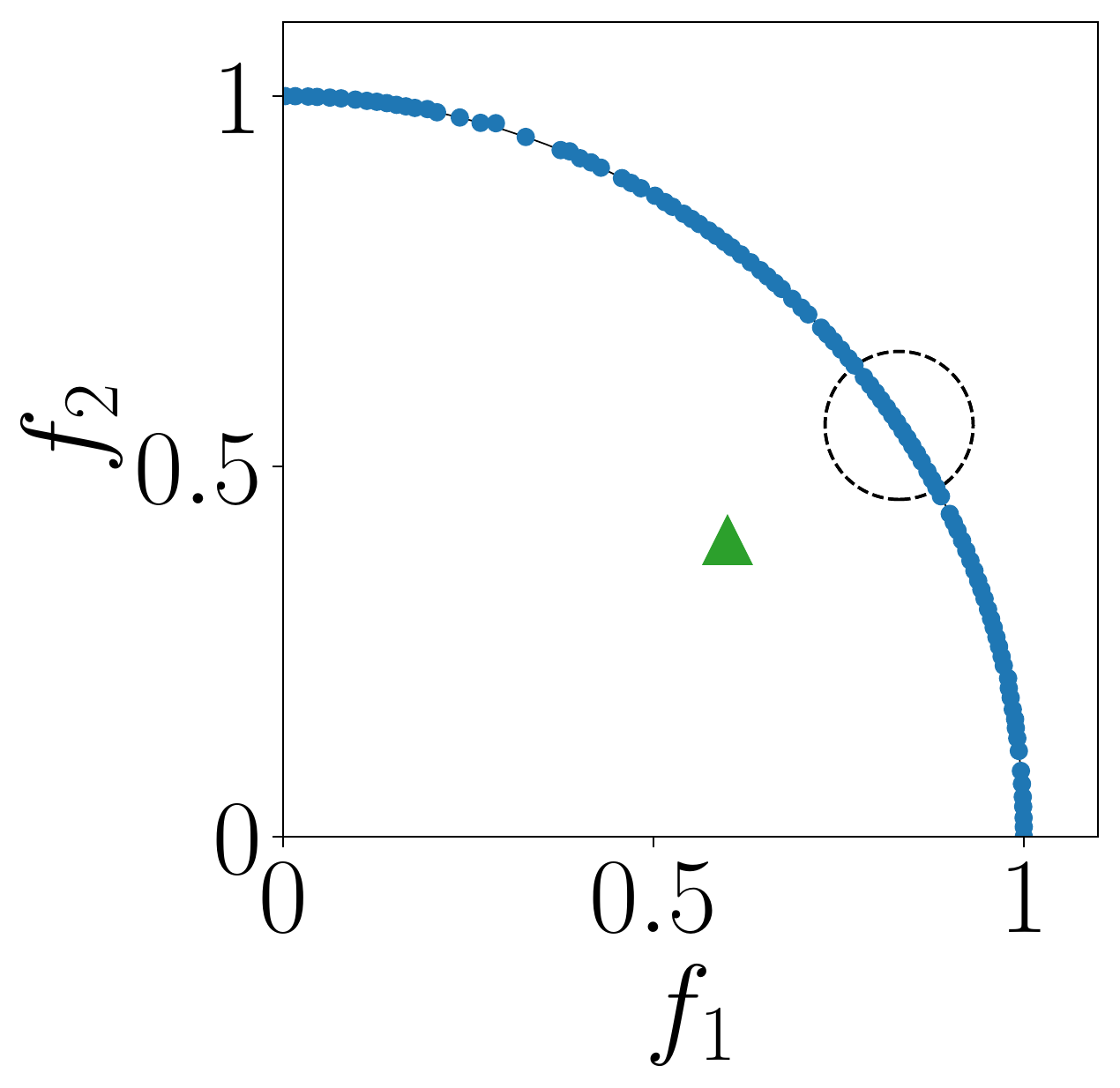}
}
\subfloat[PBEA (UA-PP)]{
  \includegraphics[width=0.15\textwidth]{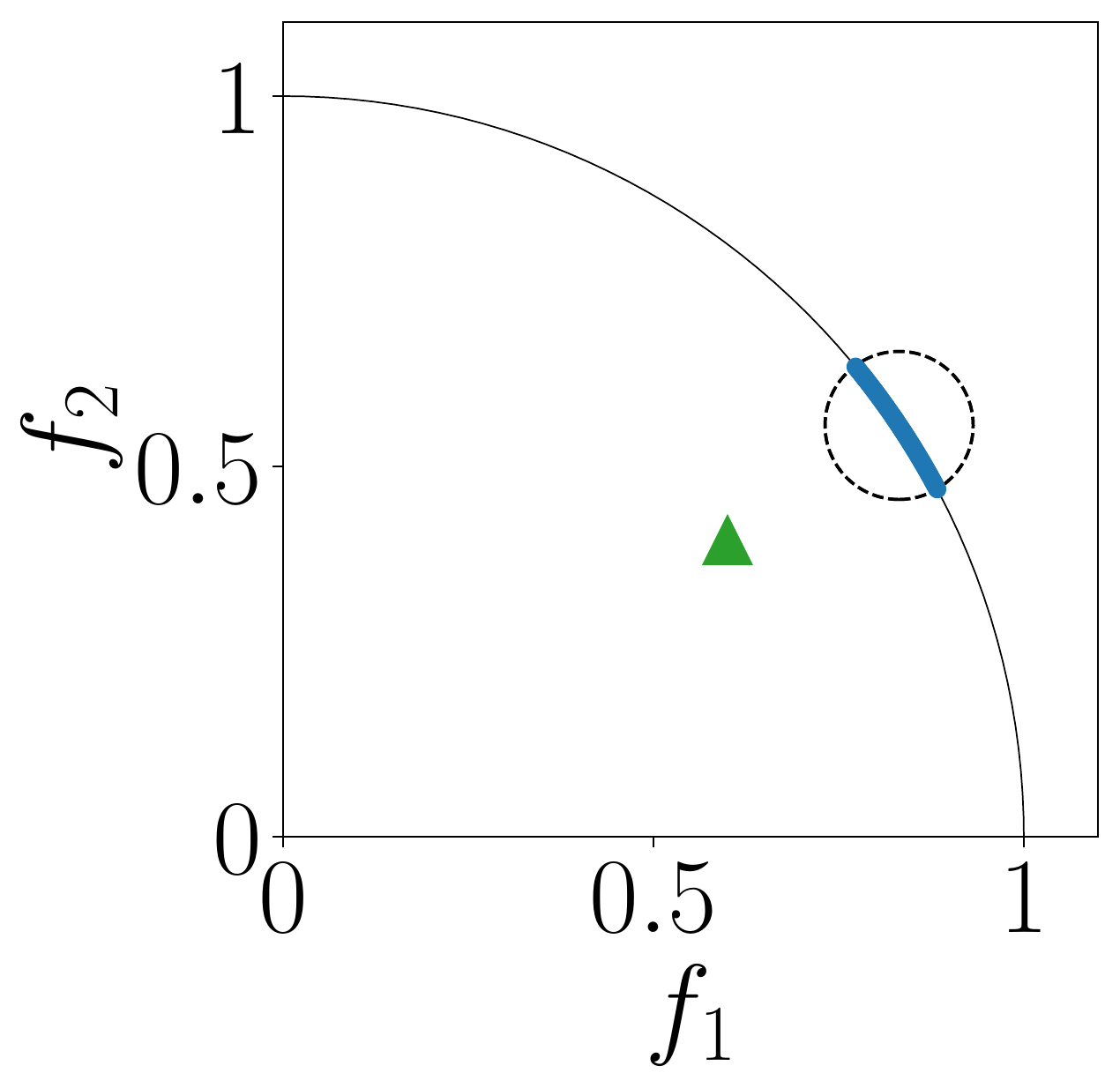}
}
\\
\subfloat[R-MEAD2 (POP)]{
     \includegraphics[width=0.15\textwidth]{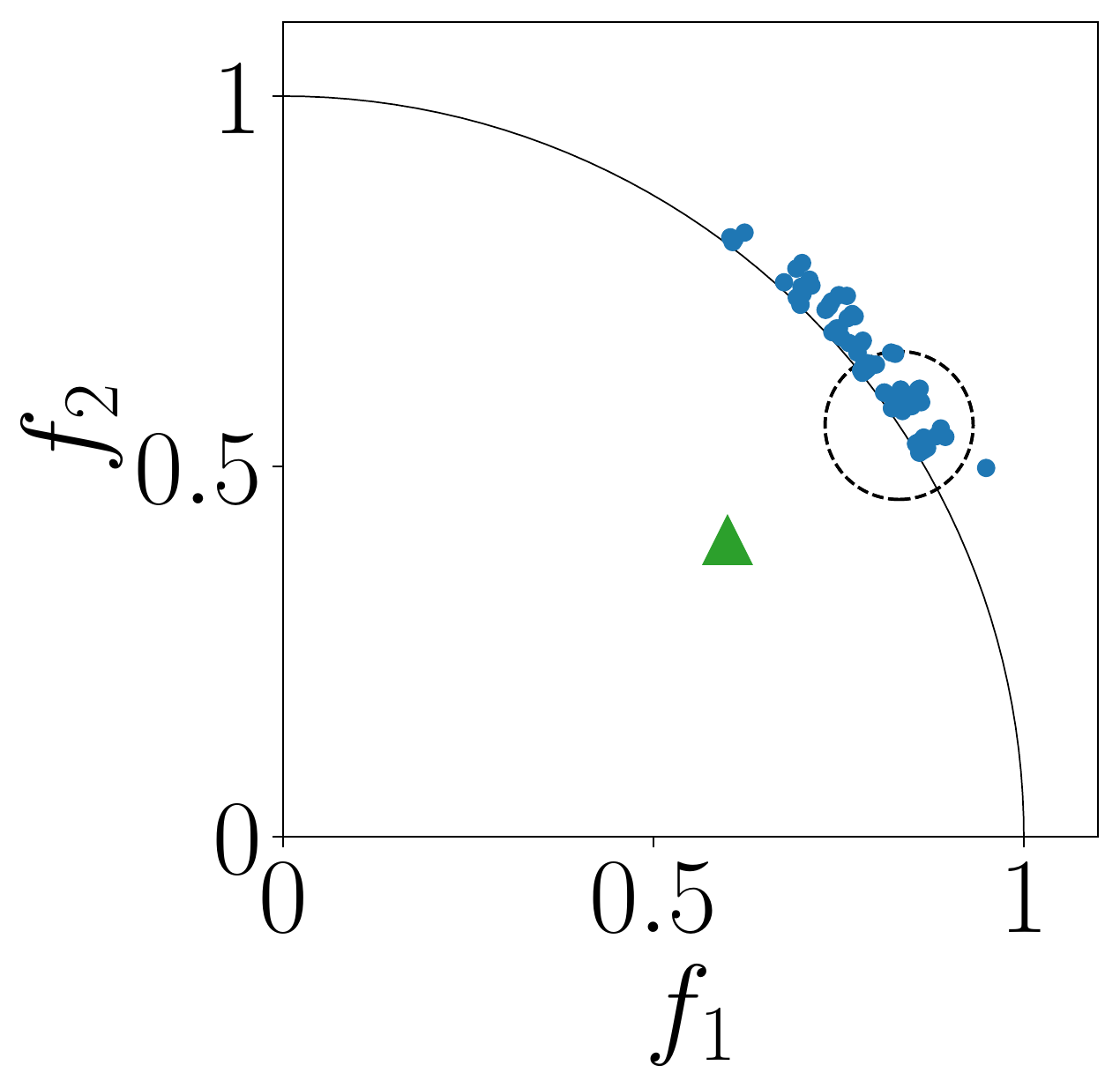}
   }
   \subfloat[R-MEAD2 (UA-IDDS)]{
  \includegraphics[width=0.15\textwidth]{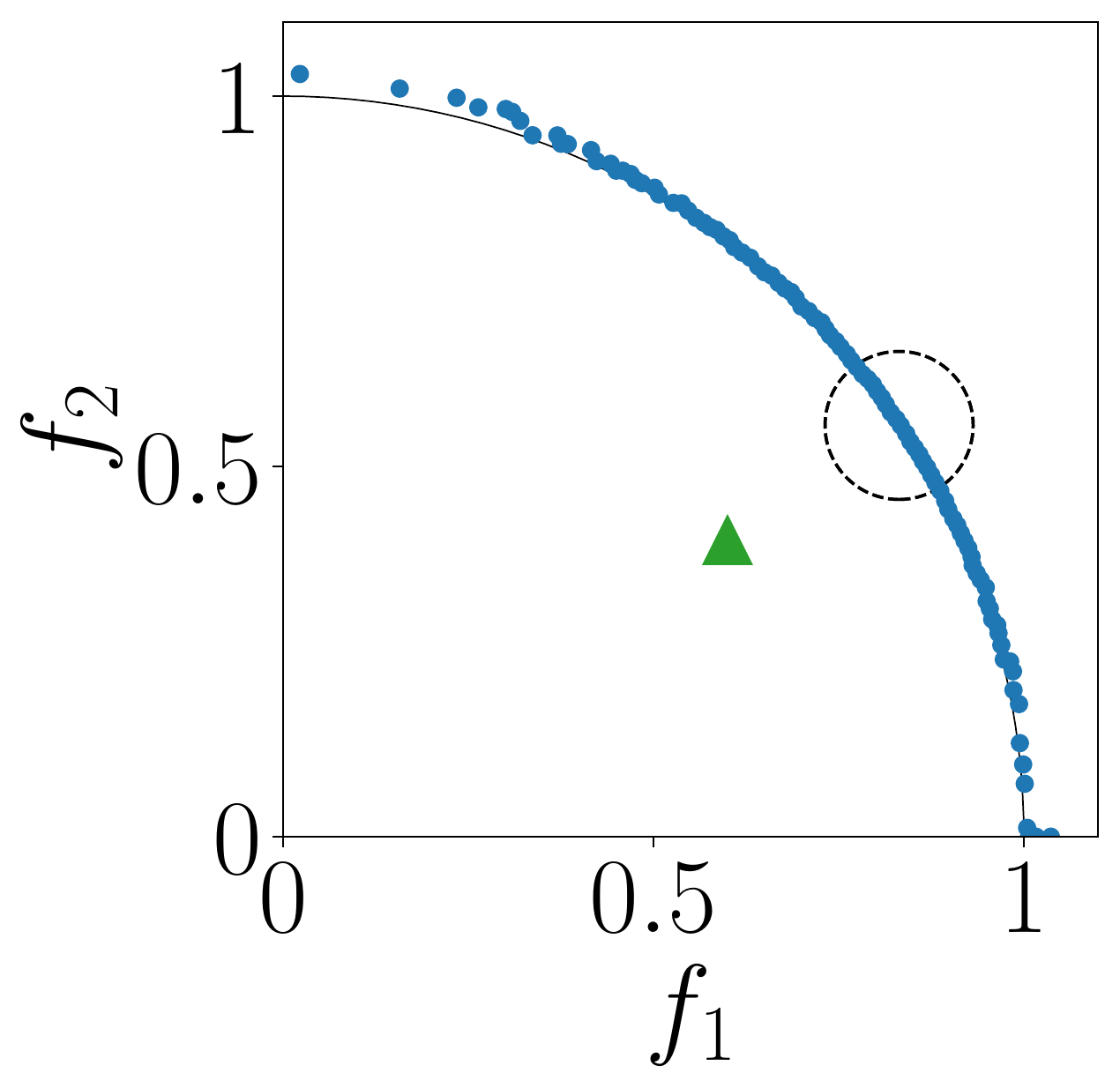}
}
\subfloat[R-MEAD2 (UA-PP)]{
  \includegraphics[width=0.15\textwidth]{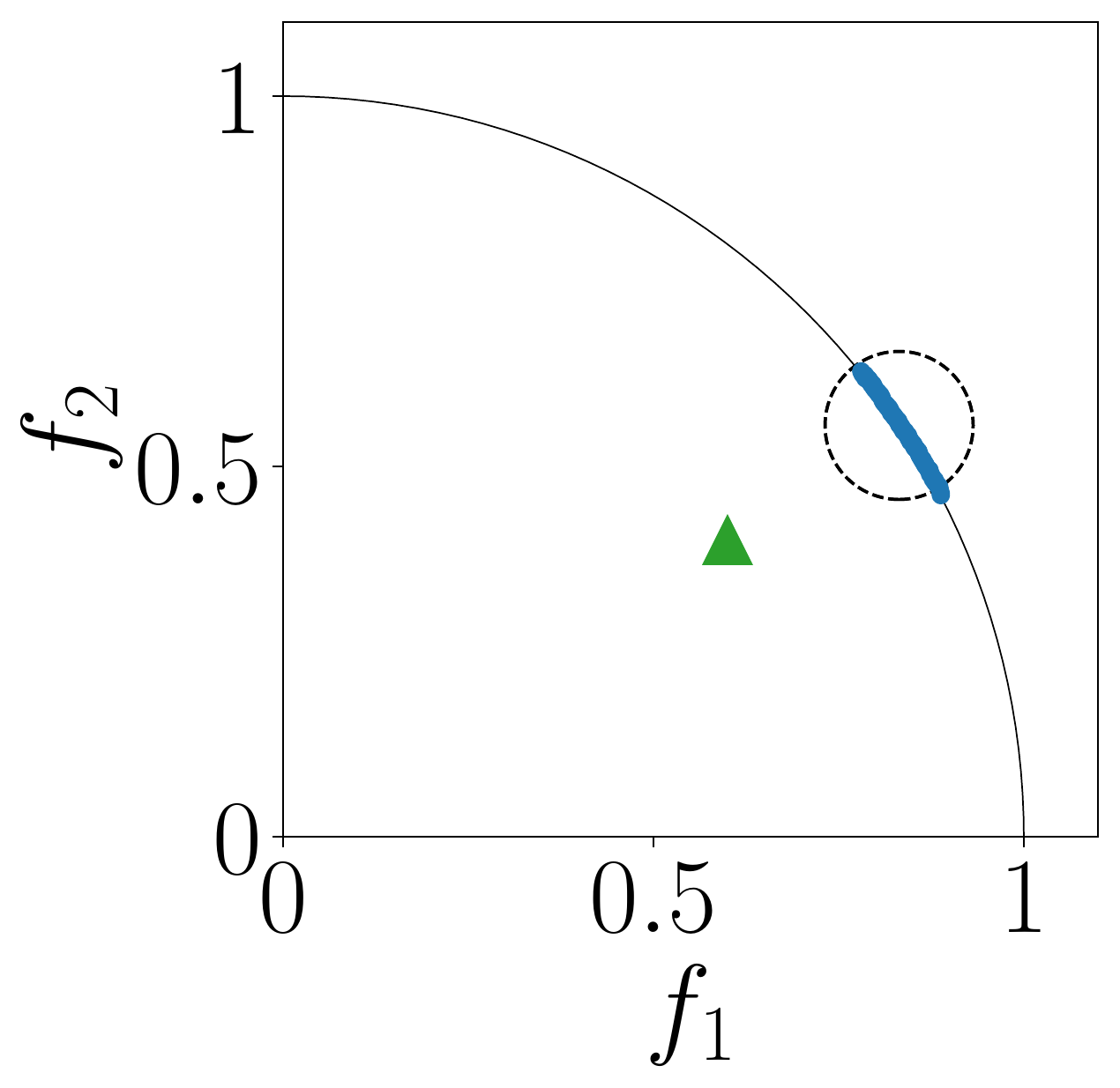}
}
\\
\subfloat[NUMS (POP)]{
     \includegraphics[width=0.15\textwidth]{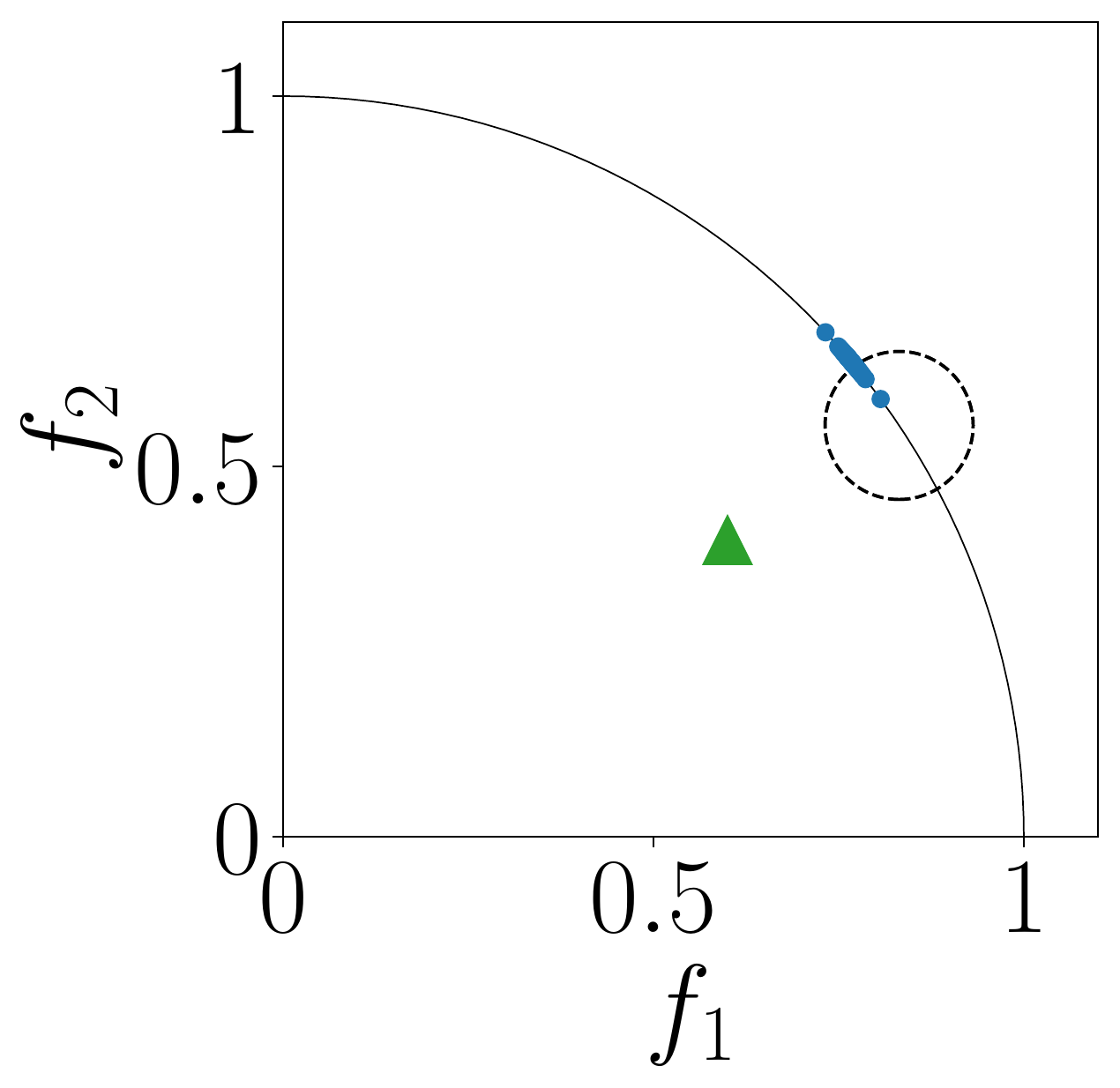}
   }
   \subfloat[NUMS (UA-IDDS)]{
  \includegraphics[width=0.15\textwidth]{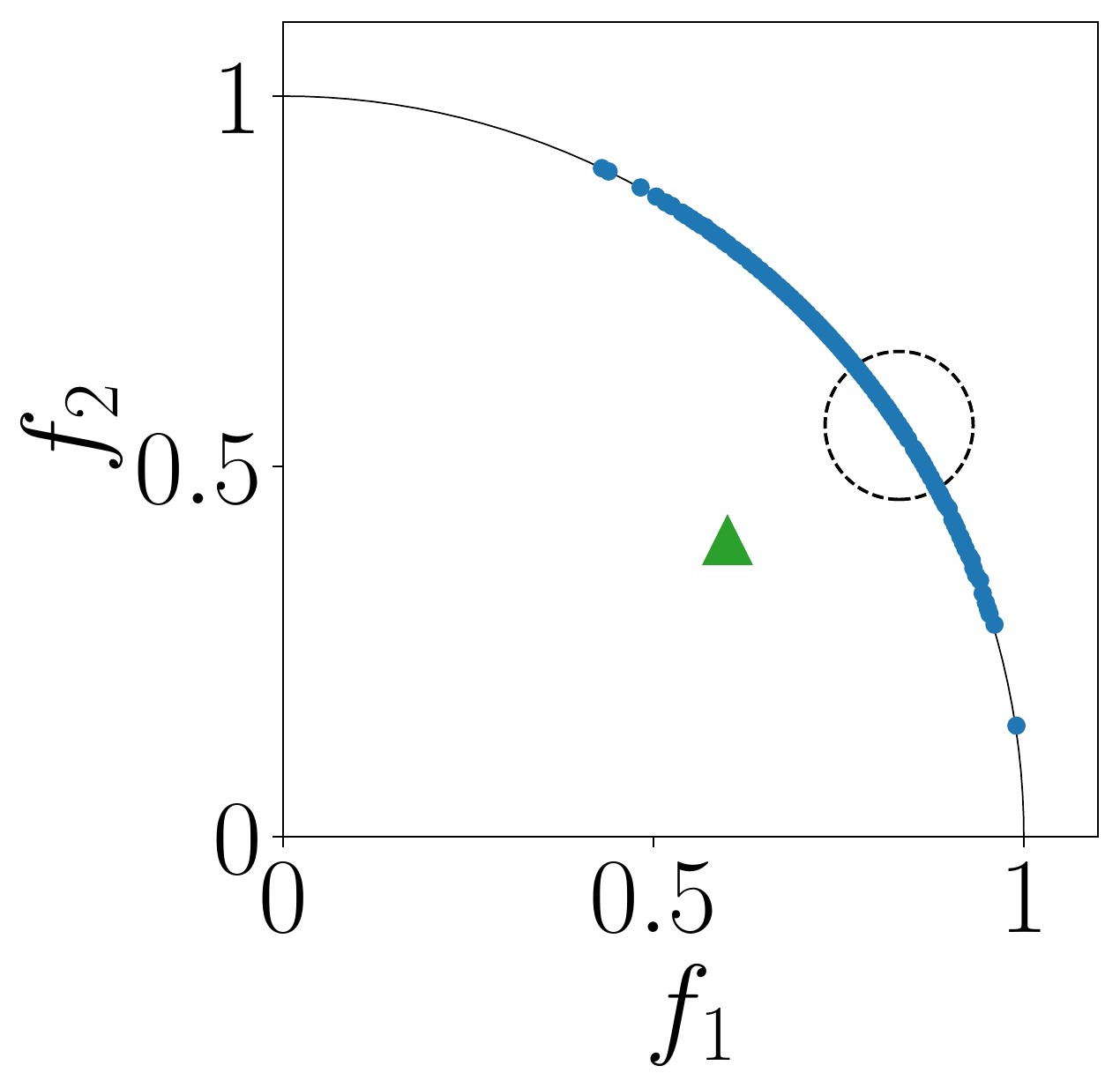}
}
\subfloat[NUMS (UA-PP)]{
  \includegraphics[width=0.15\textwidth]{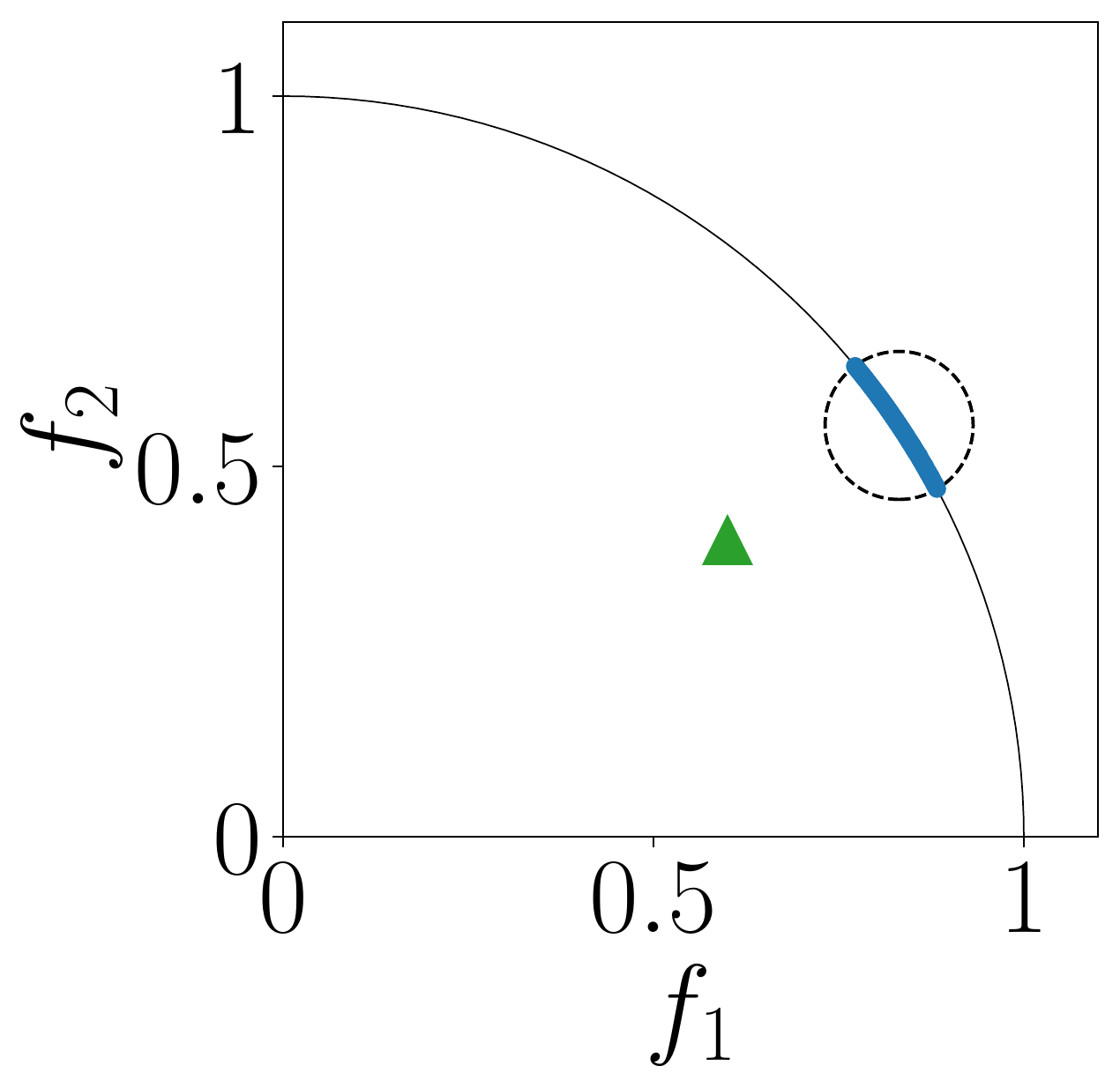}
}
\caption{Distributions of the objective vectors of the solutions in the three solution sets on DTLZ2 with $m=2$, where \tabgreen{$\blacktriangle$} is the reference point $\mathbf{z}$. The dotted circle represents the true ROI. ``NUMS'' stands for MOEA/D-NUMS.}
   \label{fig:100points}
\end{figure}

\pref{fig:100points} shows distributions of the objective vectors of the solutions in POP, UA-IDDS, and UA-PP on the bi-objective DTLZ2 problem.
\pref{fig:100points} shows the result of a single run for each PBEMO algorithm.
Figures \ref{fig:100points_dtlz1}--\ref{fig:100points_dtlz4} show the results on DTLZ1, DTLZ3, and DTLZ4 problems, respectively.

\noindent \textbf{Results of POP.}
As shown in Figures \ref{fig:100points}(a), (d), (g), (j), (m), and (p), objective vectors of some solutions in the final population of the six PBEMO algorithms are near or in the ROI.
However, the diversity of the final population in the ROI is poor for R-NSGA-II, r-NSGA-II, R-MEAD2, and MOEA/D-NUMS.
This is because none of them maintains the best-so-far non-dominated solutions. 
Although PBEA and g-NSGA-II obtain the objective vectors that cover the ROI in the objective space, they do not fit in the ROI.
Here, g-NSGA-II aims to approximate an ROI different from \pref{eq:roi} \cite{TanabeL23}.

\noindent \textbf{Results of UA-IDDS.}
As shown in Figures \ref{fig:100points}(b), (e), (h), (k), (n), and (q), objective vectors of the solution subset found by applying the IDDS method to the UA cover the ROI.
However, their distributions are too wide.
Intuitively, PBEMO algorithms are likely to generate solutions only near the ROI in the objective space.
In contrast, our results show that the six PBEMO algorithms can generate non-dominated solutions far from the ROI.
This may be because the six PBEMO algorithms do not explicitly use a mating restriction.

\noindent \textbf{Results of UA-PP.}
Figures \ref{fig:100points}(c), (f), (i), (l), (o), and (r) are similar to each other.
The objective vectors in the solution subsets obtained by applying the proposed method to the UA are densely distributed in the ROI.
In contrast to the above discussed results in the two cases, the objective vectors in the solution subsets appear to fit the ROI.
We observed similar results on the DTLZ1 and DTLZ3 problems in Figures \ref{fig:100points_dtlz1} and \ref{fig:100points_dtlz3}, where some PBEMO algorithms fail to approximate the ROI of the DTLZ4 problem.

\subsubsection{Comparison using IGD$^+$-C}

\pref{tab:res_3subsets} shows the average IGD$^+$-C values of the three solution subsets on the DTLZ1--DTLZ4 problems over 31 runs.
\pref{tab:res_3subsets} shows the results of R-NSGA-II.
\pref{tab:res_3subsets_five} shows the results of the other PBEMO algorithms.
In \pref{tab:res_3subsets}, the best and second best data are highlighted by \adjustbox{margin=0.1em, bgcolor=c1}{dark gray} and \adjustbox{margin=0.1em, bgcolor=c2}{gray}, respectively.
Let us consider the comparison of two solution sets $\mathcal{X}_1$ and $\mathcal{X}_2$ (e.g., POP and UA-PP) over 31 runs.
The symbols in parentheses in \pref{tab:res_3subsets} indicate that $\mathcal{X}_1$ performs significantly better ($+$) and significantly worse ($-$) than $\mathcal{X}_2$ in terms of  IGD$^+$-C according to the Wilcoxon rank-sum test with $p < 0.05$.
The symbol $\approx$ means that the two results are not significantly different.
We compared 1) a pair of UA-DDS and POP, 2) a pair of UA-PP and POP, and 3) a pair of UA-PP and UA-IDDS.
For example, as seen from \pref{tab:res_3subsets}, UA-IDDS performs significantly worse than POP on DTLZ1 with $m=4$.
UA-PP performs significantly better than POP on DTLZ1 with $m=3$, but UA-PP and UA-IDDS are not significantly different.

As shown in \pref{tab:res_3subsets}, UA-IDDS outperforms POP in many cases.
However, UA-IDDS performs significantly worse than POP on some problems, e.g., DTLZ1 with $m=4$.
Similar results can be found in the results of the other five PBEMO algorithms as shown in \pref{tab:res_3subsets_five}.
This is because the objective vectors in UA-IDDS are widely distributed on the PF as shown in \pref{fig:100points}.

In contrast, UA-PP performs significantly better than POP and UA-IDDS in most cases.
UA-PP does not perform worse than POP.
These results indicate the effectiveness of the UA in PBEMO and the importance of handling the DM's preference information to postprocess the UA.


\begin{tcolorbox}[title=Answers to RQ1, sharpish corners, top=2pt, bottom=2pt, left=4pt, right=4pt, boxrule=0.5pt]
    Our results based on IGD$^+$-C show that the performance of the six PBEMO algorithms can be significantly improved by using the UA.
    We observed that a better solution subset can be found by applying the proposed preference-based postprocessing method to the UA.
    This observation suggests the importance of incorporating the DM's preference information in the postprocessing of the UA.
\end{tcolorbox}

\begin{table}[t]
  \renewcommand{\arraystretch}{0.84} 
\centering
  \caption{\small Average IGD$^+$-C values of the three solution subsets found by R-NSGA-II on the DTLZ1--DTLZ4 problems.}
  \label{tab:res_3subsets}  
{\small
  \begin{tabular}{ccccc}
\toprule
Problem & $m$ & POP & UA-IDDS & UA-PP\\  
\midrule
 & 2 & 0.0236 & \cellcolor{c2}0.0018 ($+$) & \cellcolor{c1}0.0012 ($+$, $+$)\\
 & 3 & 0.0334 & \cellcolor{c1}0.0211 ($+$) & \cellcolor{c2}0.0220 ($+$, $\approx$)\\
DTLZ1 & 4 & \cellcolor{c2}0.0562 & 0.0743 ($-$) & \cellcolor{c1}0.0442 ($+$, $+$)\\
 & 5 & 0.0933 & \cellcolor{c2}0.0782 ($+$) & \cellcolor{c1}0.0558 ($+$, $+$)\\
 & 6 & 0.1131 & \cellcolor{c2}0.0779 ($+$) & \cellcolor{c1}0.0695 ($+$, $+$)\\
\midrule
 & 2 & 0.0411 & \cellcolor{c2}0.0016 ($+$) & \cellcolor{c1}0.0004 ($+$, $+$)\\
 & 3 & 0.1247 & \cellcolor{c2}0.0276 ($+$) & \cellcolor{c1}0.0114 ($+$, $+$)\\
DTLZ2 & 4 & 0.1986 & \cellcolor{c2}0.0720 ($+$) & \cellcolor{c1}0.0339 ($+$, $+$)\\
 & 5 & 0.2729 & \cellcolor{c2}0.1162 ($+$) & \cellcolor{c1}0.0600 ($+$, $+$)\\
 & 6 & 0.2840 & \cellcolor{c2}0.1540 ($+$) & \cellcolor{c1}0.0853 ($+$, $+$)\\
\midrule
 & 2 & 0.0345 & \cellcolor{c2}0.0083 ($+$) & \cellcolor{c1}0.0078 ($+$, $\approx$)\\
 & 3 & 0.1083 & \cellcolor{c2}0.0460 ($+$) & \cellcolor{c1}0.0309 ($+$, $+$)\\
DTLZ3 & 4 & 0.1988 & \cellcolor{c2}0.1736 ($\approx$) & \cellcolor{c1}0.0636 ($+$, $+$)\\
 & 5 & 0.2370 & \cellcolor{c2}0.2169 ($\approx$) & \cellcolor{c1}0.1024 ($+$, $+$)\\
 & 6 & 0.8749 & \cellcolor{c2}0.8287 ($+$) & \cellcolor{c1}0.7224 ($+$, $+$)\\
\midrule
 & 2 & 0.1014 & \cellcolor{c2}0.0829 ($\approx$) & \cellcolor{c1}0.0818 ($\approx$, $\approx$)\\
 & 3 & 0.0838 & \cellcolor{c2}0.0528 ($+$) & \cellcolor{c1}0.0375 ($+$, $+$)\\
DTLZ4 & 4 & 0.1030 & \cellcolor{c2}0.0741 ($+$) & \cellcolor{c1}0.0465 ($+$, $+$)\\
 & 5 & 0.3757 & \cellcolor{c2}0.1206 ($+$) & \cellcolor{c1}0.0759 ($+$, $+$)\\
 & 6 & 0.3214 & \cellcolor{c2}0.1144 ($+$) & \cellcolor{c1}0.0655 ($+$, $+$)\\
\bottomrule
\end{tabular}
}
\end{table}

\begin{table}[t]
  \renewcommand{\arraystretch}{0.84} 
\centering
  \caption{\small The best population size $\mu$ for each number of function evaluations and $m \in \{2, 4, 6\}$ on the DTLZ1--DTLZ4 problems.}
  \label{tab:best_mu}  
{\small
\subfloat[$m=2$]{
\begin{tabular}{lccccccc}
\toprule
PBEMO & $1$K FEs & $5$K FEs  & $10$K FEs  & $30$K FEs  & $50$K FEs\\
\midrule
R-NSGA-II & 8 & 20 & 8 & 8 & 200\\
r-NSGA-II & 20 & 20 & 20 & 40 & 40\\
g-NSGA-II & 20 & 40 & 40 & 200 & 100\\
PBEA & 20 & 40 & 8 & 100 & 200\\
R-MEAD2 & 8 & 8 & 8 & 20 & 20\\
MOEA/D-NUMS & 8 & 8 & 8 & 20 & 20\\
\bottomrule
\end{tabular}
}
\\
\subfloat[$m=4$]{
\begin{tabular}{lccccccc}
\toprule
PBEMO & $1$K FEs & $5$K FEs  & $10$K FEs  & $30$K FEs  & $50$K FEs\\
\midrule
R-NSGA-II & 8 & 40 & 20 & 20 & 20\\
r-NSGA-II & 40 & 40 & 40 & 40 & 40\\
g-NSGA-II & 40 & 100 & 200 & 400 & 400\\
PBEA & 20 & 100 & 40 & 100 & 200\\
R-MEAD2 & 20 & 100 & 100 & 100 & 20\\
MOEA/D-NUMS & 8 & 20 & 20 & 20 & 20\\
\bottomrule
\end{tabular}
}
\\
\subfloat[$m=6$]{
\begin{tabular}{lccccccc}
\toprule
PBEMO & $1$K FEs & $5$K FEs  & $10$K FEs  & $30$K FEs  & $50$K FEs\\
\midrule
R-NSGA-II & 20 & 40 & 100 & 40 & 100\\
r-NSGA-II & 40 & 40 & 100 & 100 & 100\\
g-NSGA-II & 8 & 8 & 300 & 8 & 8\\
PBEA & 20 & 40 & 100 & 300 & 300\\
R-MEAD2 & 8 & 100 & 200 & 20 & 40\\
MOEA/D-NUMS & 20 & 40 & 40 & 20 & 20\\
\bottomrule
\end{tabular}
}
}
\end{table}

\begin{figure*}[t]
  \centering
  \subfloat{\includegraphics[width=0.7\textwidth]{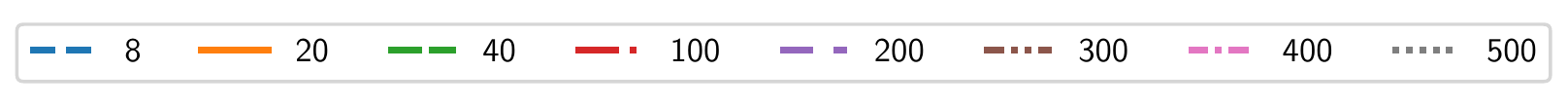}}
\vspace{-3.9mm}
  \\
   \subfloat[DTLZ2 with $m=2$]{\includegraphics[width=0.32\textwidth]{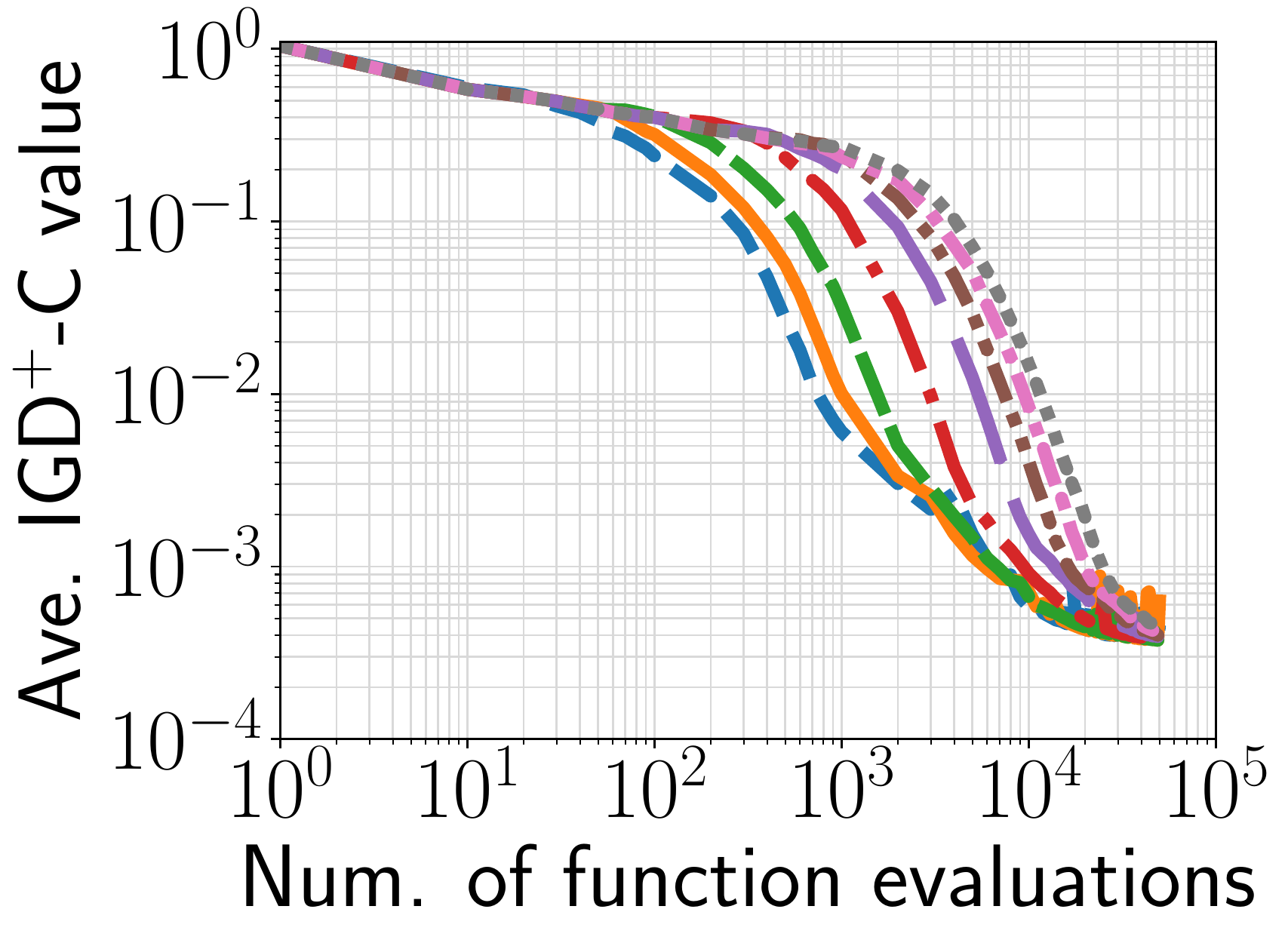}}
   \subfloat[DTLZ2 with $m=6$]{\includegraphics[width=0.32\textwidth]{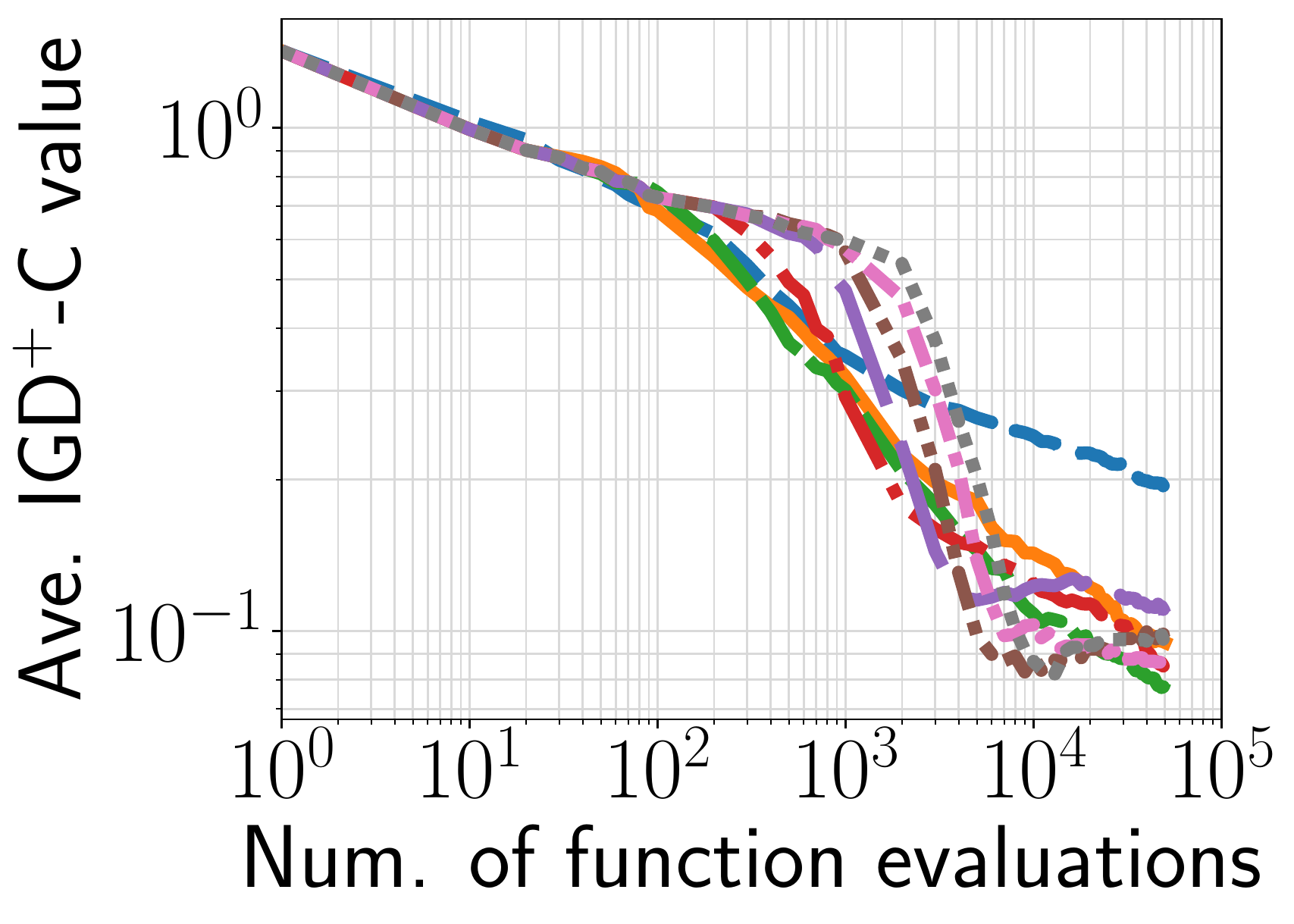}}
   \subfloat[DTLZ4 with $m=4$]{\includegraphics[width=0.32\textwidth]{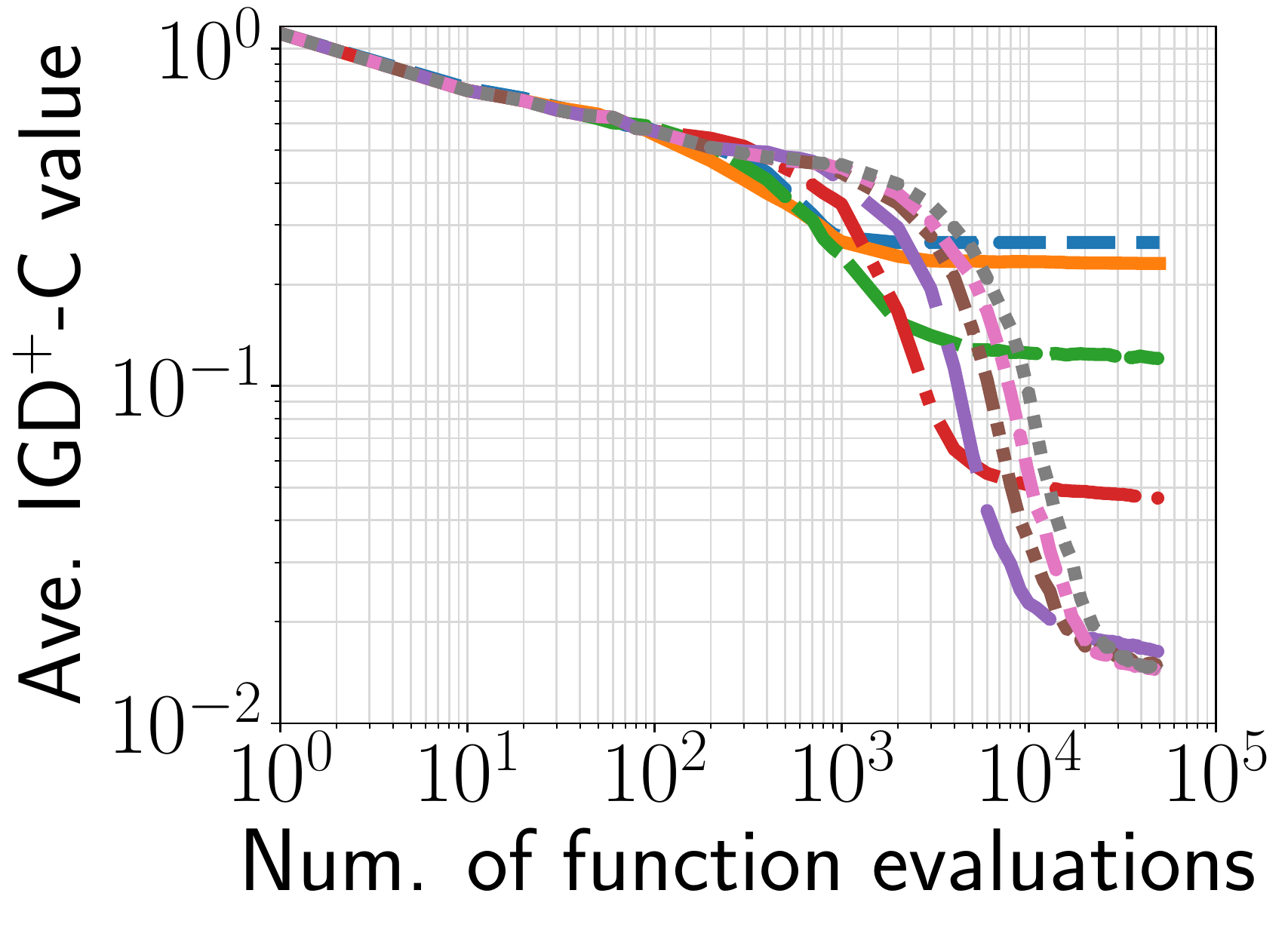}}
         \caption{Average IGD$^+$-C values of R-NSGA-II with different population sizes on three problems.}
   \label{fig:rnsga2_dtlz}
\end{figure*}

\subsection{The impact of the population size}
\label{sec:results_pop}

Based on the promising results of UA-PP in \pref{sec:results_pp}, this section and \pref{sec:results_roi_size} evaluate the performance of PBEMO algorithms using the UA and the proposed postprocessing method.
We performed 31 independent runs of the six PBEMO algorithms with the population sizes $\mu \in \{8, \ldots, 500\}$ on the DTLZ1--DTLZ4 problems.
Then, for each PBEMO algorithm, we calculated the average rankings of the population sizes $\mu \in \{8, \ldots, 500\}$ by the Friedman test \cite{DerracGMH11}.
We used the \texttt{CONTROLTEST} package (\url{https://sci2s.ugr.es/sicidm}) to calculate the rankings based on the IGD$^+$-C values.
Below, we discuss the best-ranked population size for each number of objectives $m$ and each budget of function evaluations.

\pref{tab:best_mu} shows the best population size $\mu$ for $m \in \{2, 4, 6\}$ and $10^3, 5 \times 10^3, 10^4, 3 \times 10^4, 5 \times 10^4$ function evaluations on all of the DTLZ1--DTLZ4 problems.
As seen from \pref{tab:best_mu}, $\mu=500$ is not the best for any case.
\pref{tab:sup_rankings_mu} also shows the rankings of the eight $\mu$ values.
As shown in \pref{tab:sup_rankings_mu}, $\mu=500$ is the worst in most cases.
Although some previous studies used a relatively large population size as shown in \pref{tab:popsize_literature}, this result suggests that a smaller population size than commonly used is effective for the PBEMO algorithms.
Since we used the UA, the six PBEMO algorithms do not need to maintain good solutions for decision-making.
This may be one of the reasons why the six PBEMO algorithms do not require a large population size in this study.

As shown in \pref{tab:best_mu}(a), the best population size depends on the available budget of function evaluations.
For example, $\mu=8$ and $\mu=200$ are the best for $10^3$ and $5 \times 10^4$ function evaluations for R-NSGA-II.
This property is more pronounced in the results of PBEA for many objectives.
For example, as shown in the results of PBEA in \pref{tab:best_mu}(c),  $\mu=20$ and $\mu=300$ perform the best for $10^3$ and $5 \times 10^4$ function evaluations for $m=6$.
These results are consistent with the conclusions of the previous studies that investigated the influence of the population size on EMO algorithms (e.g., \cite{DymondKH13,BrockhoffTH15,IshibuchiSMN16a,TanabeI18}).

In contrast, we observed unexpected results for some PBEMO algorithms.
For example, as shown in \pref{tab:best_mu}(c), $\mu=8$ is suitable for R-MEAD2 for $m=6$ and $10^3$ function evaluations.
For  $10^4$ function evaluations, $\mu=200$ is the best for R-MEAD2.
However, $\mu=20$ and $40$ are suitable for R-MEAD2 again for $3 \times 10^4$ and $5 \times 10^4$ function evaluations, respectively.

Below, we discuss the results on each test problem.
\pref{fig:rnsga2_dtlz} shows the average IGD$^+$-C values of R-NSGA-II with different population sizes on DTLZ2 with $m \in \{2, 6\}$ and DTLZ4 with $m=4$.
Figures \ref{fig:sup_Rnsgaii_dtlz}--\ref{fig:sup_nums_dtlz} show the results of the six PBEMO algorithms on all test problems.
We do not discuss Figures \ref{fig:sup_Rnsgaii_dtlz}--\ref{fig:sup_nums_dtlz} due to the paper length limitation, but their results are similar to \pref{fig:rnsga2_dtlz}.

As shown in \pref{fig:rnsga2_dtlz}(a), R-NSGA-II with the smallest population size ($\mu=8$) performs the best on DTLZ2 with $m=2$ for a small budget of function evaluations.
However, R-NSGA-II with $\mu=8$ performs poorly as the number of function evaluations increases.
As shown in \pref{fig:rnsga2_dtlz}(b), this observation is more pronounced for many objectives.
As seen \pref{fig:rnsga2_dtlz}(b), the average IGD$^+$-C values of R-NSGA-II with some $\mu$ values do not decrease monotonically as the number of function evaluations increases.
This may be one of the reasons why R-NSGA-II with a relatively large population size (e.g., $\mu \in \{200, 300\}$) works well for $10^4$ function evaluations, as discussed above.
A monotonic decrease in the IGD$^+$-C value may be achieved by using an indicator-based subset selection method \cite{BasseurDGL16} instead of the IDDS method.
However, indicator-based subset selection methods generally require a higher computational cost than distance-based subset selection methods (e.g., the IDDS method).
A further investigation is another future work.
As shown in \pref{fig:rnsga2_dtlz}(c), R-NSGA-II with a small population size performs poorly on DTLZ4 with $m=4$ as the search progresses.
This observation suggests that PBEMO algorithms require a large population size on problems with a biased density of solutions.

\begin{tcolorbox}[title=Answers to RQ2, sharpish corners, top=2pt, bottom=2pt, left=4pt, right=4pt, boxrule=0.5pt]
  Our results show that a smaller population size than commonly used is effective for the six PBEMO algorithms for a small budget of function evaluations even for many objectives.
We observed that the PBEMO algorithms require a large population size as the search progresses.
However, our results suggest that a general rule of thumb ``\textit{use a large population size on many-objective optimization problems with a large budget of function evaluations}'' may not be very helpful in determining the population size in PBEMO.
Note that, of course, the best population size also depends on the type of PBEMO algorithm.
\end{tcolorbox}

\subsection{How the size of the ROI affects the best population size}
\label{sec:results_roi_size}

In Sections \ref{sec:results_pp} and \ref{sec:results_pop}, we set the radius of the ROI $r$ to 0.1.
Little is also known about the influence of the size of the ROI on the performance of PBEMO algorithms.
Intuitively, PBEMO algorithms require a large-sized population to cover a large subregion of the PF.
Thus, it is likely that the best population size depends on $r$.

\pref{tab:effect_roirad} shows the best population size $\mu$ of R-NSGA-II on the DTLZ1--DTLZ4 problems with $r \in \{0.01, 0.05, 0.1, 0.2, 0.3\}$, where the results for $r=0.1$ are the same as \pref{tab:best_mu}.
Similar to \pref{tab:best_mu}, we determined the best population size according to the average rankings by the Friedman test.
Tables \ref{tab:sup_effect_roirad_rnsga2}--\ref{tab:sup_effect_roirad_nums} show the results of the other PBEMO algorithms, but they are similar to \pref{tab:effect_roirad}.

As shown in \pref{tab:effect_roirad}, the best $\mu$ values for $r \in \{0.01, 0.05, 0.1\}$ are almost the same.
In addition, a small population size is the best for a small budget of function evaluations.
In contrast, as seen from the results for $5 \times 10^4$ function evaluations in \pref{tab:effect_roirad}, the best population size becomes large as $r$ increases.
However, we believe that $r=0.3$ is too large as the size of the ROI.
Recall that $r$ is set to $0.1$ in the example in \pref{fig:roi}.
If $r=0.3$, the ROI in \pref{fig:roi} is almost the same as the PF.
In preference-based multi-objective optimization, the DM is unlikely to want the solution set that covers the whole PF in the objective space (see \pref{sec:def_PBMOPs}).

\begin{tcolorbox}[title=Answers to RQ3, sharpish corners, top=2pt, bottom=2pt, left=4pt, right=4pt, boxrule=0.5pt]
Our results show that the six PBEMO algorithms require a large-sized population when $r$, $m$, and a budget of function evaluations are large. 
  However, the DM is unlikely to prefer a large $r$ value.
  Only a limited budget of function evaluations is also available for some real-world problems \cite{ChughSHM19}.
  Thus, we believe that a relatively small population size is a reasonable first choice in practice.
\end{tcolorbox}

\begin{table}[t]
  \renewcommand{\arraystretch}{0.8} 
\centering
  \caption{\small The best population size $\mu$ of R-NSGA-II on the DTLZ1--DTLZ4 problems with $r \in \{0.01, 0.05, 0.1, 0.2, 0.3\}$.}
  \label{tab:effect_roirad}  
{\small
\subfloat[$m=2$]{
\begin{tabular}{lccccccc}
\toprule
$r$ & $1$K FEs & $5$K FEs  & $10$K FEs  & $30$K FEs  & $50$K FEs\\
\midrule
0.01 & 8 & 20 & 8 & 20 & 20\\
0.05 & 8 & 20 & 8 & 20 & 20\\
0.1 & 8 & 20 & 8 & 8 & 200\\
0.2 & 20 & 20 & 8 & 100 & 200\\
0.3 & 20 & 40 & 8 & 100 & 300\\
\bottomrule
\end{tabular}
}
\\
\subfloat[$m=4$]{
\begin{tabular}{lccccccc}
\toprule
$r$ & $1$K FEs & $5$K FEs  & $10$K FEs  & $30$K FEs  & $50$K FEs\\
\midrule
0.01 & 8 & 40 & 20 & 20 & 40\\
0.05 & 8 & 40 & 20 & 40 & 40\\
0.1 & 8 & 40 & 20 & 20 & 20\\
0.2 & 8 & 40 & 200 & 40 & 40\\
0.3 & 8 & 40 & 200 & 20 & 20\\
\bottomrule
\end{tabular}
}
\\
\subfloat[$m=6$]{
\begin{tabular}{lccccccc}
\toprule
$r$ & $1$K FEs & $5$K FEs  & $10$K FEs  & $30$K FEs  & $50$K FEs\\
\midrule
0.01 & 20 & 40 & 300 & 40 & 40\\
0.05 & 20 & 40 & 300 & 40 & 40\\
0.1 & 20 & 40 & 100 & 40 & 100\\
0.2 & 20 & 100 & 100 & 200 & 300\\
0.3 & 20 & 100 & 100 & 200 & 300\\
\bottomrule
\end{tabular}
}
}
\end{table}

\section{Conclusion}
\label{sec:conclusion}

In this paper, we first proposed the preference-based method for postprocessing the UA in PBEMO.
Unlike the existing postprocessing methods, the proposed method can handle the DM's preference information defined by the reference point $\mathbf{z}$.
Then, we investigated the effectiveness of the proposed preference-based postprocessing method on the DTLZ problems \cite{DebTLZ05} with the number of objectives $m \in \{2, \ldots, 6\}$.
The results showed that the proposed method can obtain a better subset of the UA than the IDDS method \cite{ShangIN21} in most cases.
The results also showed that the use of the UA and the proposed method can significantly improve the performance of the six PBEMO algorithms (R-NSGA-II \cite{DebSBC06}, r-NSGA-II \cite{SaidBG10}, g-NSGA-II \cite{MolinaSHCC09}, PBEA \cite{ThieleMKL09}, R-MEAD2 \cite{MohammadiOLD14}, and MOEA/D-NUMS \cite{LiCMY18}). 
Our results revealed the difficulty in selecting a suitable population size in PBEMO algorithms.
Although the best population size depends on some factors, our results suggest that a small population size (e.g., $\mu \in \{8, 20, 40\}$) for PBEMO algorithms is a good first choice, even for many objectives.

Although the benchmarking methodology has been investigated in the EMO literature (e.g., \cite{BrockhoffTH15,IshibuchiPS22,BrockhoffAHT22}), it has received little attention in the PBEMO literature.
As reviewed in \cite{AfsarMR21}, benchmarking PBEMO algorithms is a challenging task.
Our findings are helpful in establishing a systematic methodology for benchmarking PBEMO algorithms.
This paper focused only on PBEMO algorithms using a single reference point.
An investigation for multiple reference points is needed in future work.
An analysis of PBEMO algorithms using other preference expressions (e.g., a value function) is another topic for future work.

\begin{acks}
  This work was supported by JSPS KAKENHI Grant Number \seqsplit{21K17824} and LEADER, MEXT, Japan.

\end{acks}

\bibliographystyle{ACM-Reference-Format}
\bibliography{reference} 

\clearpage

\begin{figure*}
\centering
\fontsize{30pt}{100pt}\selectfont{\textbf{Supplement}}
\end{figure*}




\appendix

\setcounter{figure}{0}
\setcounter{table}{0}

\renewcommand{\thesection}{S.\arabic{section}}
\renewcommand{\thetable}{S.\arabic{table}}
\renewcommand{\thefigure}{S.\arabic{figure}}
\renewcommand\thealgocf{S.\arabic{algocf}} 
\renewcommand{\theequation}{S.\arabic{equation}}

\makeatletter
\renewcommand{\@biblabel}[1]{[S.#1]}
\renewcommand{\@cite}[1]{[S.#1]}
\makeatother

\definecolor{c1}{RGB}{150,150,150}
\definecolor{c2}{RGB}{220,220,220}

\clearpage

\begin{figure*}[t]
   \centering
   \subfloat[R-NSGA-II (POP)]{
     \includegraphics[width=0.15\textwidth]{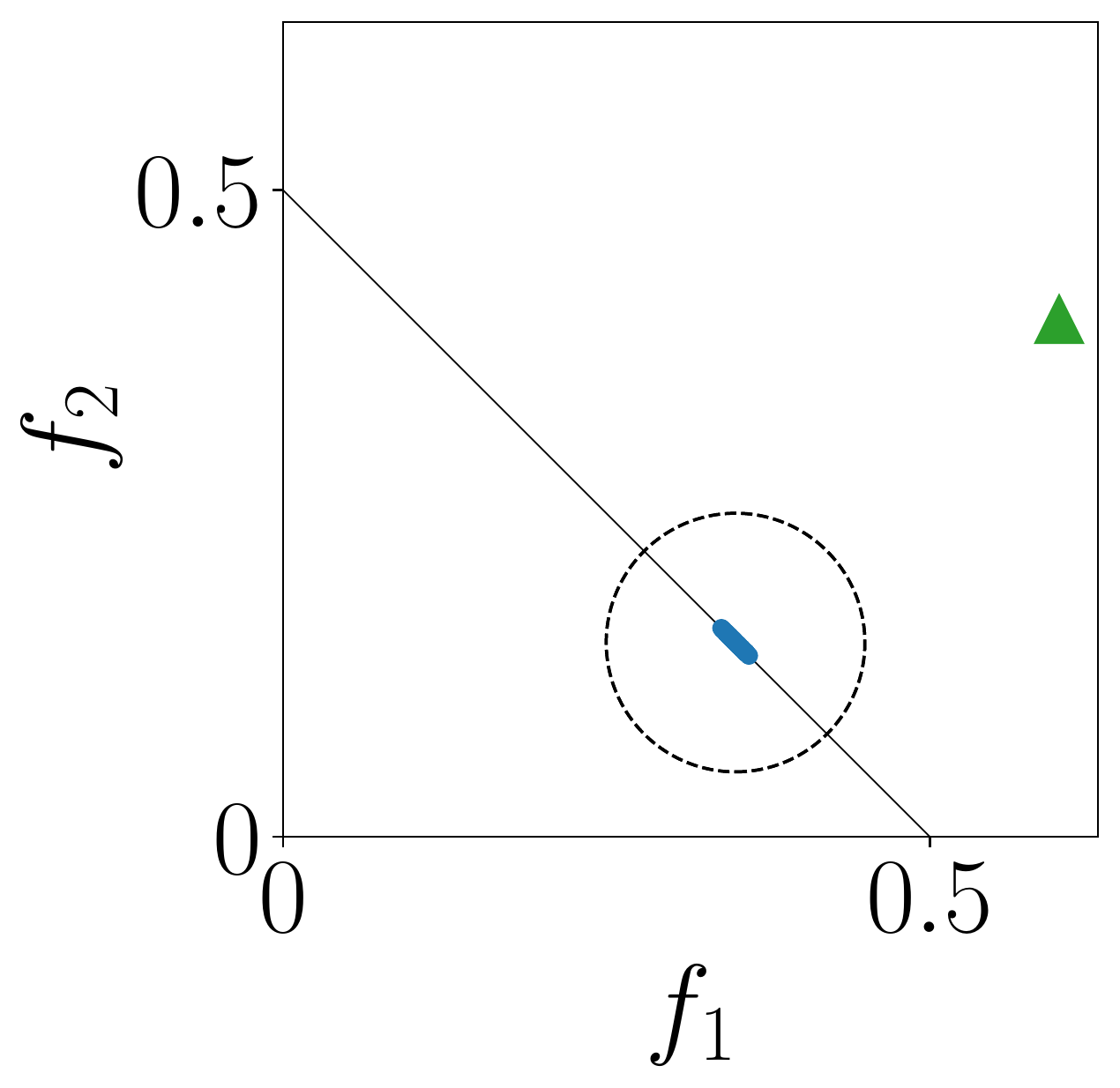}
   }
   \subfloat[R-NSGA-II (UA-IDDS)]{
  \includegraphics[width=0.15\textwidth]{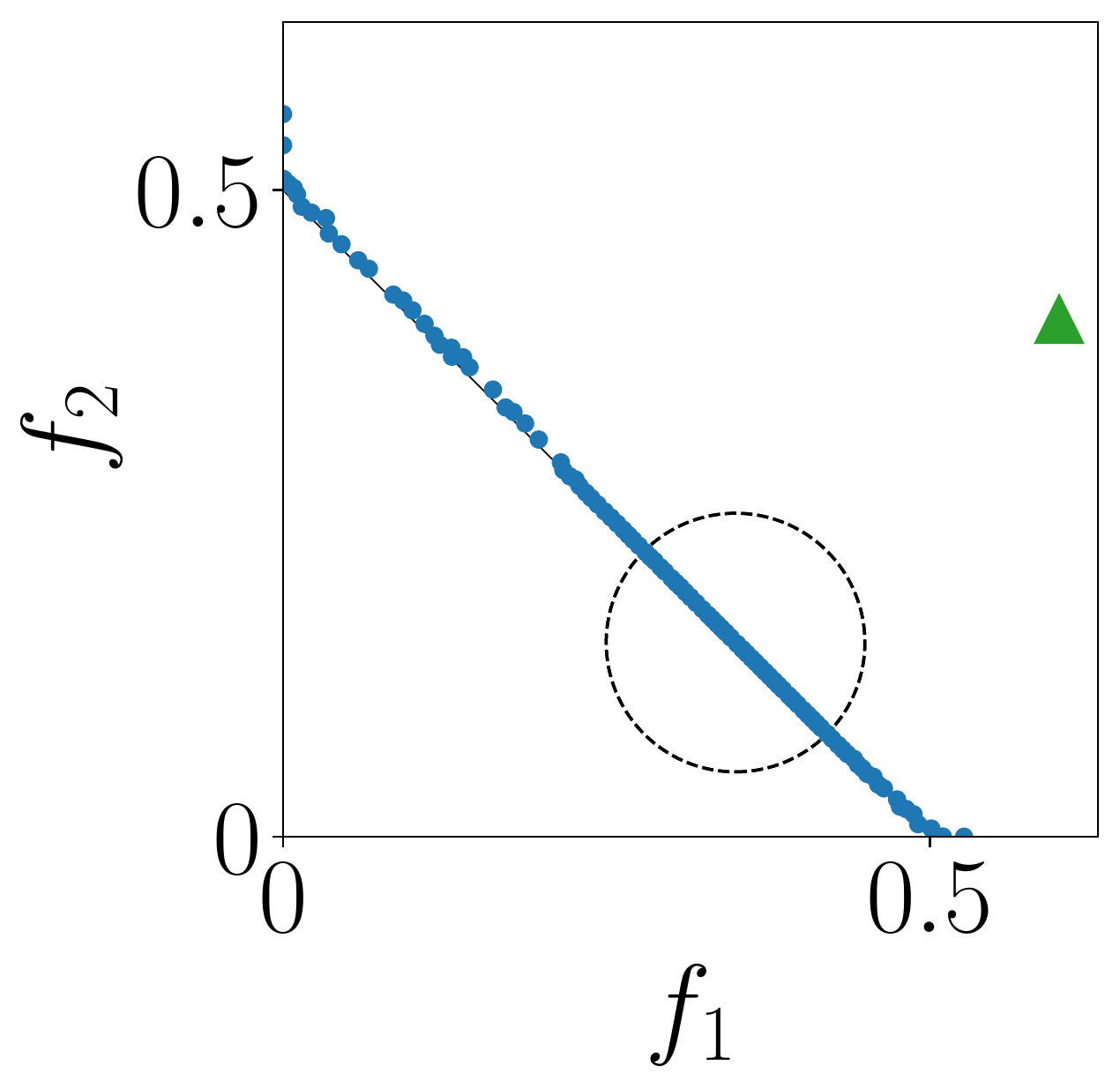}
}
\subfloat[R-NSGA-II (UA-PP)]{
  \includegraphics[width=0.15\textwidth]{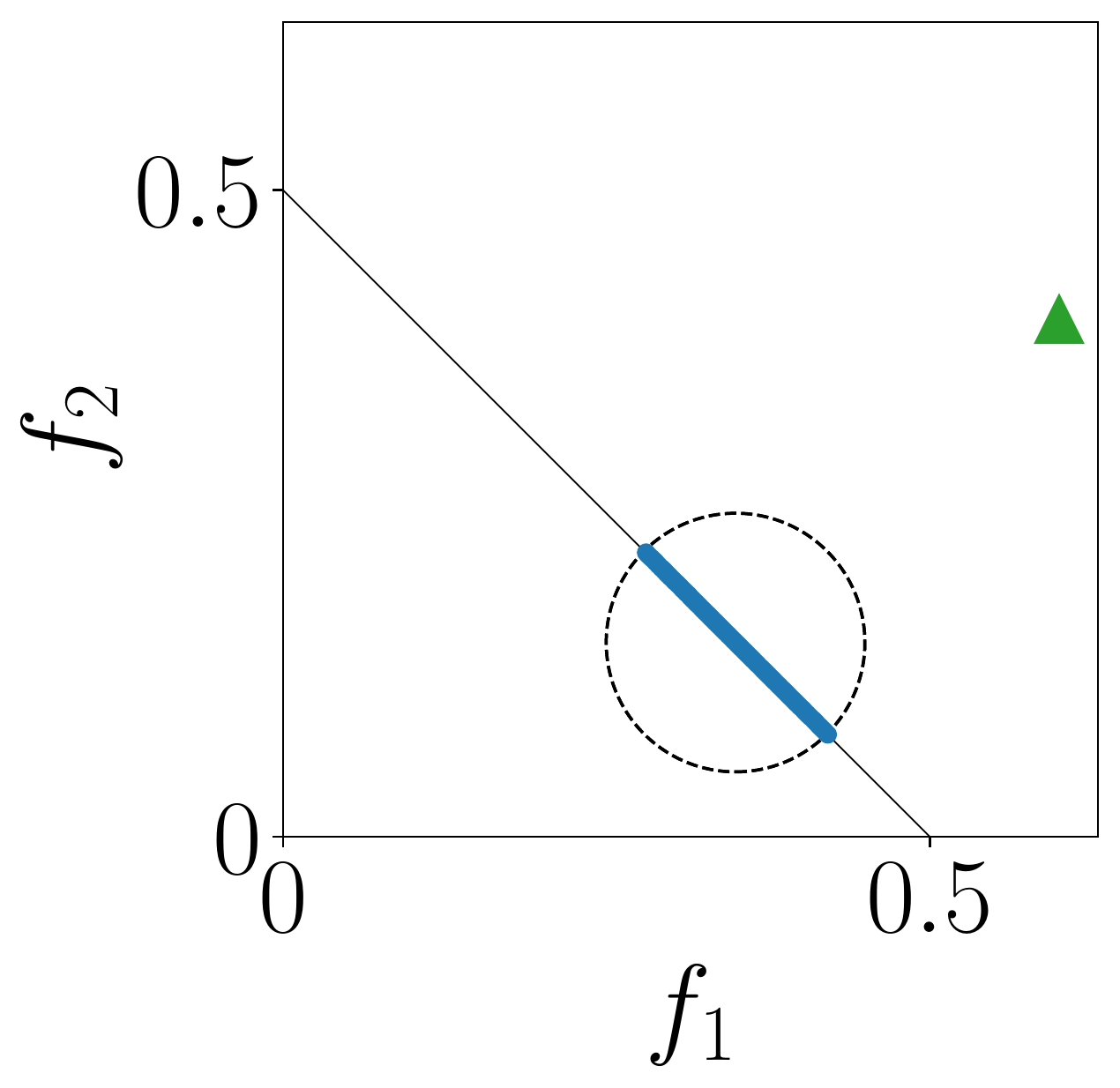}
}
\\
\subfloat[r-NSGA-II (POP)]{
     \includegraphics[width=0.15\textwidth]{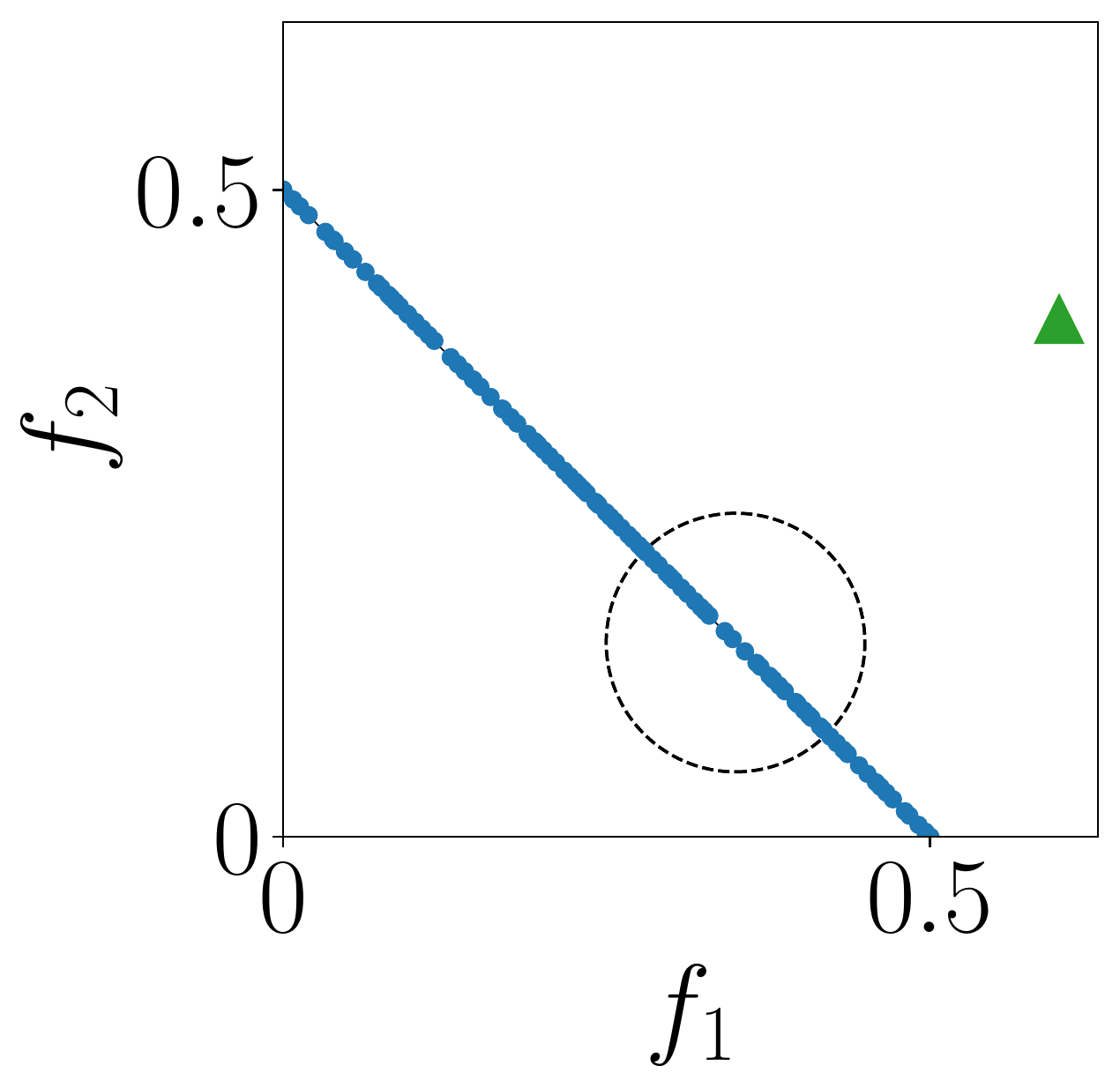}
   }
   \subfloat[r-NSGA-II (UA-IDDS)]{
  \includegraphics[width=0.15\textwidth]{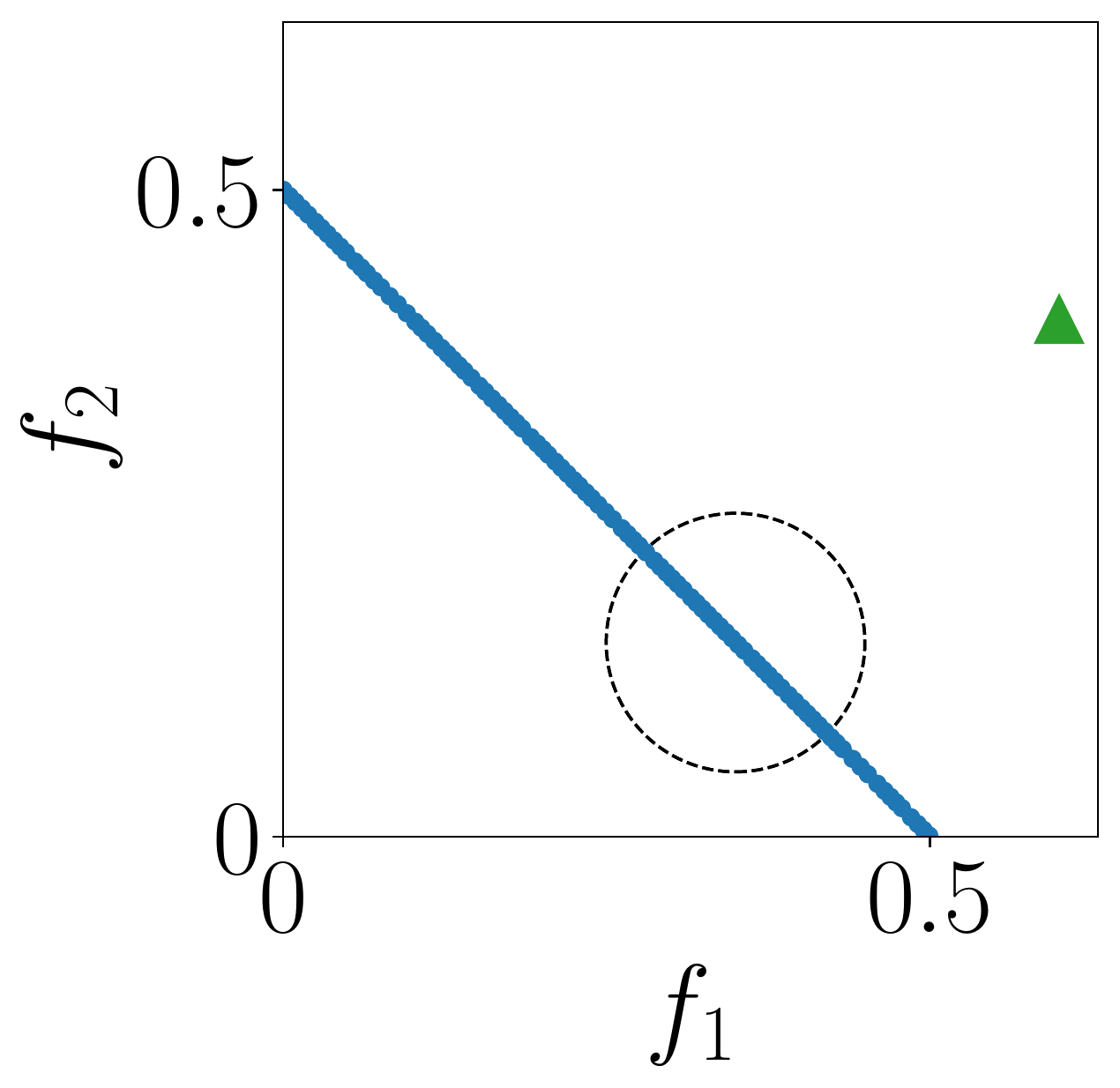}
}
\subfloat[r-NSGA-II (UA-PP)]{
  \includegraphics[width=0.15\textwidth]{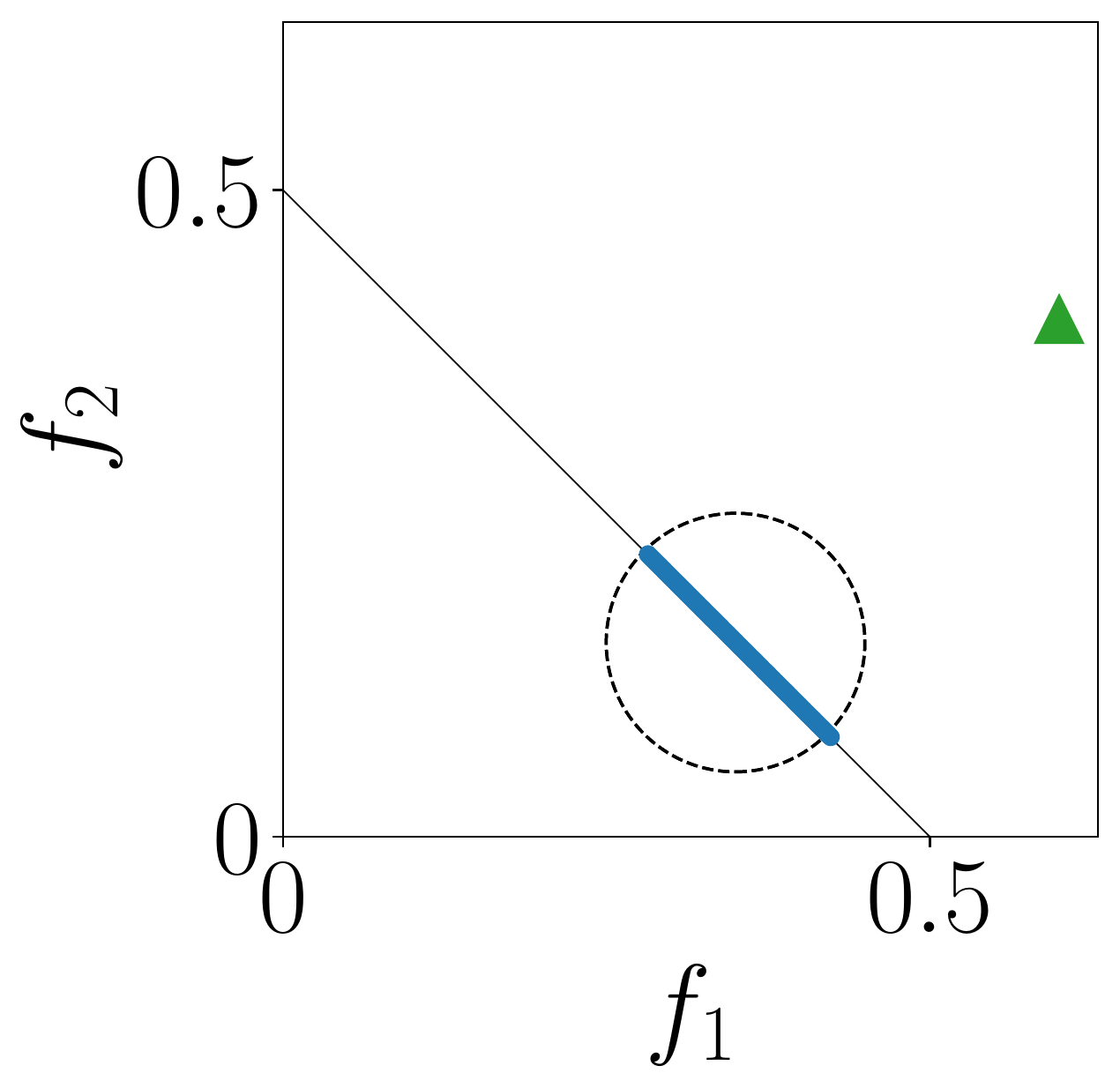}
}
\\
\subfloat[g-NSGA-II (POP)]{
     \includegraphics[width=0.15\textwidth]{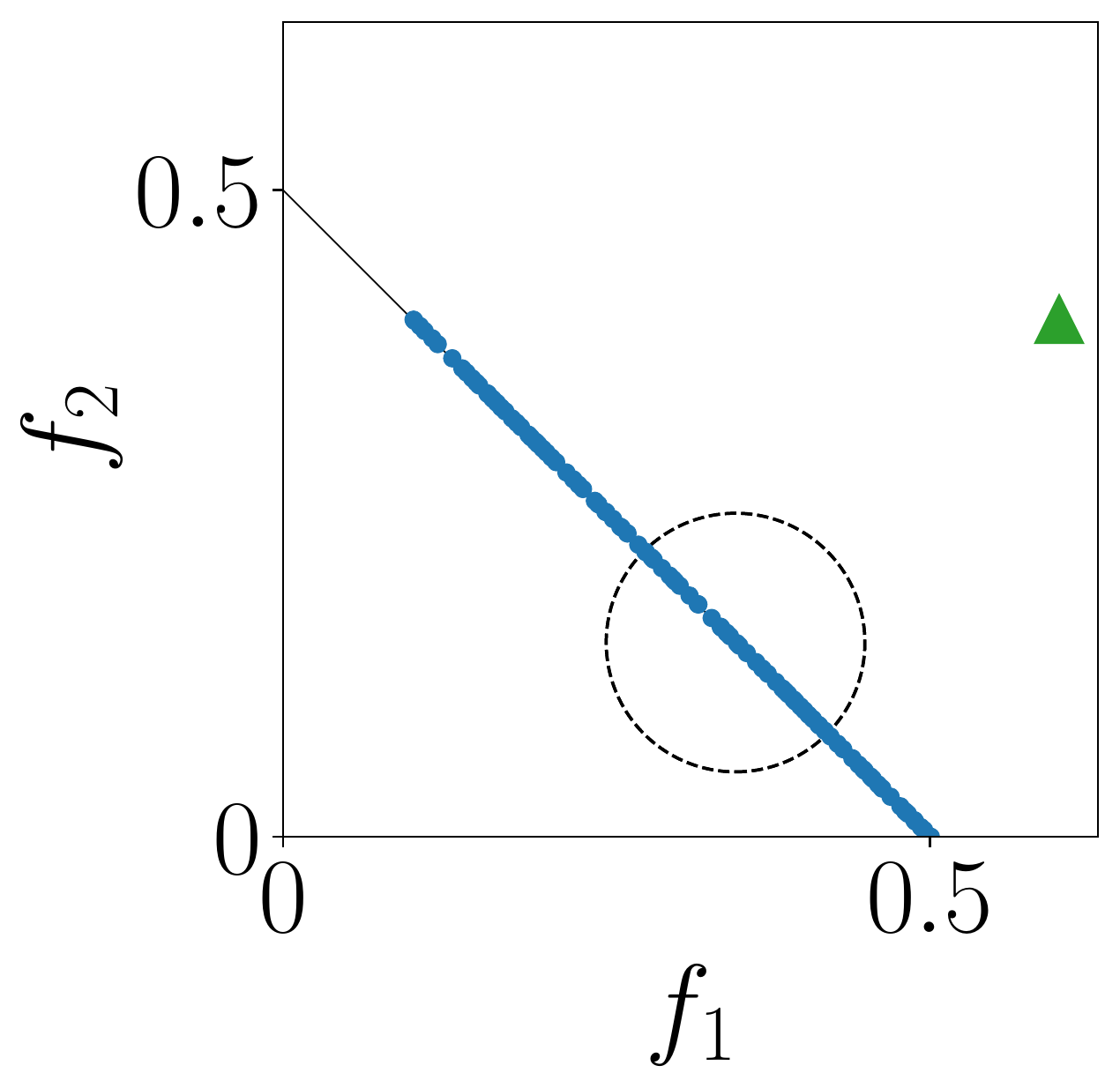}
   }
   \subfloat[g-NSGA-II (UA-IDDS)]{
  \includegraphics[width=0.15\textwidth]{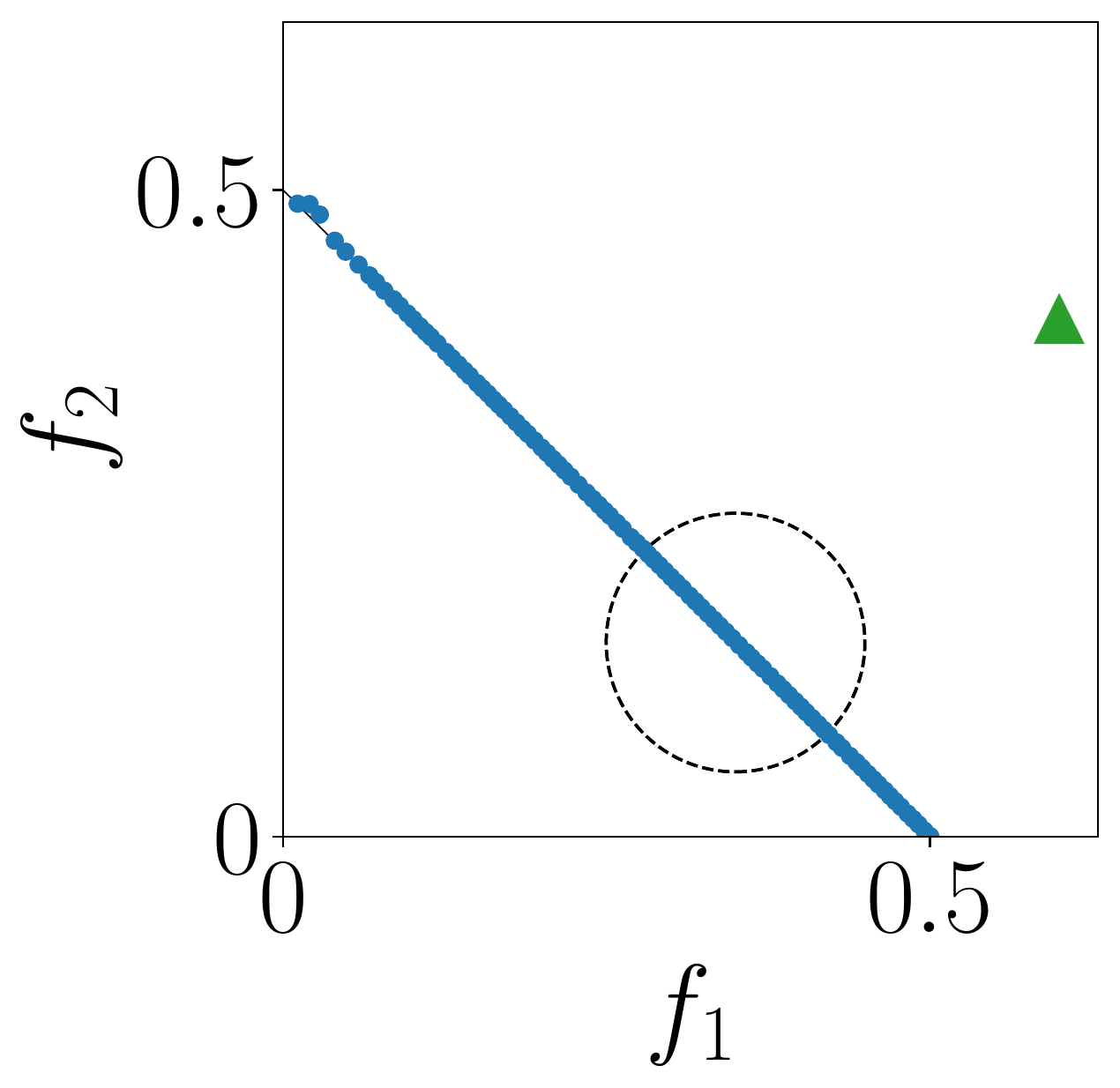}
}
\subfloat[g-NSGA-II (UA-PP)]{
  \includegraphics[width=0.15\textwidth]{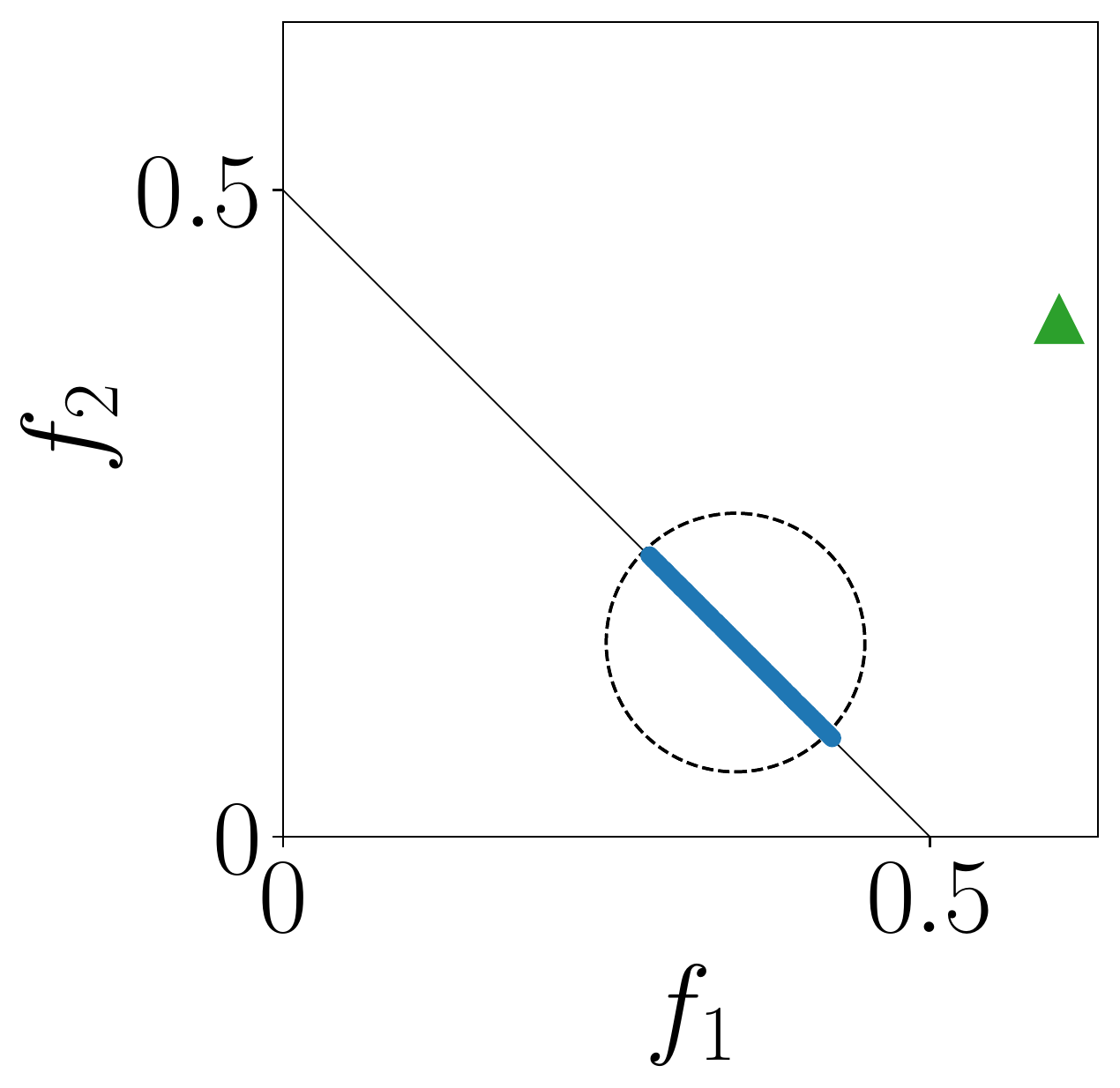}
}
\\
\subfloat[PBEA (POP)]{
     \includegraphics[width=0.15\textwidth]{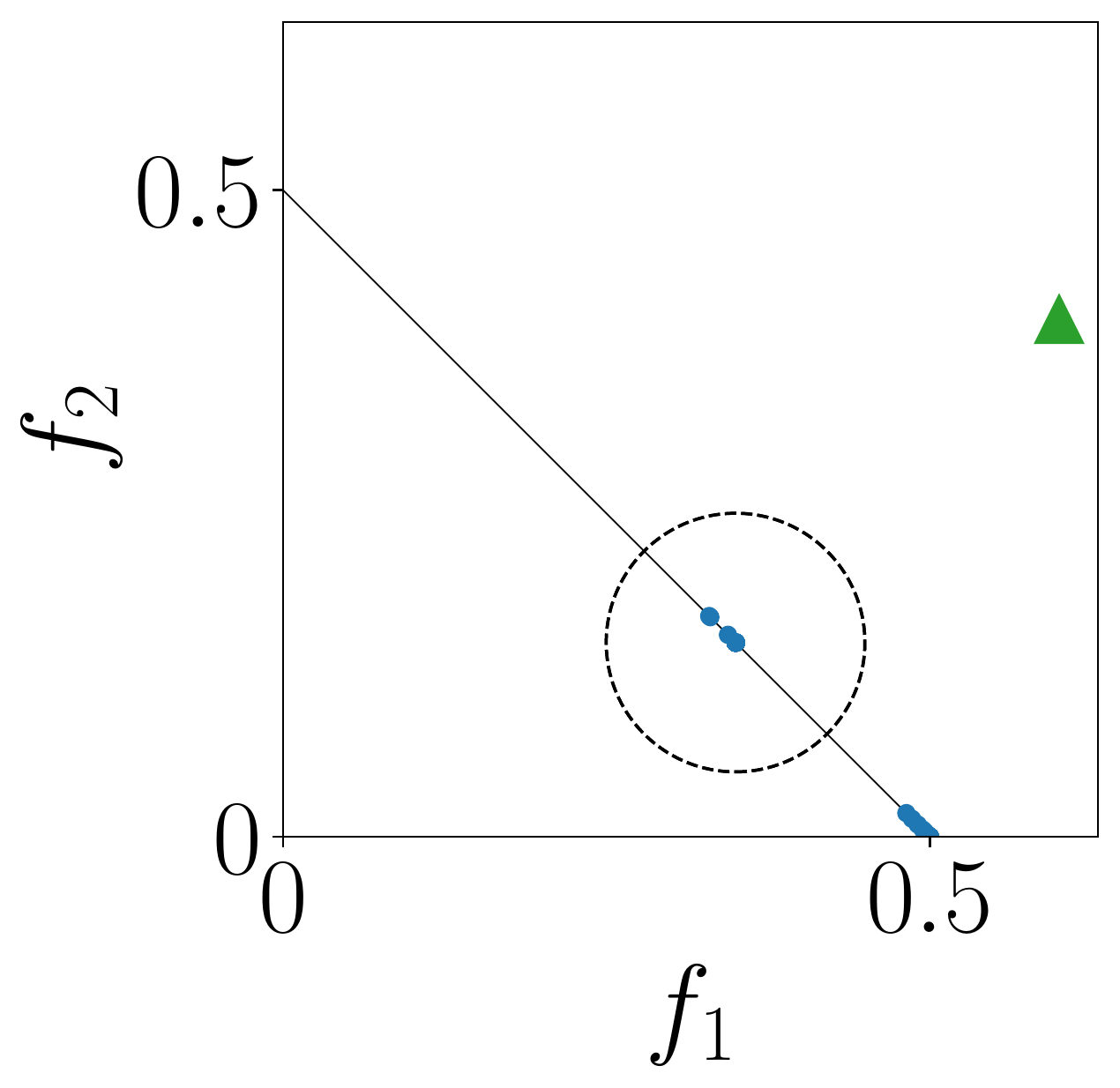}
   }
   \subfloat[PBEA (UA-IDDS)]{
  \includegraphics[width=0.15\textwidth]{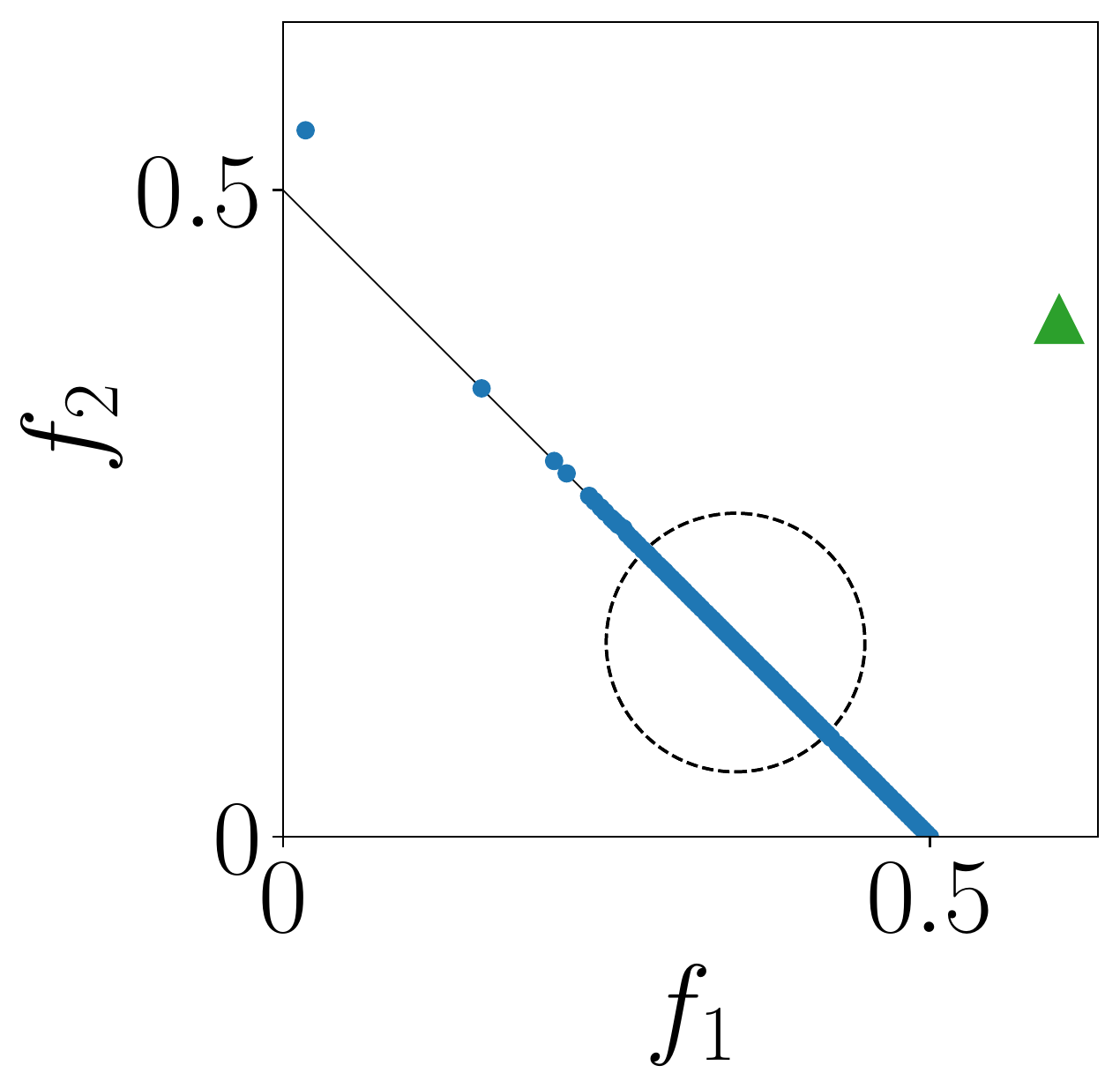}
}
\subfloat[PBEA (UA-PP)]{
  \includegraphics[width=0.15\textwidth]{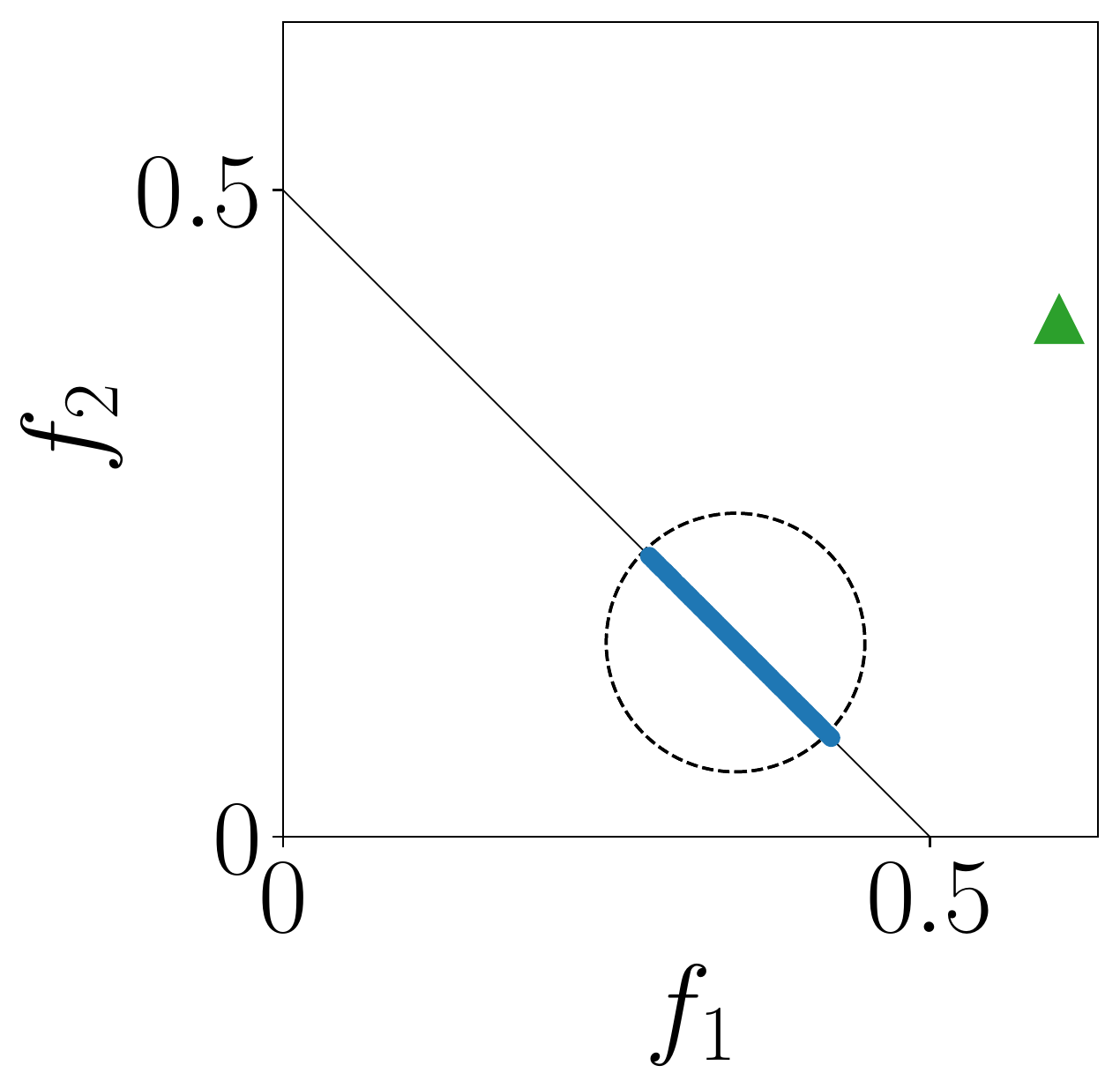}
}
\\
\subfloat[R-MEAD2 (POP]{
     \includegraphics[width=0.15\textwidth]{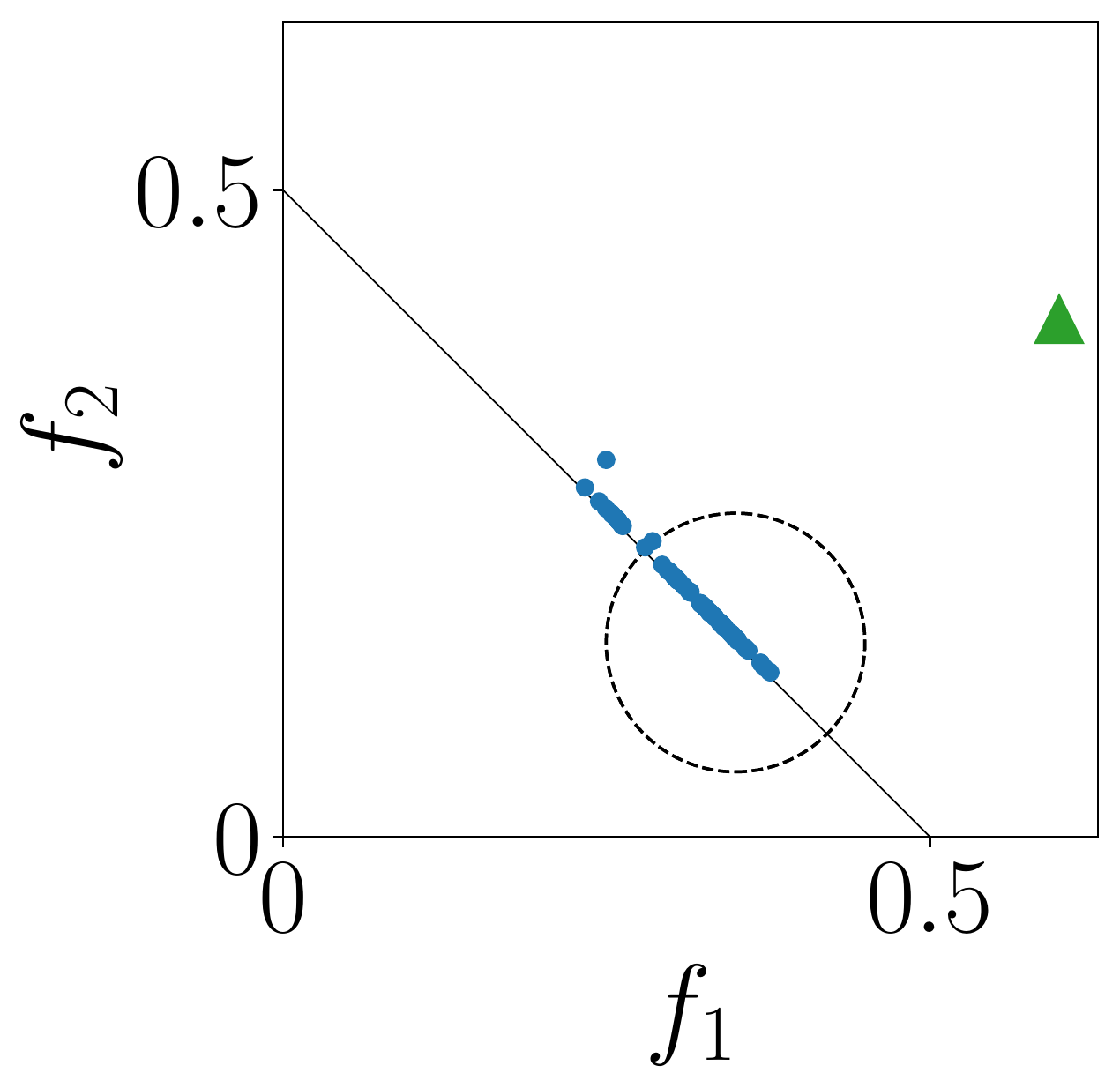}
   }
   \subfloat[R-MEAD2 (UA-IDDS)]{
  \includegraphics[width=0.15\textwidth]{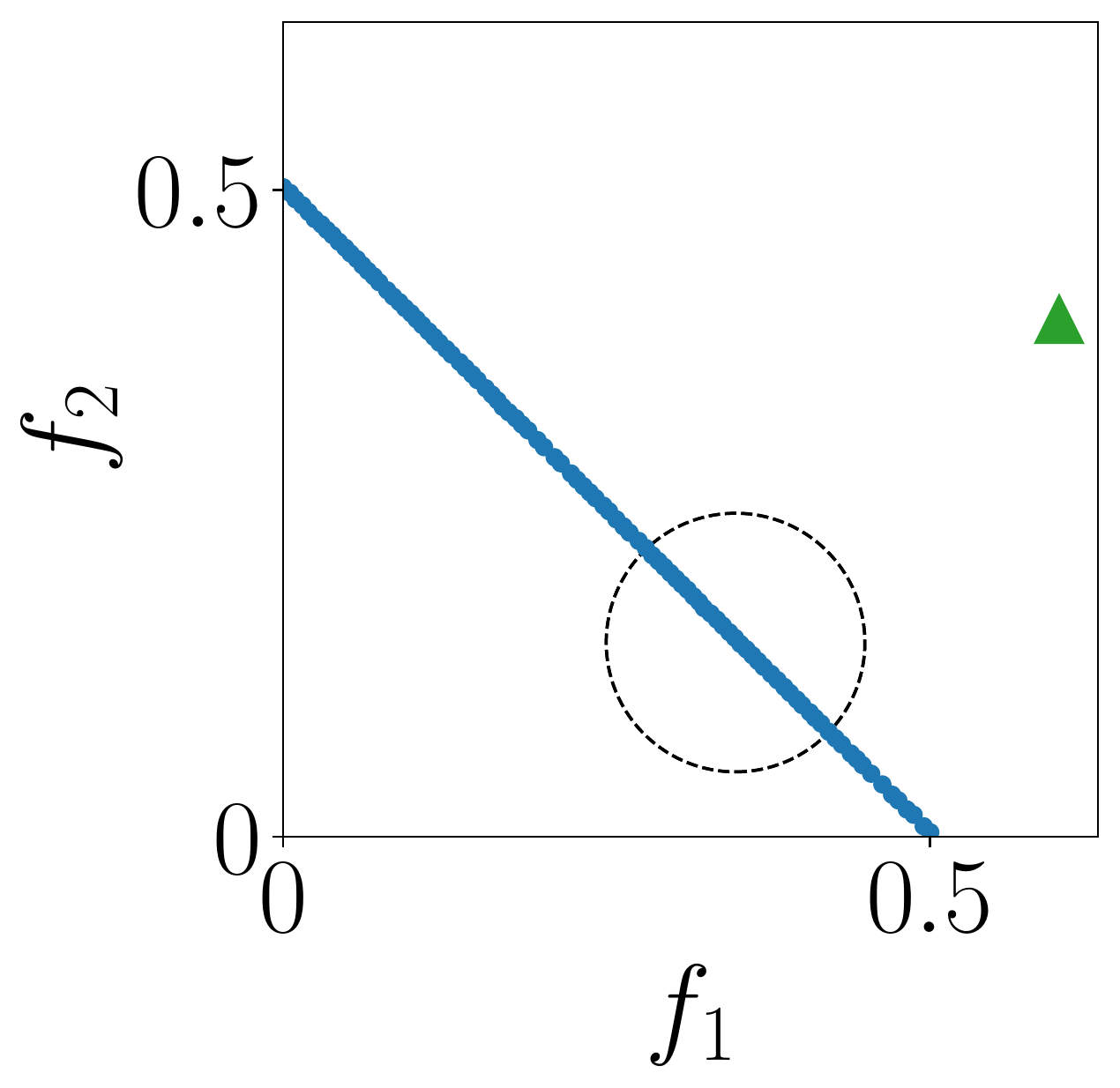}
}
\subfloat[R-MEAD2 (UA-PP)]{
  \includegraphics[width=0.15\textwidth]{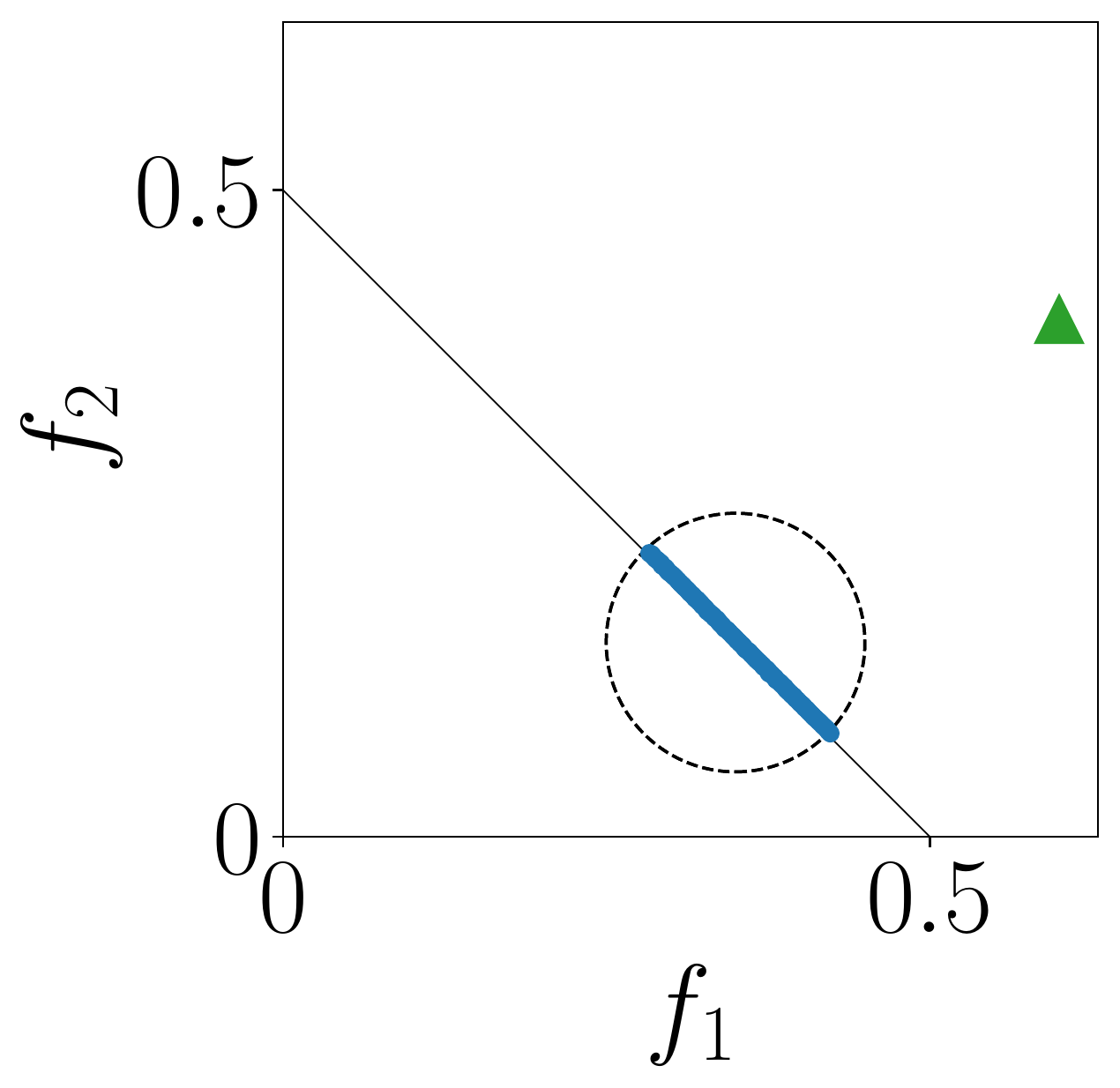}
}
\\
\subfloat[NUMS (POP)]{
     \includegraphics[width=0.15\textwidth]{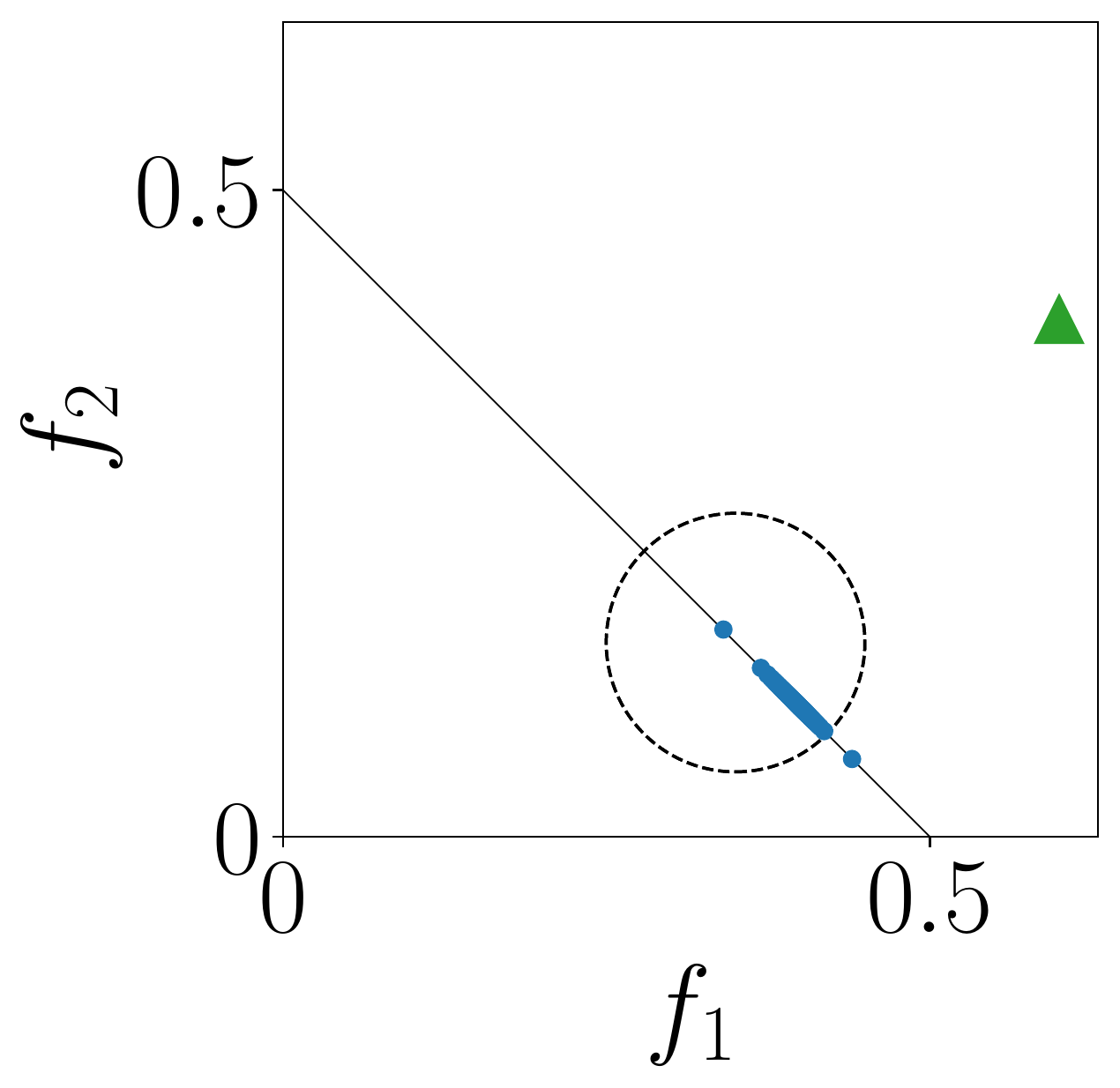}
   }
   \subfloat[NUMS (UA-IDDS)]{
  \includegraphics[width=0.15\textwidth]{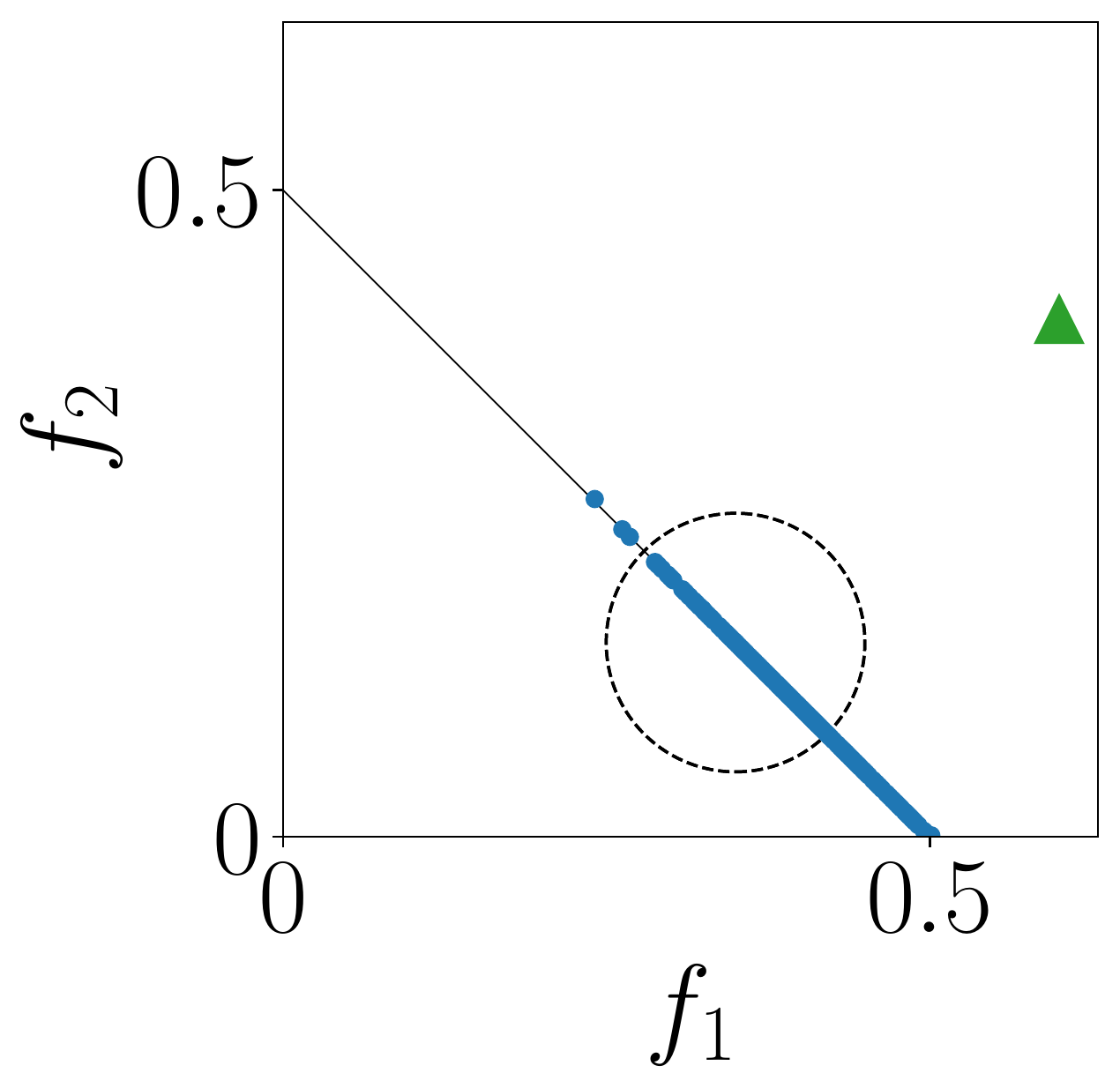}
}
\subfloat[NUMS (UA-PP)]{
  \includegraphics[width=0.15\textwidth]{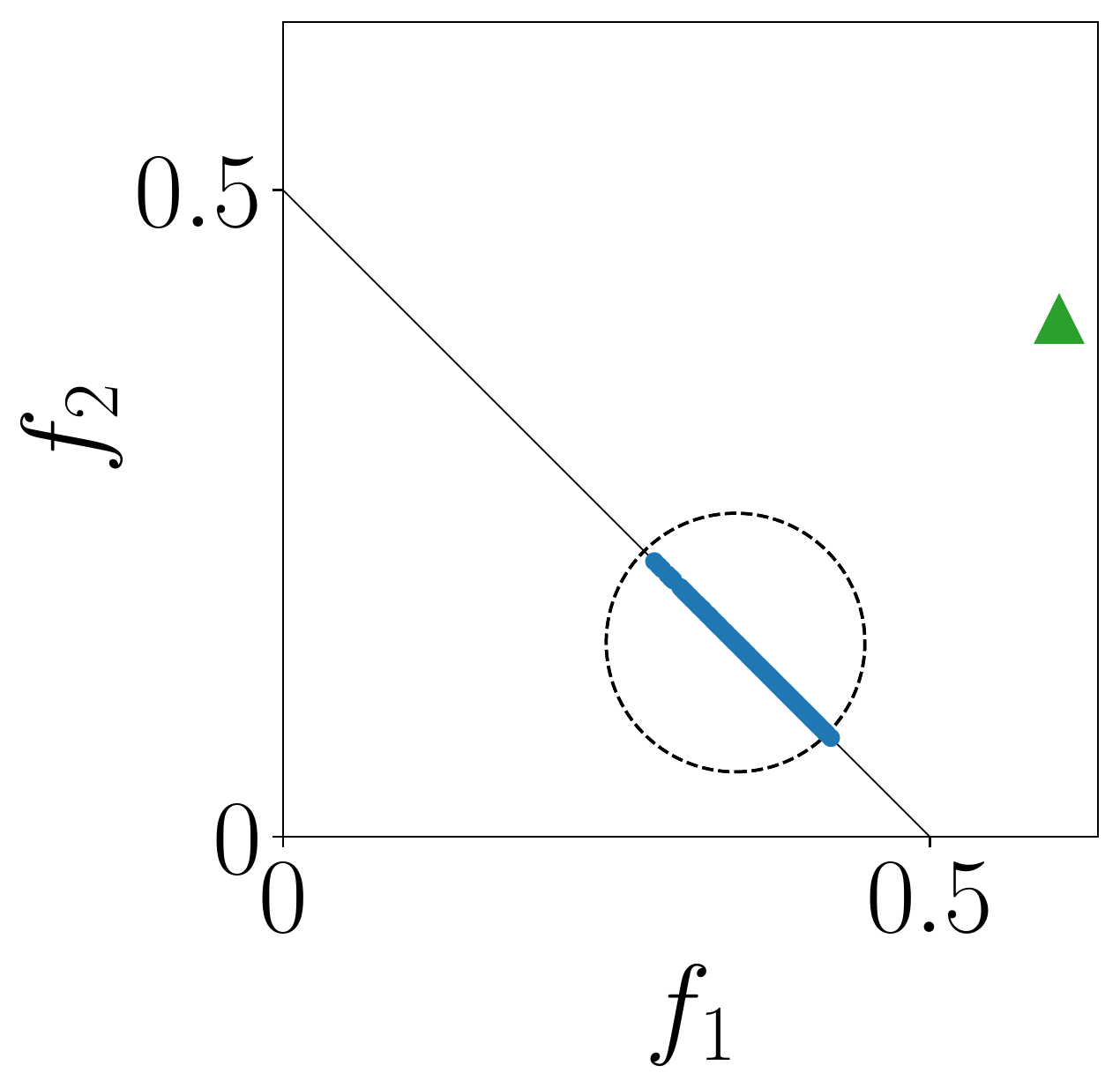}
}
\caption{Distributions of the objective vectors of the solutions in the three solution sets on DTLZ1 with $m=2$, where \tabgreen{$\blacktriangle$} is the reference point $\mathbf{z}$. The dotted circle represents the true ROI. ``NUMS'' stands for MOEA/D-NUMS.}
   \label{fig:100points_dtlz1}
\end{figure*}

\begin{figure*}[t]
   \centering
   \subfloat[R-NSGA-II (POP)]{
     \includegraphics[width=0.15\textwidth]{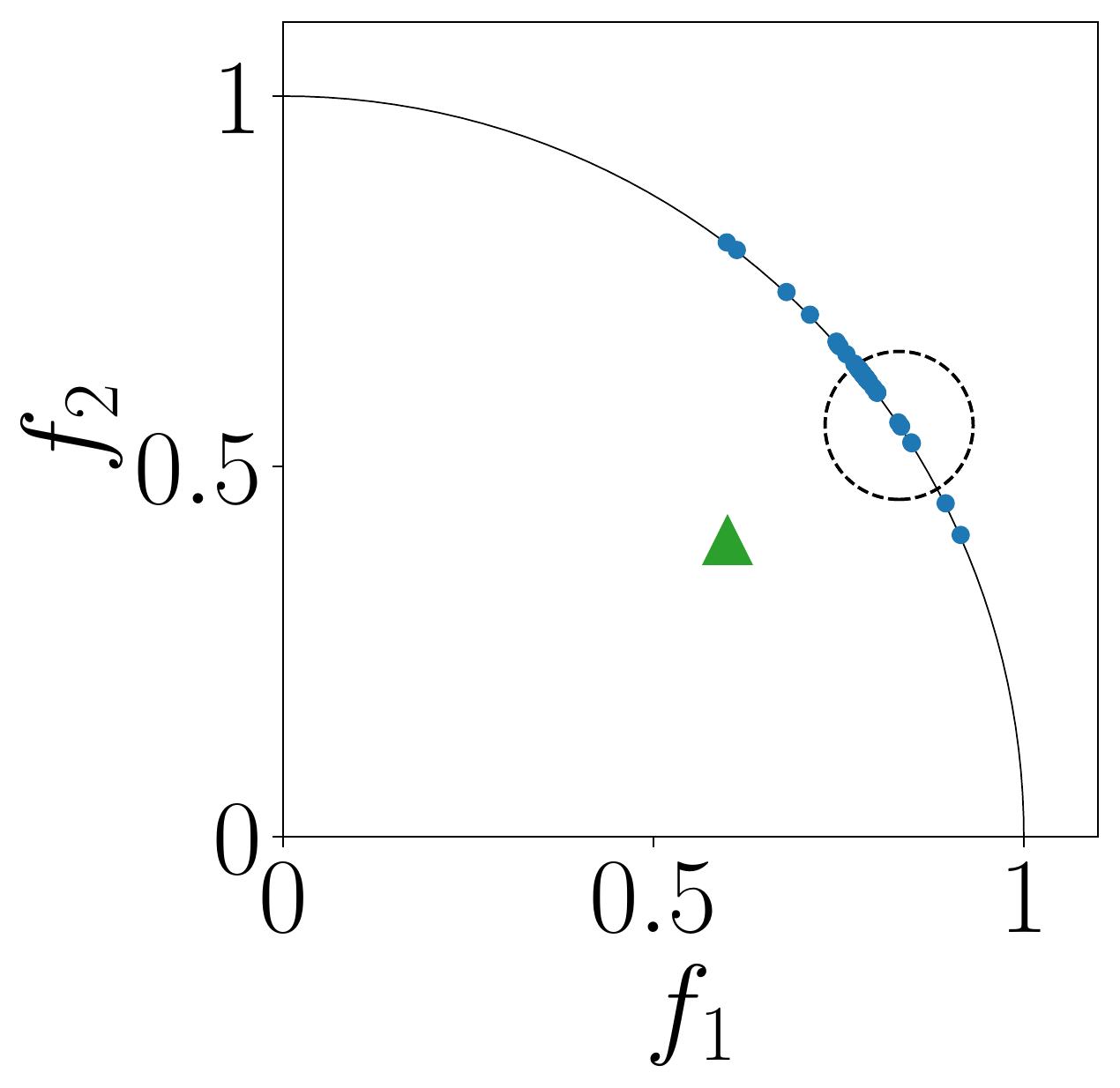}
   }
   \subfloat[R-NSGA-II (UA-IDDS)]{
  \includegraphics[width=0.15\textwidth]{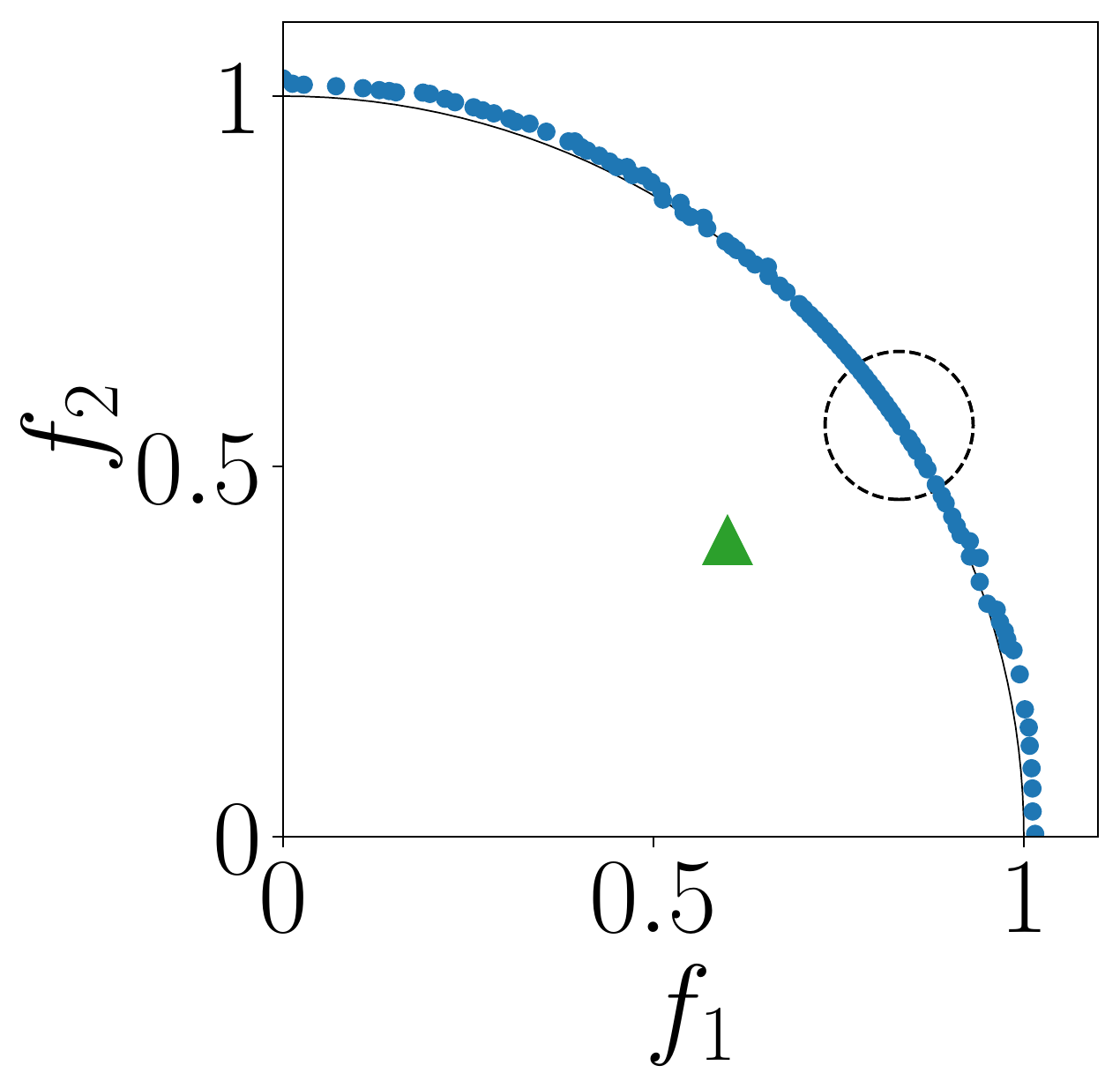}
}
\subfloat[R-NSGA-II (UA-PP)]{
  \includegraphics[width=0.15\textwidth]{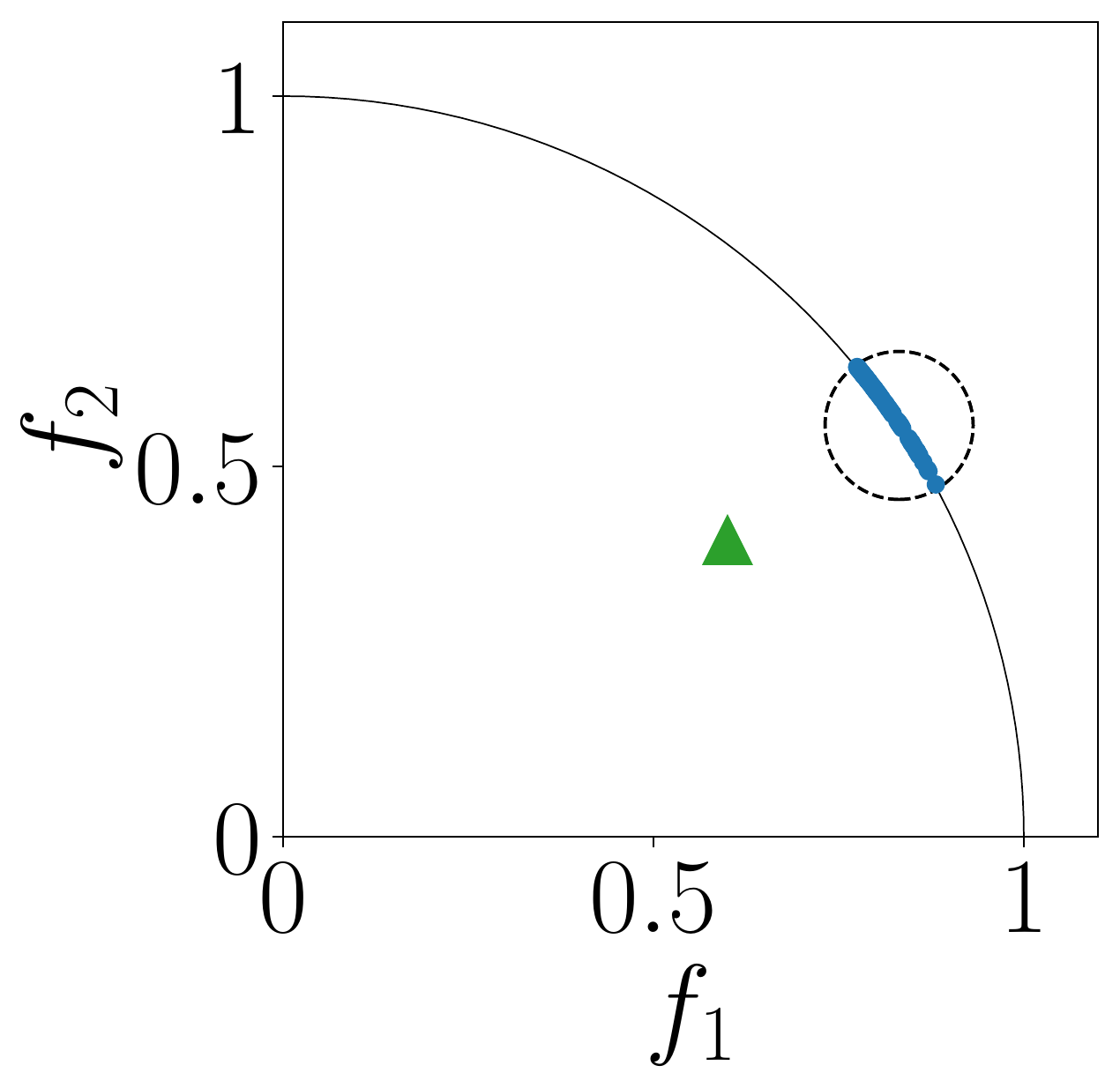}
}
\\
\subfloat[r-NSGA-II (POP)]{
     \includegraphics[width=0.15\textwidth]{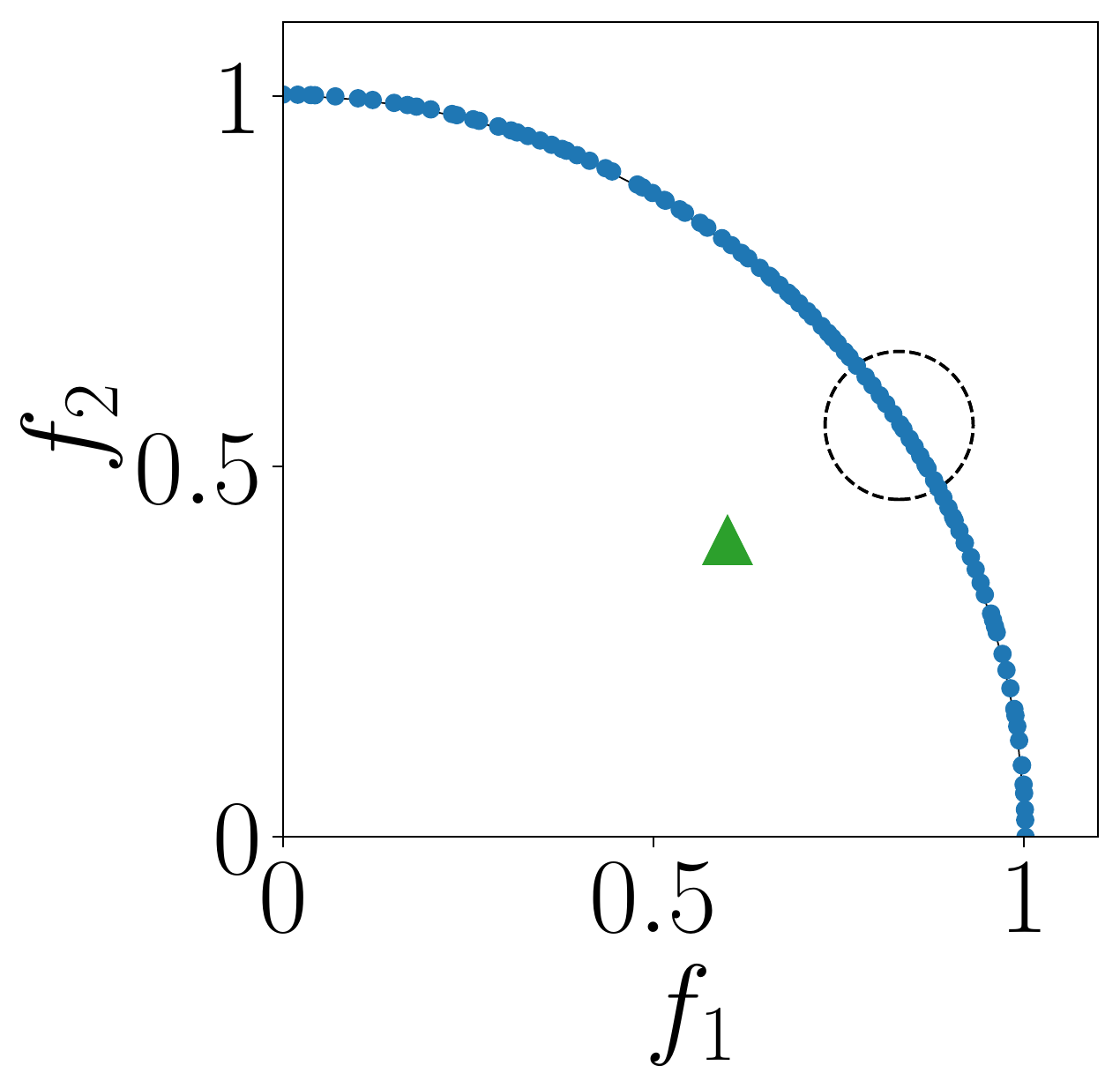}
   }
   \subfloat[r-NSGA-II (UA-IDDS)]{
  \includegraphics[width=0.15\textwidth]{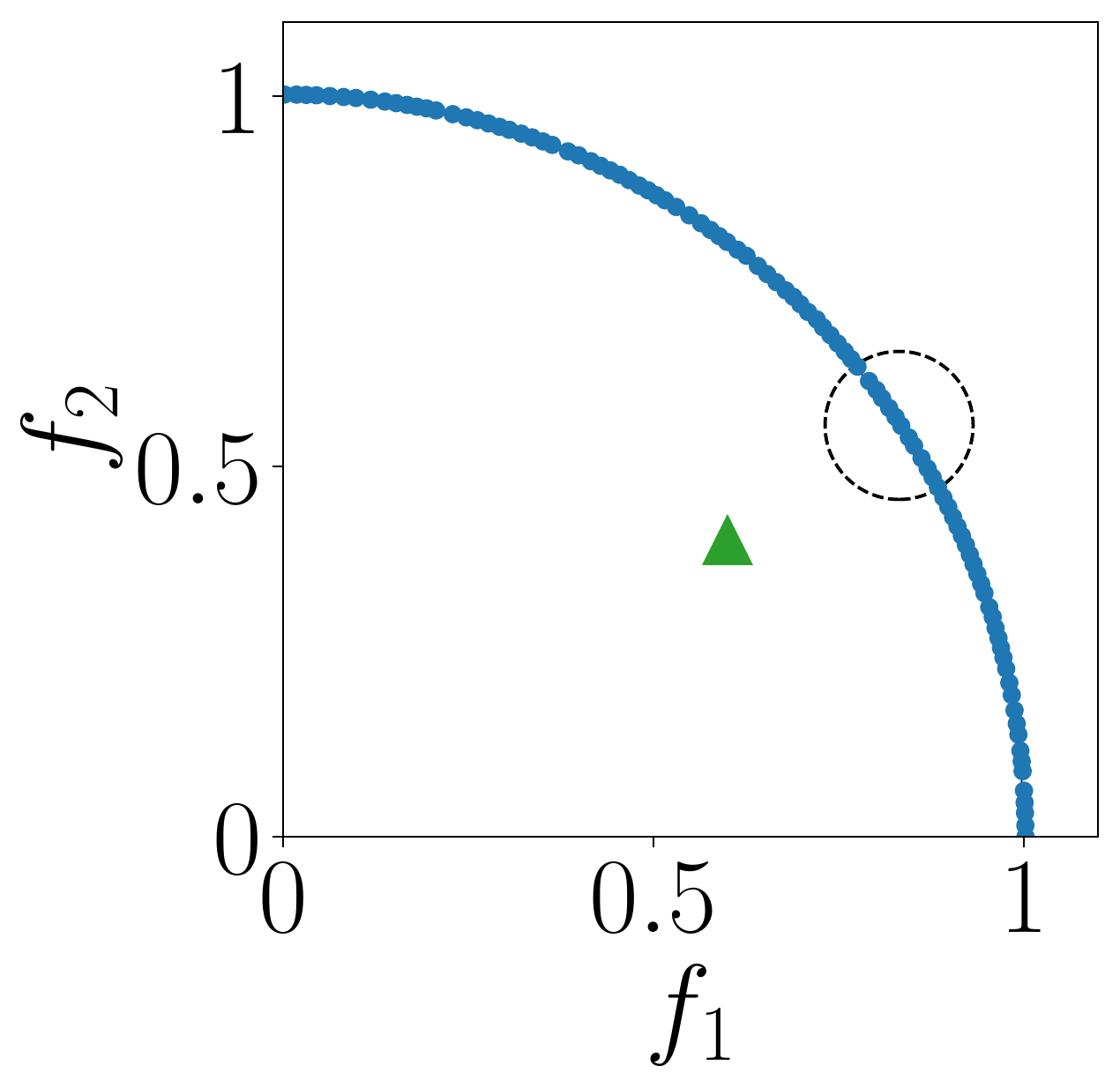}
}
\subfloat[r-NSGA-II (UA-PP)]{
  \includegraphics[width=0.15\textwidth]{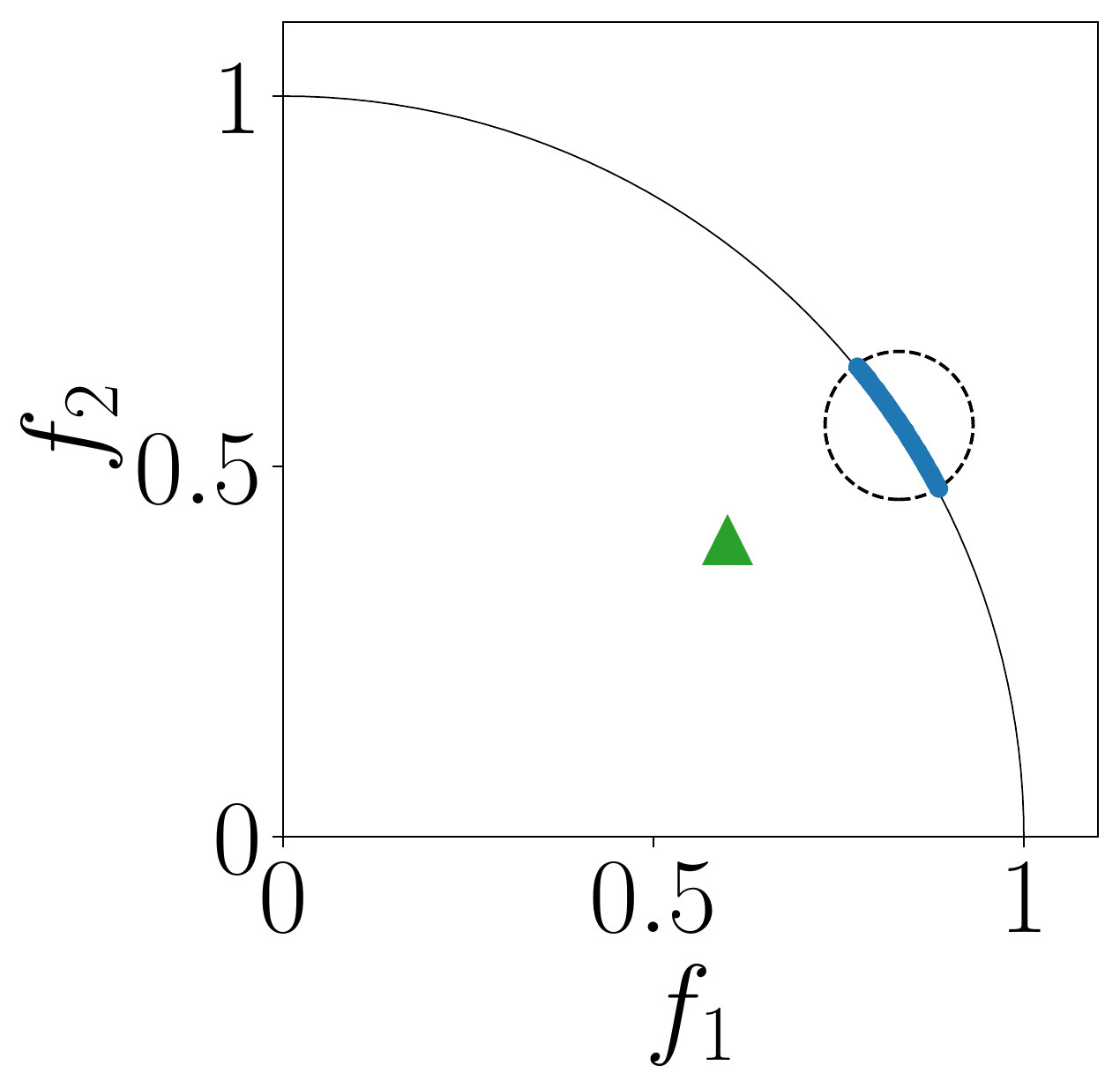}
}
\\
\subfloat[g-NSGA-II (POP)]{
     \includegraphics[width=0.15\textwidth]{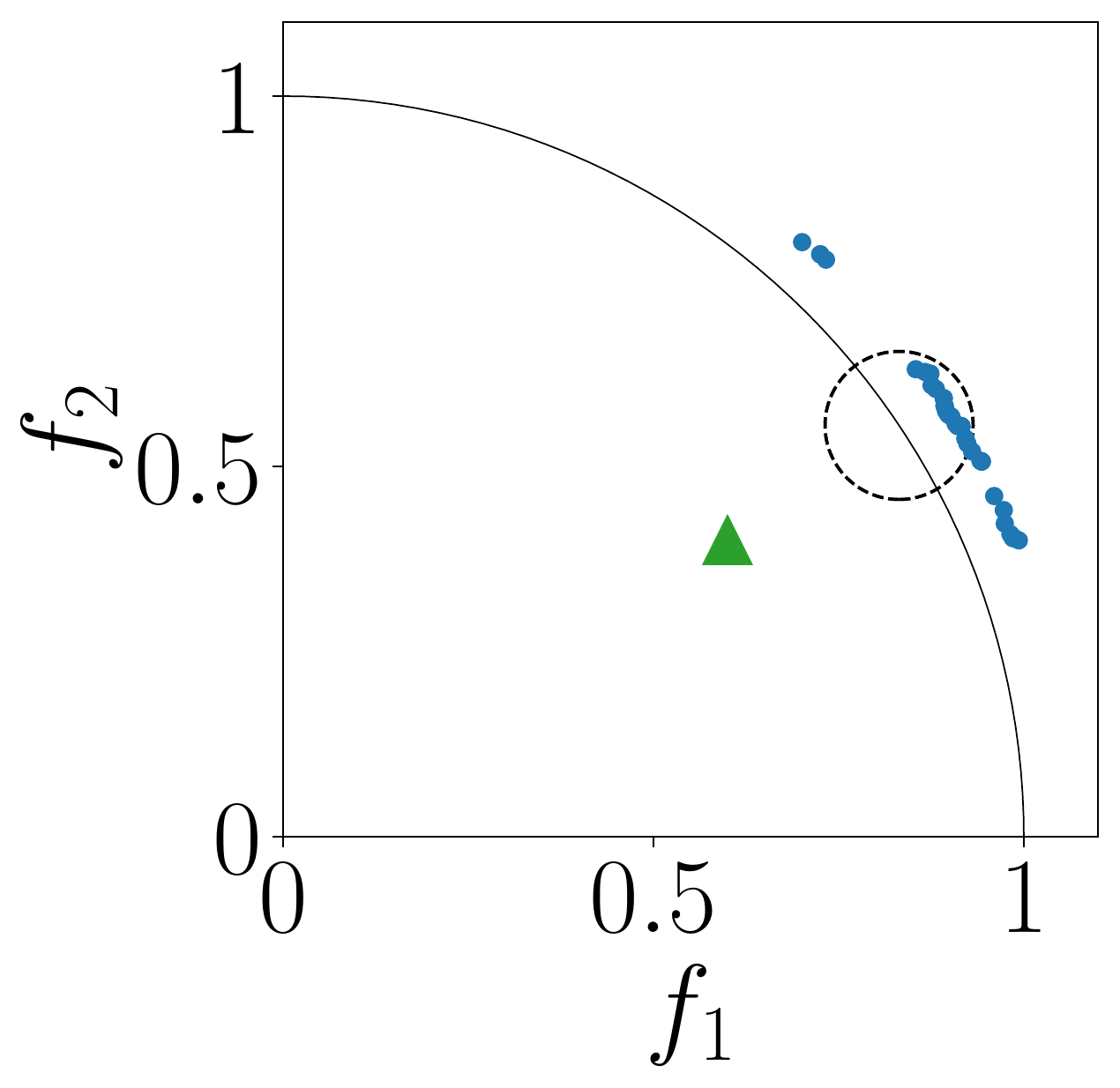}
   }
   \subfloat[g-NSGA-II (UA-IDDS)]{
  \includegraphics[width=0.15\textwidth]{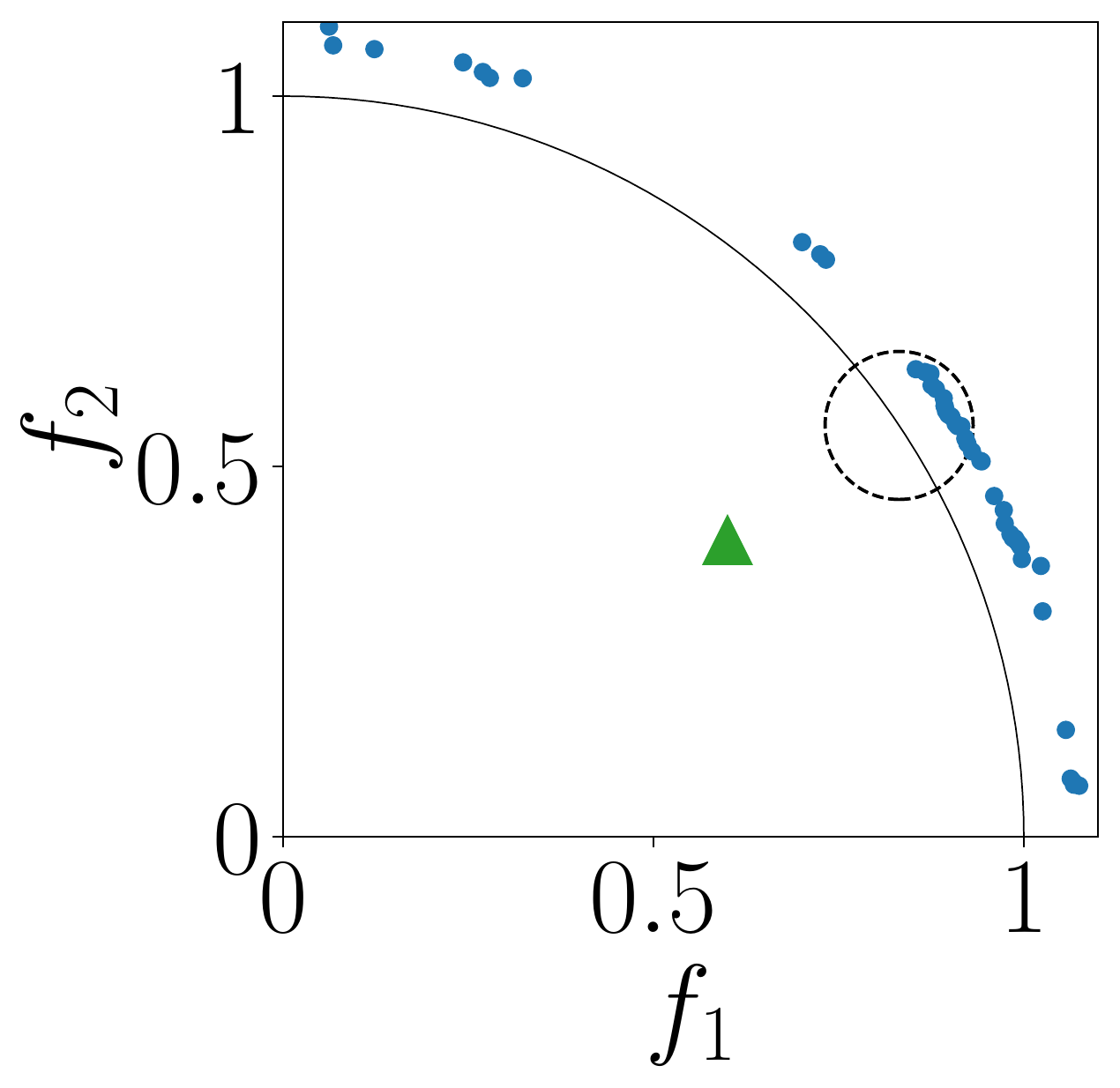}
}
\subfloat[g-NSGA-II (UA-PP)]{
  \includegraphics[width=0.15\textwidth]{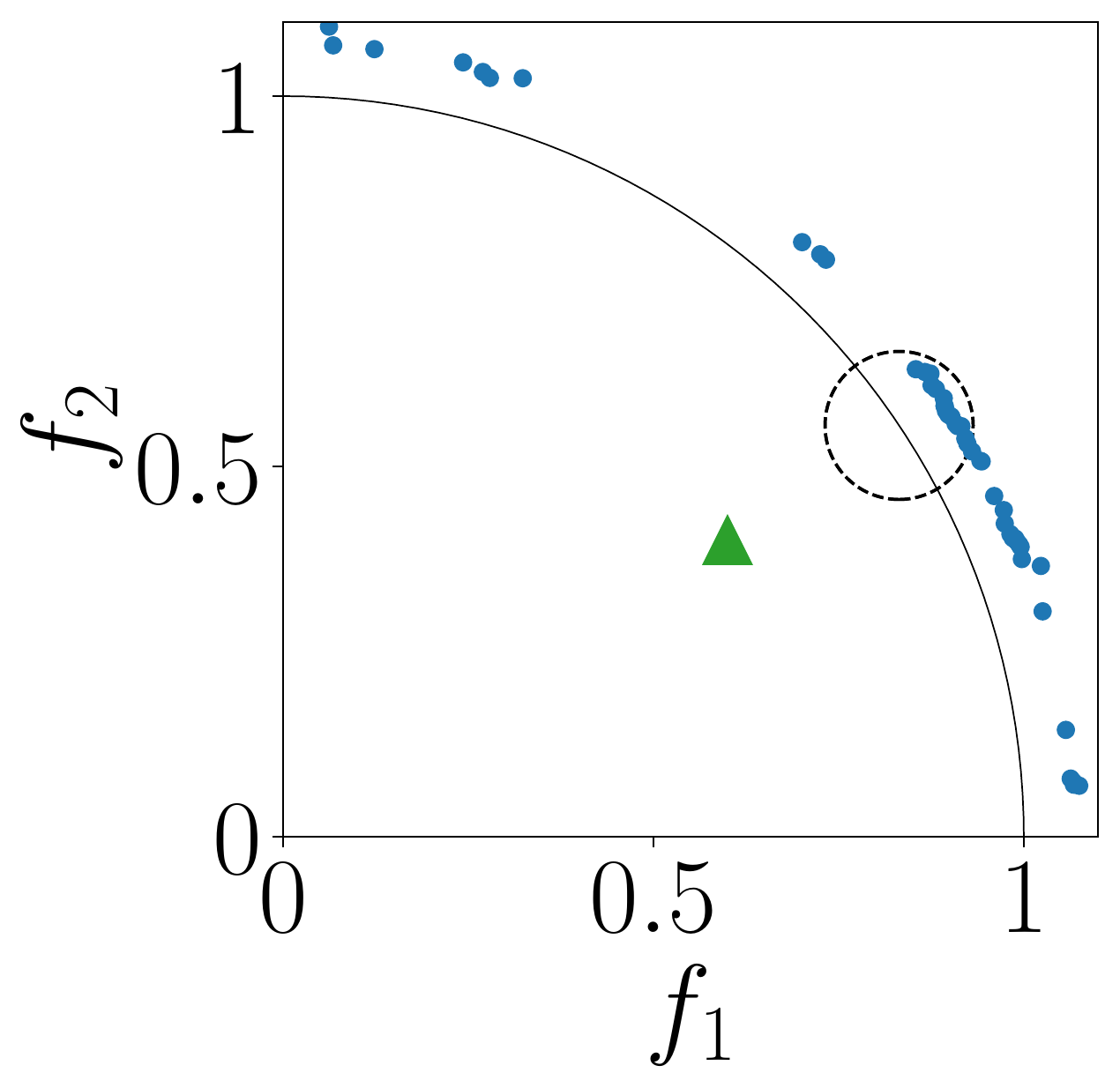}
}
\\
\subfloat[PBEA (POP)]{
     \includegraphics[width=0.15\textwidth]{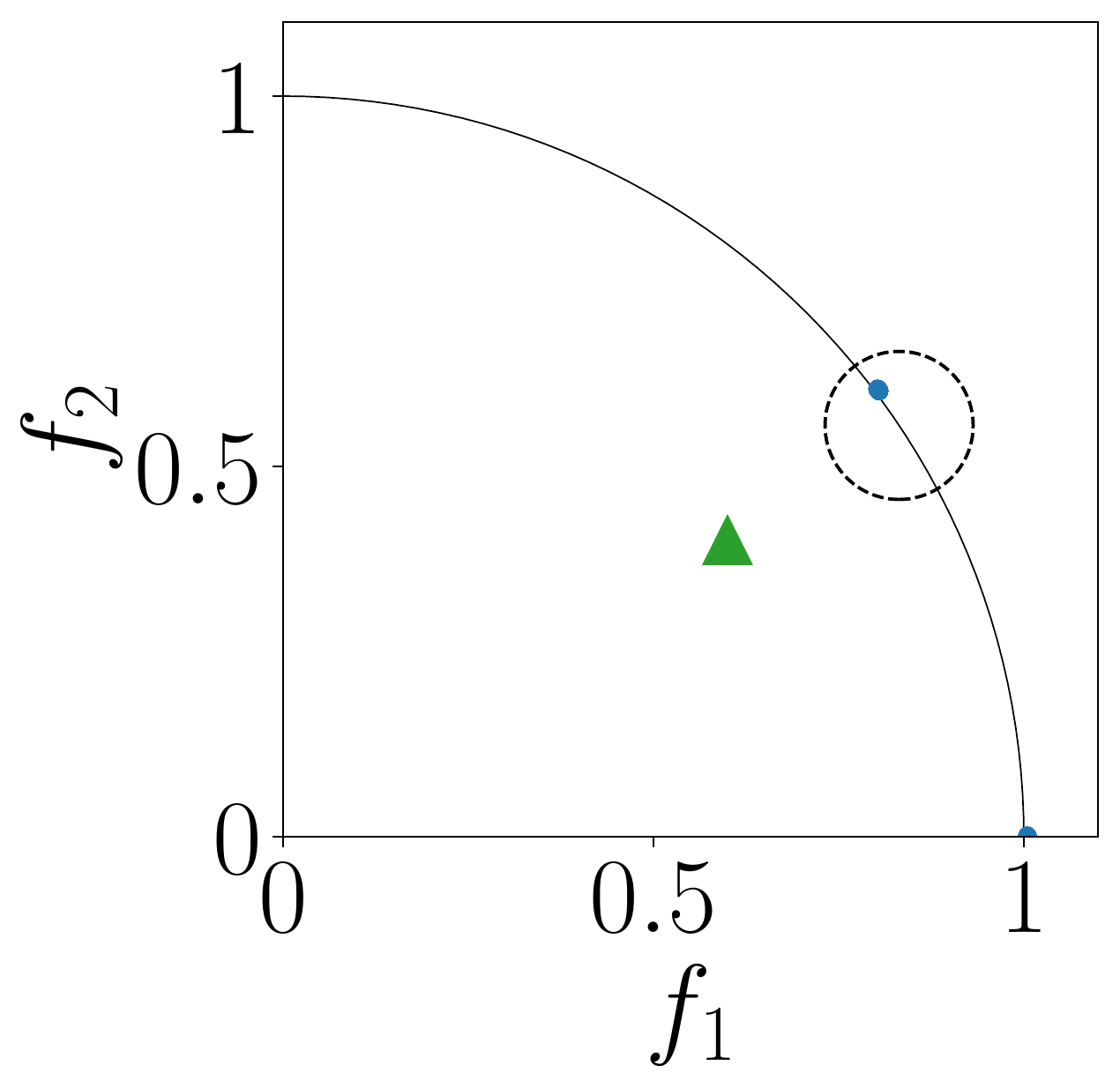}
   }
   \subfloat[PBEA (UA-IDDS)]{
  \includegraphics[width=0.15\textwidth]{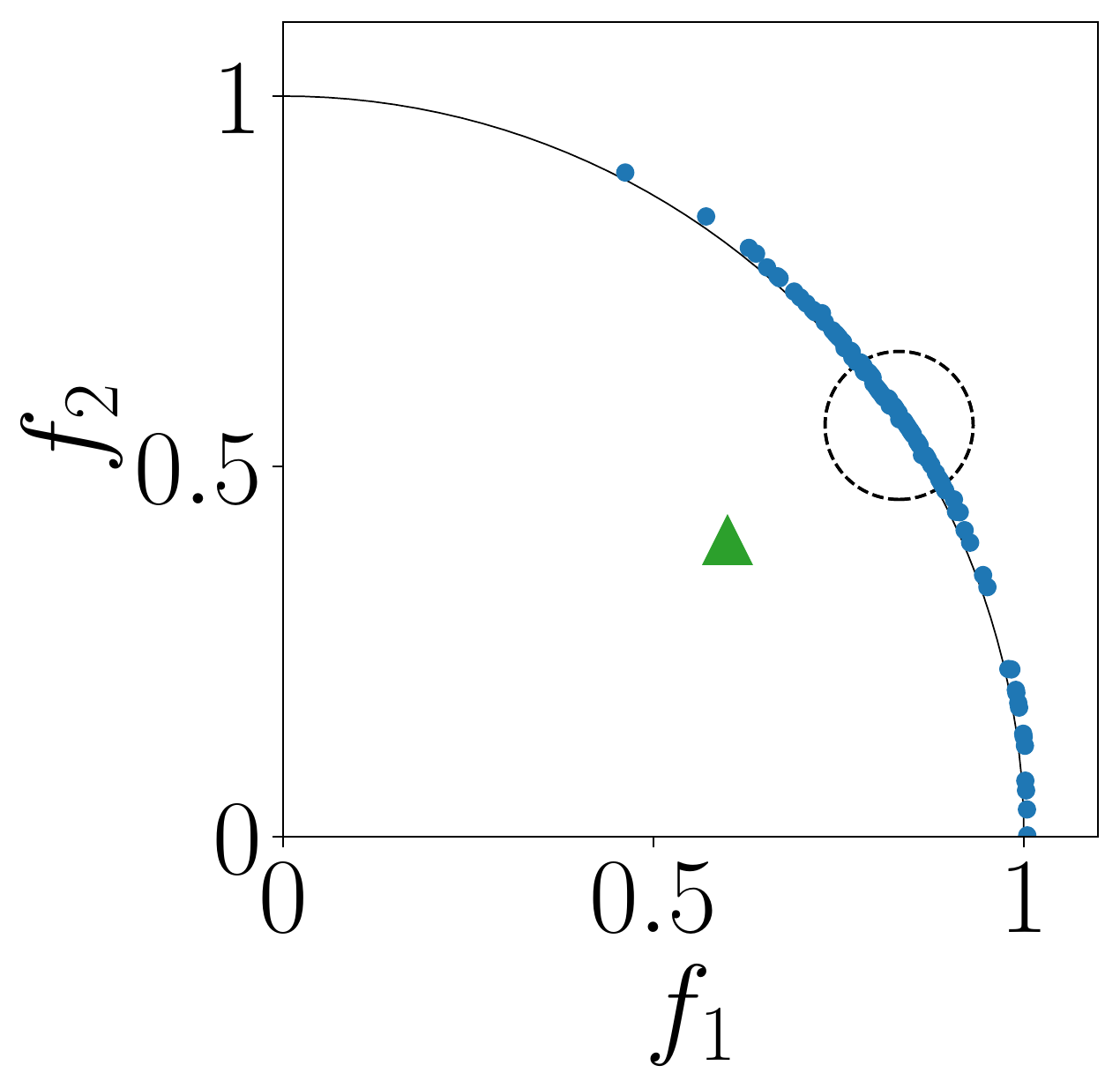}
}
\subfloat[PBEA (UA-PP)]{
  \includegraphics[width=0.15\textwidth]{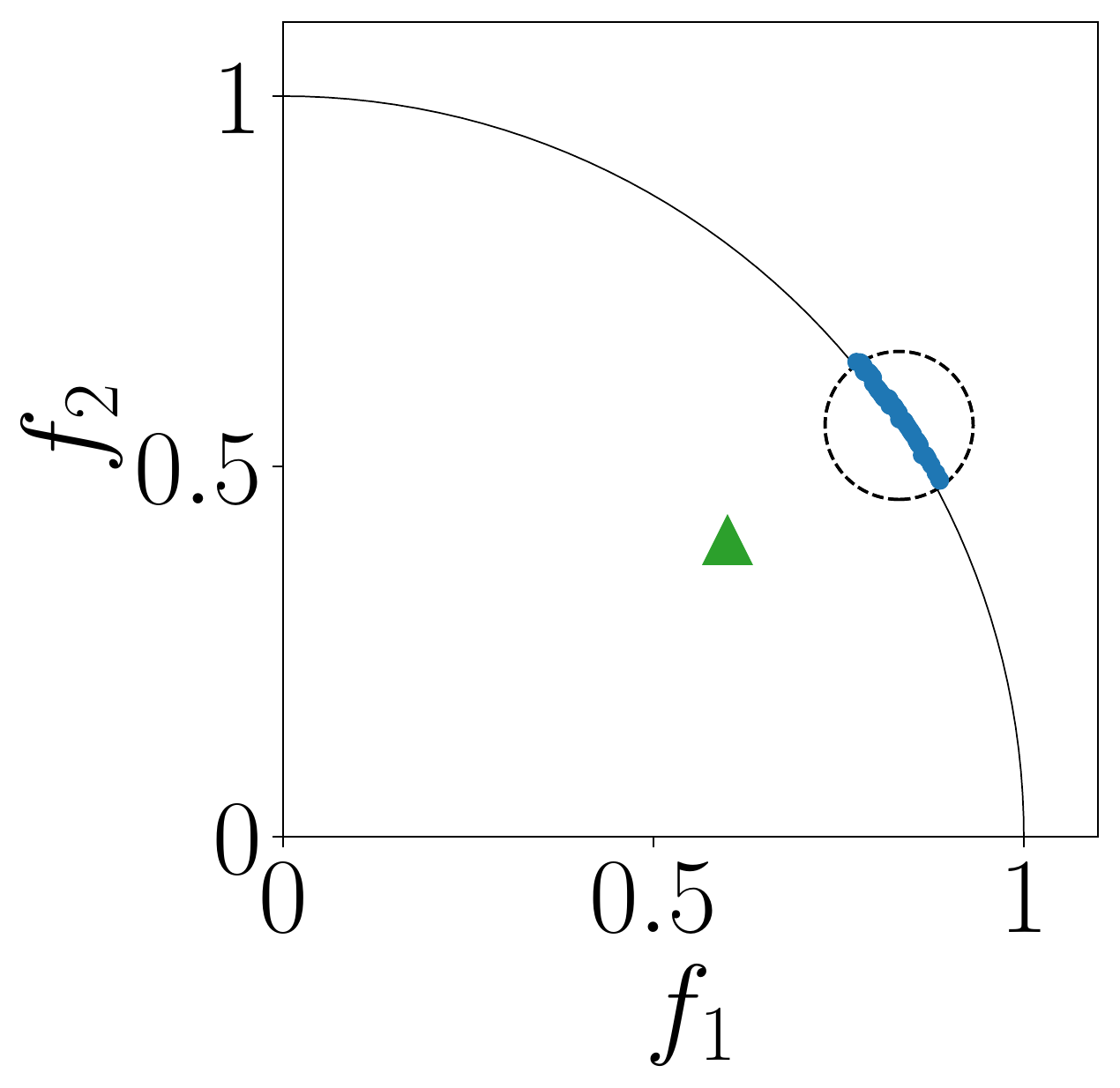}
}
\\
\subfloat[R-MEAD2 (POP]{
     \includegraphics[width=0.15\textwidth]{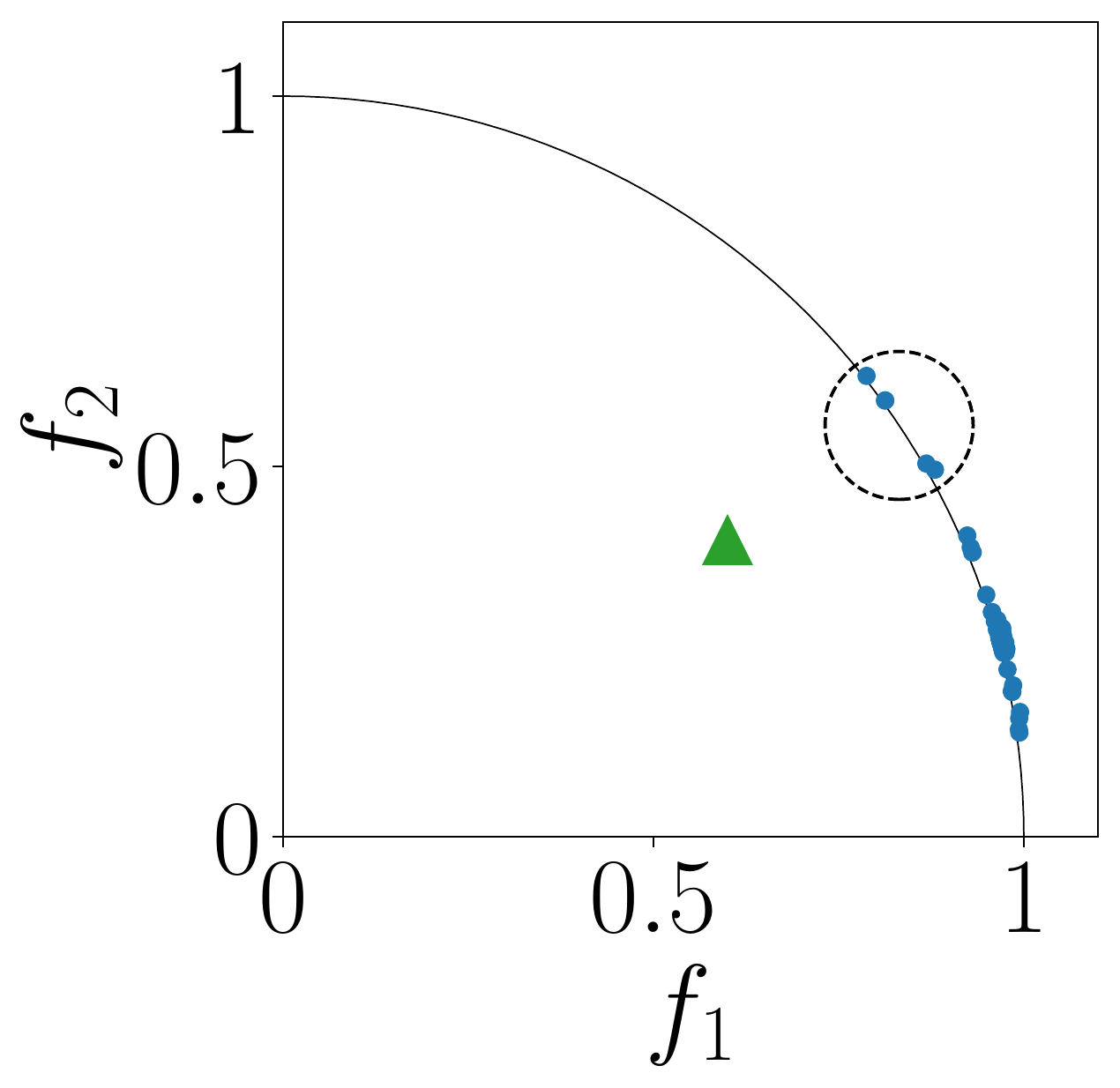}
   }
   \subfloat[R-MEAD2 (UA-IDDS)]{
  \includegraphics[width=0.15\textwidth]{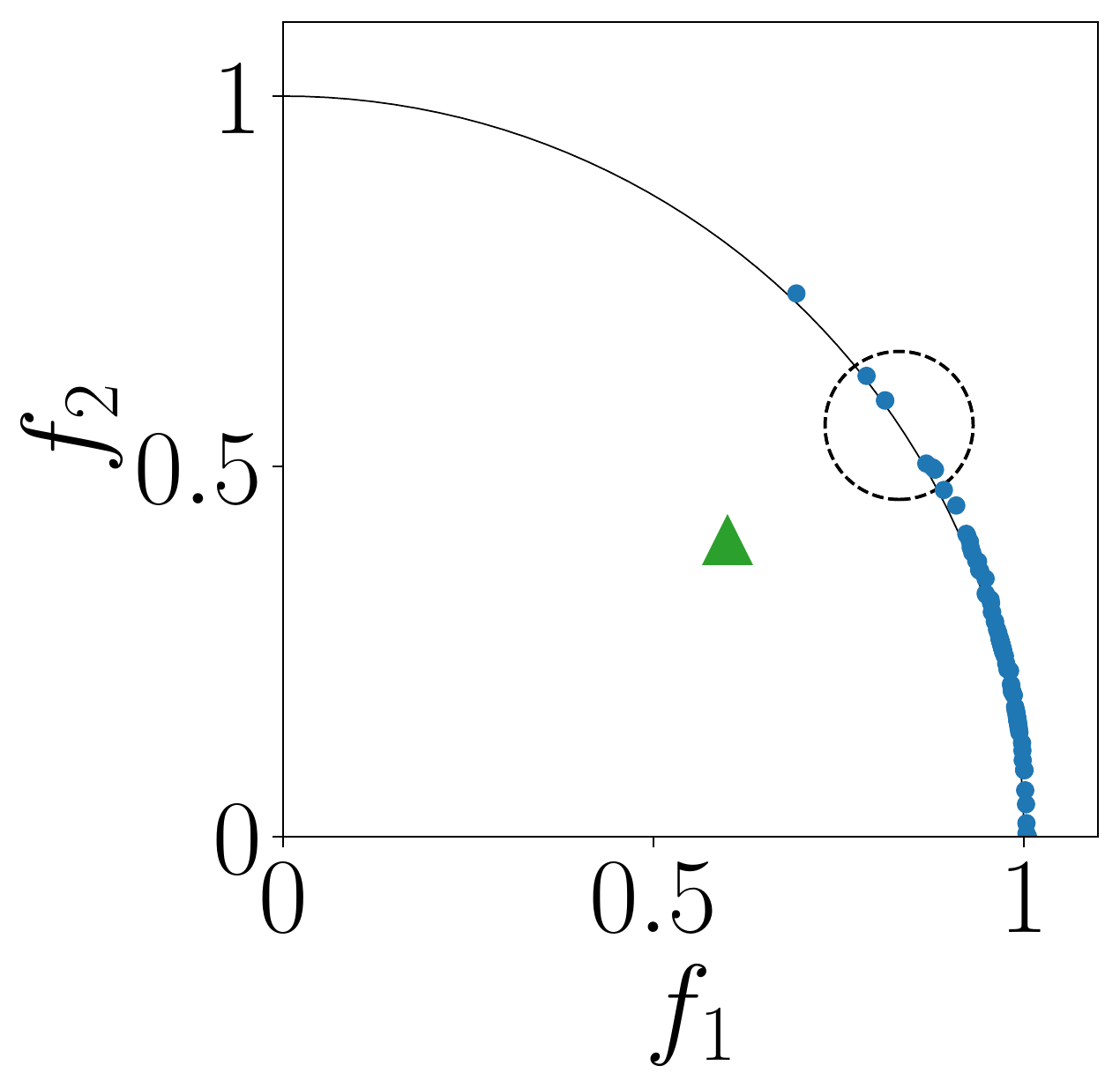}
}
\subfloat[R-MEAD2 (UA-PP)]{
  \includegraphics[width=0.15\textwidth]{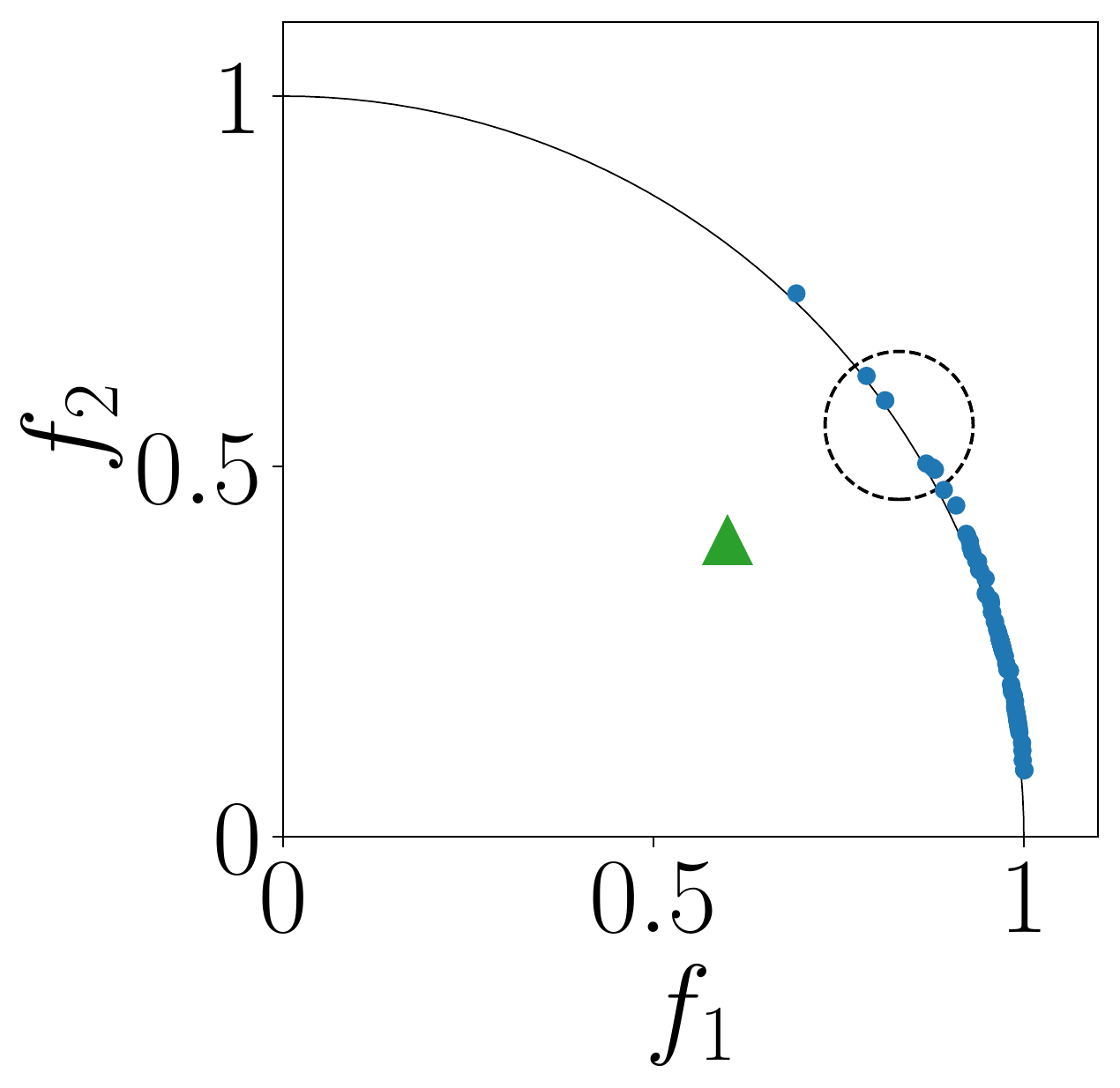}
}
\\
\subfloat[NUMS (POP)]{
     \includegraphics[width=0.15\textwidth]{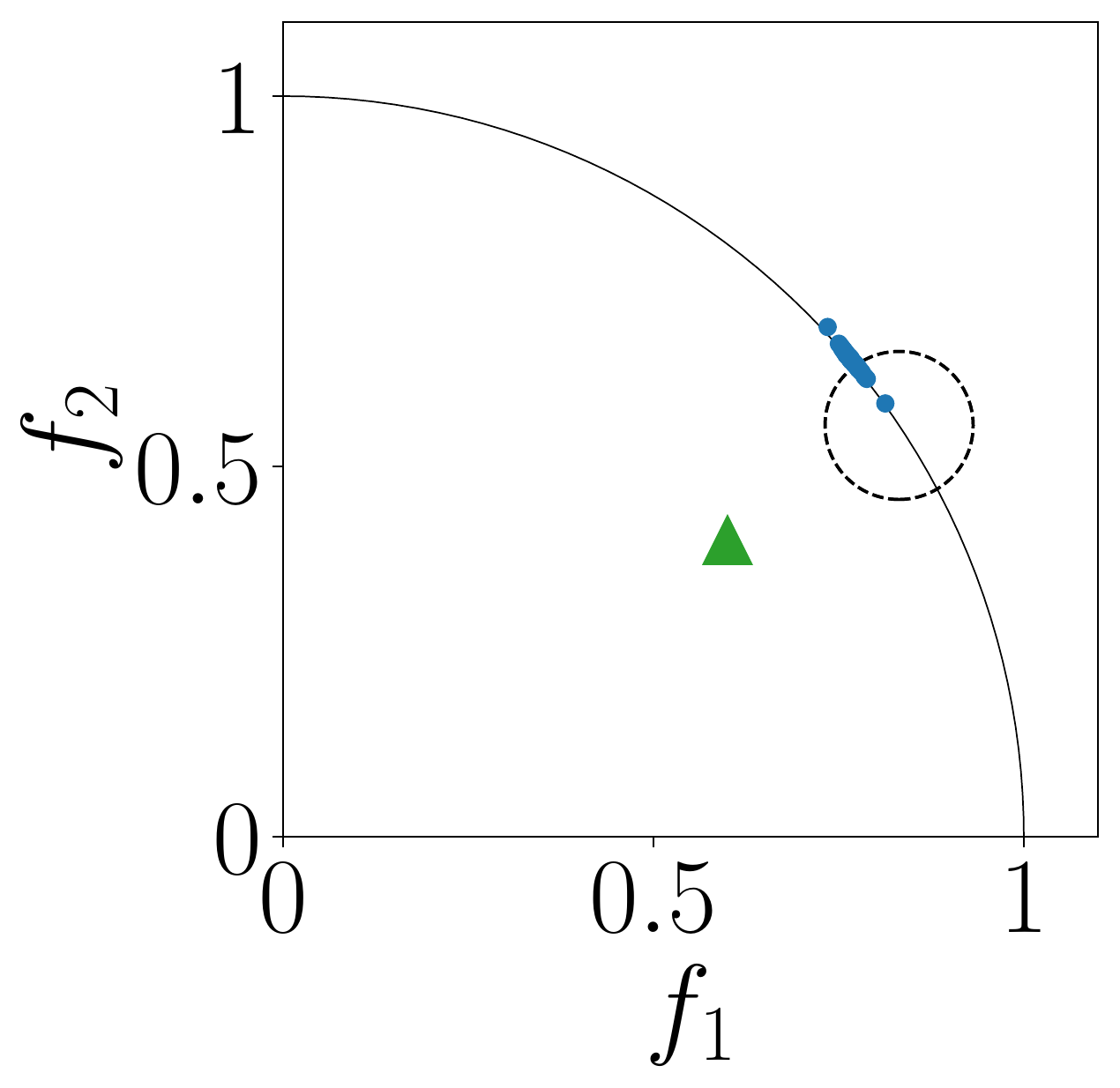}
   }
   \subfloat[NUMS (UA-IDDS)]{
  \includegraphics[width=0.15\textwidth]{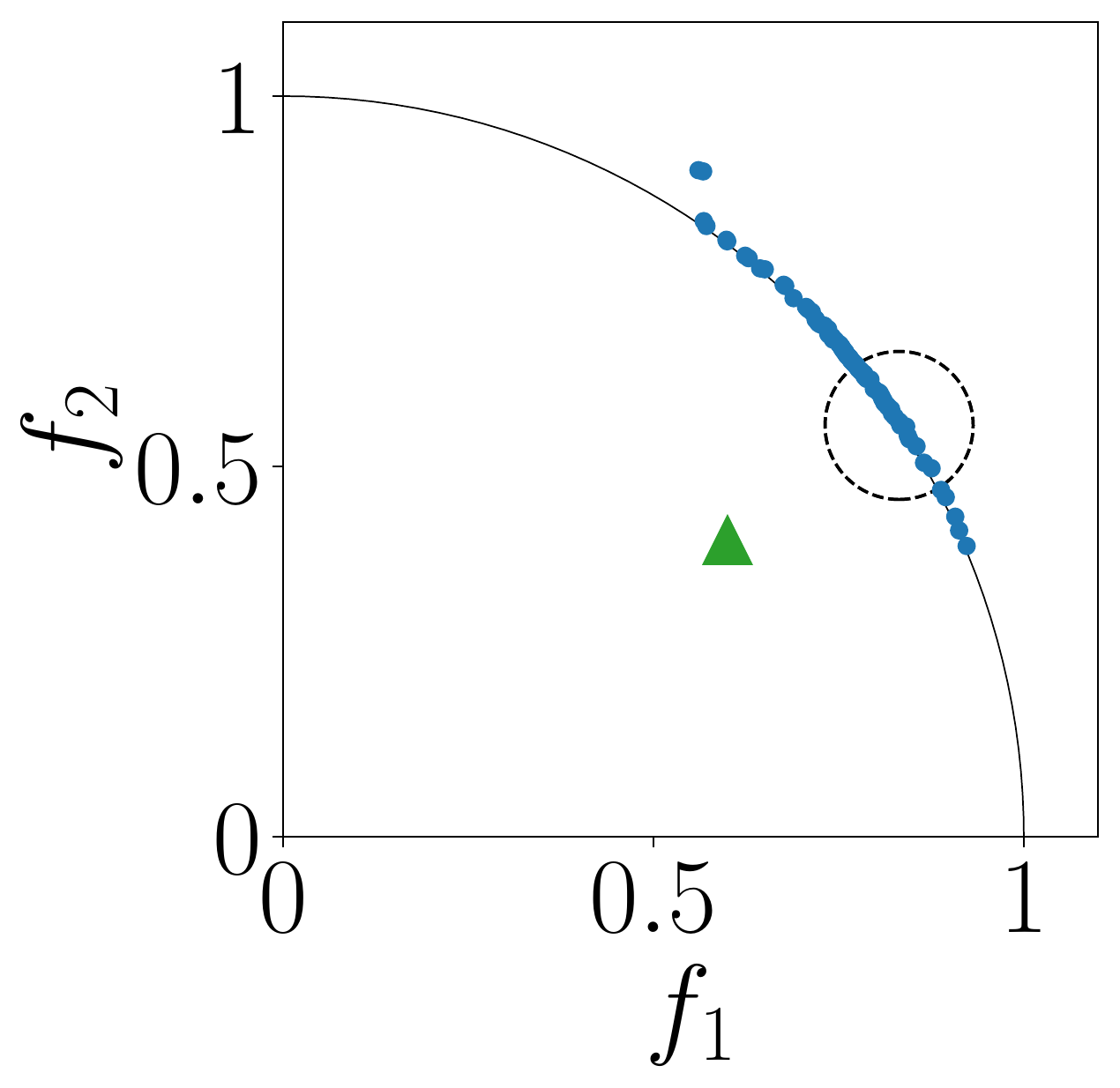}
}
\subfloat[NUMS (UA-PP)]{
  \includegraphics[width=0.15\textwidth]{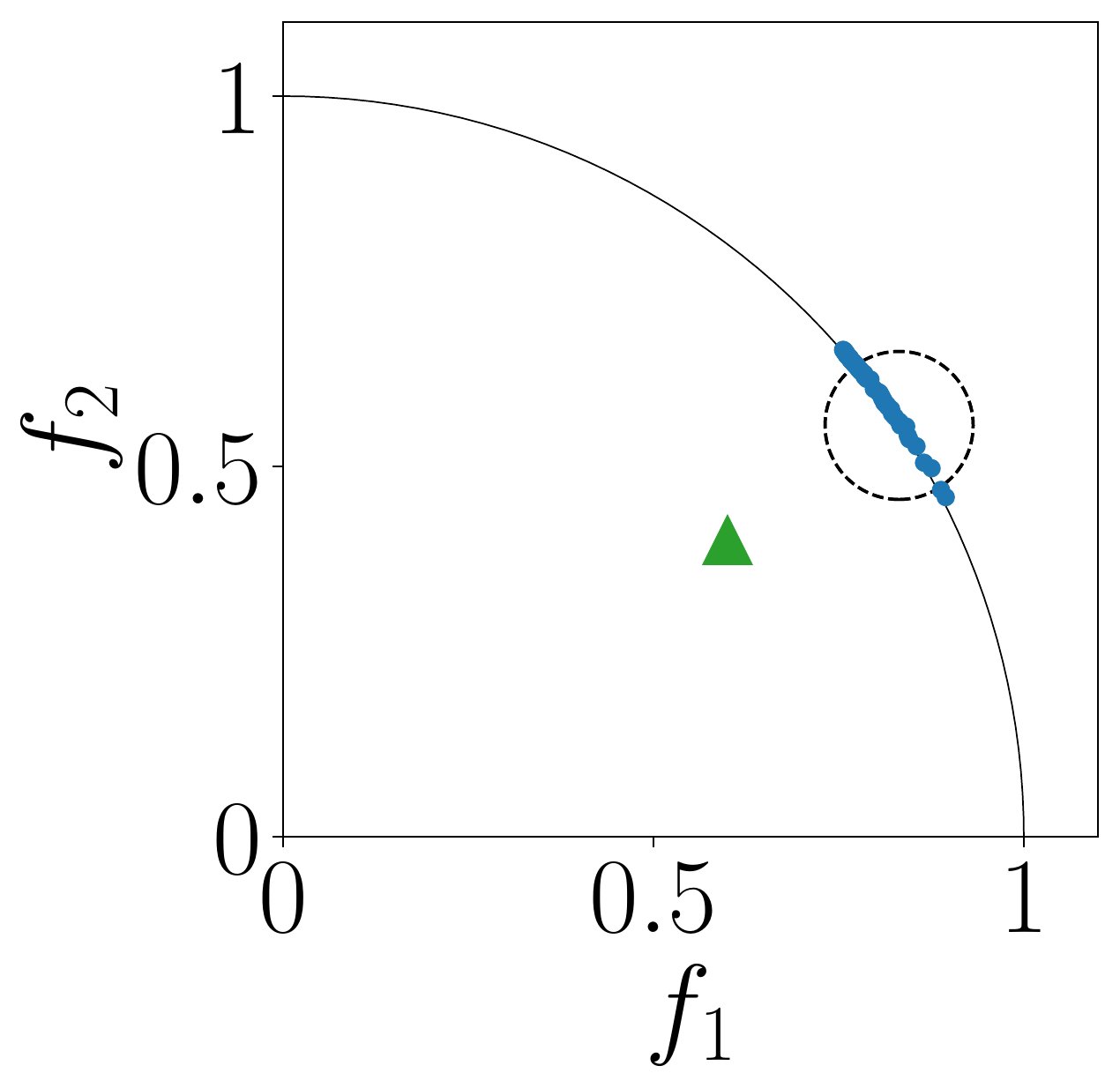}
}
\caption{Distributions of the objective vectors of the solutions in the three solution sets on DTLZ3 with $m=2$, where \tabgreen{$\blacktriangle$} is the reference point $\mathbf{z}$. The dotted circle represents the true ROI. ``NUMS'' stands for MOEA/D-NUMS.}
   \label{fig:100points_dtlz3}
\end{figure*}

\begin{figure*}[t]
   \centering
   \subfloat[R-NSGA-II (POP)]{
     \includegraphics[width=0.15\textwidth]{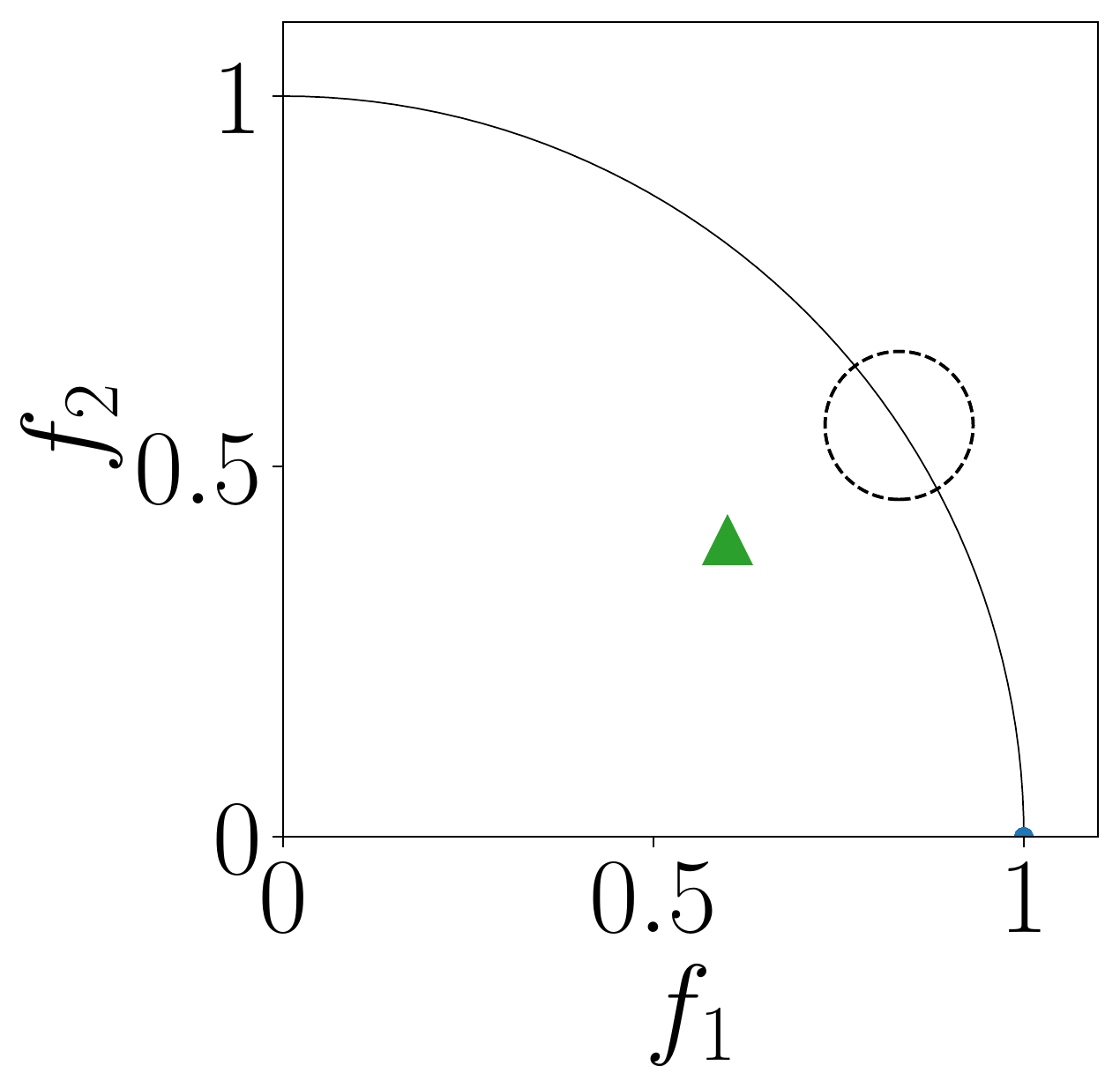}
   }
   \subfloat[R-NSGA-II (UA-IDDS)]{
  \includegraphics[width=0.15\textwidth]{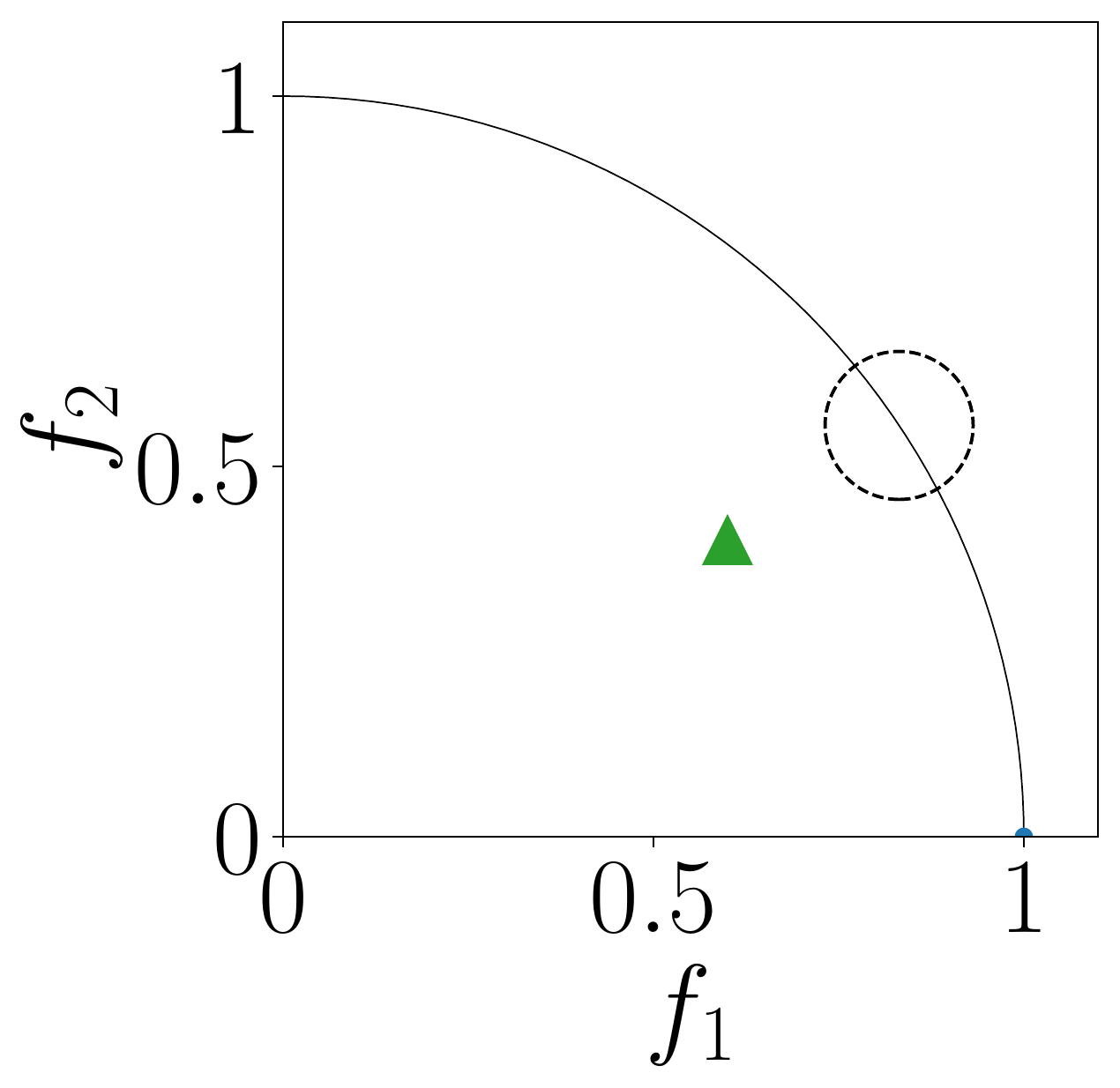}
}
\subfloat[R-NSGA-II (UA-PP)]{
  \includegraphics[width=0.15\textwidth]{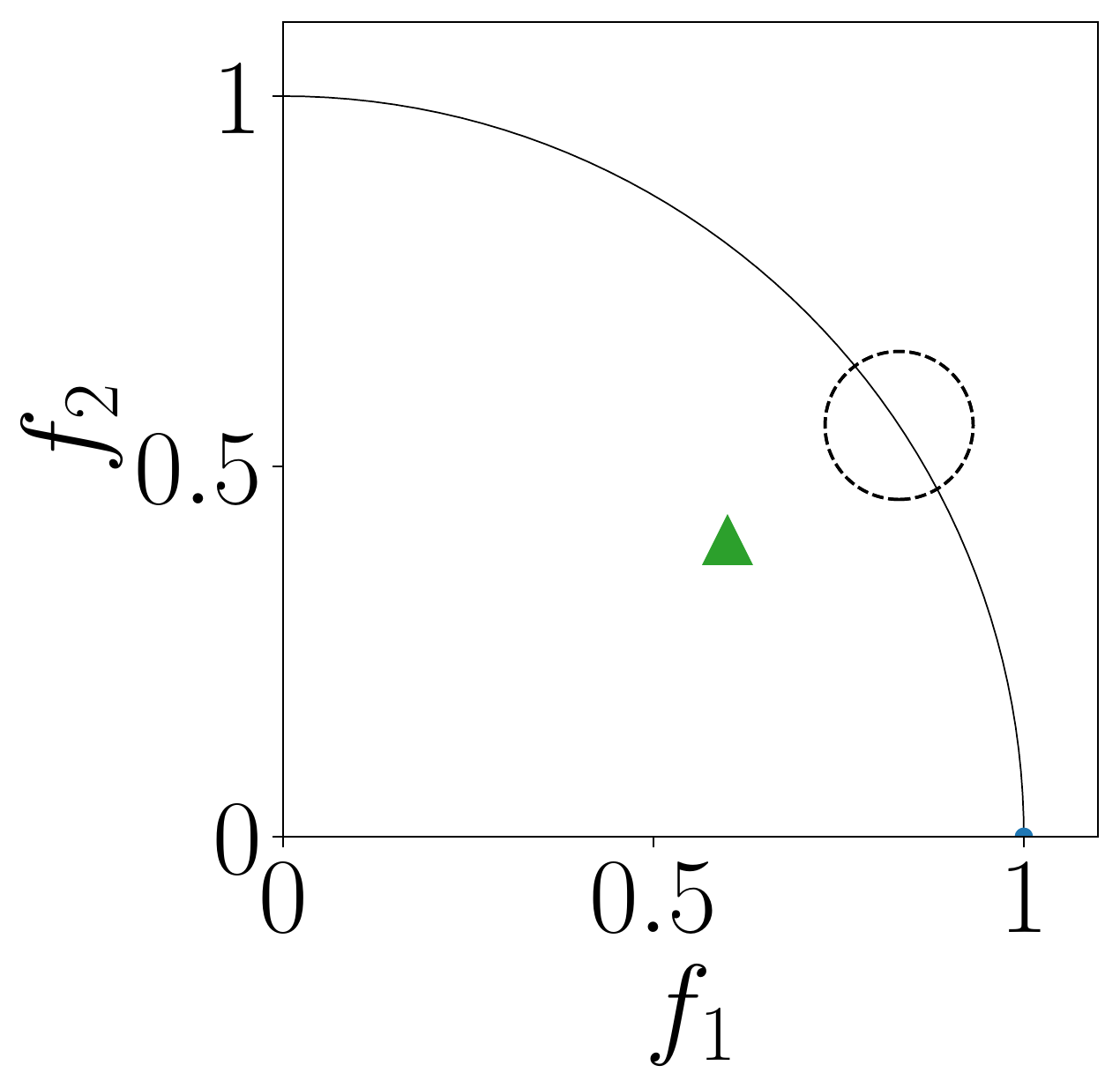}
}
\\
\subfloat[r-NSGA-II (POP)]{
     \includegraphics[width=0.15\textwidth]{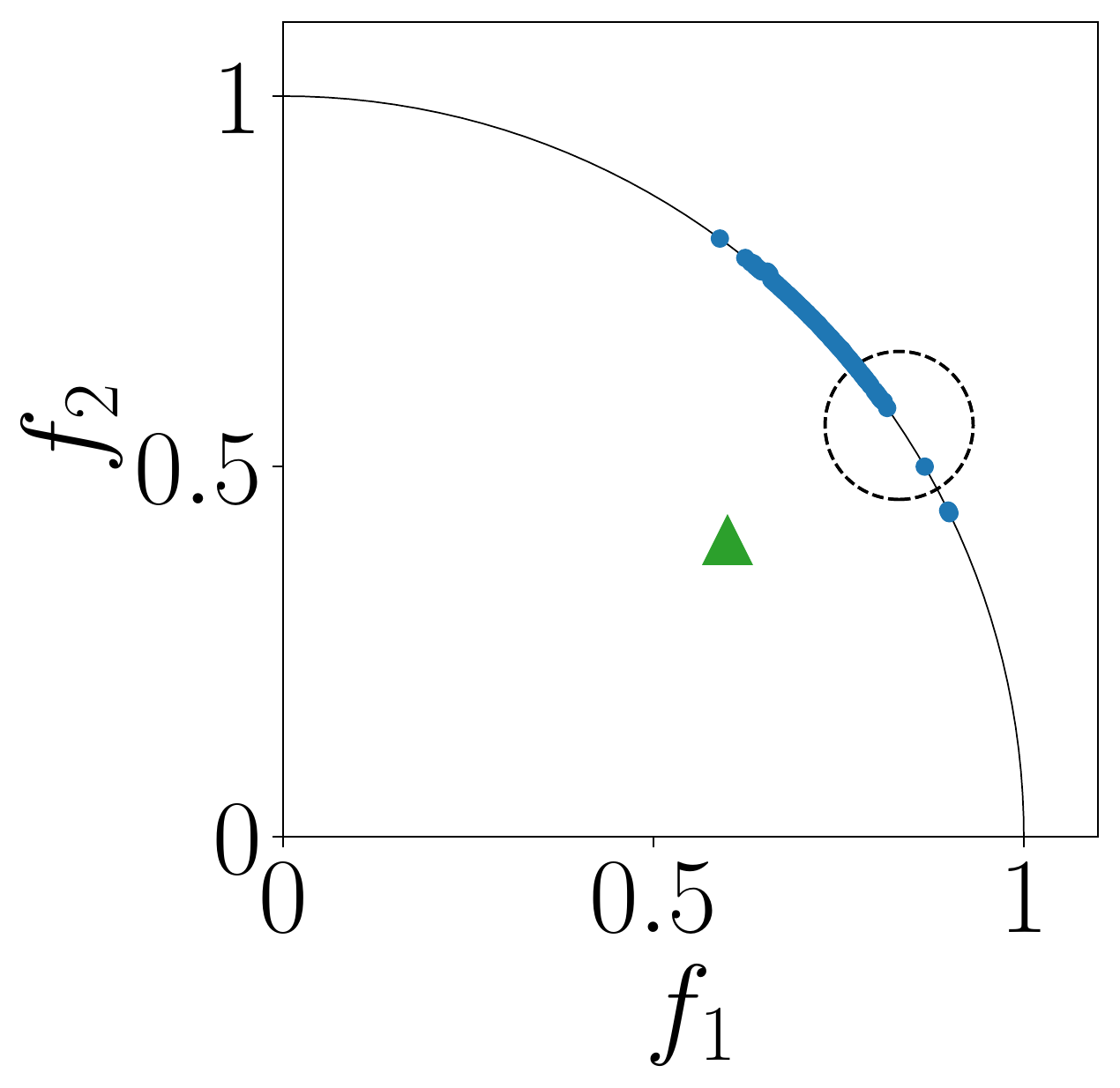}
   }
   \subfloat[r-NSGA-II (UA-IDDS)]{
  \includegraphics[width=0.15\textwidth]{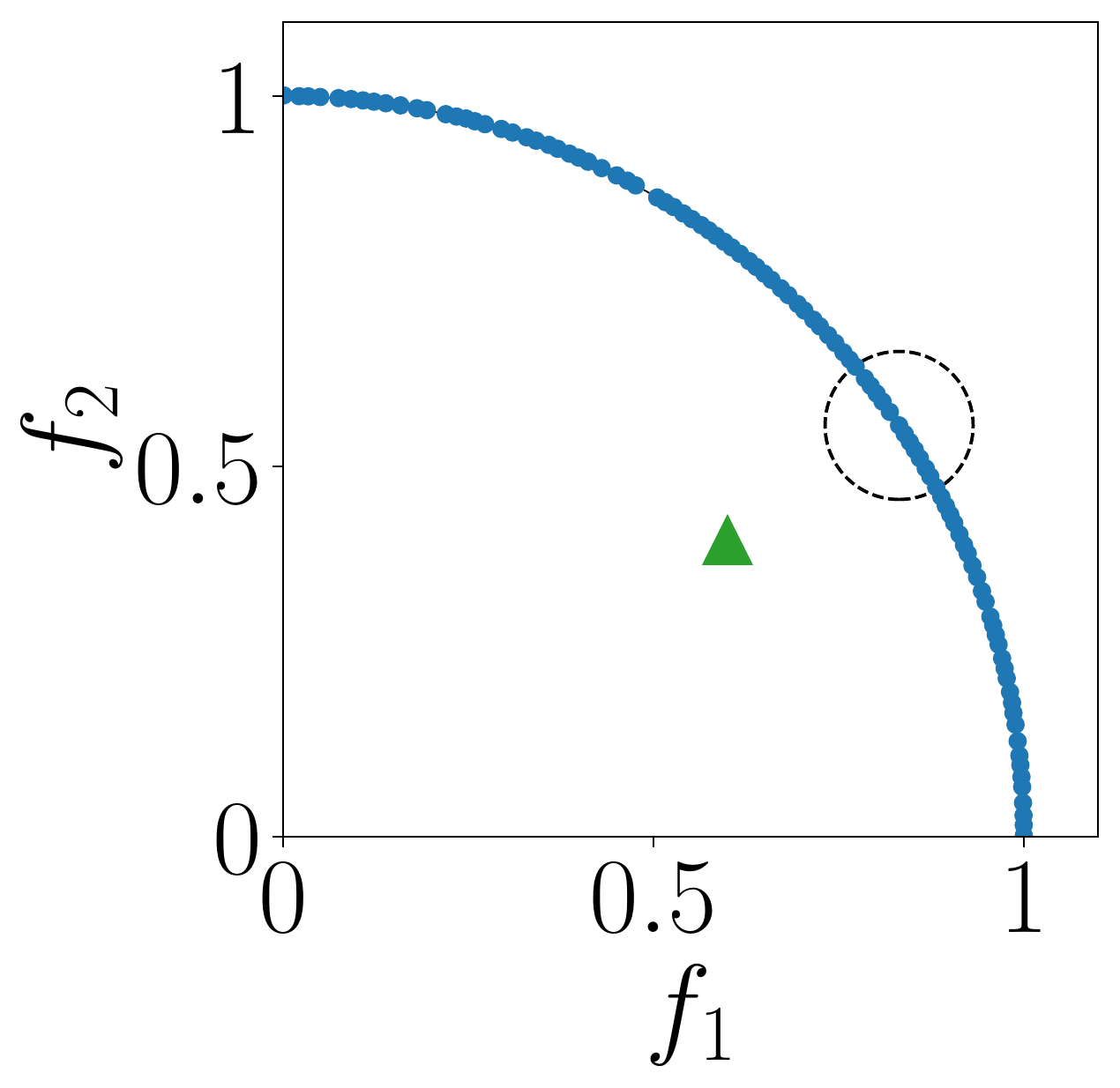}
}
\subfloat[r-NSGA-II (UA-PP)]{
  \includegraphics[width=0.15\textwidth]{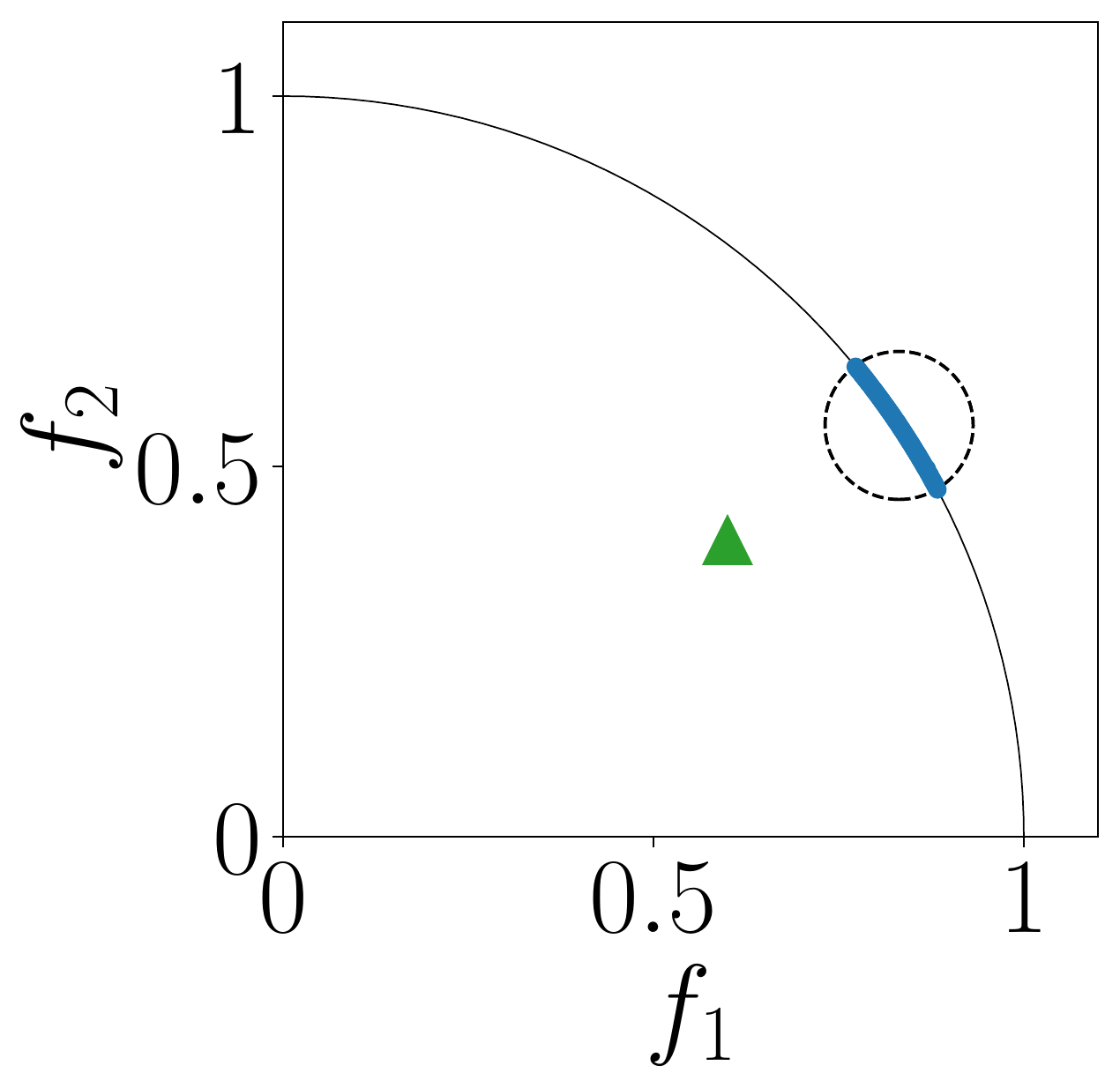}
}
\\
\subfloat[g-NSGA-II (POP)]{
     \includegraphics[width=0.15\textwidth]{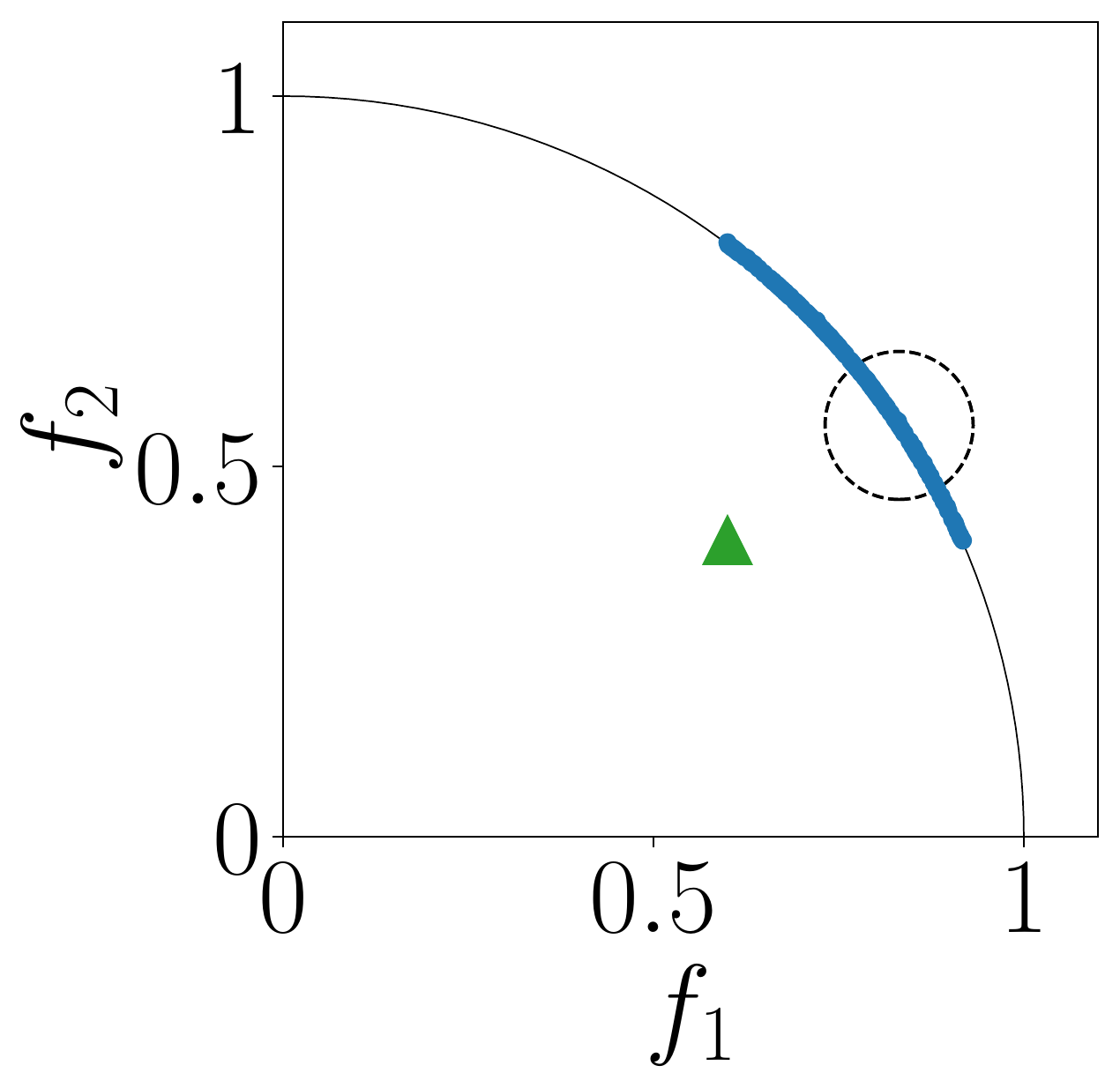}
   }
   \subfloat[g-NSGA-II (UA-IDDS)]{
  \includegraphics[width=0.15\textwidth]{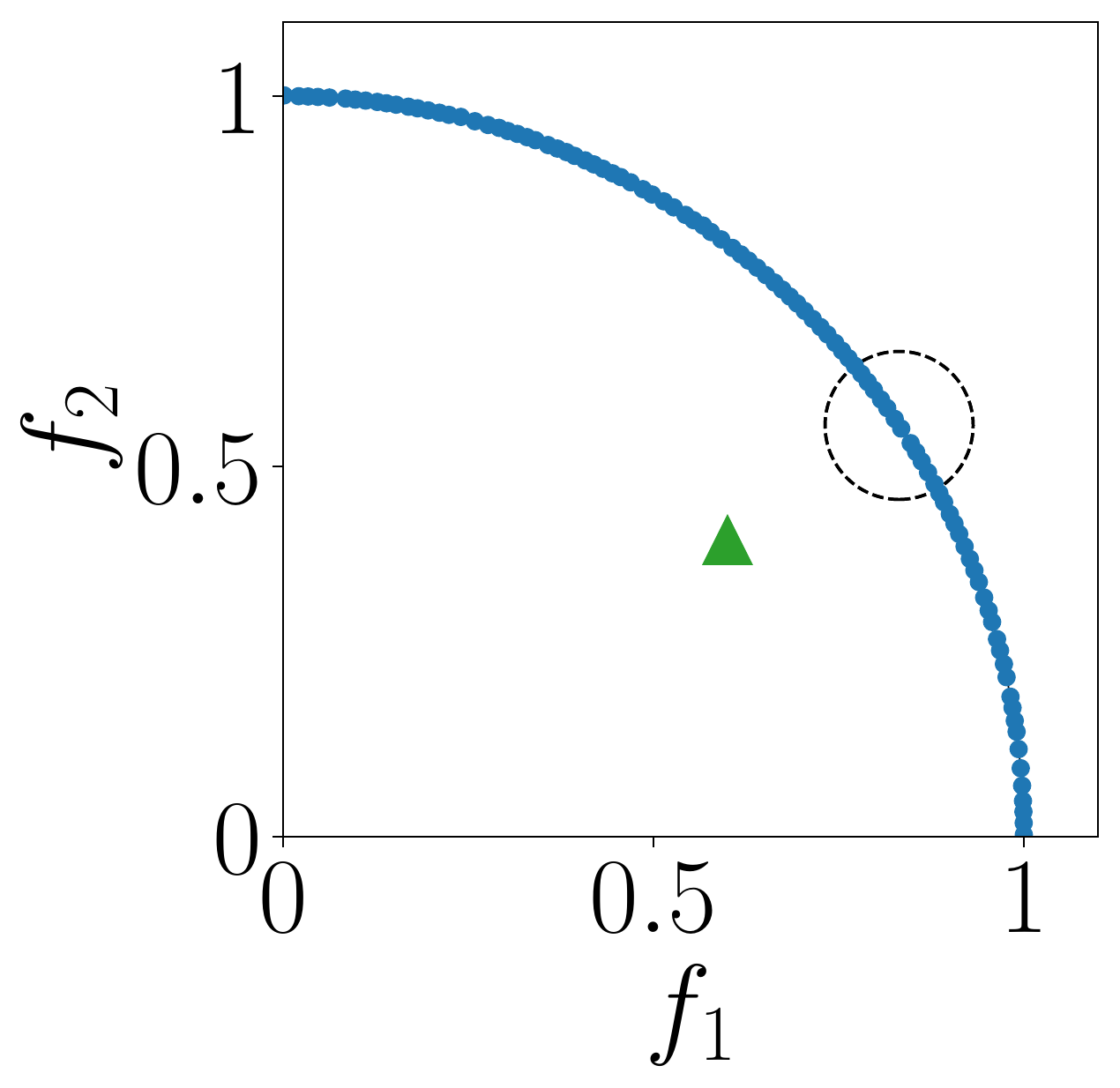}
}
\subfloat[g-NSGA-II (UA-PP)]{
  \includegraphics[width=0.15\textwidth]{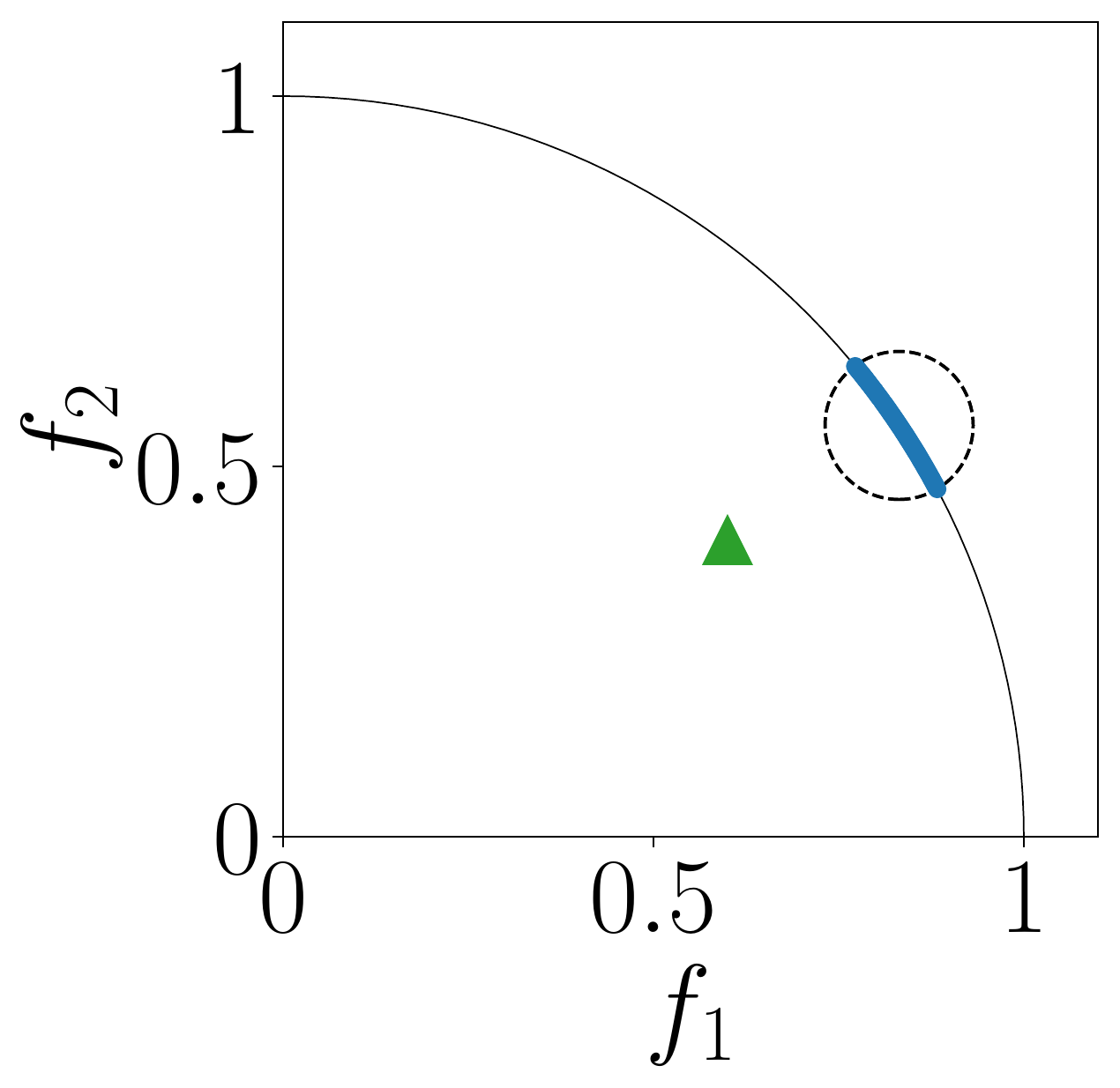}
}
\\
\subfloat[PBEA (POP)]{
     \includegraphics[width=0.15\textwidth]{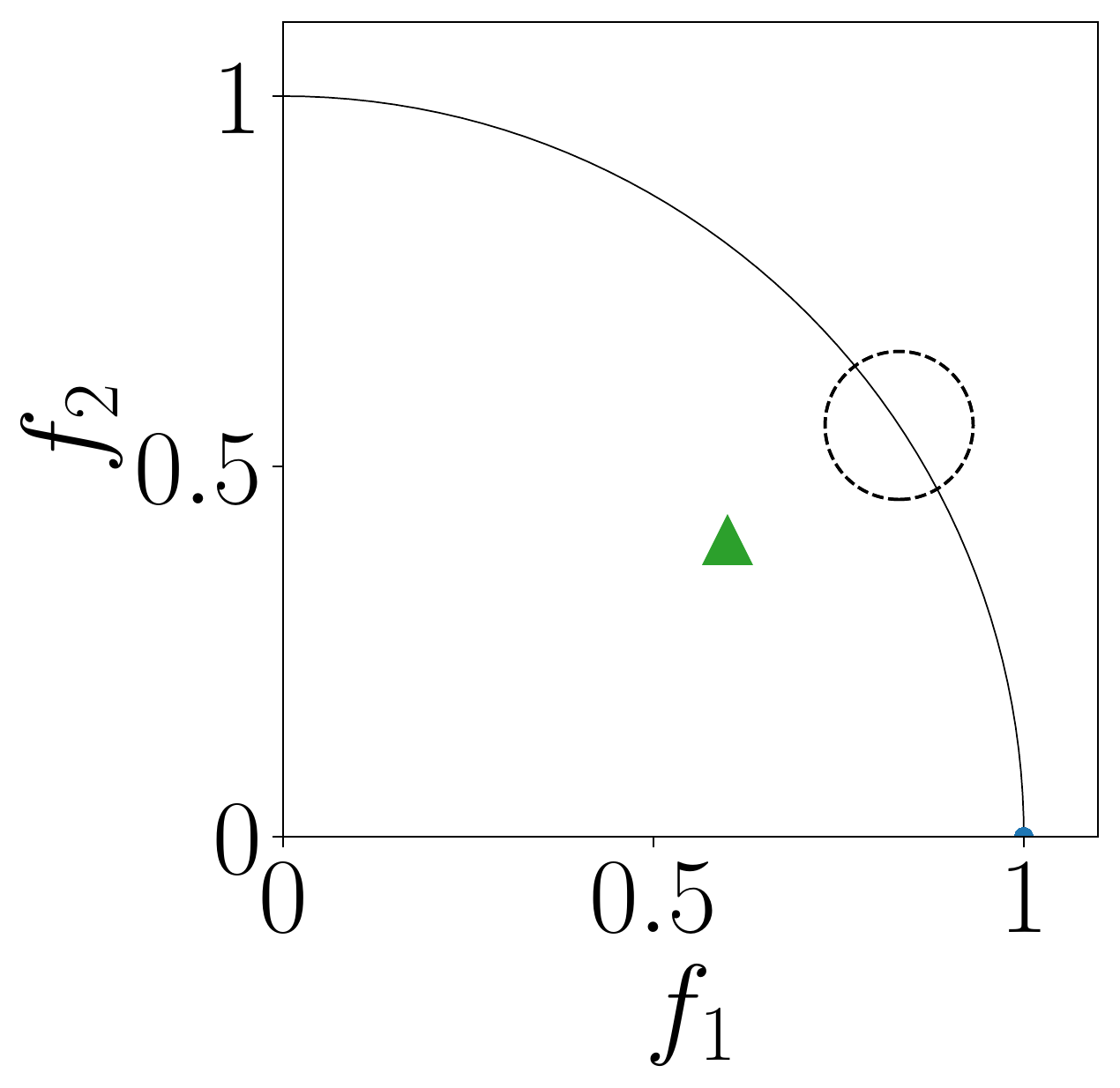}
   }
   \subfloat[PBEA (UA-IDDS)]{
  \includegraphics[width=0.15\textwidth]{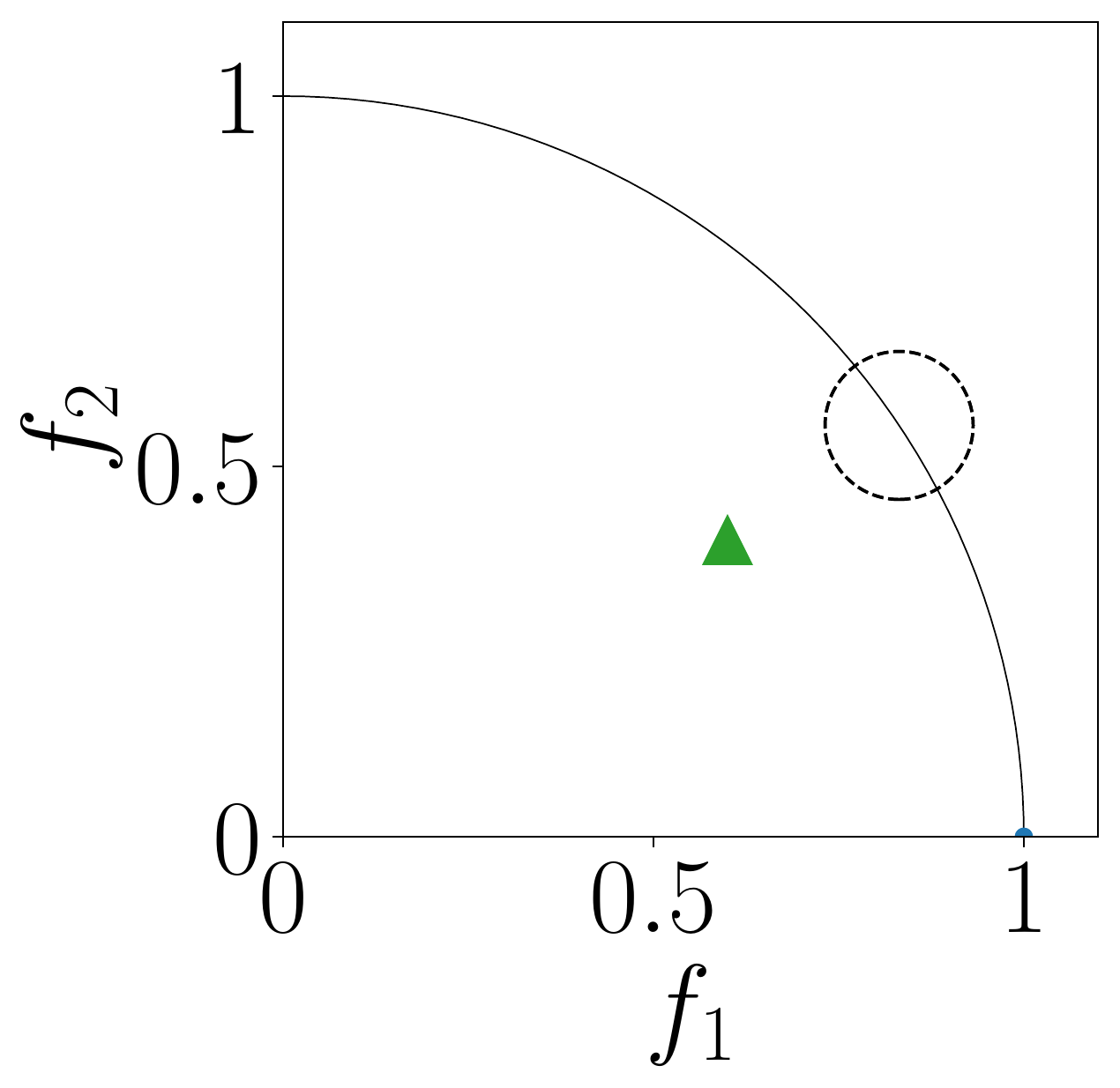}
}
\subfloat[PBEA (UA-PP)]{
  \includegraphics[width=0.15\textwidth]{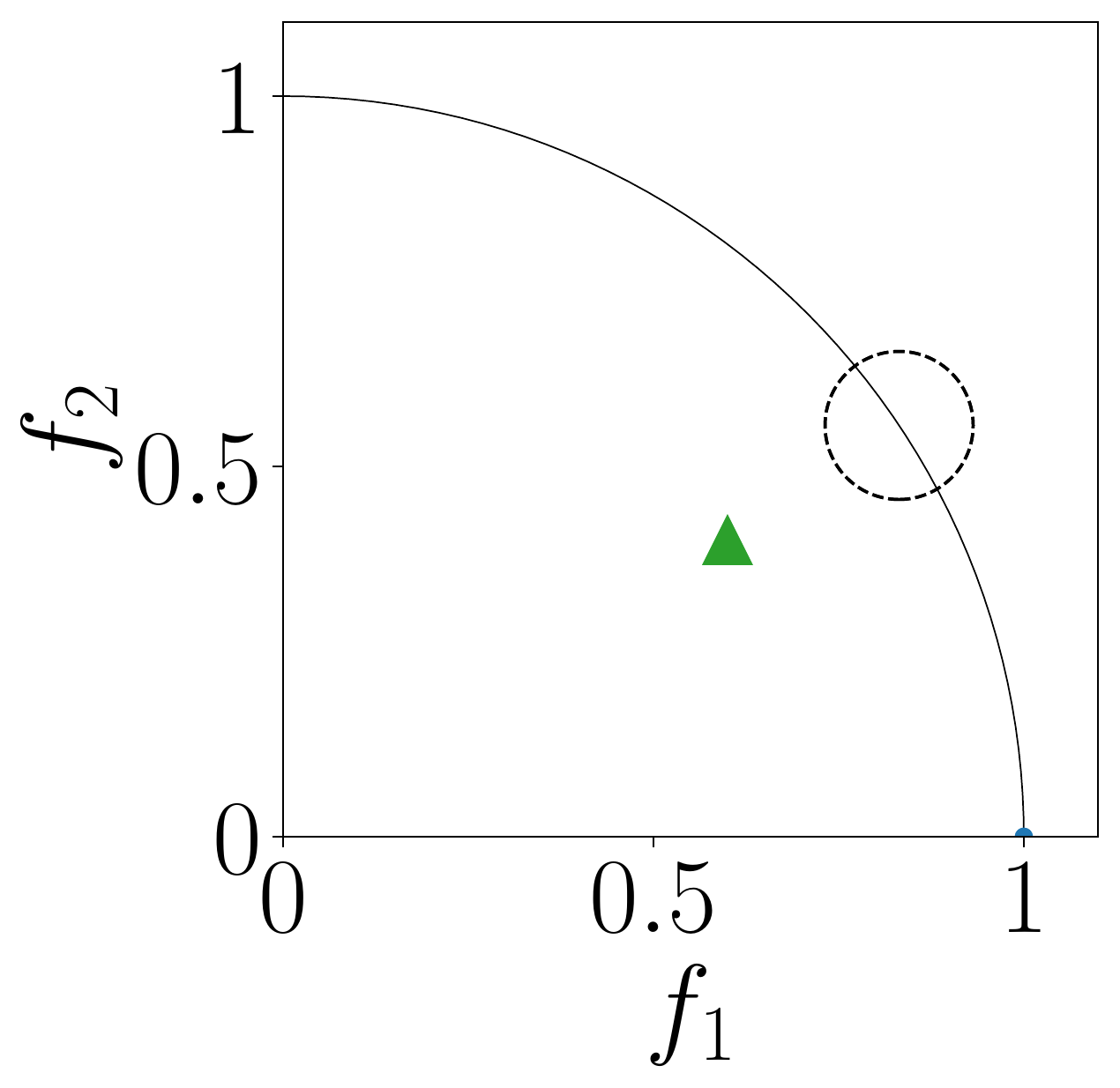}
}
\\
\subfloat[R-MEAD2 (POP]{
     \includegraphics[width=0.15\textwidth]{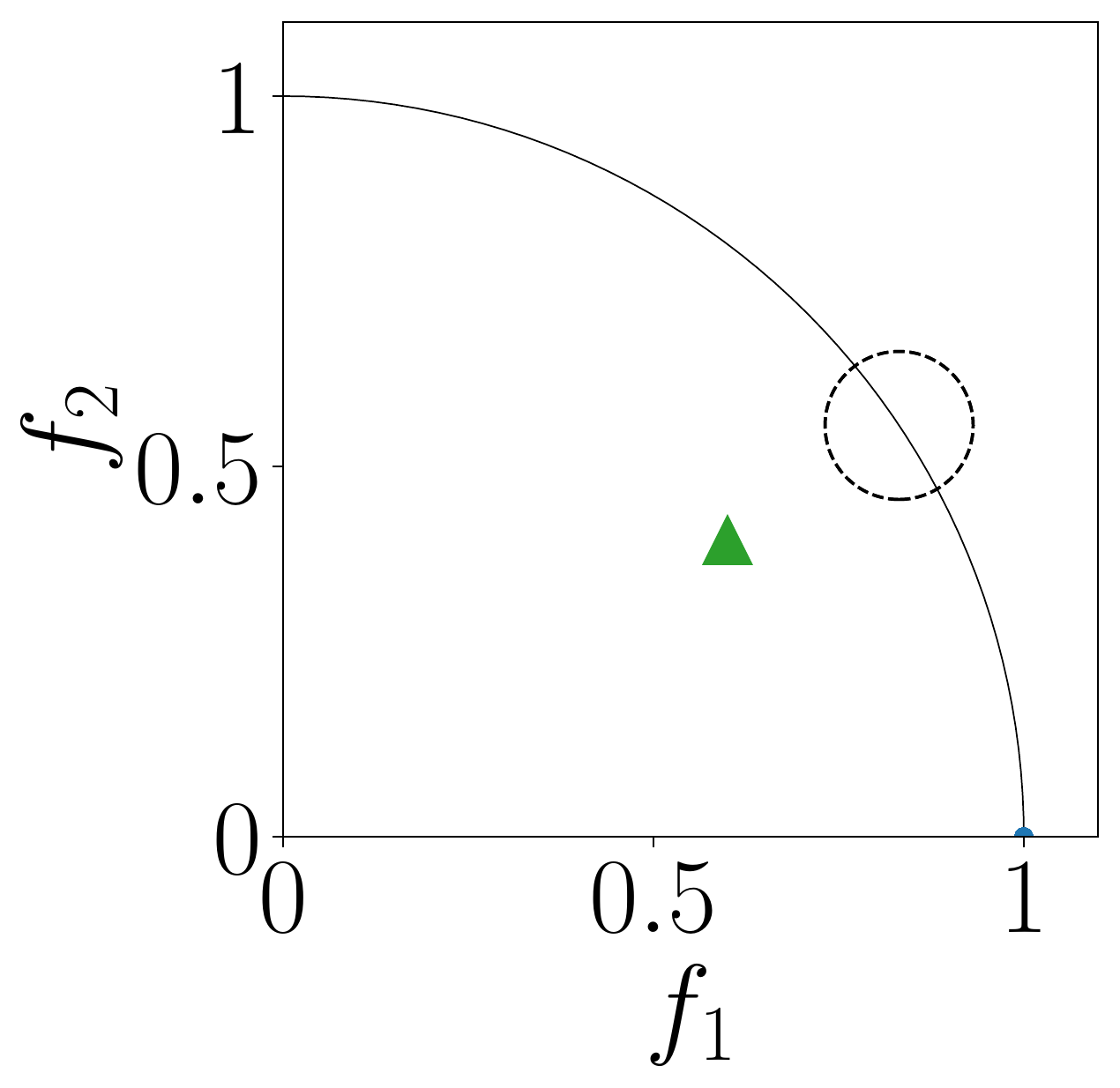}
   }
   \subfloat[R-MEAD2 (UA-IDDS)]{
  \includegraphics[width=0.15\textwidth]{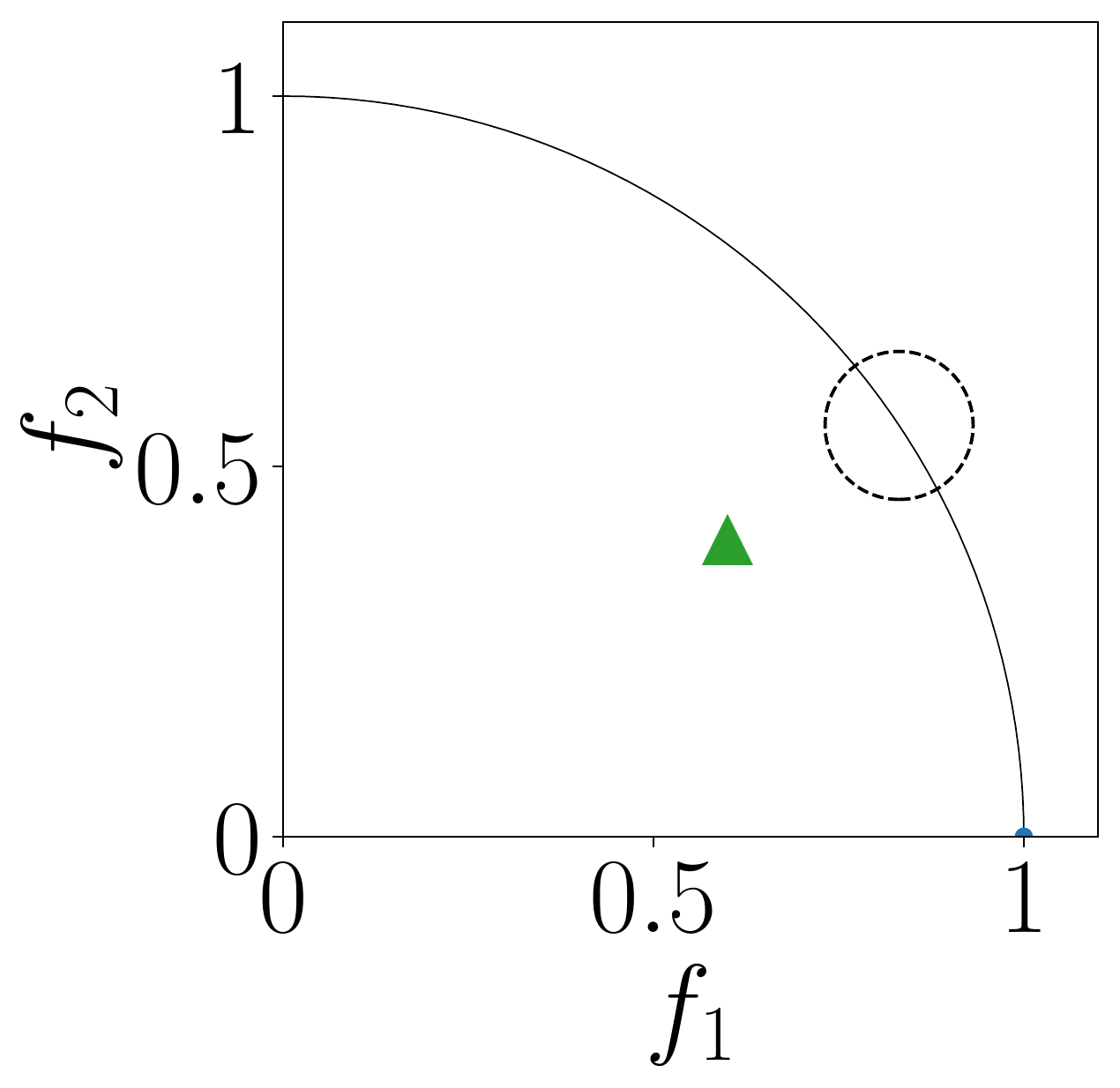}
}
\subfloat[R-MEAD2 (UA-PP)]{
  \includegraphics[width=0.15\textwidth]{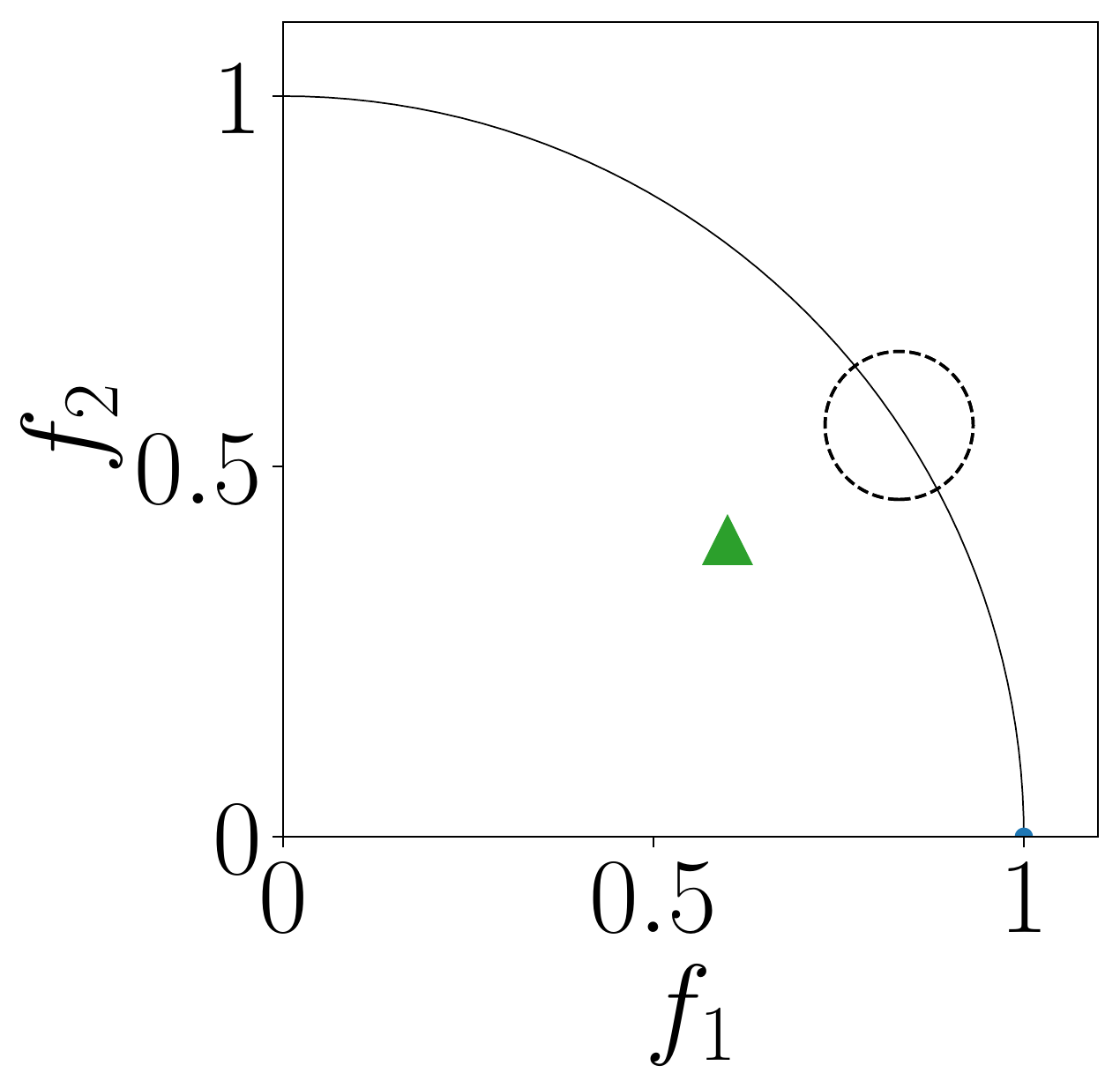}
}
\\
\subfloat[NUMS (POP)]{
     \includegraphics[width=0.15\textwidth]{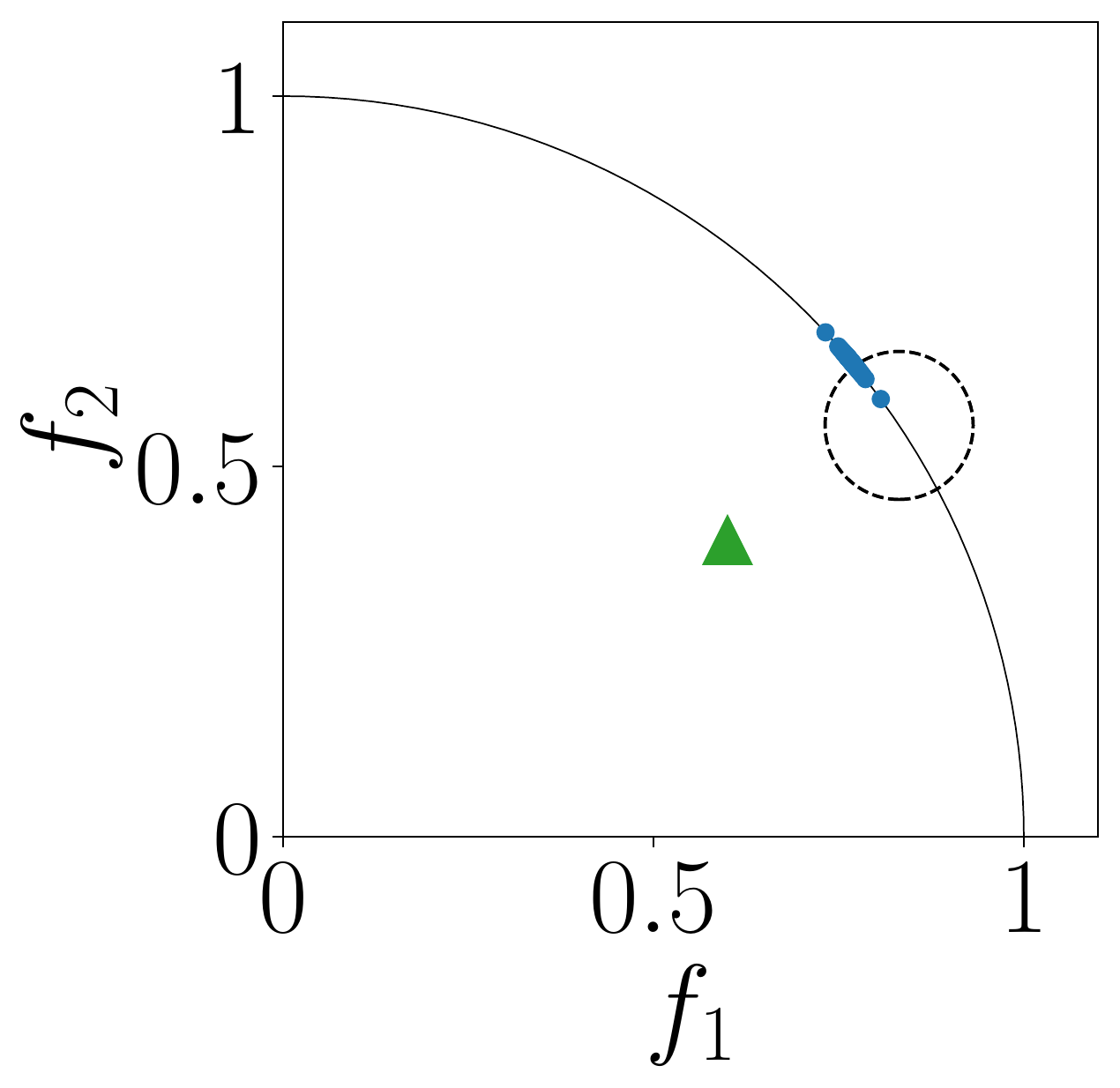}
   }
   \subfloat[NUMS (UA-IDDS)]{
  \includegraphics[width=0.15\textwidth]{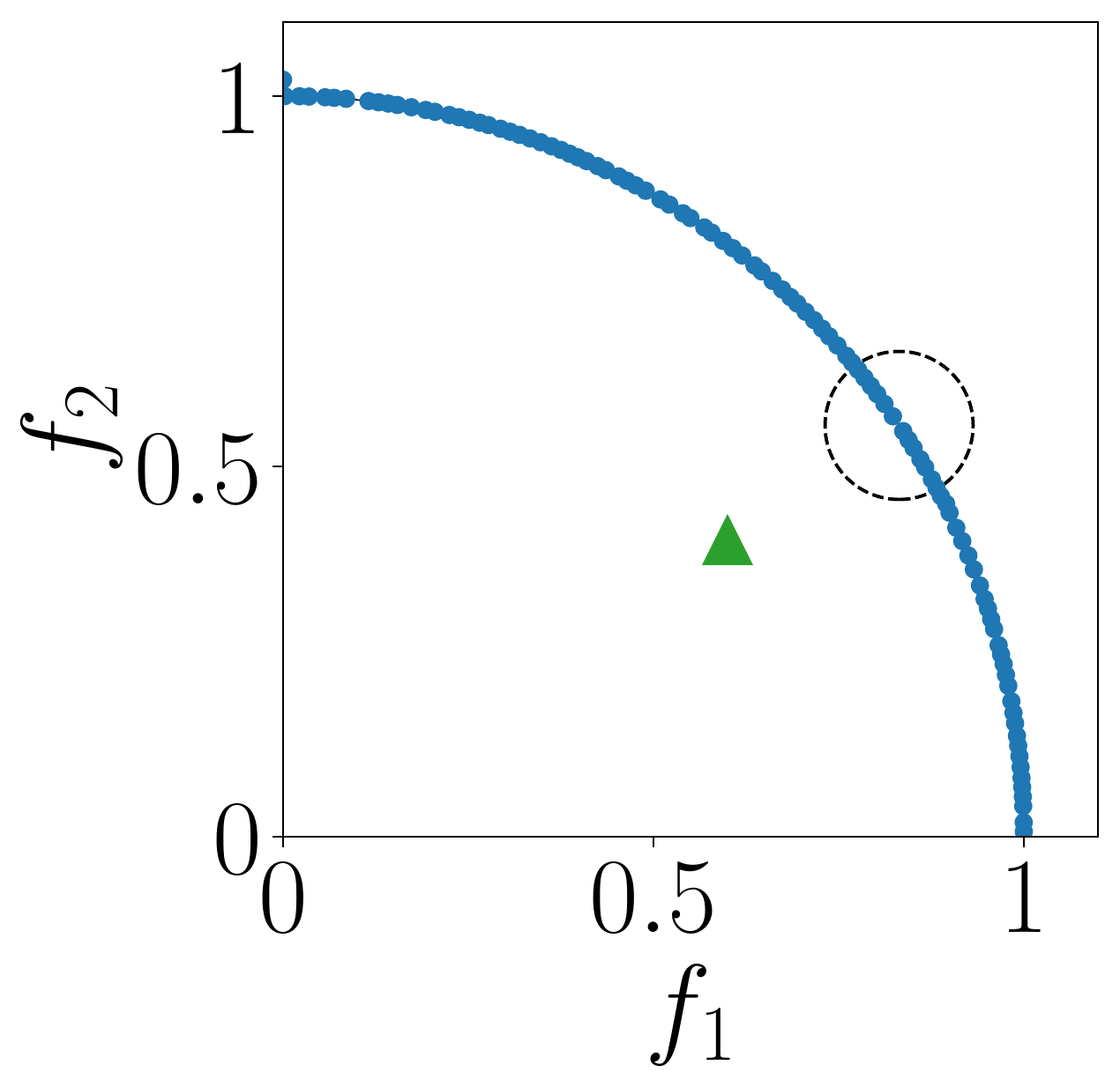}
}
\subfloat[NUMS (UA-PP)]{
  \includegraphics[width=0.15\textwidth]{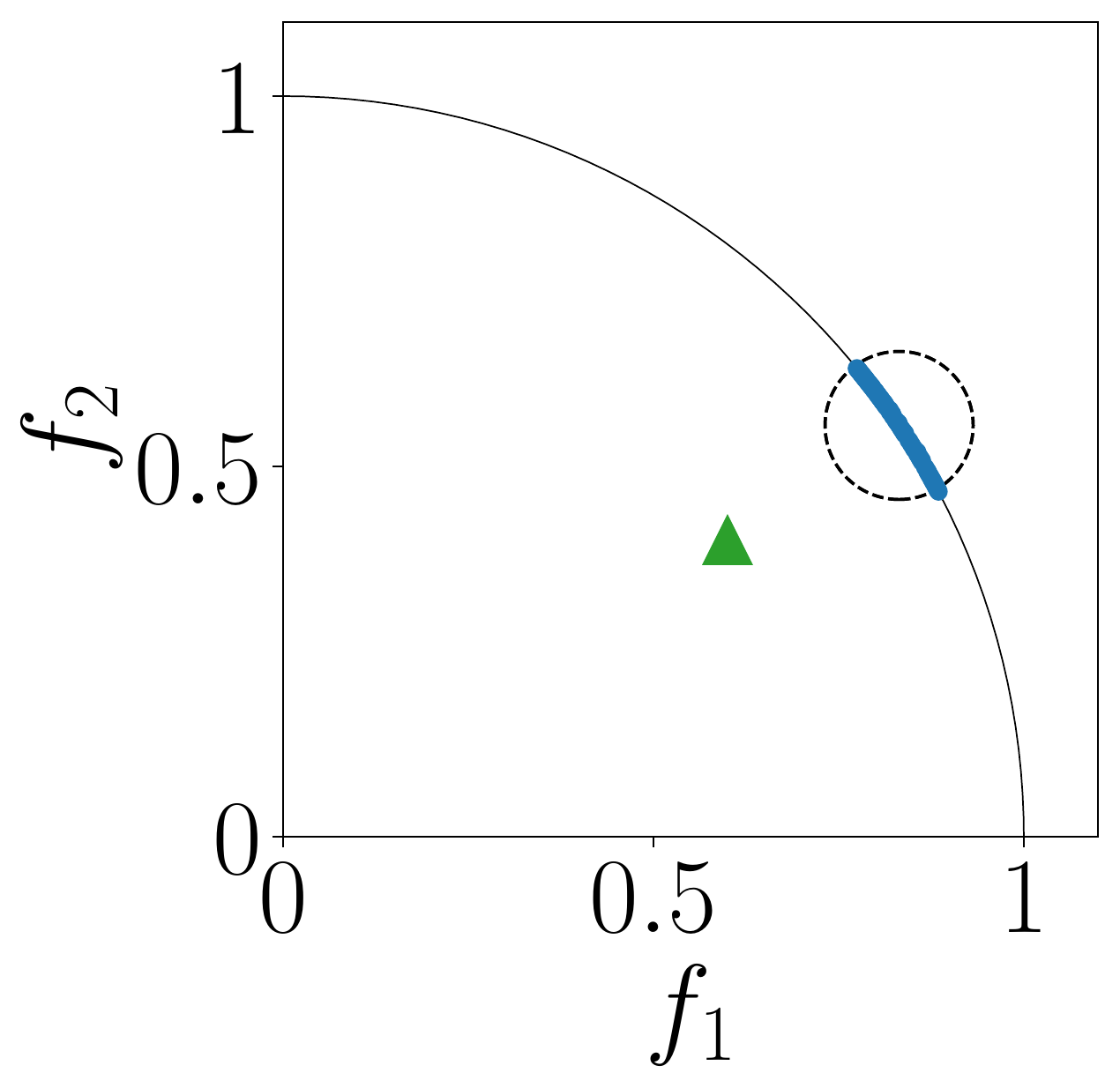}
}
\caption{Distributions of the objective vectors of the solutions in the three solution sets on DTLZ4 with $m=2$, where \tabgreen{$\blacktriangle$} is the reference point $\mathbf{z}$. The dotted circle represents the true ROI. ``NUMS'' stands for MOEA/D-NUMS.}
   \label{fig:100points_dtlz4}
\end{figure*}


\begin{table*}[t]
  \renewcommand{\arraystretch}{0.7} 
\centering
  \caption{\small Average IGD$^+$-C values of the three solution subsets found by r-NSGA-II, g-NSGA-II, PBEA, R-MEAD2, and MOEA/D-NUMS on the DTLZ1--DTLZ4 problems.}
  \label{tab:res_3subsets_five}  
{\small
\subfloat[r-NSGA-II]{
  \begin{tabular}{ccccc}
\toprule
Problem & $m$ & POP & UA-IDDS & UA-PP\\  
\midrule
 & 2 & 0.0018 & \cellcolor{c2}0.0015 ($+$) & \cellcolor{c1}0.0007 ($+$, $+$)\\
 & 3 & 0.0221 & \cellcolor{c2}0.0152 ($+$) & \cellcolor{c1}0.0079 ($+$, $+$)\\
DTLZ1 & 4 & 0.6009 & \cellcolor{c2}0.4523 ($\approx$) & \cellcolor{c1}0.4361 ($\approx$, $\approx$)\\
 & 5 & 2.5538 & \cellcolor{c2}1.8340 ($+$) & \cellcolor{c1}1.2448 ($+$, $+$)\\
 & 6 & \cellcolor{c2}2.9928 & 3.0074 ($\approx$) & \cellcolor{c1}1.3174 ($+$, $+$)\\
\midrule
 & 2 & 0.0312 & \cellcolor{c2}0.0015 ($+$) & \cellcolor{c1}0.0004 ($+$, $+$)\\
 & 3 & 0.0979 & \cellcolor{c2}0.0255 ($+$) & \cellcolor{c1}0.0085 ($+$, $+$)\\
DTLZ2 & 4 & 0.1700 & \cellcolor{c2}0.0618 ($+$) & \cellcolor{c1}0.0222 ($+$, $+$)\\
 & 5 & 0.1569 & \cellcolor{c2}0.0967 ($+$) & \cellcolor{c1}0.0306 ($+$, $+$)\\
 & 6 & 0.2044 & \cellcolor{c2}0.1223 ($+$) & \cellcolor{c1}0.0505 ($+$, $+$)\\
\midrule
 & 2 & 0.0076 & \cellcolor{c2}0.0075 ($\approx$) & \cellcolor{c1}0.0060 ($+$, $+$)\\
 & 3 & 0.0905 & \cellcolor{c2}0.0861 ($\approx$) & \cellcolor{c1}0.0713 ($+$, $+$)\\
DTLZ3 & 4 & 5.7675 & \cellcolor{c2}5.5092 ($\approx$) & \cellcolor{c1}5.3614 ($\approx$, $\approx$)\\
 & 5 & 11.0159 & \cellcolor{c2}10.3284 ($\approx$) & \cellcolor{c1}9.1448 ($\approx$, $\approx$)\\
 & 6 & 12.4904 & \cellcolor{c2}11.4337 ($\approx$) & \cellcolor{c1}9.7152 ($\approx$, $\approx$)\\
\midrule
 & 2 & 0.0553 & \cellcolor{c2}0.0454 ($+$) & \cellcolor{c1}0.0438 ($+$, $+$)\\
 & 3 & \cellcolor{c2}0.0241 & 0.0280 ($-$) & \cellcolor{c1}0.0062 ($+$, $+$)\\
DTLZ4 & 4 & \cellcolor{c2}0.0306 & 0.0552 ($-$) & \cellcolor{c1}0.0153 ($+$, $+$)\\
 & 5 & \cellcolor{c2}0.0657 & 0.0941 ($-$) & \cellcolor{c1}0.0278 ($+$, $+$)\\
 & 6 & \cellcolor{c2}0.1120 & 0.1155 ($\approx$) & \cellcolor{c1}0.0367 ($+$, $+$)\\
\bottomrule
\end{tabular}
}
\subfloat[g-NSGA-II]{
  \begin{tabular}{ccccc}
\toprule
Problem & $m$ & POP & UA-IDDS & UA-PP\\  
\midrule
 & 2 & 0.0267 & \cellcolor{c2}0.0265 ($\approx$) & \cellcolor{c1}0.0259 ($+$, $+$)\\
 & 3 & 10.7277 & \cellcolor{c2}3.7391 ($+$) & \cellcolor{c1}3.7306 ($+$, $\approx$)\\
DTLZ1 & 4 & 77.7690 & \cellcolor{c2}20.7678 ($+$) & \cellcolor{c1}15.7325 ($+$, $+$)\\
 & 5 & 123.8328 & \cellcolor{c2}50.6386 ($+$) & \cellcolor{c1}23.8382 ($+$, $+$)\\
 & 6 & 145.9085 & \cellcolor{c2}74.0552 ($+$) & \cellcolor{c1}30.8597 ($+$, $+$)\\
\midrule
 & 2 & \cellcolor{c2}0.0013 & 0.0019 ($-$) & \cellcolor{c1}0.0004 ($+$, $+$)\\
 & 3 & \cellcolor{c2}0.0264 & 0.0332 ($-$) & \cellcolor{c1}0.0090 ($+$, $+$)\\
DTLZ2 & 4 & 0.2756 & \cellcolor{c2}0.1978 ($+$) & \cellcolor{c1}0.0999 ($+$, $+$)\\
 & 5 & 1.1044 & \cellcolor{c2}0.5408 ($+$) & \cellcolor{c1}0.3901 ($+$, $+$)\\
 & 6 & 1.3364 & \cellcolor{c2}0.7812 ($+$) & \cellcolor{c1}0.4915 ($+$, $+$)\\
\midrule
 & 2 & 5.4893 & \cellcolor{c2}5.3352 ($\approx$) & \cellcolor{c1}5.3352 ($\approx$, $\approx$)\\
 & 3 & 30.6638 & \cellcolor{c2}27.3401 ($\approx$) & \cellcolor{c1}27.3305 ($\approx$, $\approx$)\\
DTLZ3 & 4 & 266.2339 & \cellcolor{c2}127.7268 ($+$) & \cellcolor{c1}119.3209 ($+$, $\approx$)\\
 & 5 & 608.8715 & \cellcolor{c2}352.6806 ($+$) & \cellcolor{c1}257.7681 ($+$, $+$)\\
 & 6 & 851.7529 & \cellcolor{c2}481.3857 ($+$) & \cellcolor{c1}340.8093 ($+$, $+$)\\
\midrule
 & 2 & 1.3208 & \cellcolor{c2}0.0347 ($+$) & \cellcolor{c1}0.0329 ($+$, $+$)\\
 & 3 & 0.2915 & \cellcolor{c2}0.0346 ($+$) & \cellcolor{c1}0.0116 ($+$, $+$)\\
DTLZ4 & 4 & 0.0884 & \cellcolor{c2}0.0851 ($\approx$) & \cellcolor{c1}0.0344 ($+$, $+$)\\
 & 5 & 0.4019 & \cellcolor{c2}0.2268 ($+$) & \cellcolor{c1}0.1172 ($+$, $+$)\\
 & 6 & 0.7191 & \cellcolor{c2}0.3082 ($+$) & \cellcolor{c1}0.2047 ($+$, $+$)\\
\bottomrule
\end{tabular}
}
\\
\subfloat[PBEA]{
  \begin{tabular}{ccccc}
\toprule
Problem & $m$ & POP & UA-IDDS & UA-PP\\  
\midrule
 & 2 & 0.0339 & \cellcolor{c2}0.0009 ($+$) & \cellcolor{c1}0.0006 ($+$, $+$)\\
 & 3 & 0.0411 & \cellcolor{c2}0.0121 ($+$) & \cellcolor{c1}0.0081 ($+$, $+$)\\
DTLZ1 & 4 & 0.0609 & \cellcolor{c2}0.0258 ($+$) & \cellcolor{c1}0.0175 ($+$, $+$)\\
 & 5 & 0.0628 & \cellcolor{c2}0.0385 ($+$) & \cellcolor{c1}0.0285 ($+$, $+$)\\
 & 6 & 0.0915 & \cellcolor{c2}0.0647 ($+$) & \cellcolor{c1}0.0535 ($+$, $+$)\\
\midrule
 & 2 & \cellcolor{c2}0.0012 & 0.0023 ($-$) & \cellcolor{c1}0.0003 ($+$, $+$)\\
 & 3 & 0.0420 & \cellcolor{c2}0.0263 ($+$) & \cellcolor{c1}0.0053 ($+$, $+$)\\
DTLZ2 & 4 & \cellcolor{c2}0.0550 & 0.0590 ($-$) & \cellcolor{c1}0.0296 ($+$, $+$)\\
 & 5 & \cellcolor{c2}0.0799 & 0.1030 ($-$) & \cellcolor{c1}0.0527 ($+$, $+$)\\
 & 6 & \cellcolor{c2}0.1494 & 0.1536 ($\approx$) & \cellcolor{c1}0.1412 ($+$, $+$)\\
\midrule
 & 2 & 0.0571 & \cellcolor{c2}0.0061 ($+$) & \cellcolor{c1}0.0059 ($+$, $\approx$)\\
 & 3 & 0.1553 & \cellcolor{c1}0.0804 ($+$) & \cellcolor{c2}0.0963 ($+$, $\approx$)\\
DTLZ3 & 4 & 0.2022 & \cellcolor{c1}0.1328 ($+$) & \cellcolor{c2}0.1664 ($+$, $\approx$)\\
 & 5 & 0.2503 & \cellcolor{c1}0.2114 ($\approx$) & \cellcolor{c2}0.2389 ($\approx$, $\approx$)\\
 & 6 & 0.3300 & \cellcolor{c1}0.2740 ($+$) & \cellcolor{c2}0.3157 ($\approx$, $-$)\\
\midrule
 & 2 & \cellcolor{c2}0.0768 & 0.0775 ($-$) & \cellcolor{c1}0.0763 ($+$, $+$)\\
 & 3 & 0.0580 & \cellcolor{c2}0.0470 ($+$) & \cellcolor{c1}0.0281 ($+$, $+$)\\
DTLZ4 & 4 & 0.0897 & \cellcolor{c2}0.0886 ($\approx$) & \cellcolor{c1}0.0660 ($+$, $+$)\\
 & 5 & \cellcolor{c2}0.1183 & 0.1309 ($-$) & \cellcolor{c1}0.0884 ($+$, $+$)\\
 & 6 & 0.1506 & \cellcolor{c2}0.1488 ($\approx$) & \cellcolor{c1}0.1174 ($+$, $+$)\\
\bottomrule
\end{tabular}
}
\subfloat[R-MEAD2]{
  \begin{tabular}{ccccc}
\toprule
Problem & $m$ & POP & UA-IDDS & UA-PP\\  
\midrule
 & 2 & 0.0140 & \cellcolor{c2}0.0072 ($+$) & \cellcolor{c1}0.0067 ($+$, $\approx$)\\
 & 3 & 0.1785 & \cellcolor{c2}0.1344 ($\approx$) & \cellcolor{c1}0.1324 ($\approx$, $\approx$)\\
DTLZ1 & 4 & 0.2730 & \cellcolor{c2}0.2247 ($+$) & \cellcolor{c1}0.2235 ($+$, $\approx$)\\
 & 5 & 0.2630 & \cellcolor{c2}0.2297 ($\approx$) & \cellcolor{c1}0.2286 ($+$, $\approx$)\\
 & 6 & 0.2874 & \cellcolor{c2}0.2567 ($\approx$) & \cellcolor{c1}0.2554 ($\approx$, $\approx$)\\
\midrule
 & 2 & 0.0112 & \cellcolor{c2}0.0032 ($+$) & \cellcolor{c1}0.0019 ($+$, $+$)\\
 & 3 & 0.0931 & \cellcolor{c2}0.0352 ($+$) & \cellcolor{c1}0.0141 ($+$, $+$)\\
DTLZ2 & 4 & 0.2894 & \cellcolor{c2}0.1176 ($+$) & \cellcolor{c1}0.0948 ($+$, $+$)\\
 & 5 & 0.3439 & \cellcolor{c2}0.2084 ($+$) & \cellcolor{c1}0.1837 ($+$, $\approx$)\\
 & 6 & 0.4480 & \cellcolor{c2}0.3216 ($+$) & \cellcolor{c1}0.3077 ($+$, $\approx$)\\
\midrule
 & 2 & 0.1931 & \cellcolor{c1}0.1437 ($\approx$) & \cellcolor{c2}0.1439 ($\approx$, $\approx$)\\
 & 3 & 0.6909 & \cellcolor{c1}0.6211 ($+$) & \cellcolor{c2}0.6211 ($+$, $\approx$)\\
DTLZ3 & 4 & 0.8050 & \cellcolor{c2}0.7381 ($+$) & \cellcolor{c1}0.7381 ($+$, $\approx$)\\
 & 5 & 0.7892 & \cellcolor{c1}0.7068 ($+$) & \cellcolor{c2}0.7093 ($+$, $\approx$)\\
 & 6 & 0.8368 & \cellcolor{c1}0.7823 ($+$) & \cellcolor{c2}0.7824 ($+$, $\approx$)\\
\midrule
 & 2 & 0.1639 & \cellcolor{c2}0.1580 ($\approx$) & \cellcolor{c1}0.1579 ($\approx$, $\approx$)\\
 & 3 & 0.1734 & \cellcolor{c2}0.1711 ($\approx$) & \cellcolor{c1}0.1710 ($\approx$, $\approx$)\\
DTLZ4 & 4 & 0.2456 & \cellcolor{c2}0.2404 ($\approx$) & \cellcolor{c1}0.2401 ($\approx$, $\approx$)\\
 & 5 & 0.2958 & \cellcolor{c2}0.2662 ($+$) & \cellcolor{c1}0.2629 ($+$, $\approx$)\\
 & 6 & 0.2598 & \cellcolor{c1}0.2409 ($+$) & \cellcolor{c2}0.2443 ($\approx$, $\approx$)\\
\bottomrule
\end{tabular}
}
\\
\subfloat[MOEA/D-NUMS]{
  \begin{tabular}{ccccc}
\toprule
Problem & $m$ & POP & UA-IDDS & UA-PP\\  
\midrule
 & 2 & 0.0148 & \cellcolor{c1}0.0007 ($+$) & \cellcolor{c2}0.0014 ($+$, $-$)\\
 & 3 & 0.0278 & \cellcolor{c2}0.0092 ($+$) & \cellcolor{c1}0.0054 ($+$, $+$)\\
DTLZ1 & 4 & 0.0423 & \cellcolor{c2}0.0263 ($+$) & \cellcolor{c1}0.0149 ($+$, $+$)\\
 & 5 & \cellcolor{c2}0.0339 & 0.0341 ($\approx$) & \cellcolor{c1}0.0231 ($+$, $+$)\\
 & 6 & \cellcolor{c2}0.0378 & 0.0449 ($-$) & \cellcolor{c1}0.0311 ($+$, $+$)\\
\midrule
 & 2 & 0.0452 & \cellcolor{c2}0.0014 ($+$) & \cellcolor{c1}0.0004 ($+$, $+$)\\
 & 3 & 0.1464 & \cellcolor{c2}0.0426 ($+$) & \cellcolor{c1}0.0345 ($+$, $+$)\\
DTLZ2 & 4 & 0.2161 & \cellcolor{c2}0.1256 ($+$) & \cellcolor{c1}0.1012 ($+$, $+$)\\
 & 5 & 0.1696 & \cellcolor{c2}0.1360 ($+$) & \cellcolor{c1}0.0965 ($+$, $+$)\\
 & 6 & \cellcolor{c2}0.2392 & 0.2457 ($-$) & \cellcolor{c1}0.1780 ($+$, $+$)\\
\midrule
 & 2 & 0.0834 & \cellcolor{c1}0.0424 ($+$) & \cellcolor{c2}0.0426 ($+$, $\approx$)\\
 & 3 & 0.2056 & \cellcolor{c1}0.1373 ($+$) & \cellcolor{c2}0.1405 ($+$, $\approx$)\\
DTLZ3 & 4 & 0.2958 & \cellcolor{c2}0.2411 ($+$) & \cellcolor{c1}0.2257 ($+$, $\approx$)\\
 & 5 & \cellcolor{c2}0.4276 & 0.4491 ($\approx$) & \cellcolor{c1}0.3911 ($+$, $+$)\\
 & 6 & \cellcolor{c2}0.7108 & 0.7999 ($-$) & \cellcolor{c1}0.6825 ($\approx$, $+$)\\
\midrule
 & 2 & 0.0453 & \cellcolor{c2}0.0030 ($+$) & \cellcolor{c1}0.0007 ($+$, $+$)\\
 & 3 & 0.1476 & \cellcolor{c1}0.0425 ($+$) & \cellcolor{c2}0.0430 ($+$, $\approx$)\\
DTLZ4 & 4 & 0.2308 & \cellcolor{c2}0.0929 ($+$) & \cellcolor{c1}0.0831 ($+$, $+$)\\
 & 5 & 0.1802 & \cellcolor{c2}0.1255 ($+$) & \cellcolor{c1}0.0926 ($+$, $+$)\\
 & 6 & 0.2639 & \cellcolor{c2}0.2134 ($+$) & \cellcolor{c1}0.2046 ($+$, $+$)\\
\bottomrule
\end{tabular}
}
}
\end{table*}

\begin{table*}[t]
  \renewcommand{\arraystretch}{0.84} 
\centering
  \caption{\small The rankings of the eight $\mu$ values for each number of function evaluations and $m \in \{2, 4, 6\}$ on the DTLZ1--DTLZ4 problems (1st/2nd/3rd/4th/5th/6th/7th/8th).}
  \label{tab:sup_rankings_mu}  
{\footnotesize
\subfloat[$m=2$]{
\begin{tabular}{lccccccc}
\toprule
PBEMO & $1$K FEs & $5$K FEs  & $10$K FEs  & $30$K FEs  & $50$K FEs\\
\midrule
R-NSGA-II & 8/20/40/100/200/300/400/500 & 20/40/8/100/200/300/400/500 & 8/20/40/100/200/300/400/500 & 8/100/40/200/20/300/400/500 & 200/100/300/40/400/8/20/500\\
r-NSGA-II & 20/40/8/100/200/300/400/500 & 20/40/8/100/200/300/400/500 & 20/8/40/100/200/300/400/500 & 40/20/100/8/200/300/400/500 & 40/20/100/8/200/300/400/500\\
g-NSGA-II & 20/8/40/100/200/300/400/500 & 40/100/20/200/8/300/400/500 & 40/100/200/20/300/400/8/500 & 200/100/40/300/400/20/500/8 & 100/200/40/300/20/400/500/8\\
PBEA & 20/40/8/100/200/300/400/500 & 40/20/8/200/100/300/400/500 & 8/20/40/100/200/300/400/500 & 100/40/20/300/200/8/400/500 & 200/40/100/300/20/400/8/500\\
R-MEAD2 & 8/20/40/100/200/300/400/500 & 8/20/40/100/200/300/400/500 & 8/20/40/100/200/300/400/500 & 20/40/100/8/200/300/400/500 & 20/40/100/200/300/8/400/500\\
MOEA/D-NUMS & 8/20/40/100/200/300/400/500 & 8/20/100/40/200/300/400/500 & 8/20/40/200/100/300/400/500 & 20/100/40/8/200/300/400/500 & 20/100/40/200/8/300/400/500\\
\bottomrule
\end{tabular}
}
\\
\subfloat[$m=4$]{
\begin{tabular}{lccccccc}
\toprule
PBEMO & $1$K FEs & $5$K FEs  & $10$K FEs  & $30$K FEs  & $50$K FEs\\
\midrule
R-NSGA-II & 8/20/40/100/200/300/400/500 & 40/100/20/200/8/300/400/500 & 20/200/300/40/100/8/400/500 & 20/40/300/400/8/100/200/500 & 20/40/100/400/300/500/8/200\\
r-NSGA-II & 40/20/8/100/200/300/400/500 & 40/20/100/200/8/300/400/500 & 40/20/100/200/8/300/400/500 & 40/20/100/8/200/300/400/500 & 40/20/8/100/200/300/400/500\\
g-NSGA-II & 40/20/8/100/200/300/400/500 & 100/200/300/40/20/400/500/8 & 200/300/100/400/500/40/20/8 & 400/500/300/200/100/40/20/8 & 400/500/200/300/100/40/20/8\\
PBEA & 20/40/8/100/200/300/400/500 & 100/40/20/200/300/8/400/500 & 40/100/20/200/300/8/400/500 & 100/200/300/400/40/500/20/8 & 200/300/400/40/100/500/20/8\\
R-MEAD2 & 20/8/40/100/300/200/400/500 & 100/20/200/8/40/300/400/500 & 100/200/8/20/300/40/400/500 & 100/20/8/200/400/40/300/500 & 20/8/100/300/500/40/200/400\\
MOEA/D-NUMS & 8/20/40/100/200/400/300/500 & 20/8/40/100/200/300/400/500 & 20/8/40/100/200/300/400/500 & 20/40/8/200/100/300/400/500 & 20/40/8/100/200/300/400/500\\
\bottomrule
\end{tabular}
}
\\
\subfloat[$m=6$]{
\begin{tabular}{lccccccc}
\toprule
PBEMO & $1$K FEs & $5$K FEs  & $10$K FEs  & $30$K FEs  & $50$K FEs\\
\midrule
R-NSGA-II & 20/40/100/8/200/300/400/500 & 40/100/200/20/300/8/400/500 & 100/200/300/20/40/8/400/500 & 40/20/200/400/100/300/500/8 & 100/40/400/20/500/8/200/300\\
r-NSGA-II & 40/20/100/8/200/300/400/500 & 40/100/200/8/20/300/400/500 & 100/40/20/200/8/300/400/500 & 100/20/40/200/8/300/400/500 & 100/40/20/200/8/300/400/500\\
g-NSGA-II & 8/20/200/100/400/40/300/500 & 8/500/300/400/200/20/40/100 & 300/8/500/400/200/20/40/100 & 8/500/400/300/20/200/40/100 & 8/500/400/20/300/200/40/100\\
PBEA & 20/40/8/100/200/300/400/500 & 40/100/200/20/300/400/8/500 & 100/200/40/300/20/400/8/500 & 300/100/200/400/500/40/20/8 & 300/400/200/100/500/40/20/8\\
R-MEAD2 & 8/40/20/100/200/300/400/500 & 100/200/8/20/40/300/400/500 & 200/300/8/20/100/400/40/500 & 20/200/40/100/400/500/300/8 & 40/20/100/200/400/500/8/300\\
MOEA/D-NUMS & 20/8/40/100/200/300/400/500 & 40/100/8/20/200/300/400/500 & 40/20/100/8/200/400/300/500 & 20/40/100/8/200/300/400/500 & 20/40/100/8/200/300/400/500\\
\bottomrule
\end{tabular}
}
}
\end{table*}

\begin{figure*}[t]
  \centering
  \subfloat{\includegraphics[width=0.9\textwidth]{./figs/comp_mu/legend.pdf}}
  \\
  \subfloat[DTLZ1 ($m=2$)]{\includegraphics[width=0.32\textwidth]{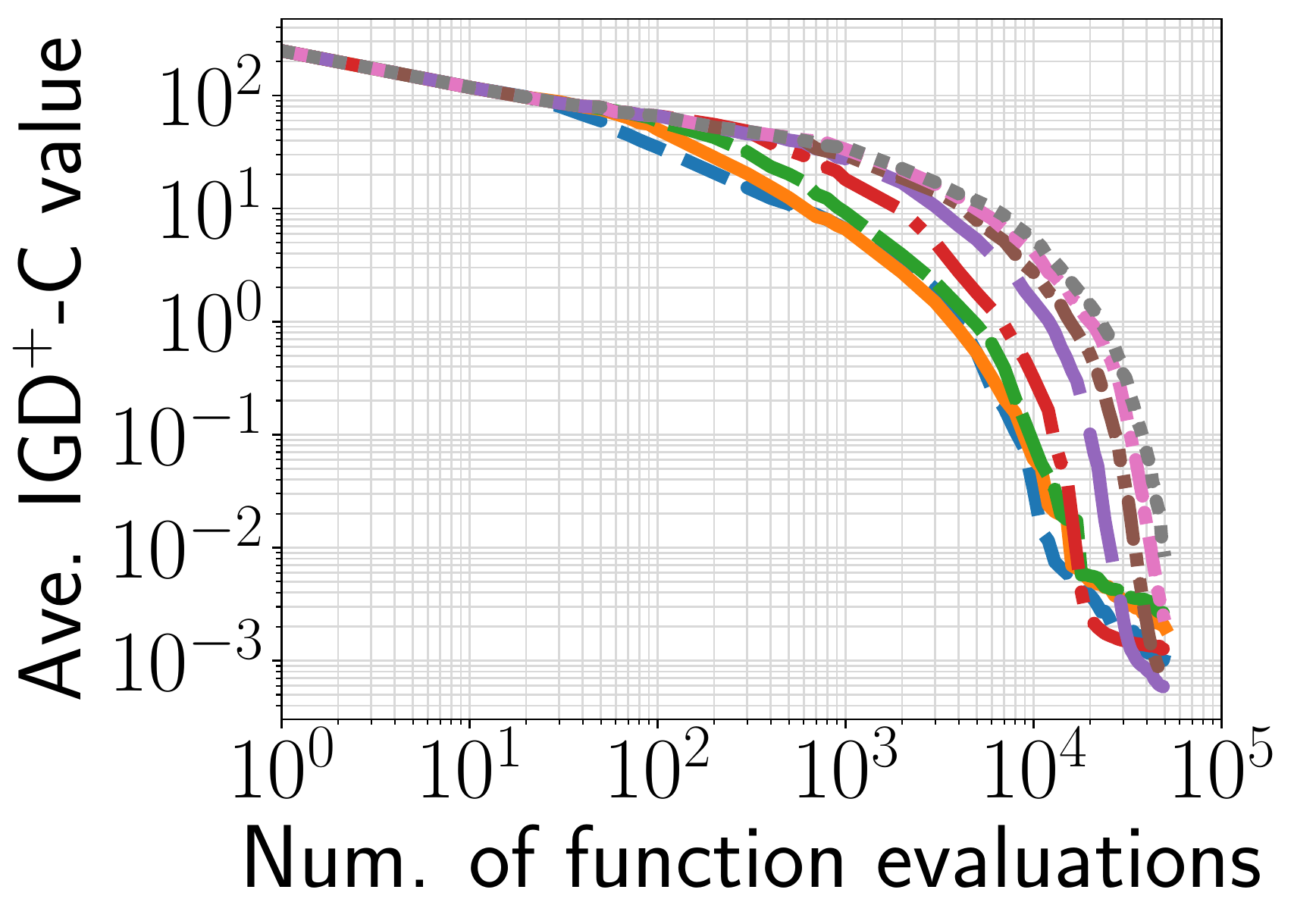}}
  \subfloat[DTLZ1 ($m=4$)]{\includegraphics[width=0.32\textwidth]{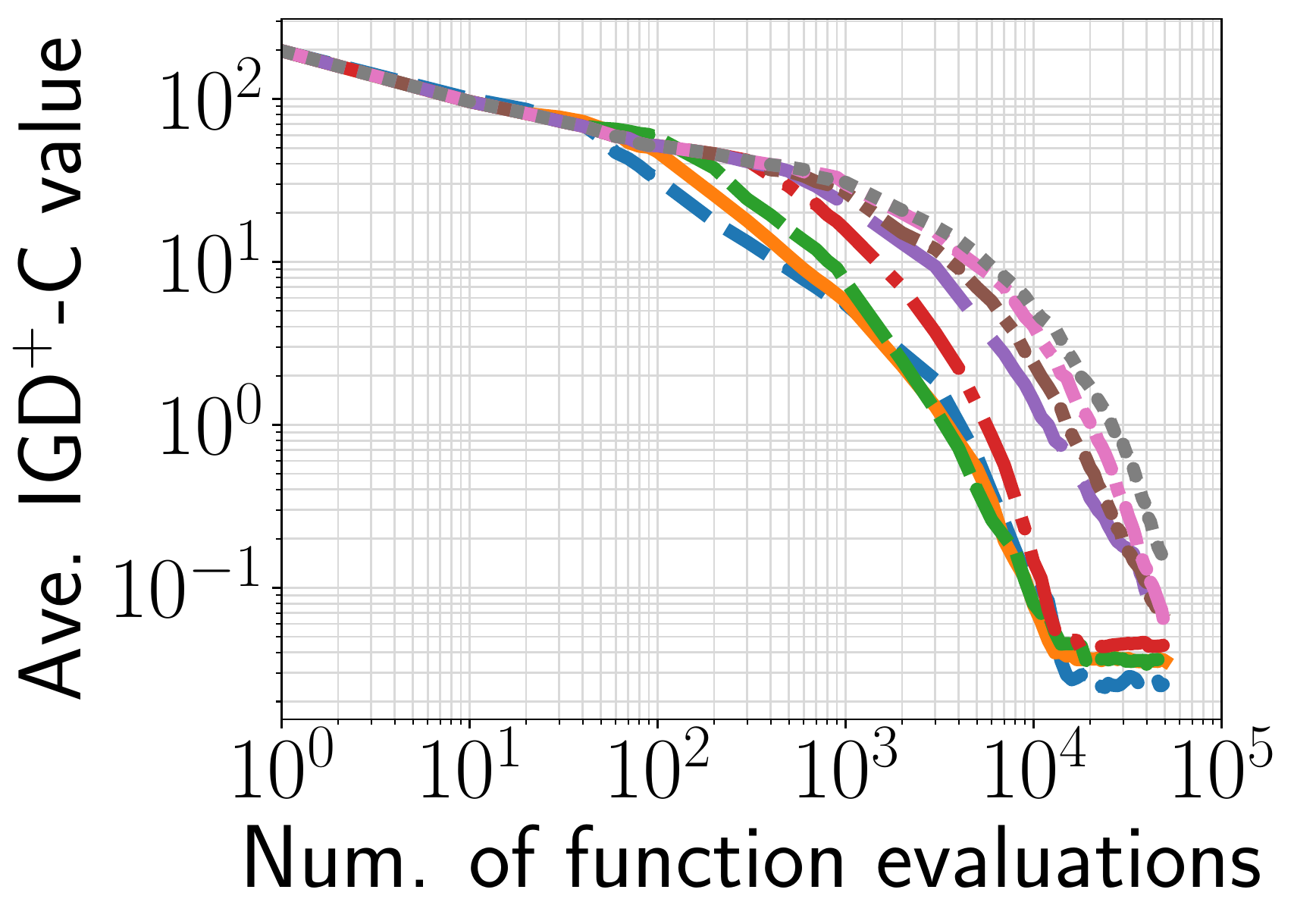}}
  \subfloat[DTLZ1 ($m=6$)]{\includegraphics[width=0.32\textwidth]{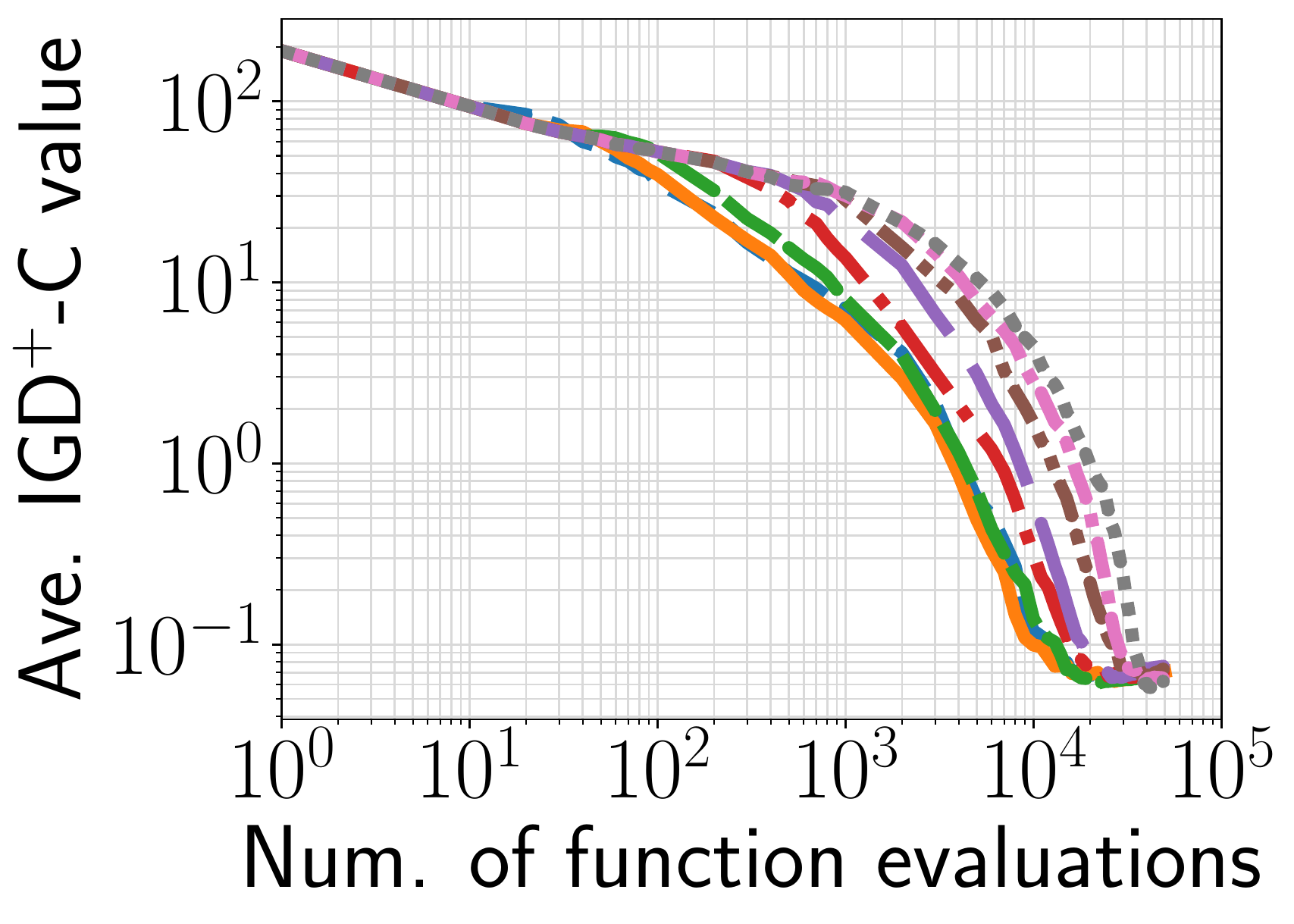}}
  \\
  \subfloat[DTLZ2 ($m=2$)]{\includegraphics[width=0.32\textwidth]{./figs/comp_mu/RNSGA2/DTLZ2_m2_z0.6_0.4.pdf}}
  \subfloat[DTLZ2 ($m=4$)]{\includegraphics[width=0.32\textwidth]{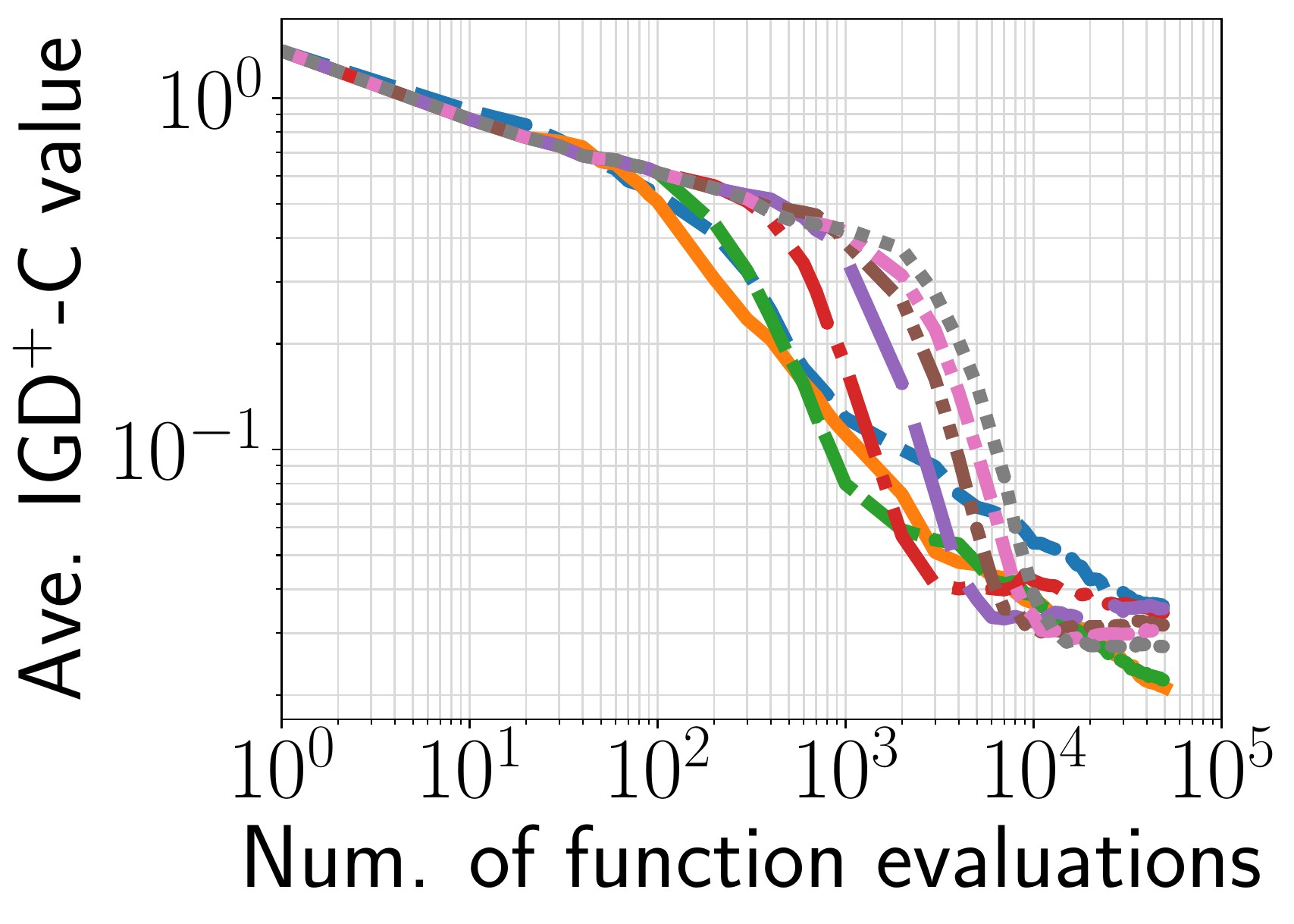}}
  \subfloat[DTLZ2 ($m=6$)]{\includegraphics[width=0.32\textwidth]{./figs/comp_mu/RNSGA2/DTLZ2_m6_z0.3_0.2_0.15_0.13_0.12_0.1.pdf}}
  \\
  \subfloat[DTLZ3 ($m=2$)]{\includegraphics[width=0.32\textwidth]{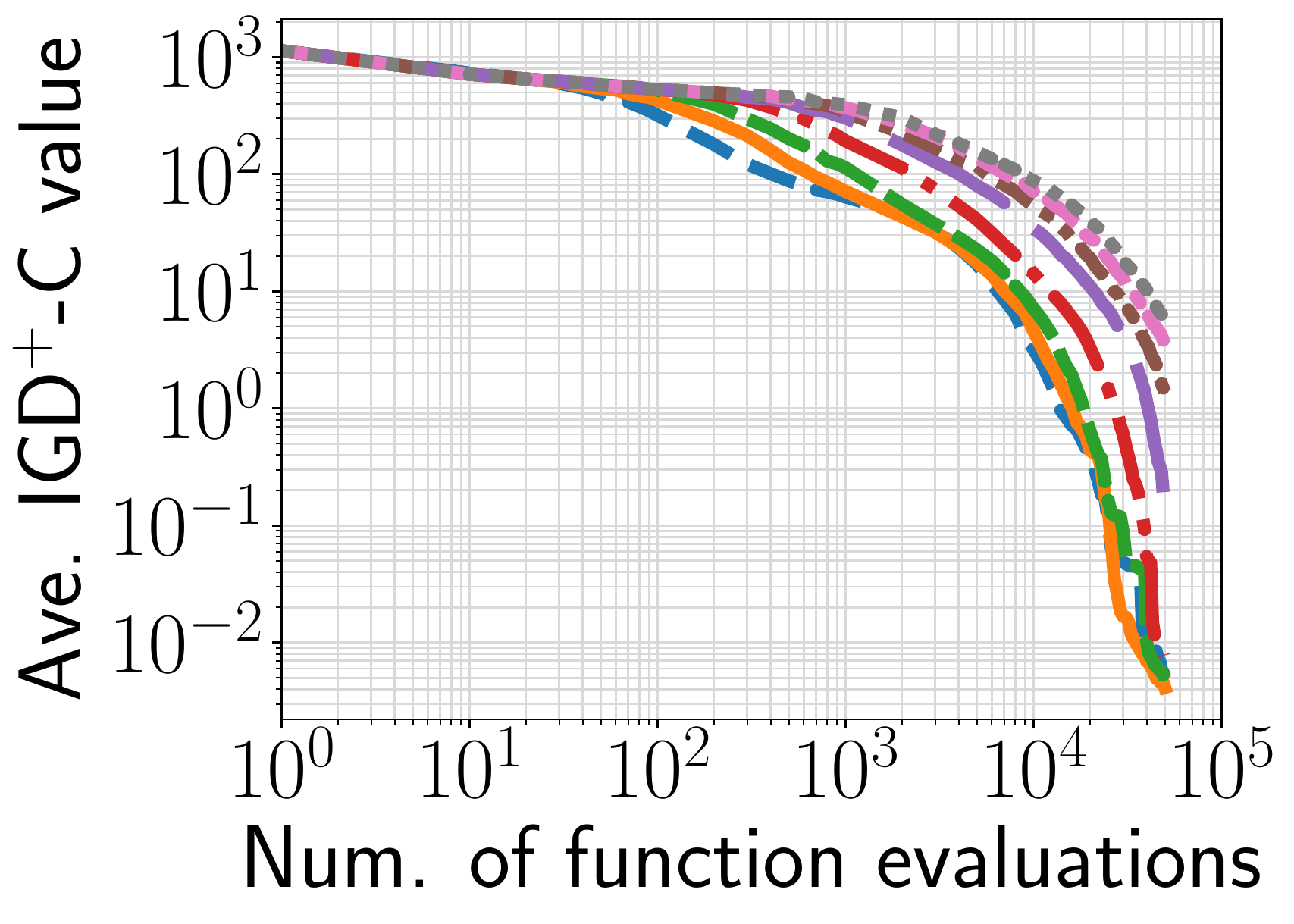}}
  \subfloat[DTLZ3 ($m=4$)]{\includegraphics[width=0.32\textwidth]{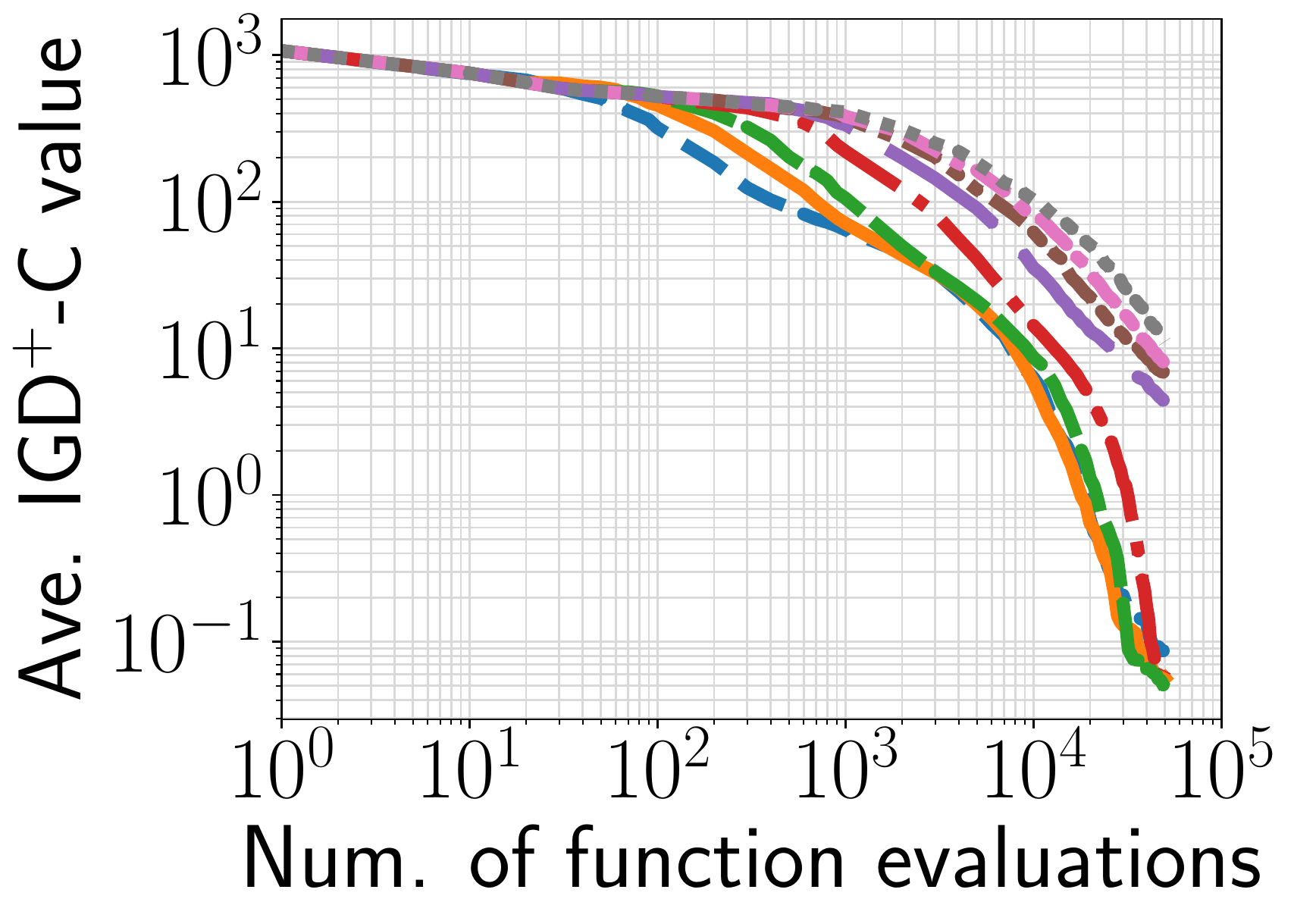}}
  \subfloat[DTLZ3 ($m=6$)]{\includegraphics[width=0.32\textwidth]{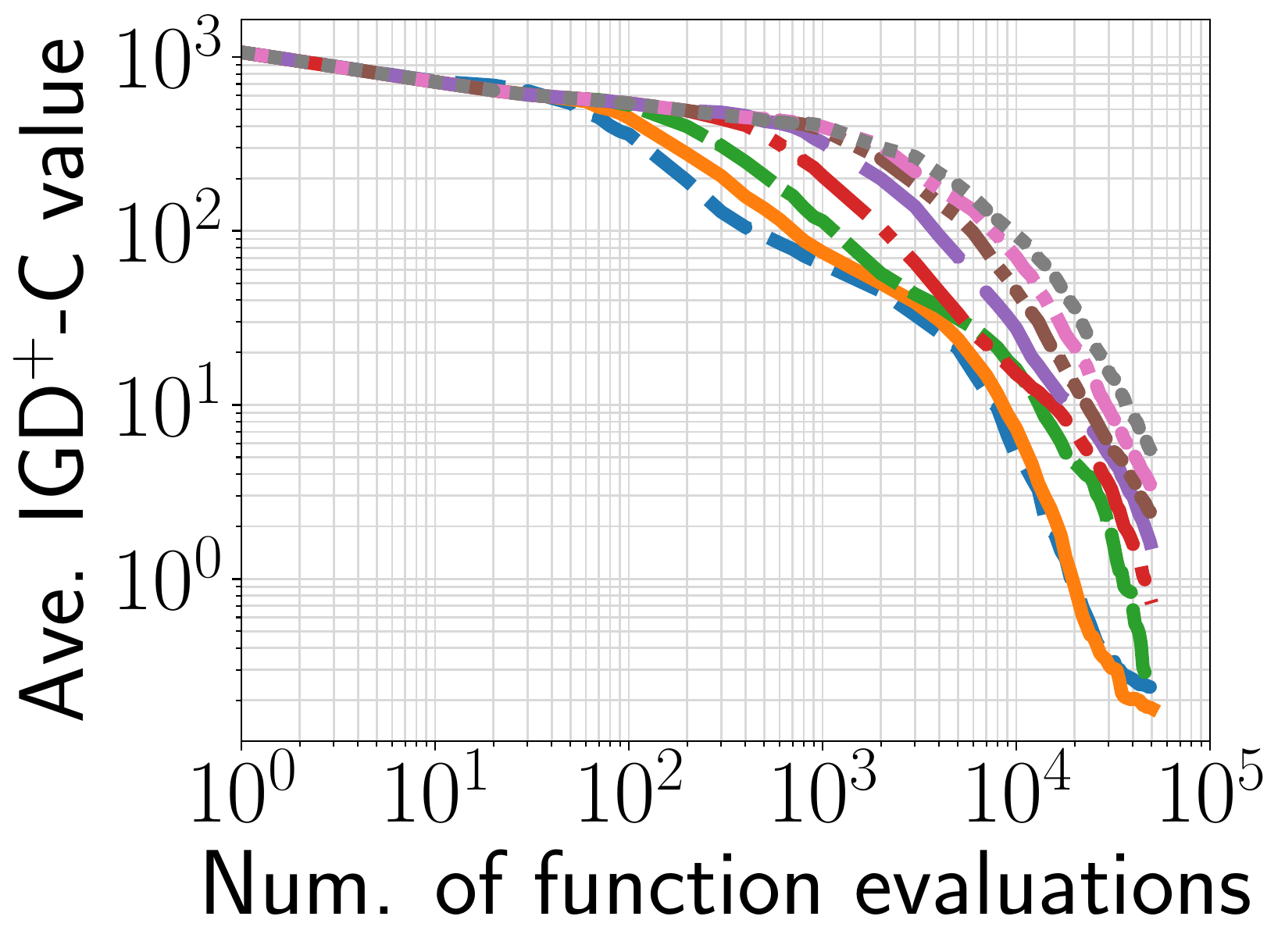}}
  \\
  \subfloat[DTLZ4 ($m=2$)]{\includegraphics[width=0.32\textwidth]{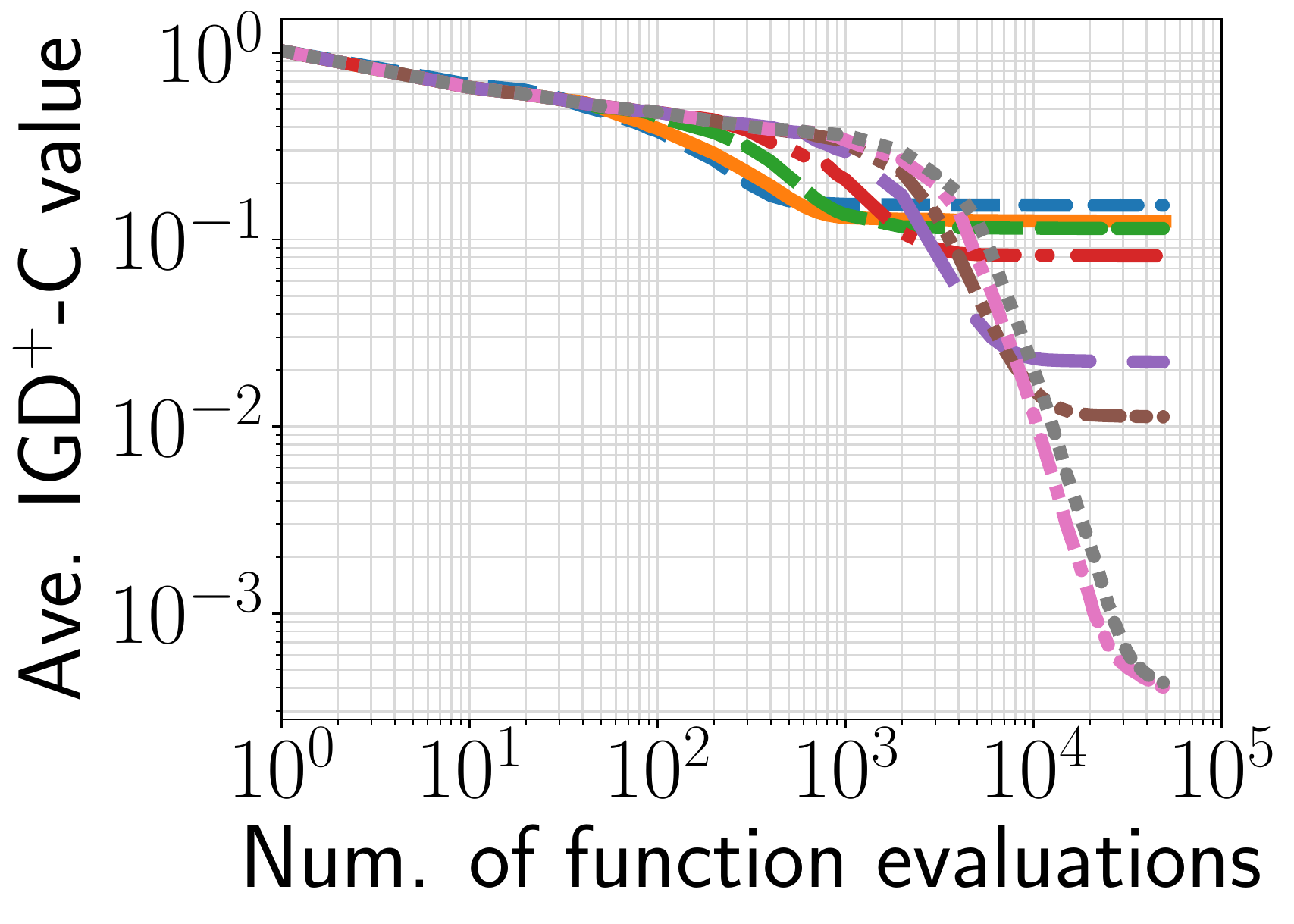}}
  \subfloat[DTLZ4 ($m=4$)]{\includegraphics[width=0.32\textwidth]{./figs/comp_mu/RNSGA2/DTLZ4_m4_z0.4_0.3_0.2_0.1.pdf}}
  \subfloat[DTLZ4 ($m=6$)]{\includegraphics[width=0.32\textwidth]{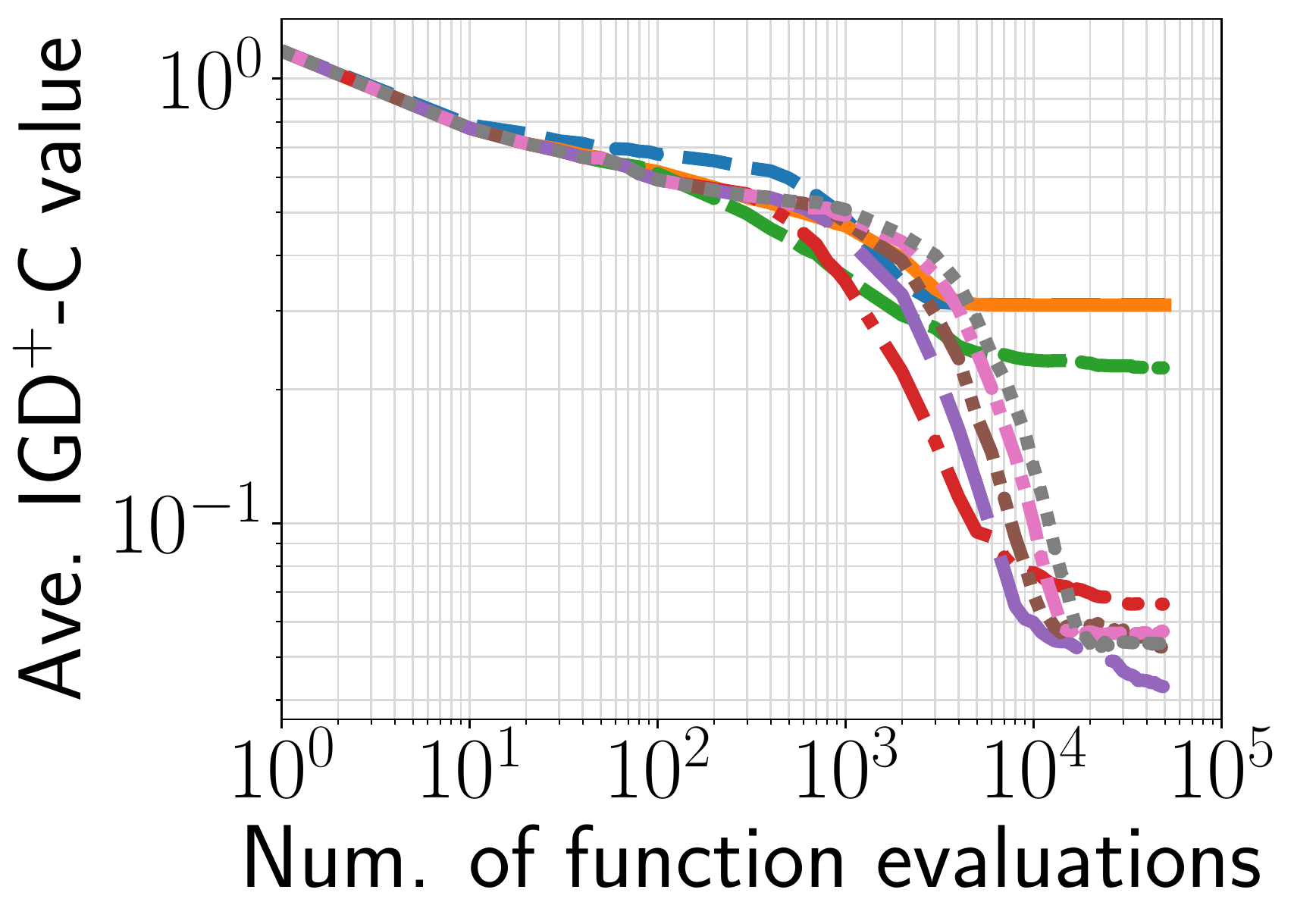}}
      \caption{Average IGD$^+$-C values of R-NSGA-II with different population sizes on the DTLZ1--DTLZ4 problems with $m \in \{2, 4, 6\}$.}
   \label{fig:sup_Rnsgaii_dtlz}
\end{figure*}

\begin{figure*}[t]
  \centering
  \subfloat{\includegraphics[width=0.9\textwidth]{./figs/comp_mu/legend.pdf}}
  \\
  \subfloat[DTLZ1 ($m=2$)]{\includegraphics[width=0.32\textwidth]{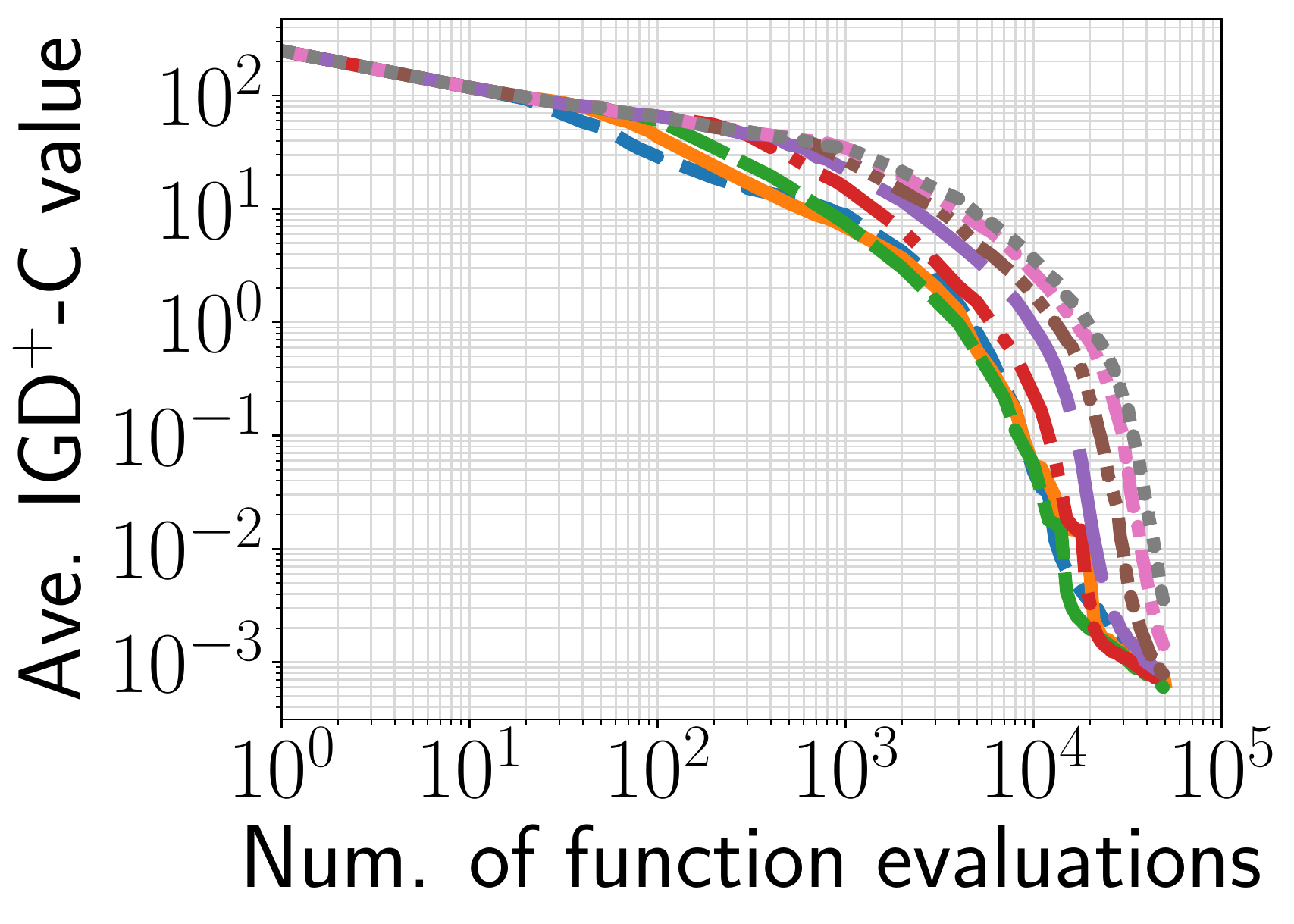}}
  \subfloat[DTLZ1 ($m=4$)]{\includegraphics[width=0.32\textwidth]{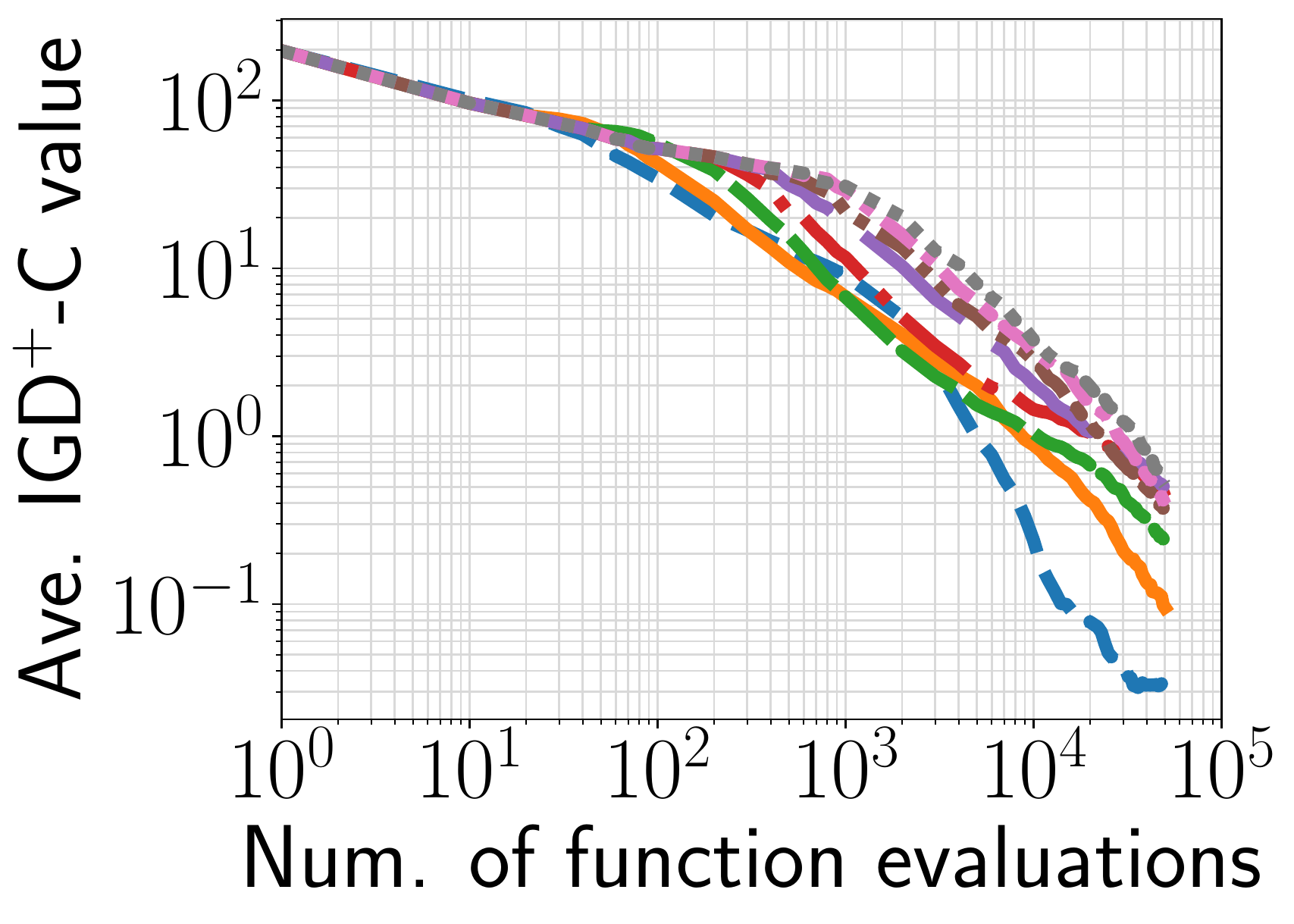}}
  \subfloat[DTLZ1 ($m=6$)]{\includegraphics[width=0.32\textwidth]{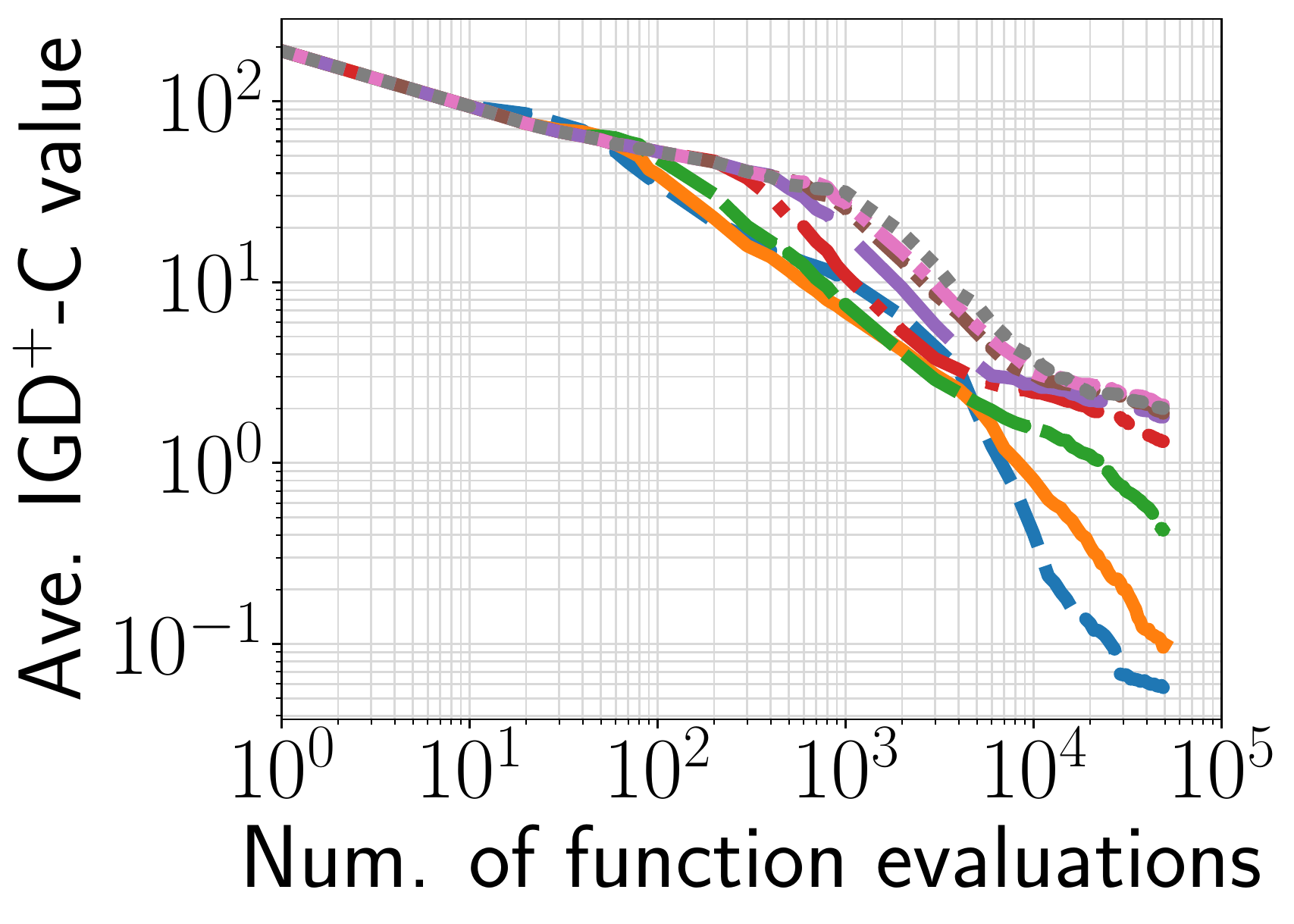}}
  \\
  \subfloat[DTLZ2 ($m=2$)]{\includegraphics[width=0.32\textwidth]{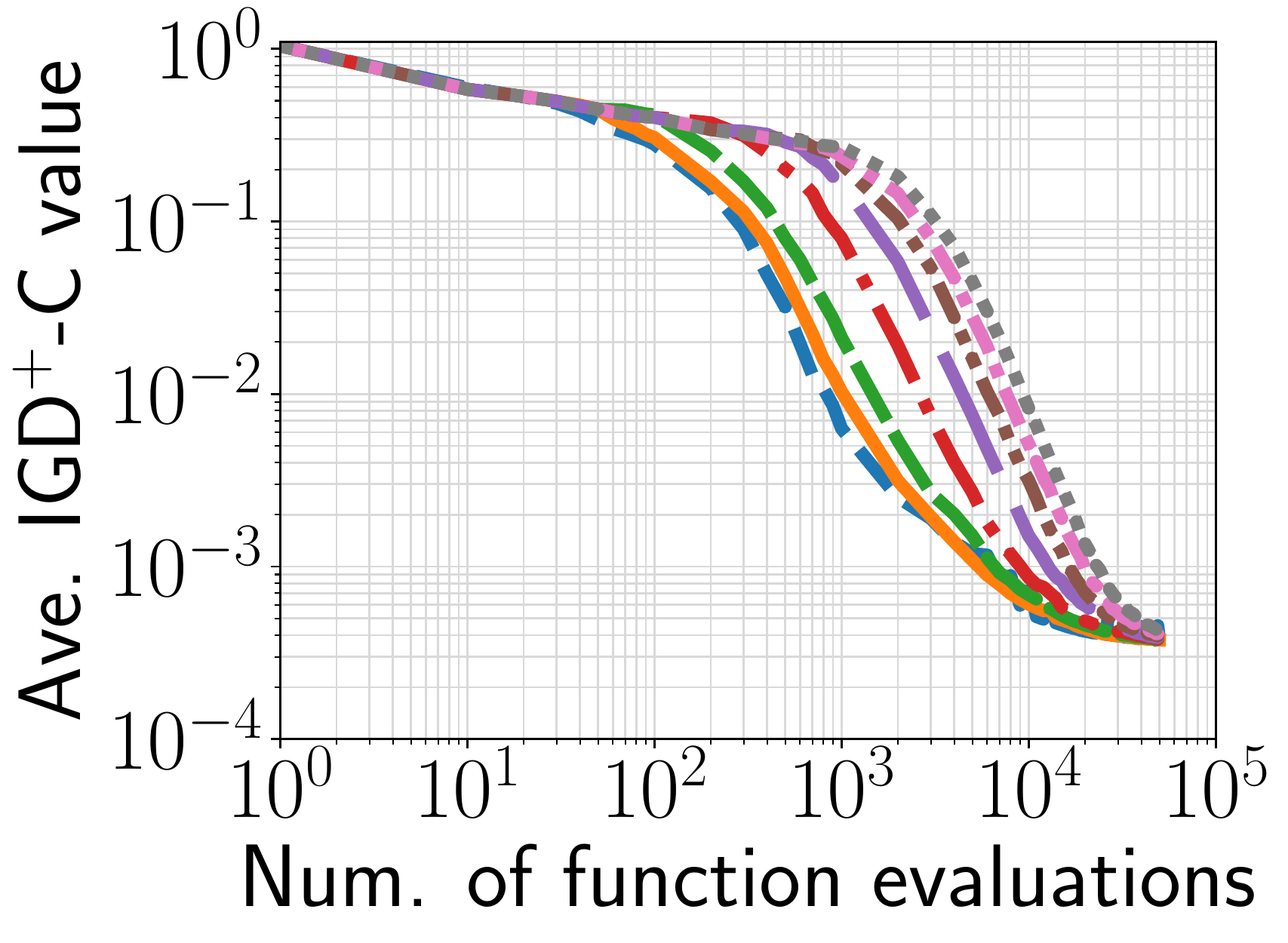}}
  \subfloat[DTLZ2 ($m=4$)]{\includegraphics[width=0.32\textwidth]{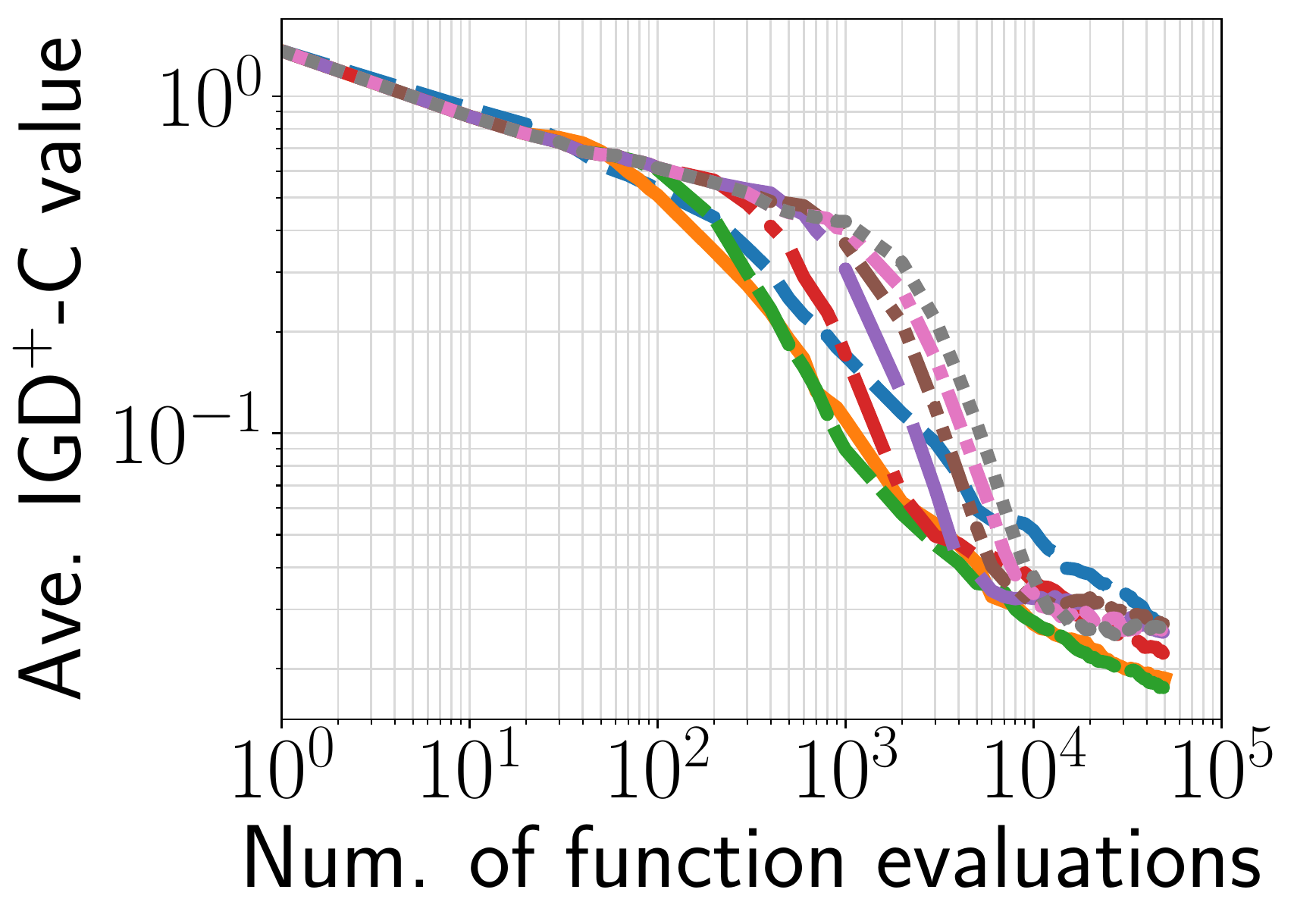}}
  \subfloat[DTLZ2 ($m=6$)]{\includegraphics[width=0.32\textwidth]{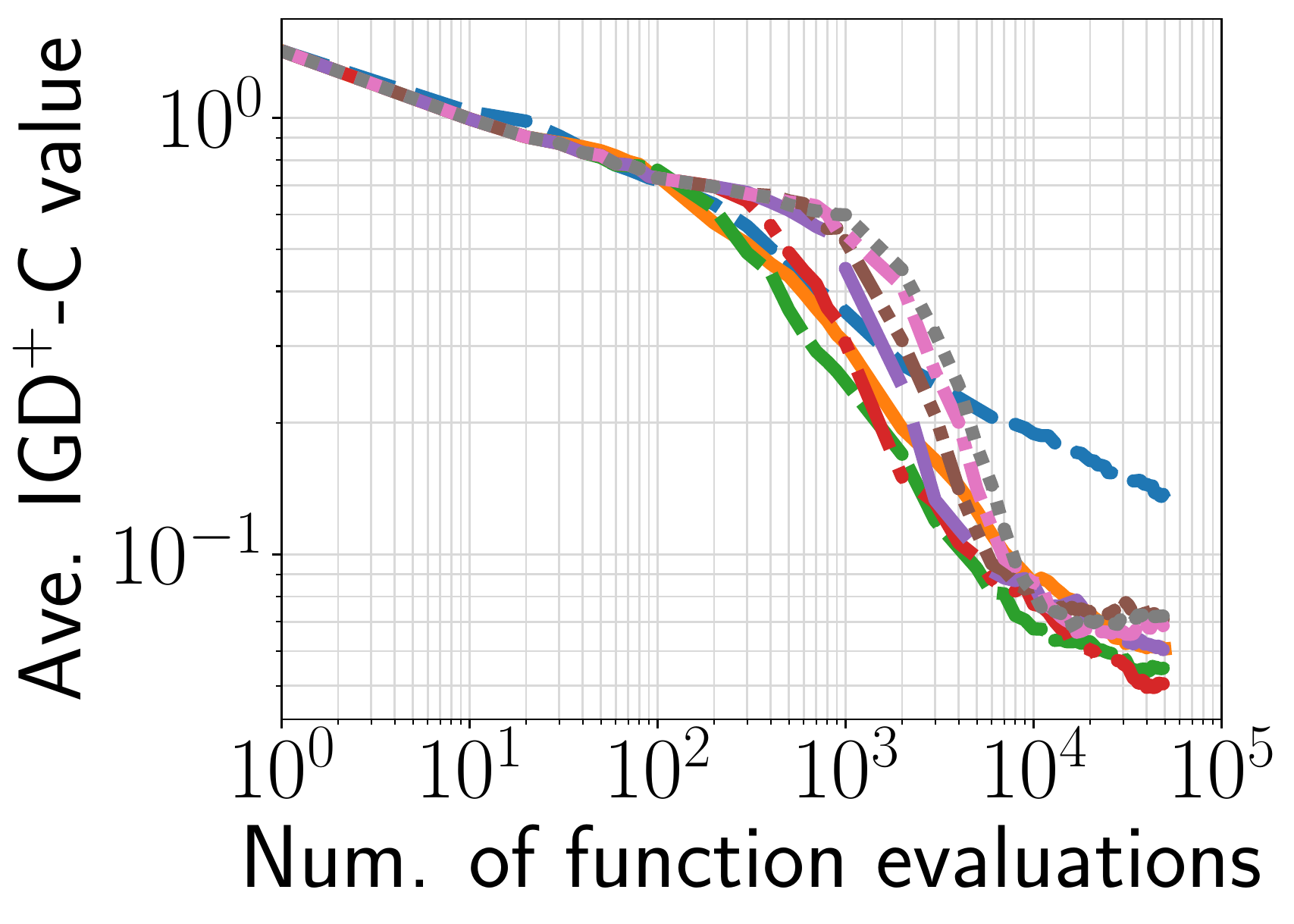}}
  \\
  \subfloat[DTLZ3 ($m=2$)]{\includegraphics[width=0.32\textwidth]{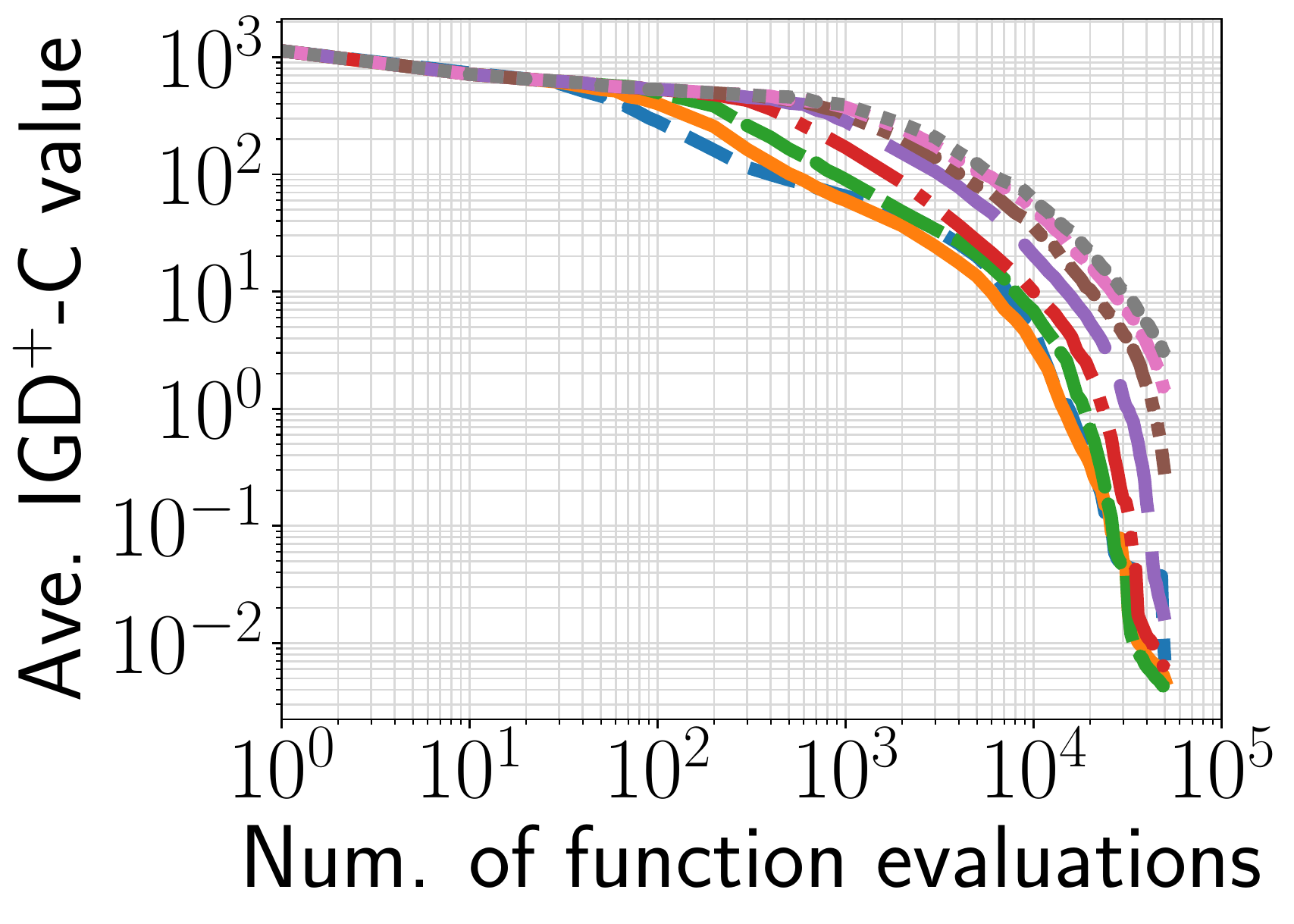}}
  \subfloat[DTLZ3 ($m=4$)]{\includegraphics[width=0.32\textwidth]{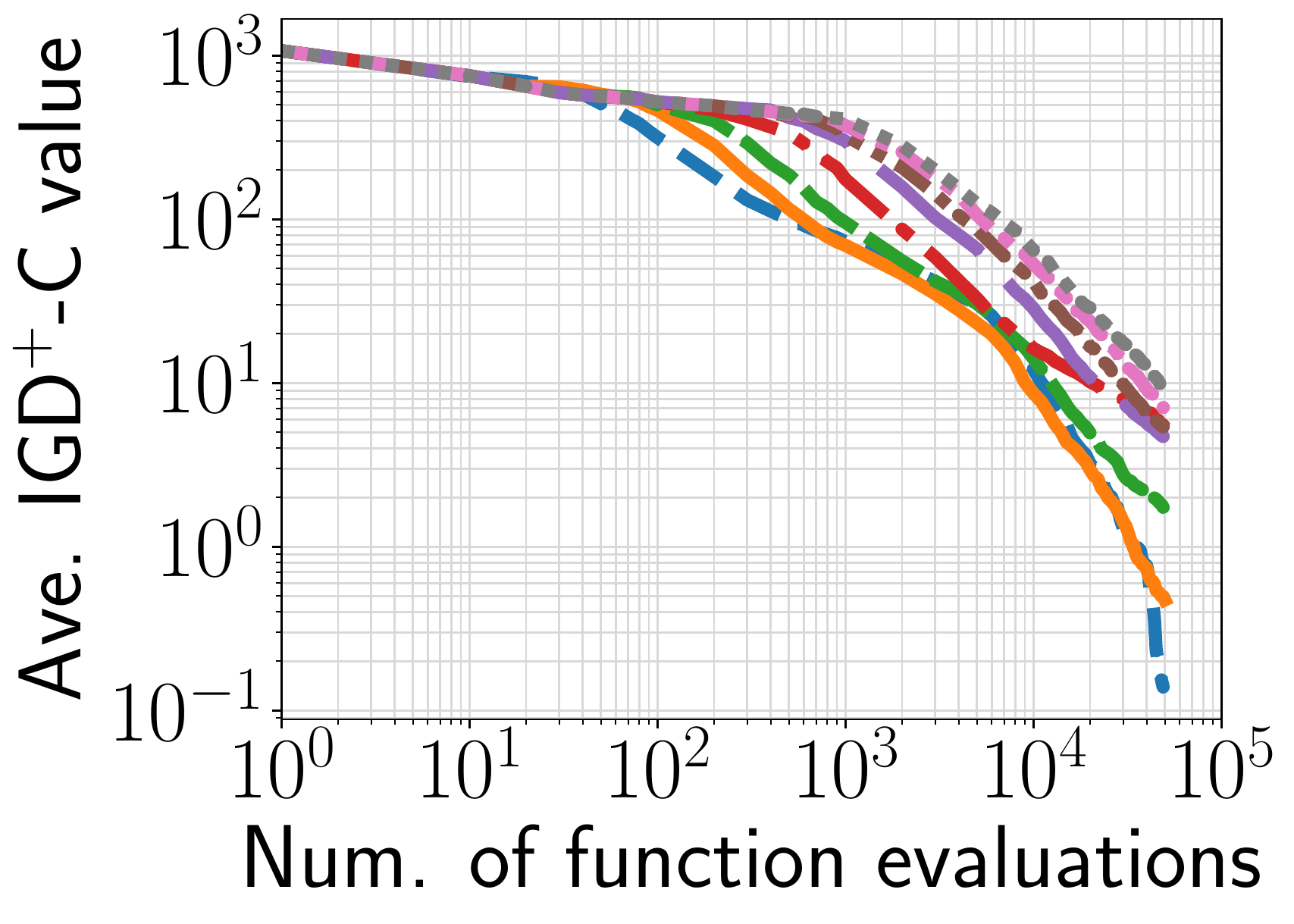}}
  \subfloat[DTLZ3 ($m=6$)]{\includegraphics[width=0.32\textwidth]{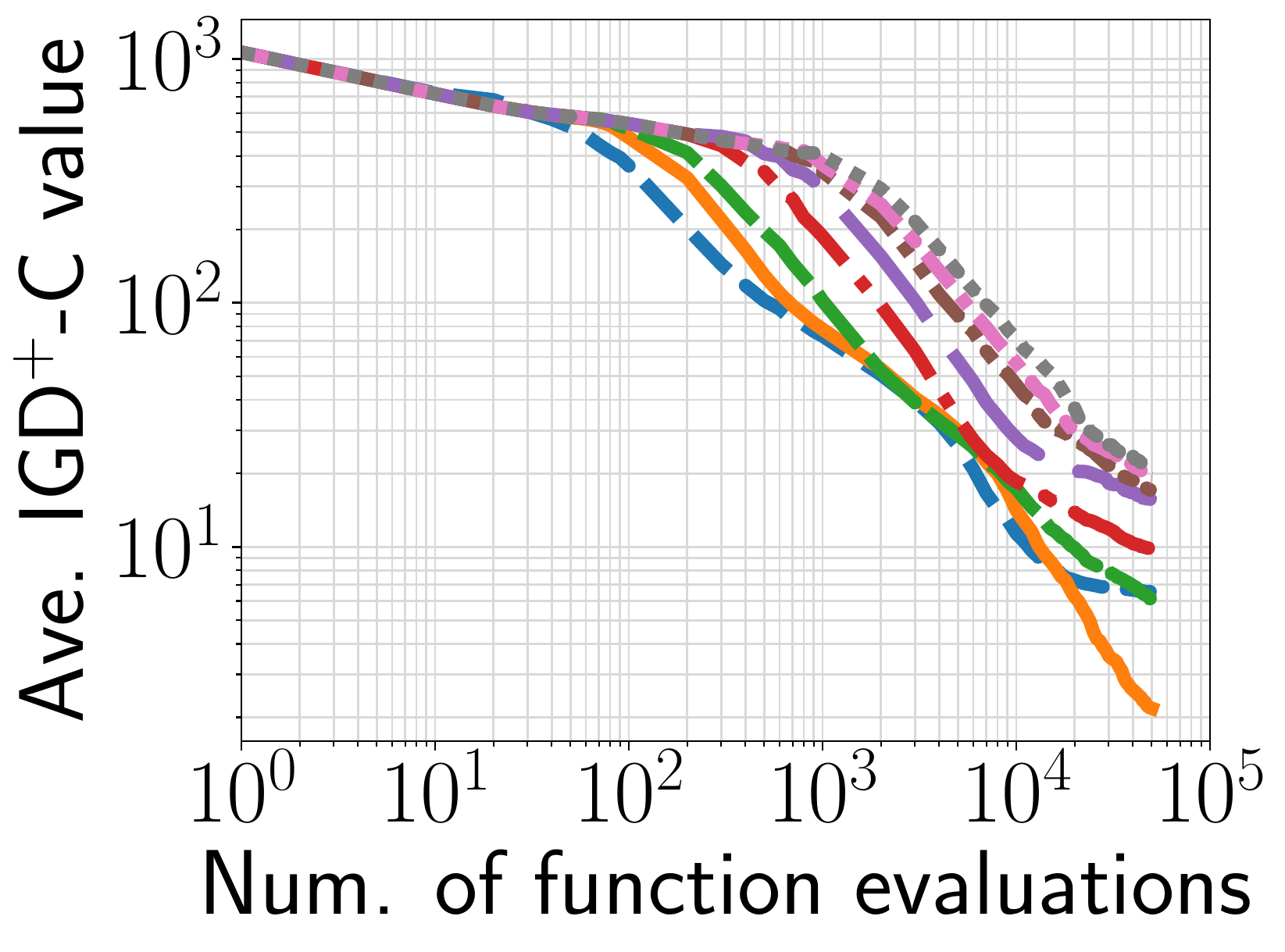}}
  \\
  \subfloat[DTLZ4 ($m=2$)]{\includegraphics[width=0.32\textwidth]{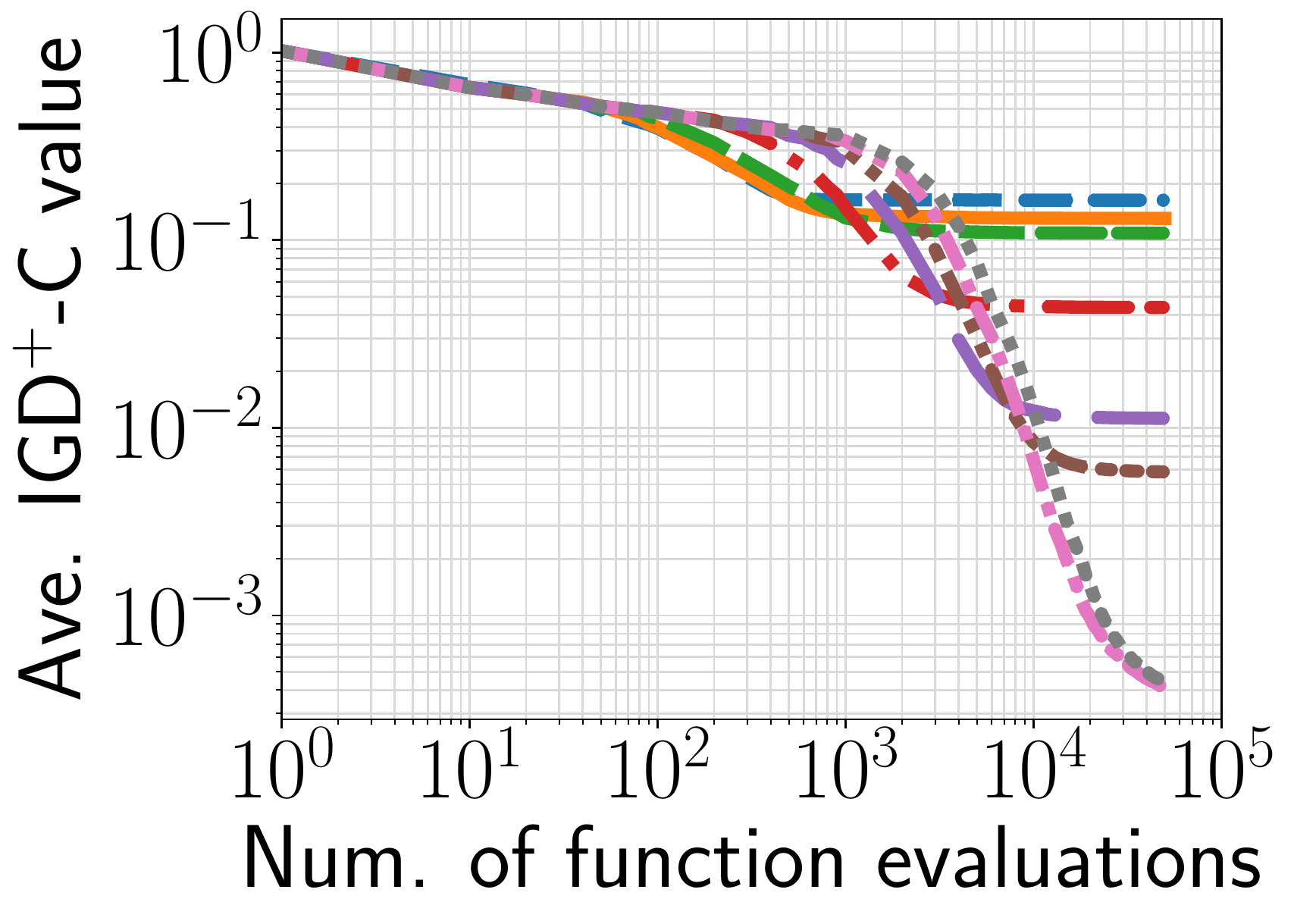}}
  \subfloat[DTLZ4 ($m=4$)]{\includegraphics[width=0.32\textwidth]{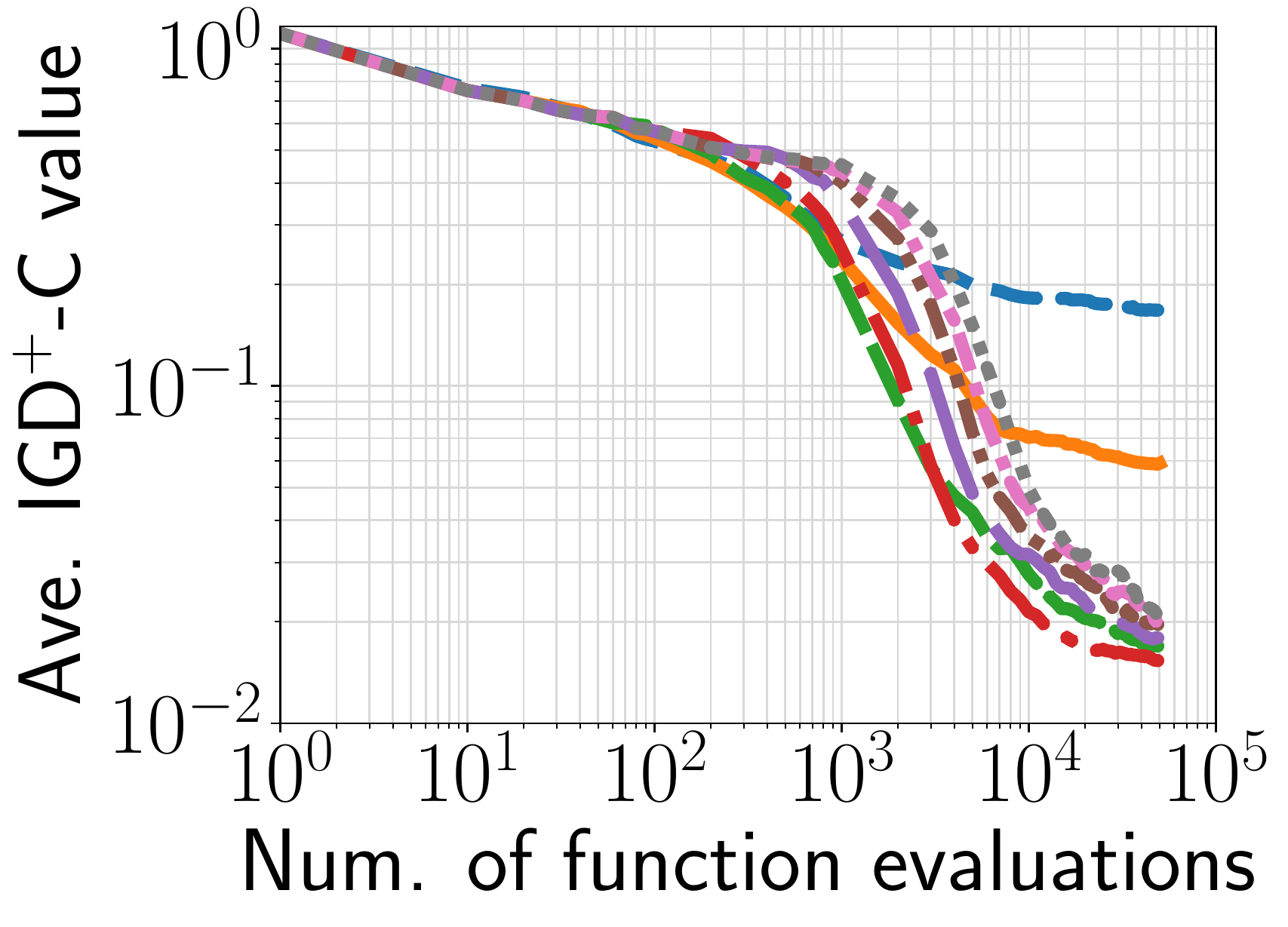}}
  \subfloat[DTLZ4 ($m=6$)]{\includegraphics[width=0.32\textwidth]{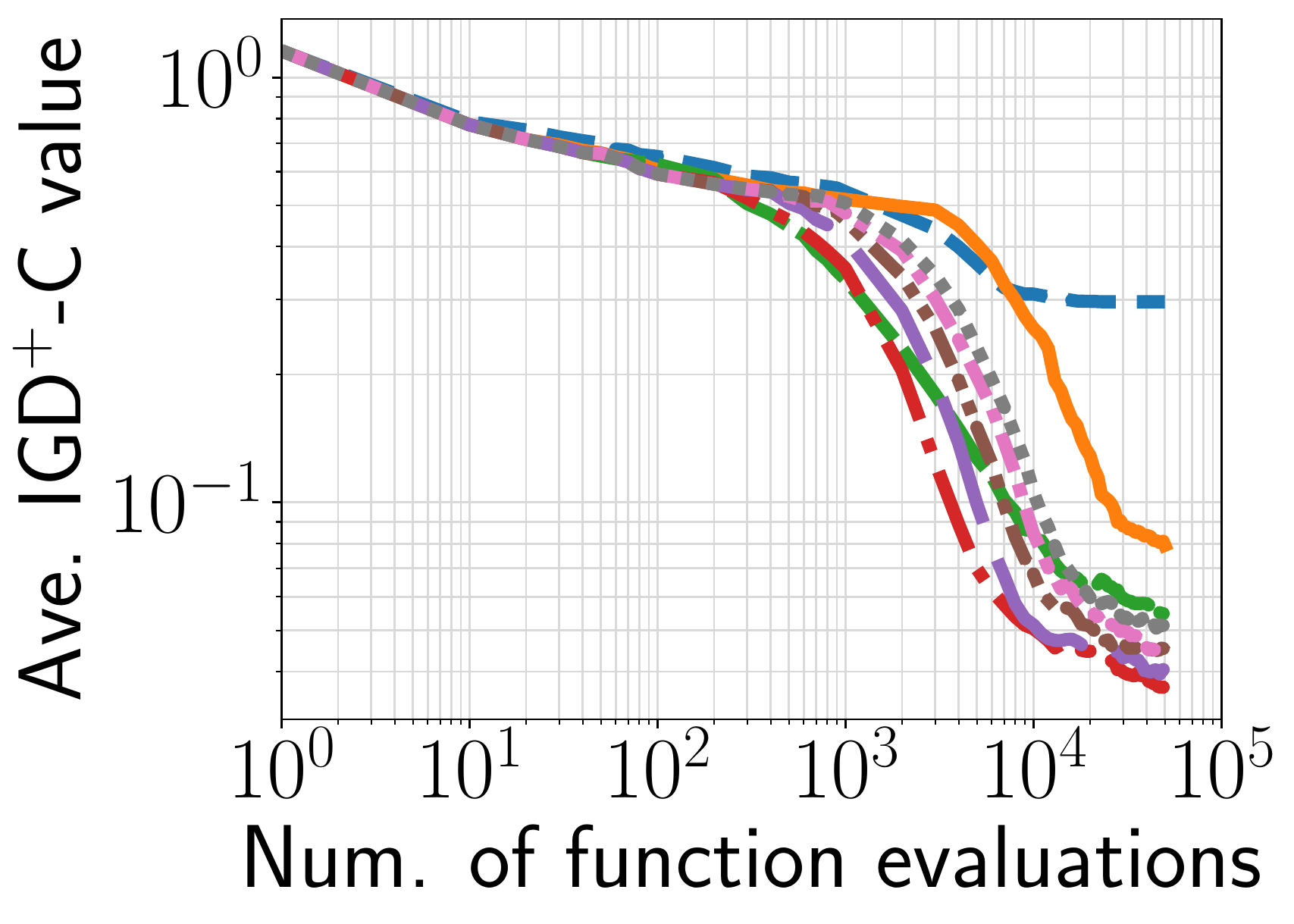}}
      \caption{Average IGD$^+$-C values of r-NSGA-II with different population sizes on the DTLZ1--DTLZ4 problems with $m \in \{2, 4, 6\}$.}
   \label{fig:sup_rnsgaii_dtlz}
\end{figure*}

\begin{figure*}[t]
  \centering
  \subfloat{\includegraphics[width=0.9\textwidth]{./figs/comp_mu/legend.pdf}}
  \\
  \subfloat[DTLZ1 ($m=2$)]{\includegraphics[width=0.32\textwidth]{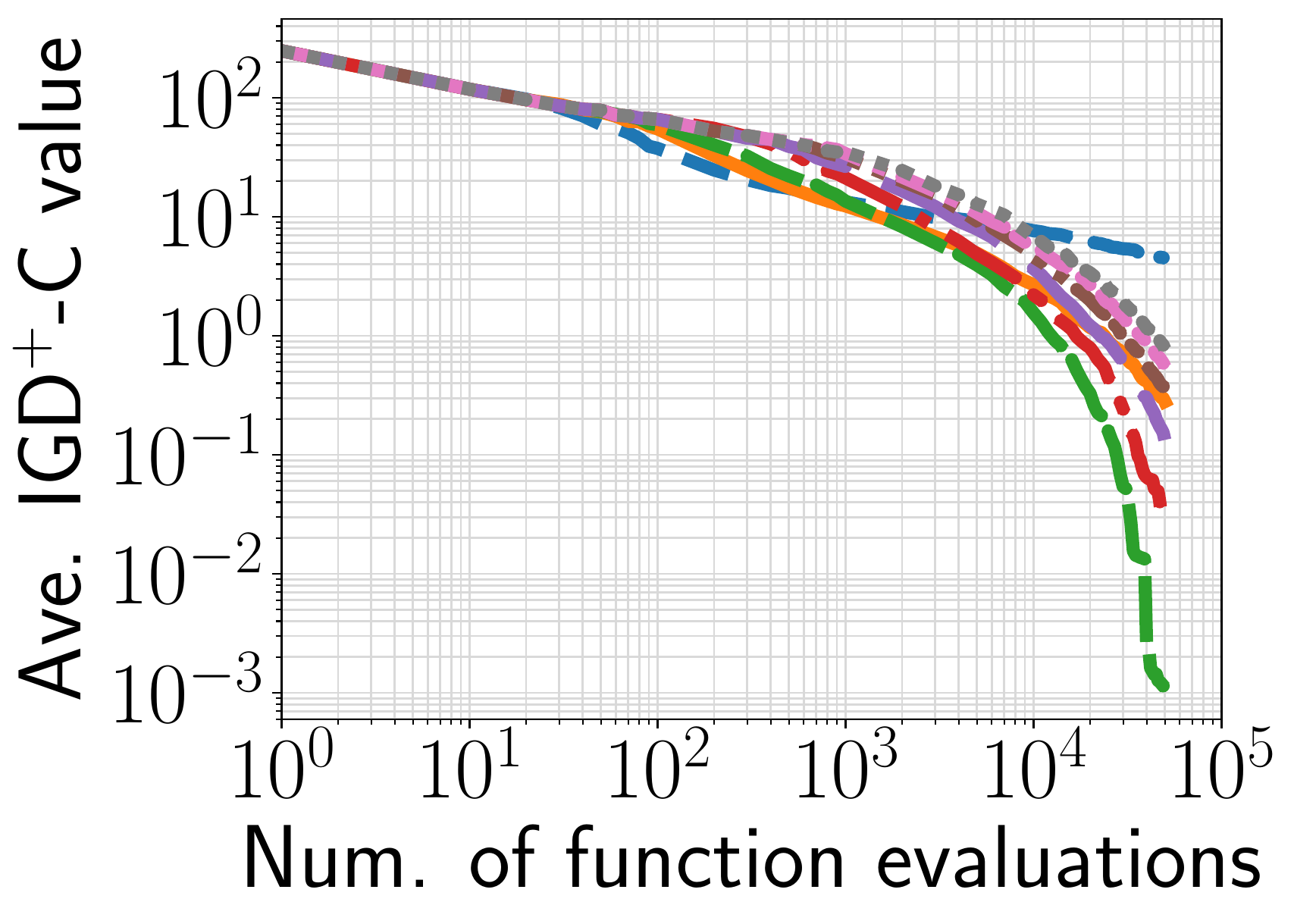}}
  \subfloat[DTLZ1 ($m=4$)]{\includegraphics[width=0.32\textwidth]{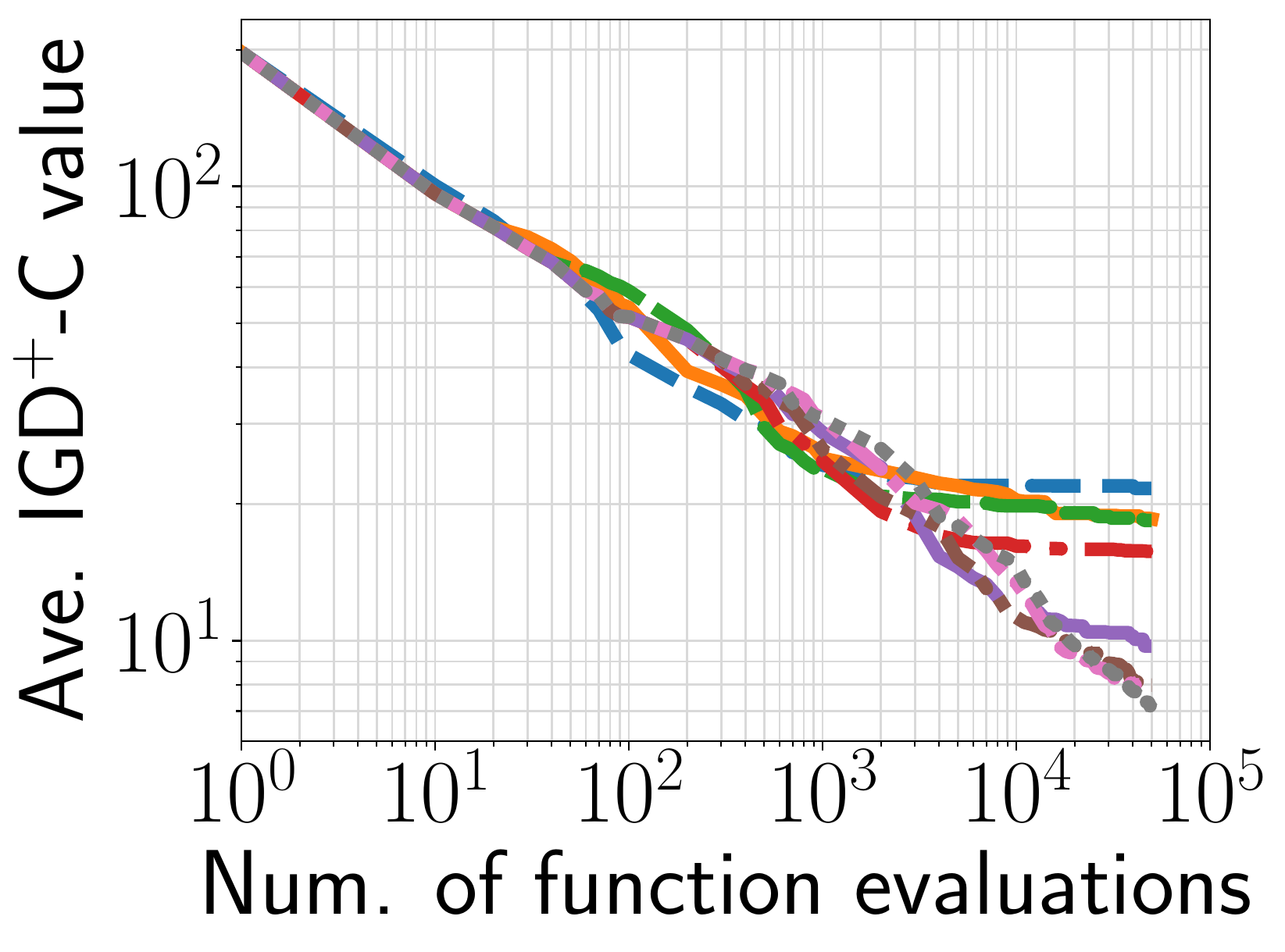}}
  \subfloat[DTLZ1 ($m=6$)]{\includegraphics[width=0.32\textwidth]{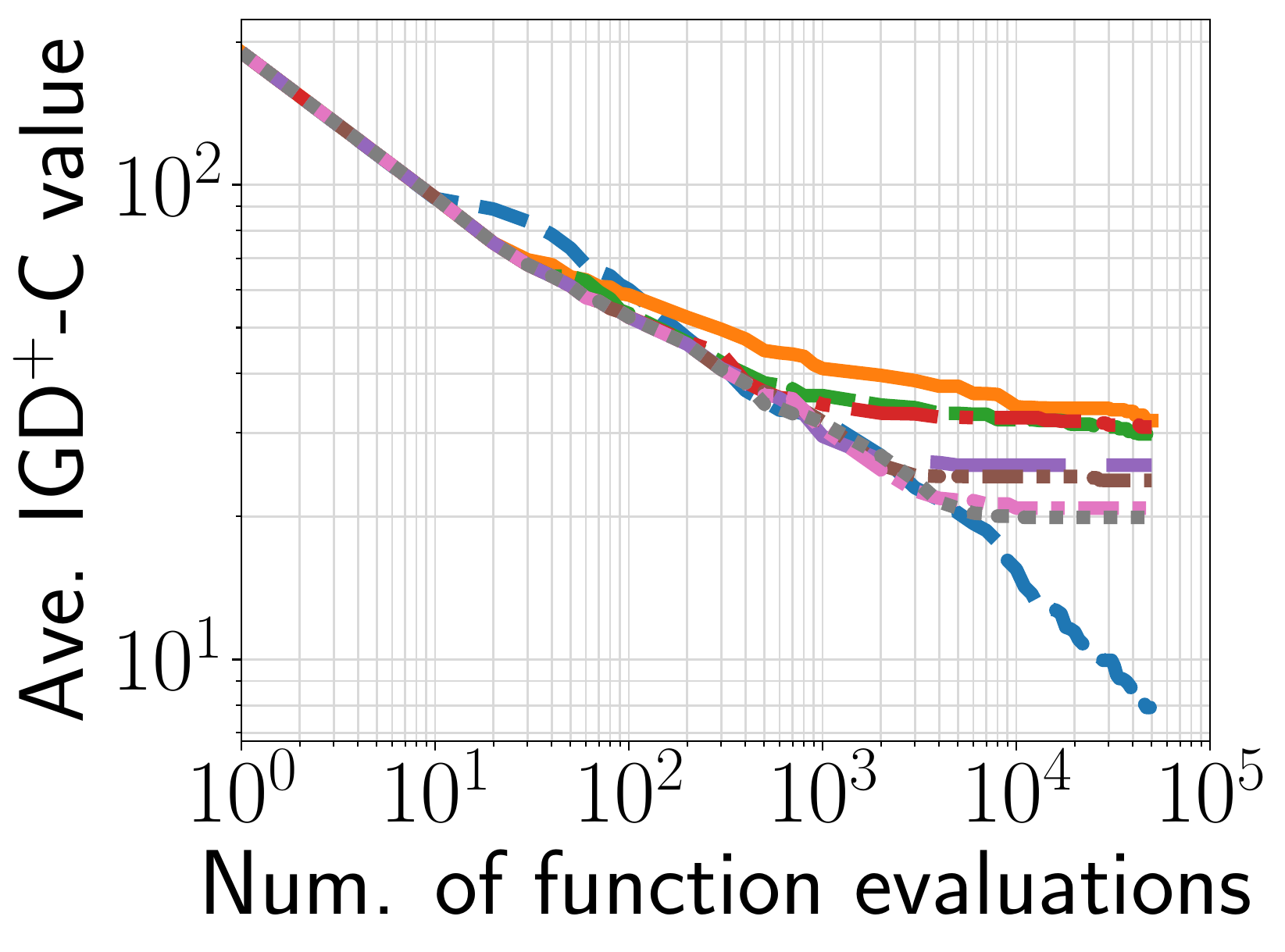}}
  \\
  \subfloat[DTLZ2 ($m=2$)]{\includegraphics[width=0.32\textwidth]{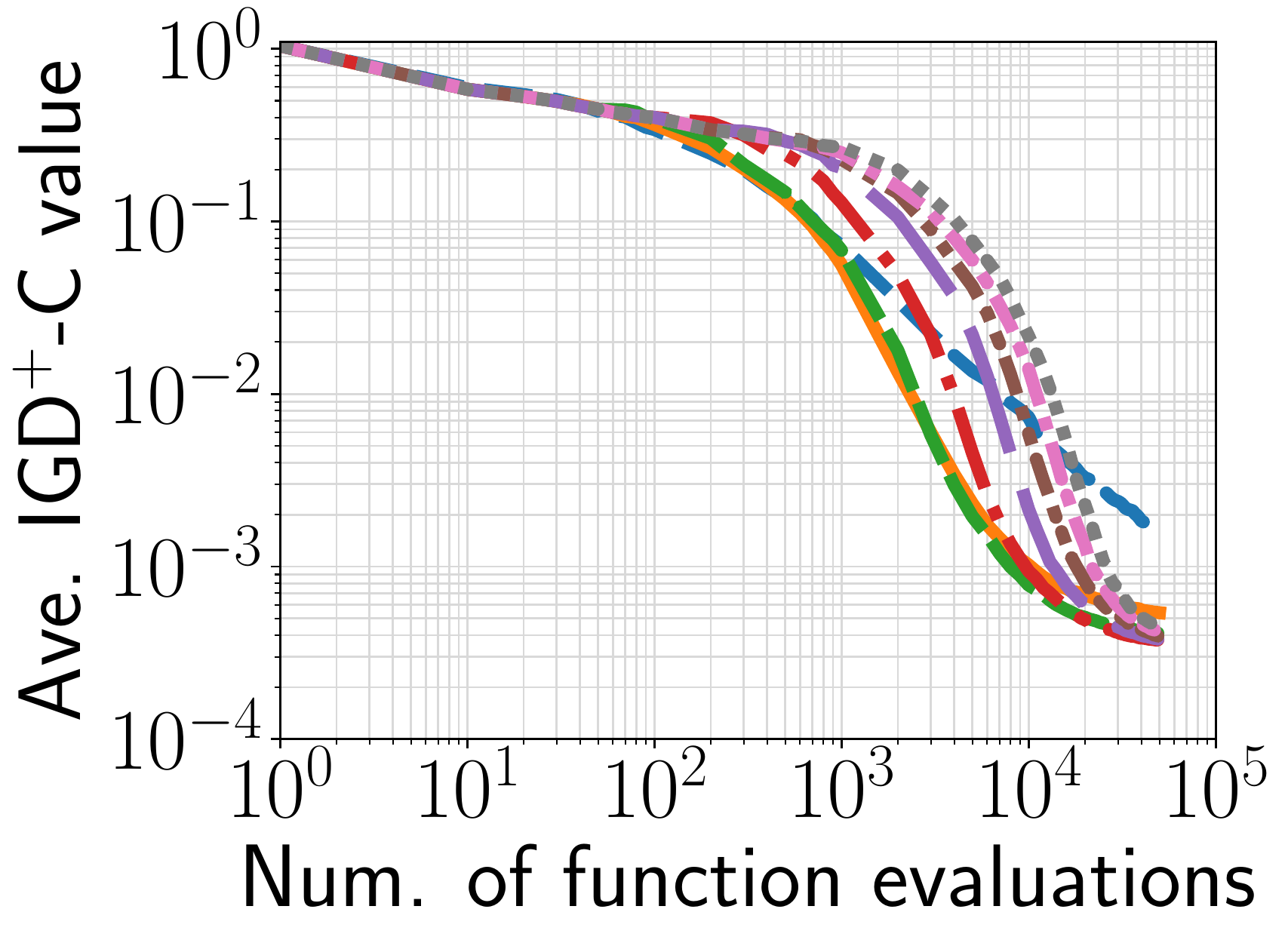}}
  \subfloat[DTLZ2 ($m=4$)]{\includegraphics[width=0.32\textwidth]{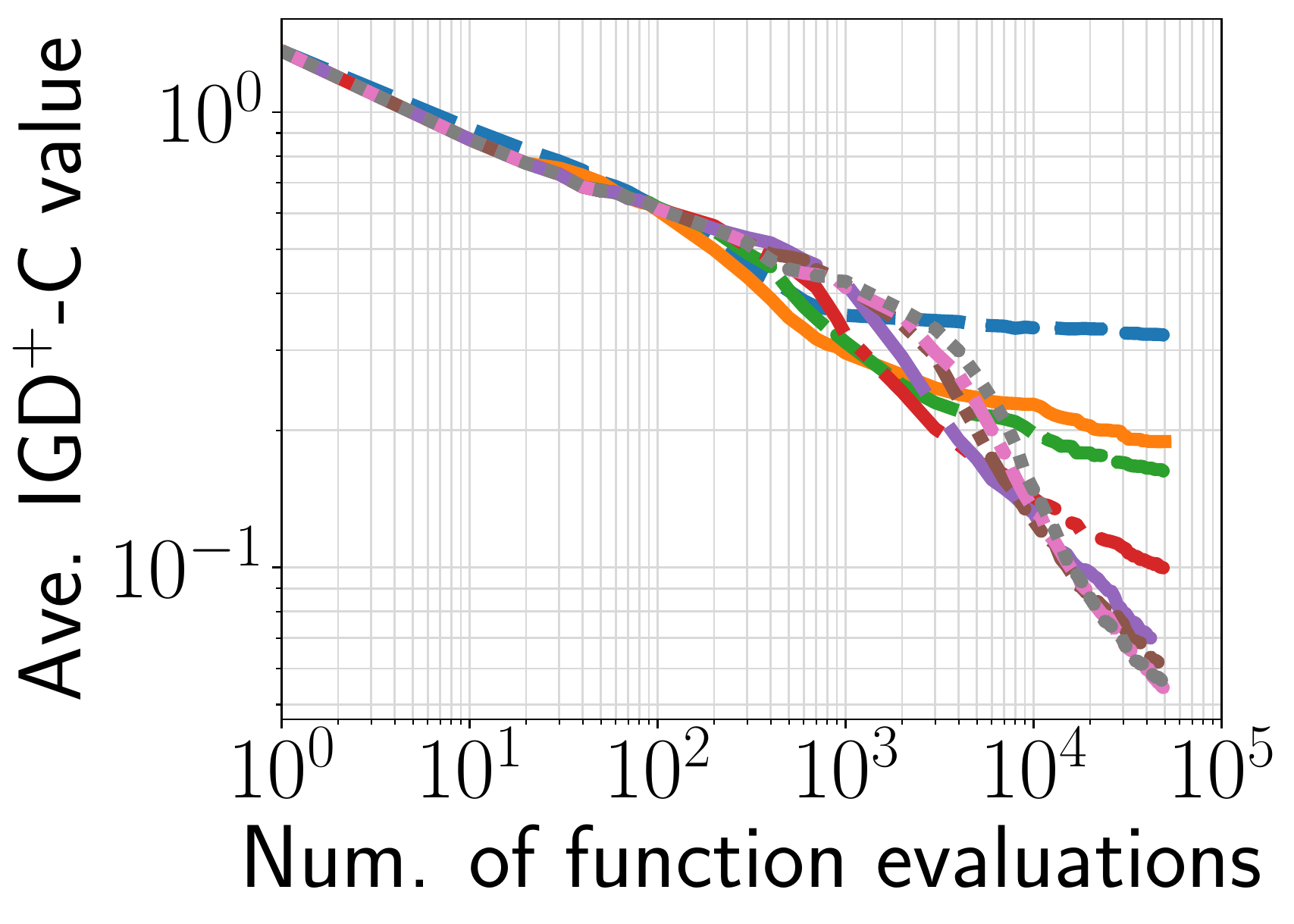}}
  \subfloat[DTLZ2 ($m=6$)]{\includegraphics[width=0.32\textwidth]{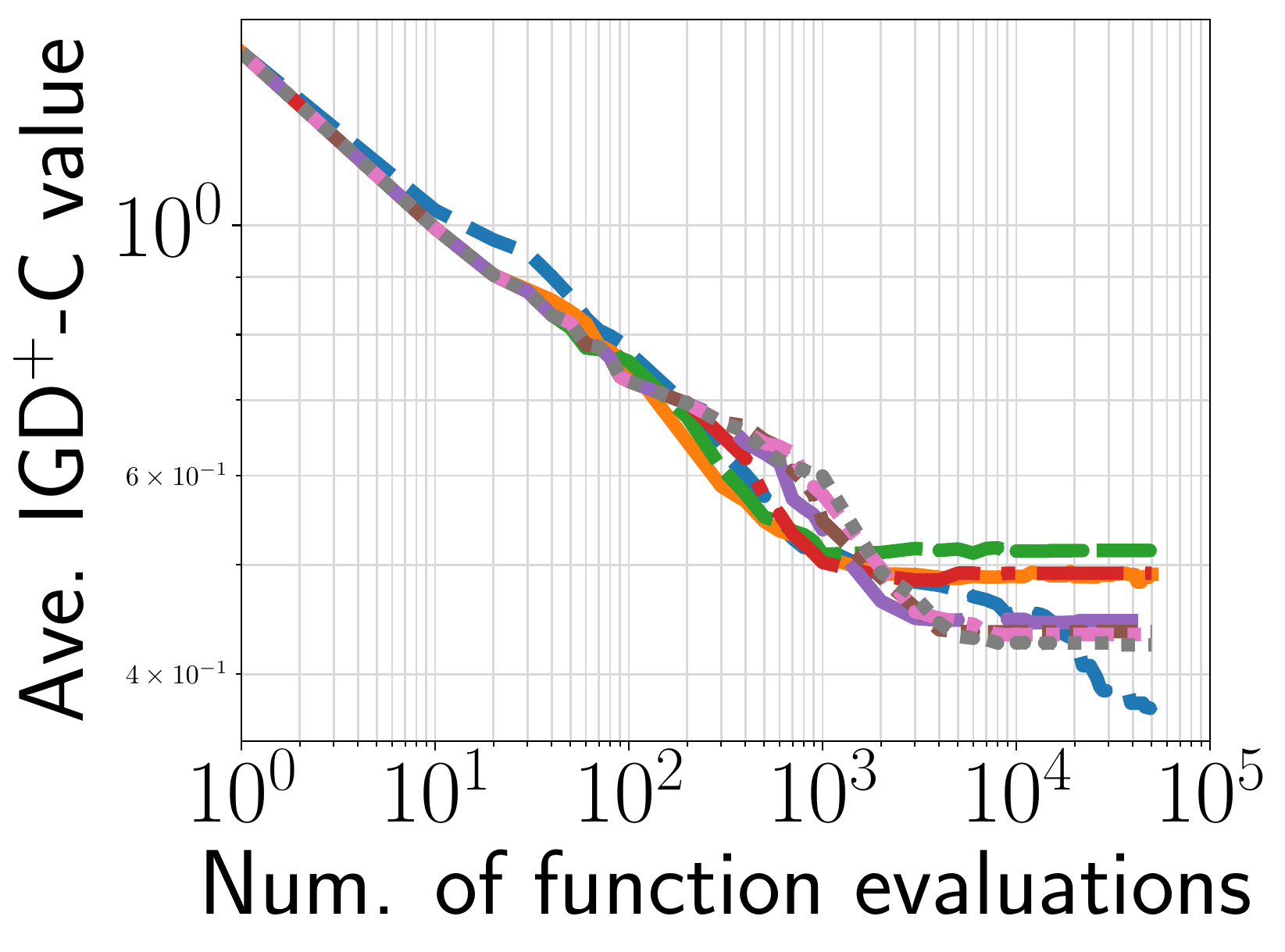}}
  \\
  \subfloat[DTLZ3 ($m=2$)]{\includegraphics[width=0.32\textwidth]{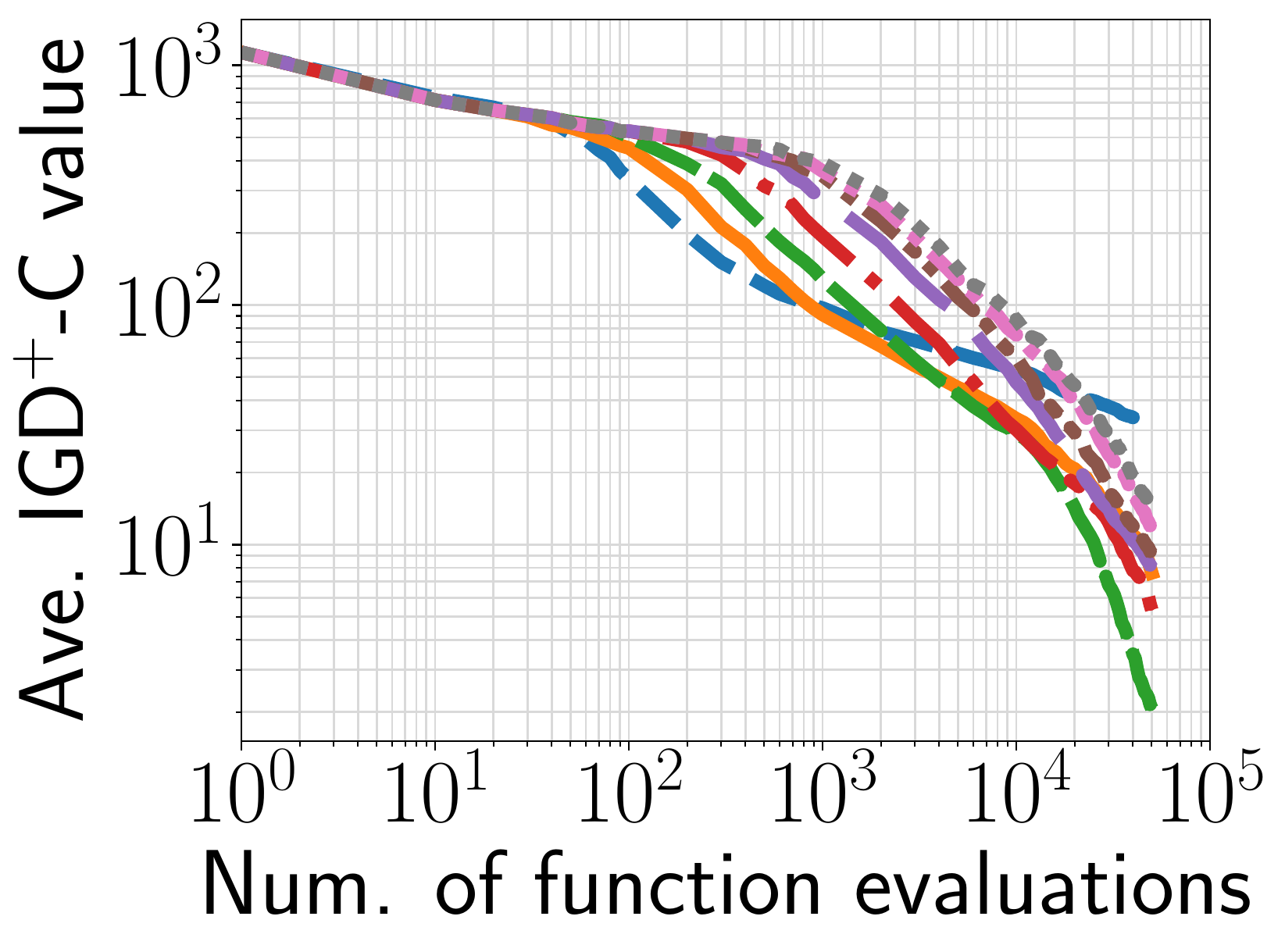}}
  \subfloat[DTLZ3 ($m=4$)]{\includegraphics[width=0.32\textwidth]{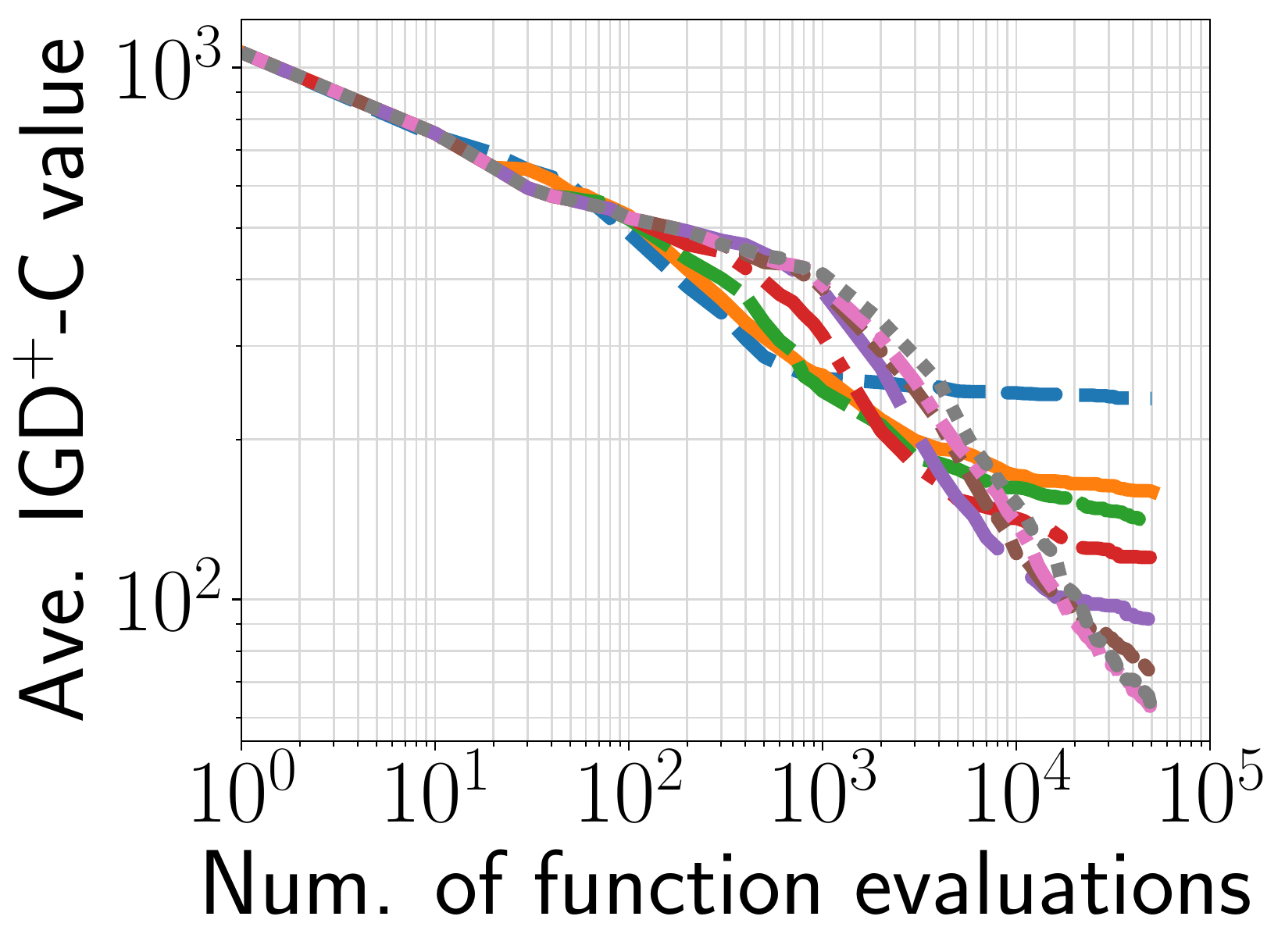}}
  \subfloat[DTLZ3 ($m=6$)]{\includegraphics[width=0.32\textwidth]{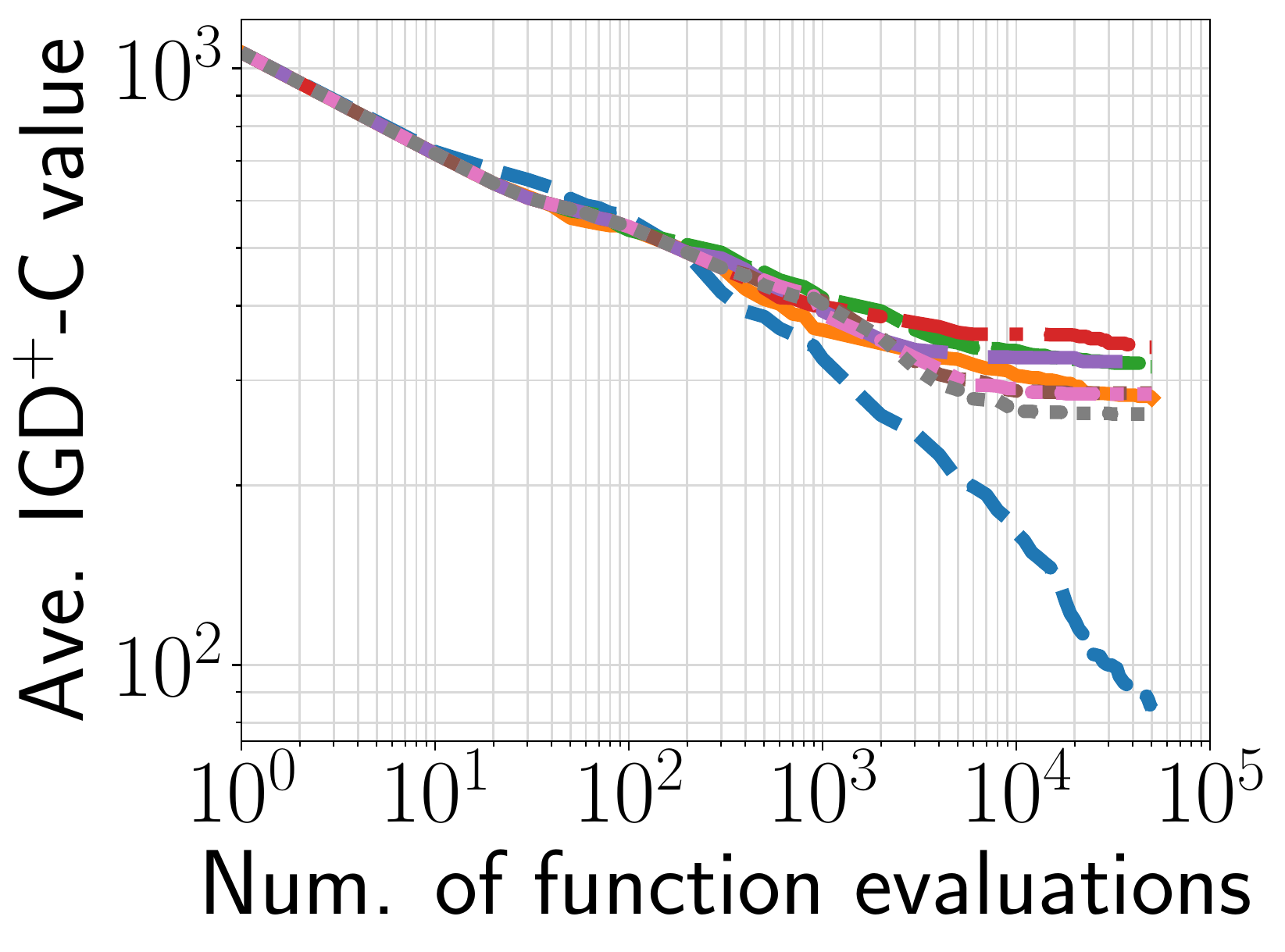}}
  \\
  \subfloat[DTLZ4 ($m=2$)]{\includegraphics[width=0.32\textwidth]{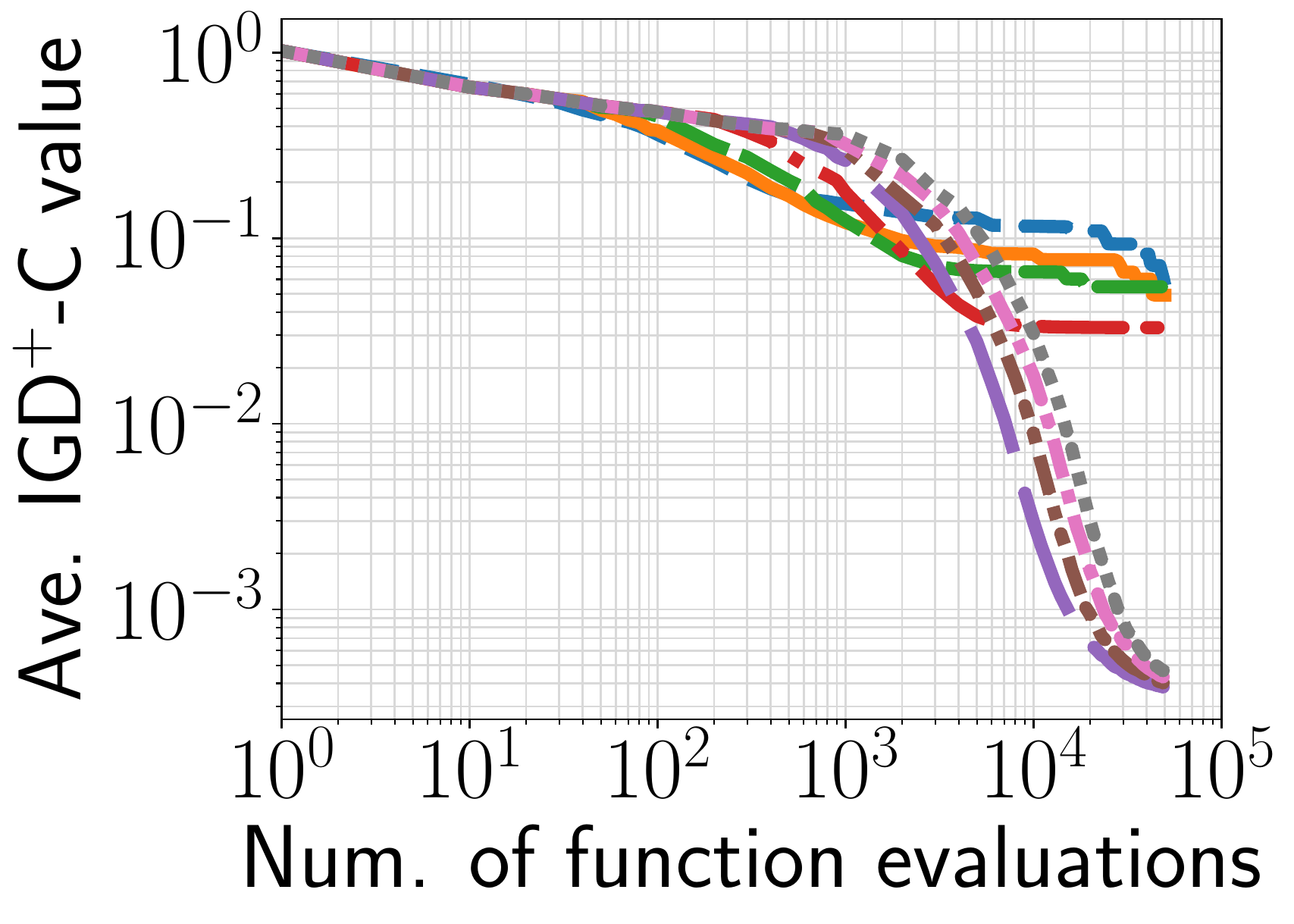}}
  \subfloat[DTLZ4 ($m=4$)]{\includegraphics[width=0.32\textwidth]{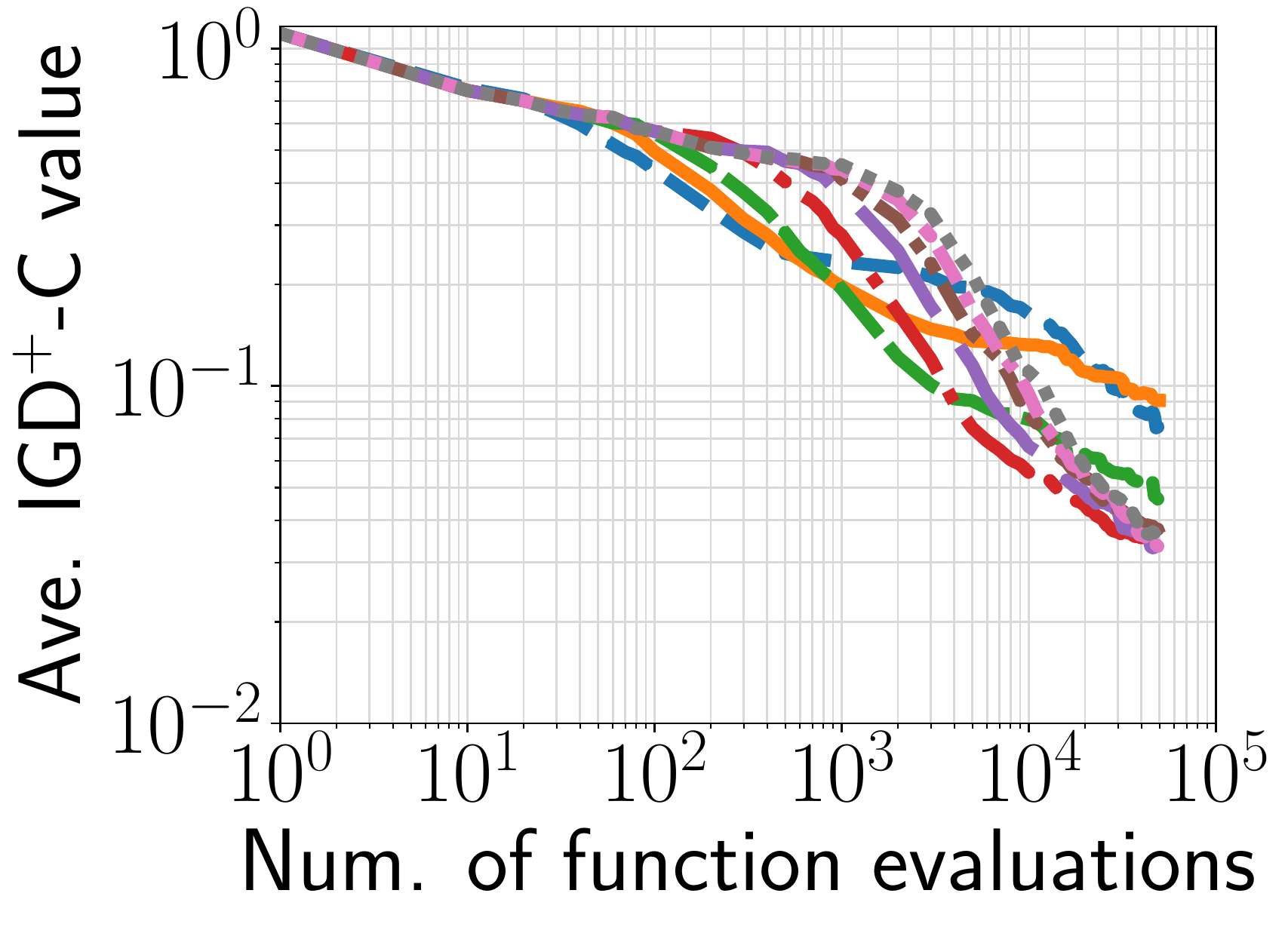}}
  \subfloat[DTLZ4 ($m=6$)]{\includegraphics[width=0.32\textwidth]{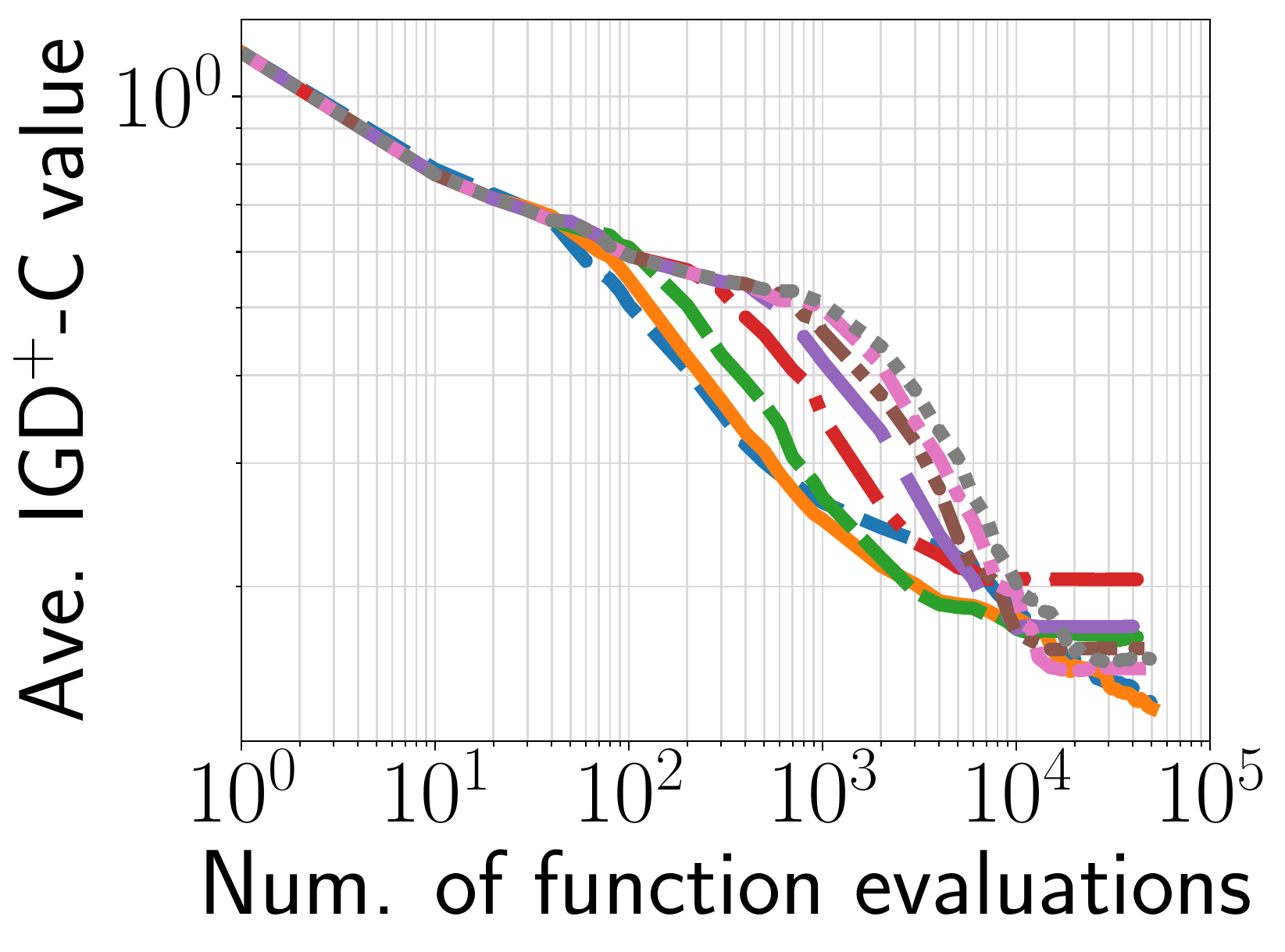}}
      \caption{Average IGD$^+$-C values of g-NSGA-II with different population sizes on the DTLZ1--DTLZ4 problems with $m \in \{2, 4, 6\}$.}
   \label{fig:sup_gnsgaii_dtlz}
\end{figure*}

\begin{figure*}[t]
  \centering
  \subfloat{\includegraphics[width=0.9\textwidth]{./figs/comp_mu/legend.pdf}}
  \\
  \subfloat[DTLZ1 ($m=2$)]{\includegraphics[width=0.32\textwidth]{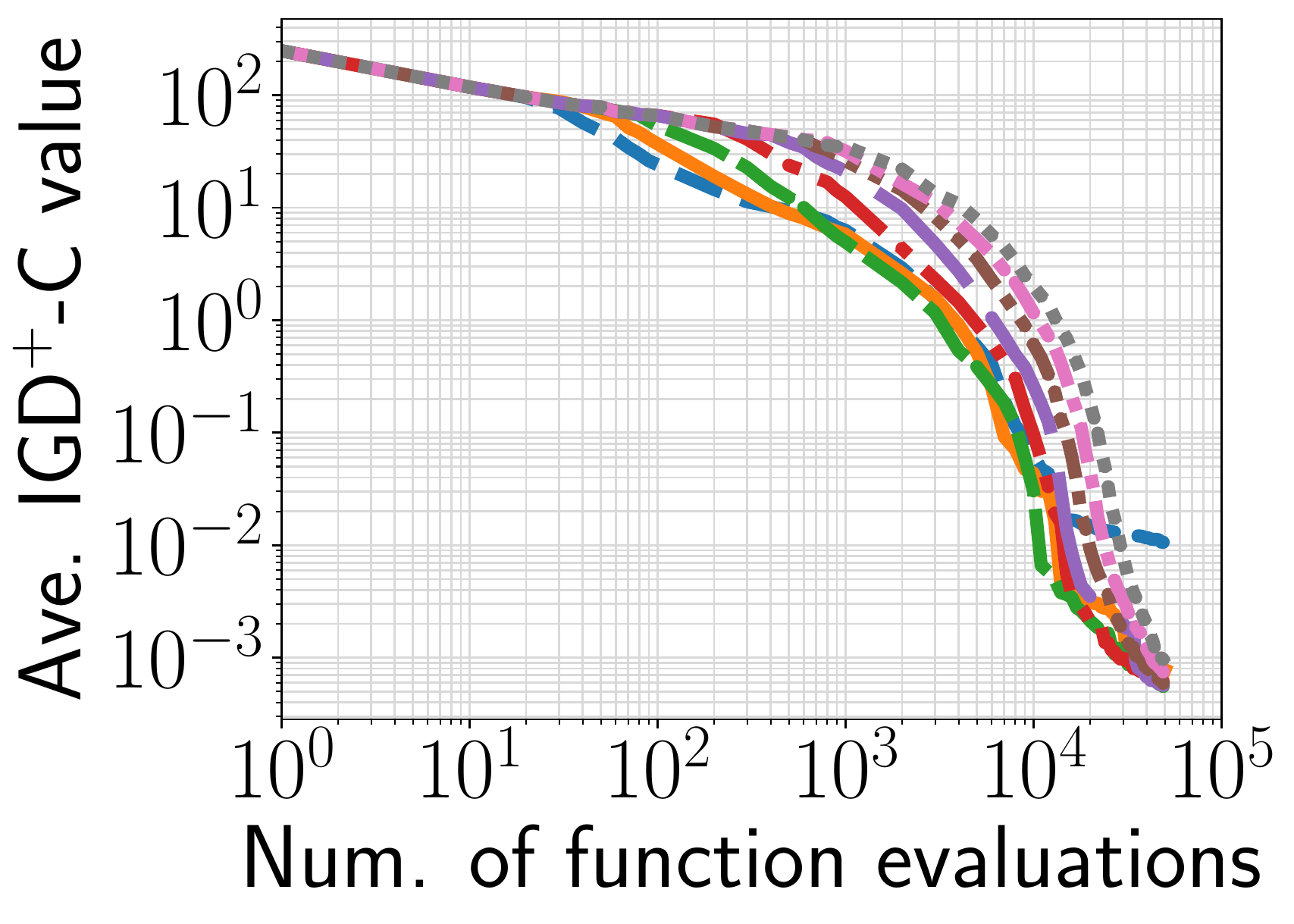}}
  \subfloat[DTLZ1 ($m=4$)]{\includegraphics[width=0.32\textwidth]{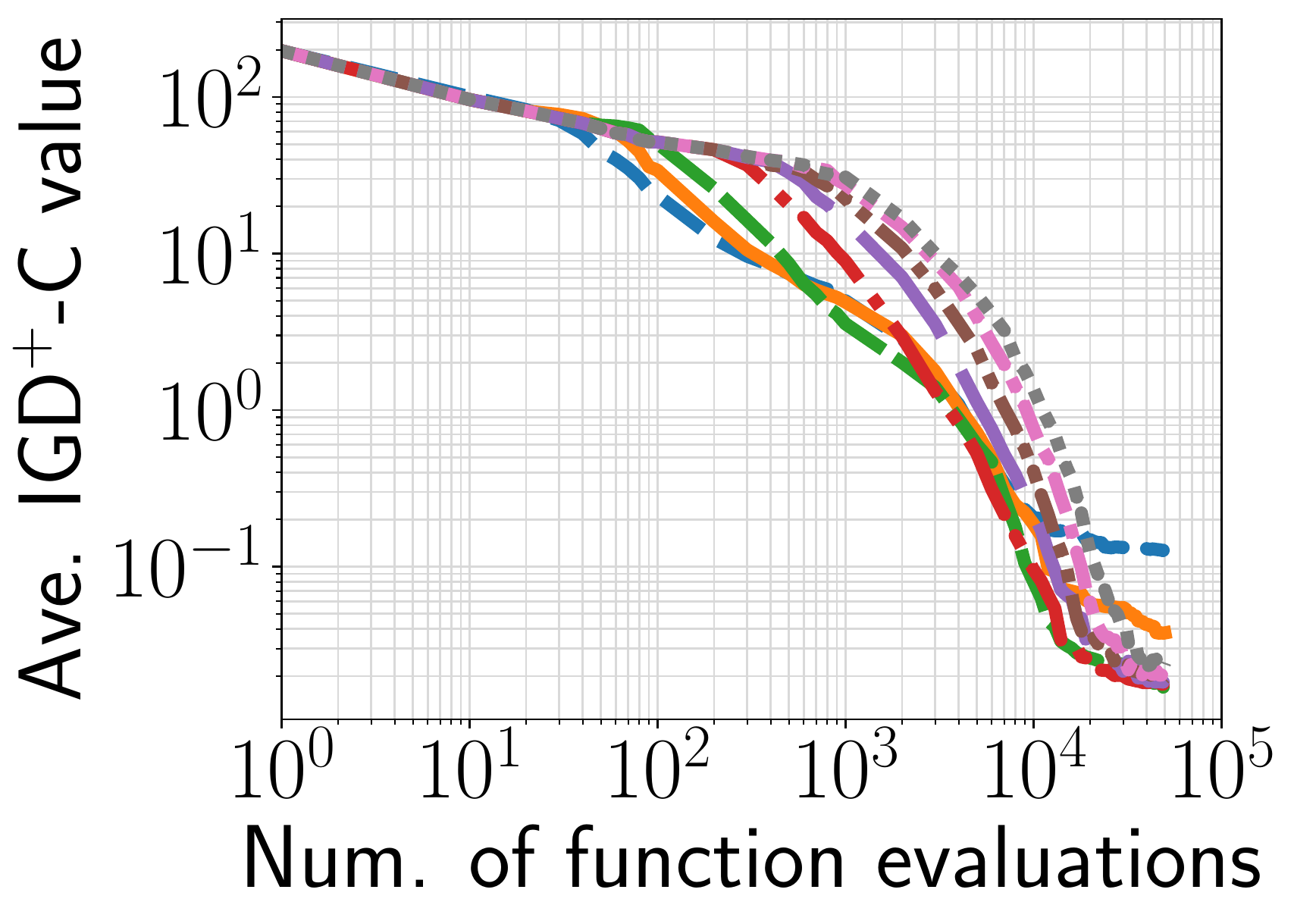}}
  \subfloat[DTLZ1 ($m=6$)]{\includegraphics[width=0.32\textwidth]{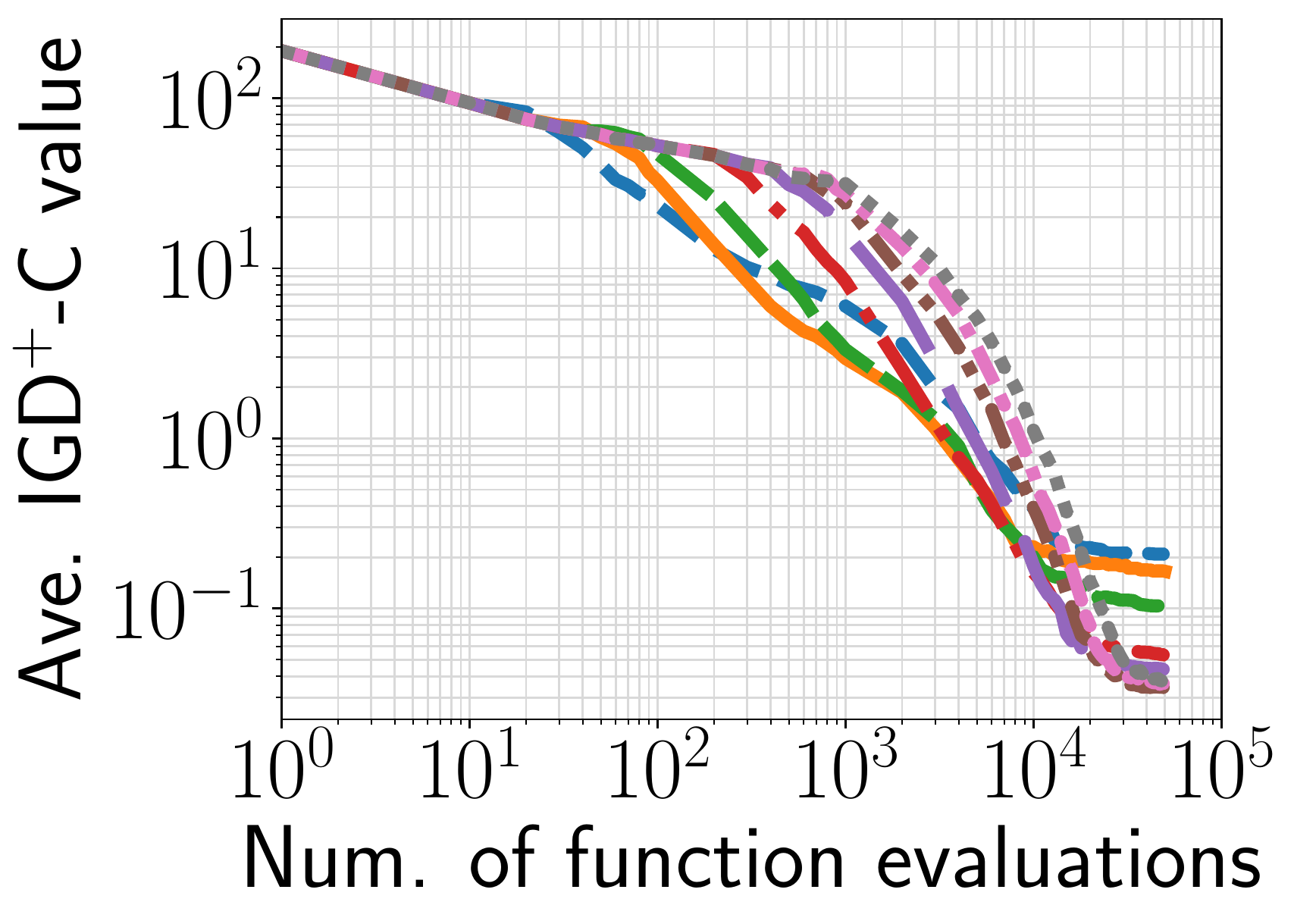}}
  \\
  \subfloat[DTLZ2 ($m=2$)]{\includegraphics[width=0.32\textwidth]{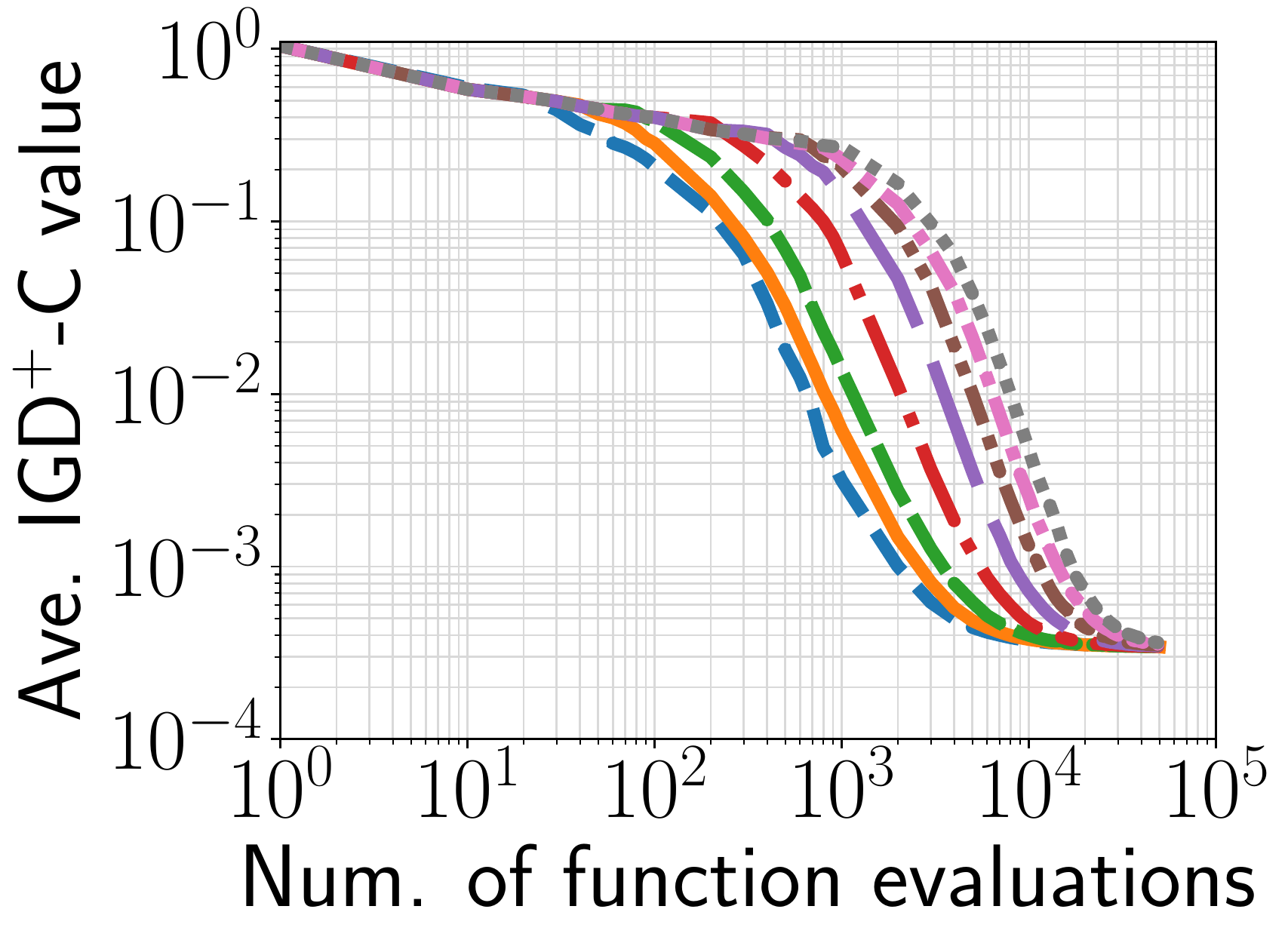}}
  \subfloat[DTLZ2 ($m=4$)]{\includegraphics[width=0.32\textwidth]{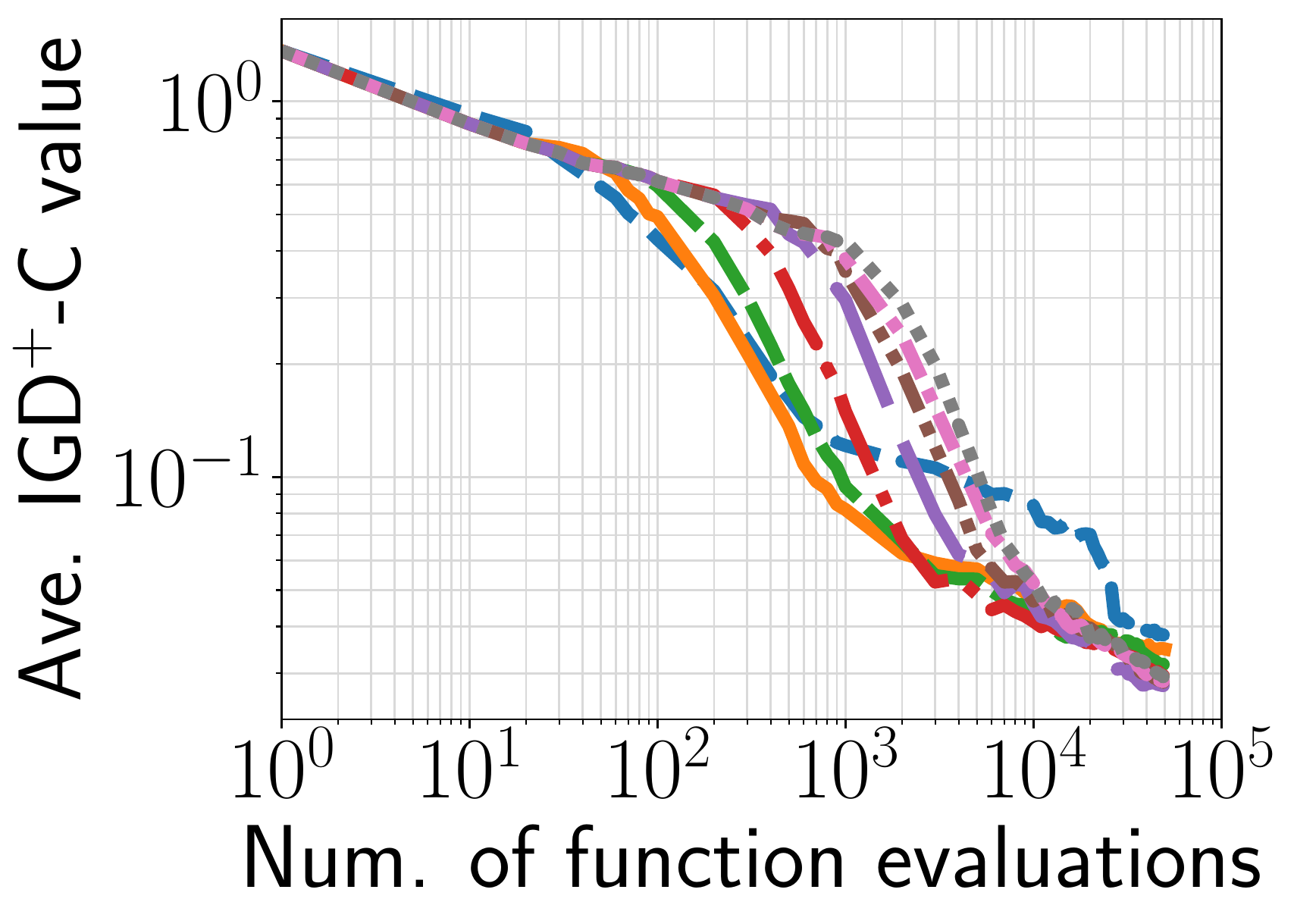}}
  \subfloat[DTLZ2 ($m=6$)]{\includegraphics[width=0.32\textwidth]{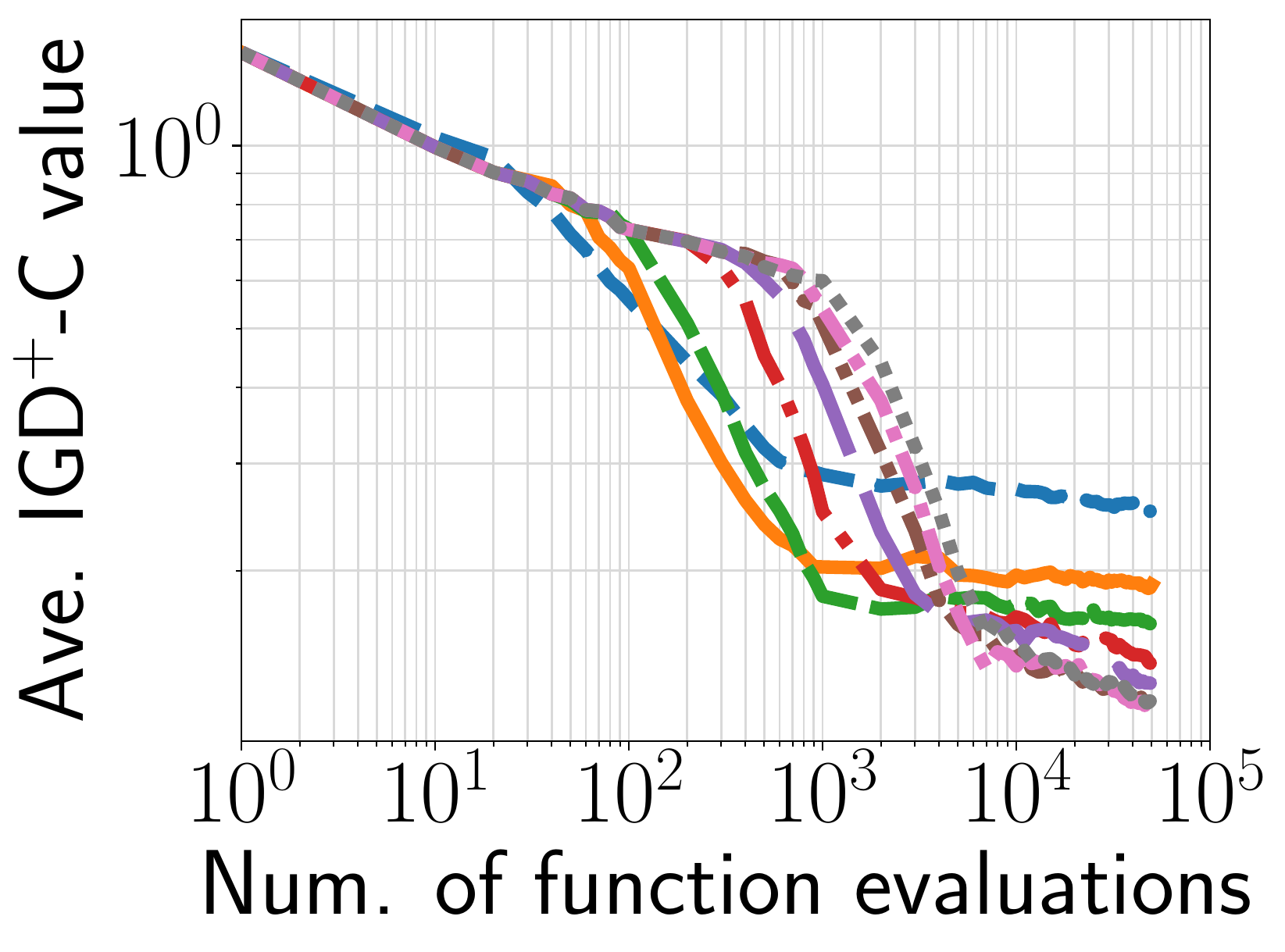}}
  \\
  \subfloat[DTLZ3 ($m=2$)]{\includegraphics[width=0.32\textwidth]{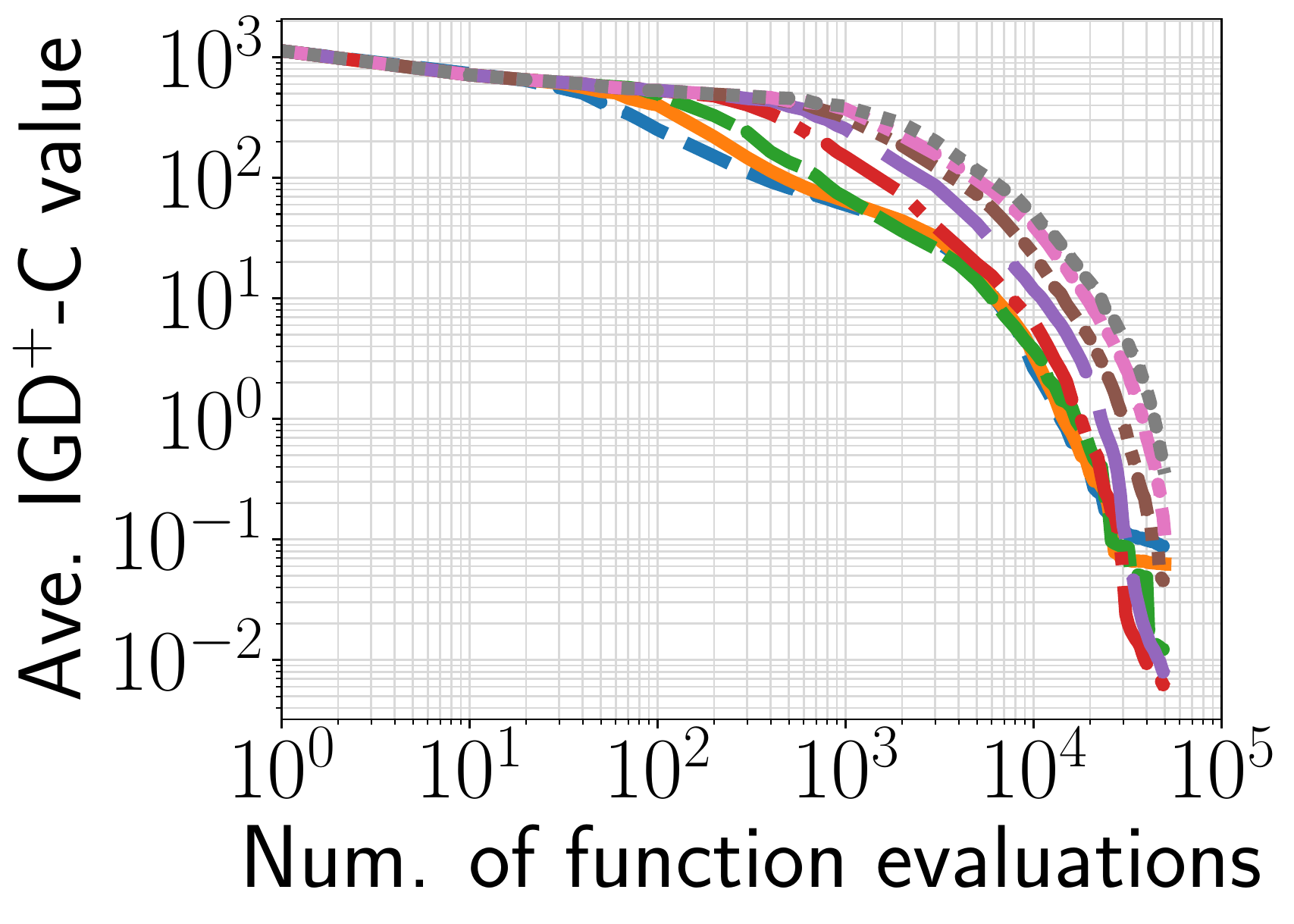}}
  \subfloat[DTLZ3 ($m=4$)]{\includegraphics[width=0.32\textwidth]{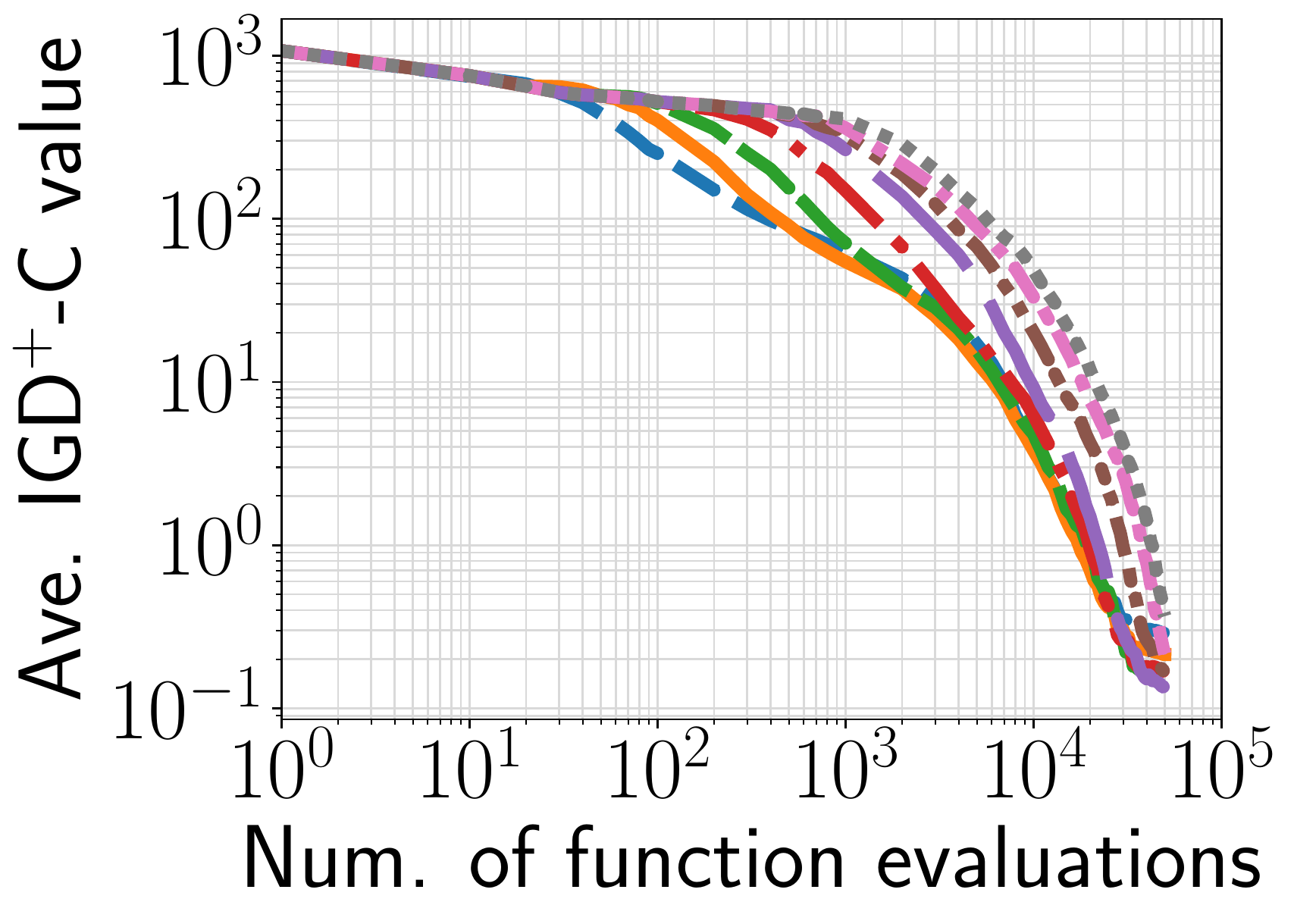}}
  \subfloat[DTLZ3 ($m=6$)]{\includegraphics[width=0.32\textwidth]{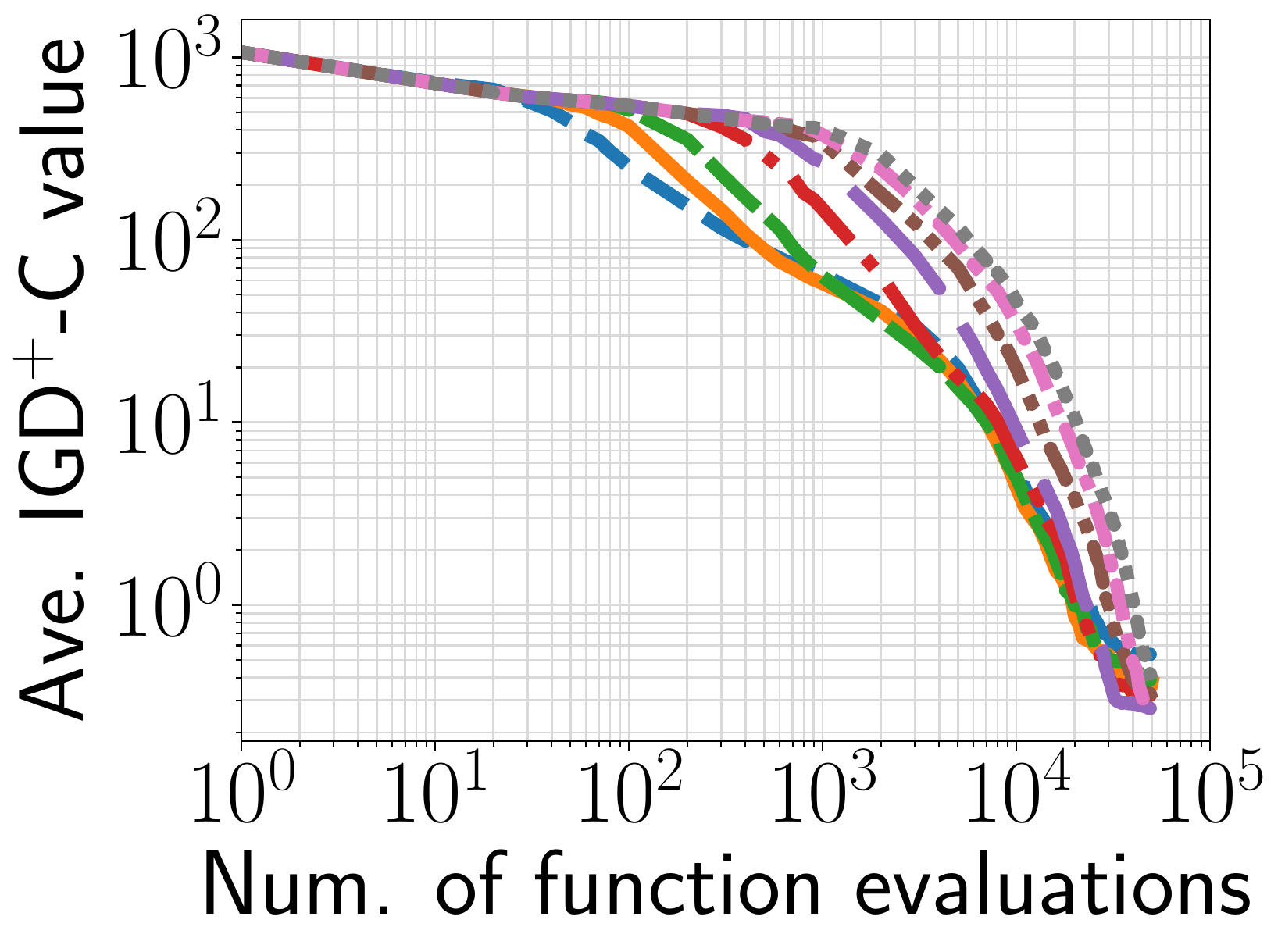}}
  \\
  \subfloat[DTLZ4 ($m=2$)]{\includegraphics[width=0.32\textwidth]{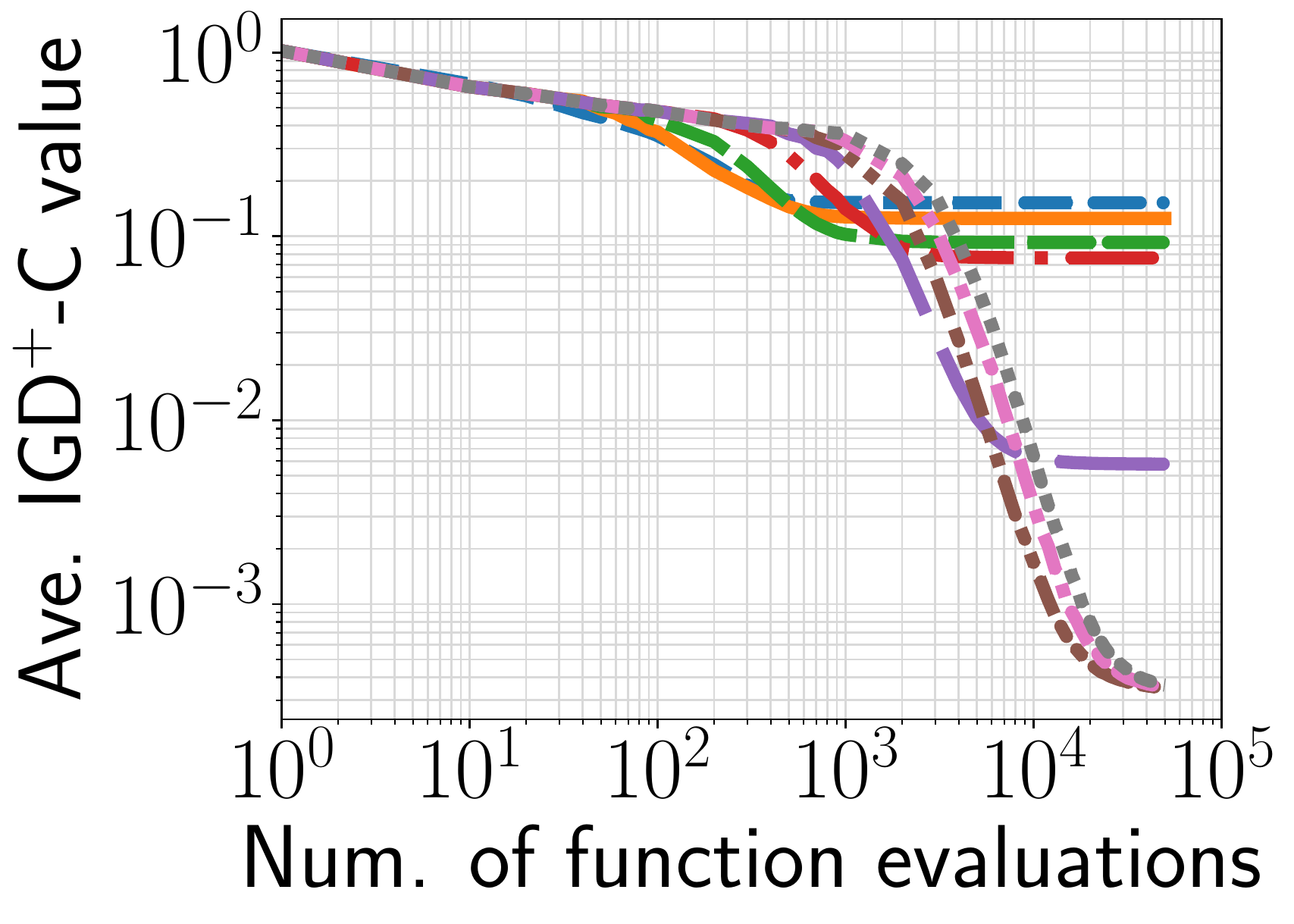}}
  \subfloat[DTLZ4 ($m=4$)]{\includegraphics[width=0.32\textwidth]{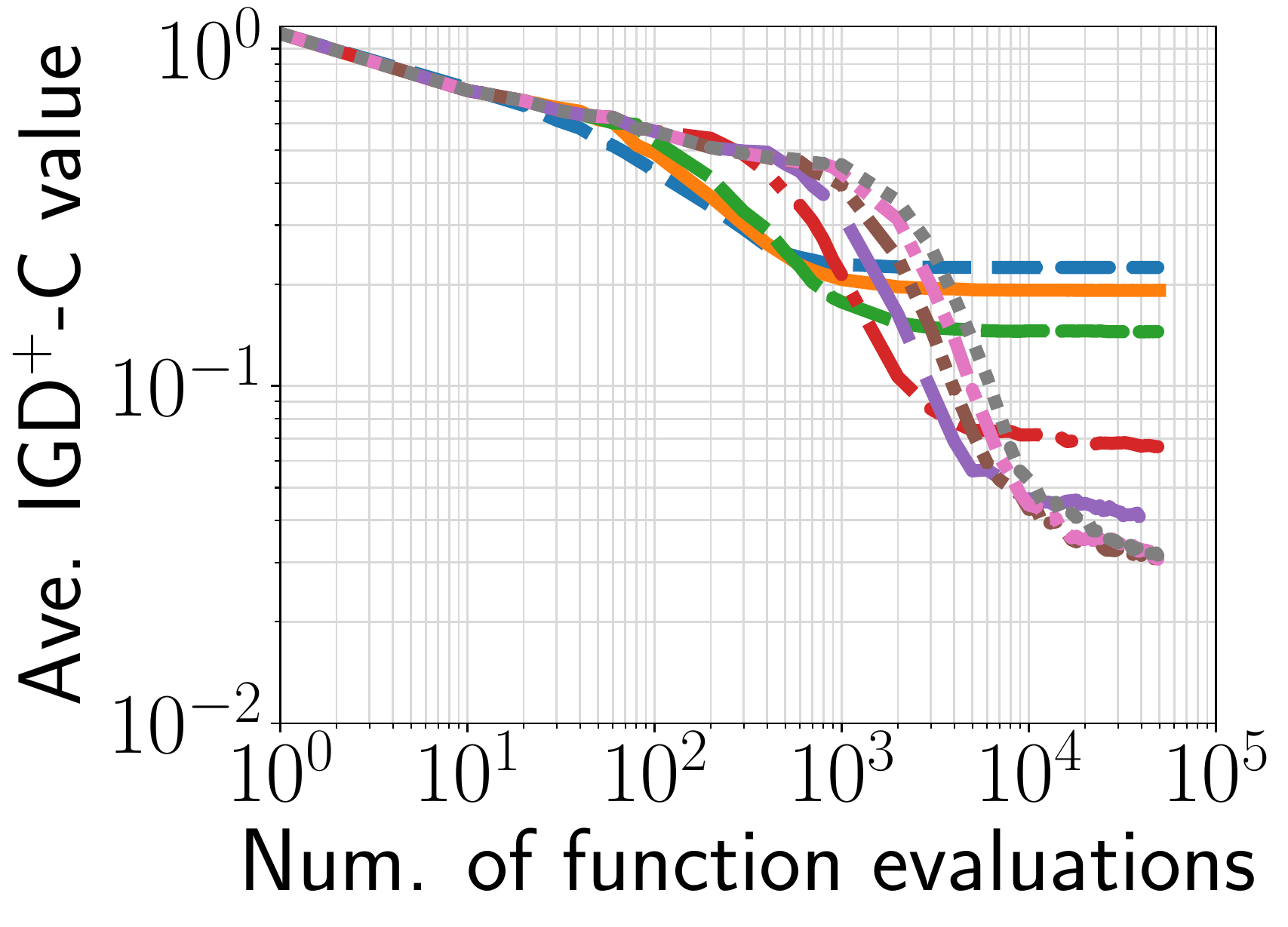}}
  \subfloat[DTLZ4 ($m=6$)]{\includegraphics[width=0.32\textwidth]{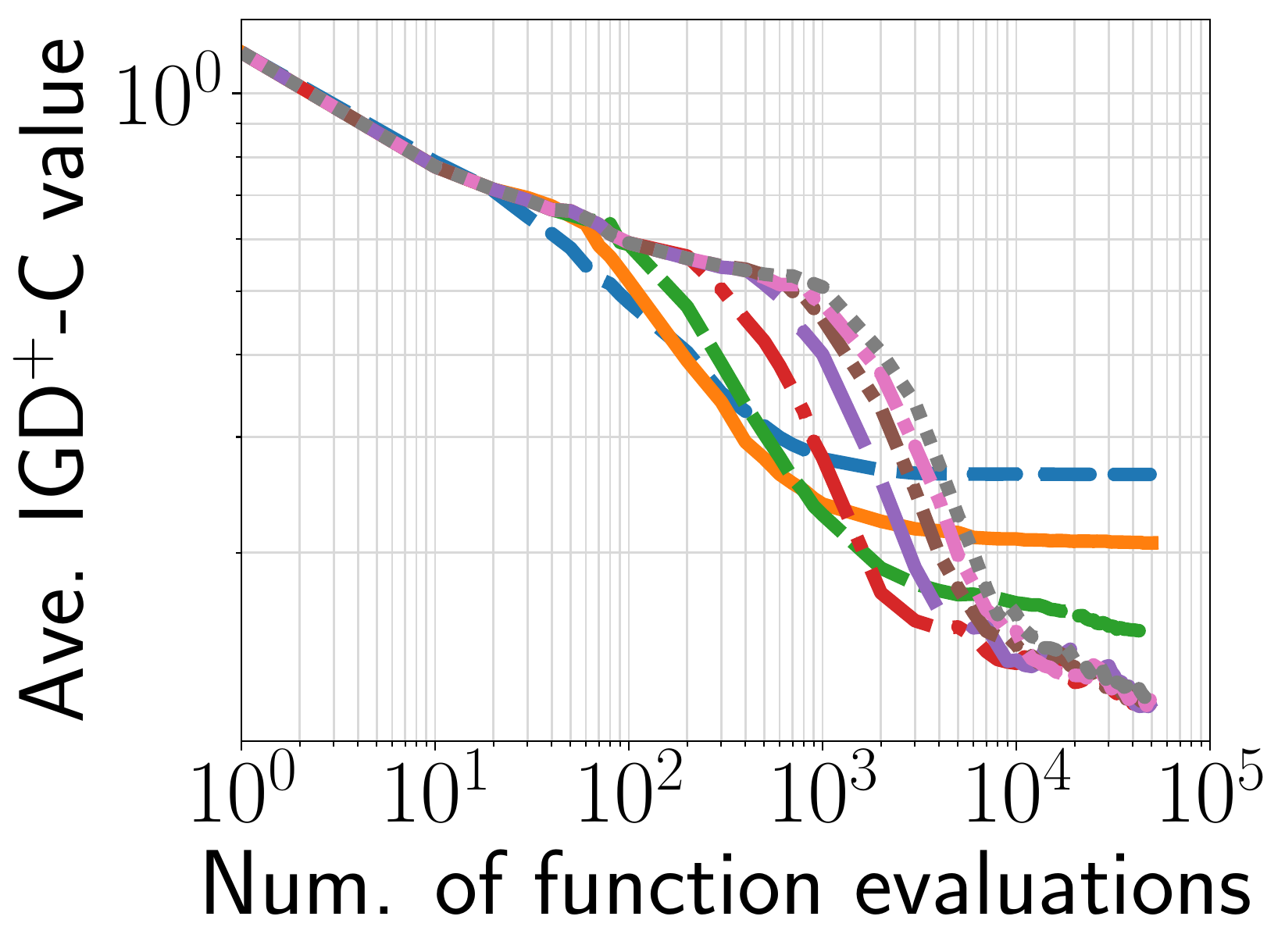}}
      \caption{Average IGD$^+$-C values of PBEA with different population sizes on the DTLZ1--DTLZ4 problems with $m \in \{2, 4, 6\}$.}
   \label{fig:sup_pbea_dtlz}
\end{figure*}

\begin{figure*}[t]
  \centering
  \subfloat{\includegraphics[width=0.9\textwidth]{./figs/comp_mu/legend.pdf}}
  \\
  \subfloat[DTLZ1 ($m=2$)]{\includegraphics[width=0.32\textwidth]{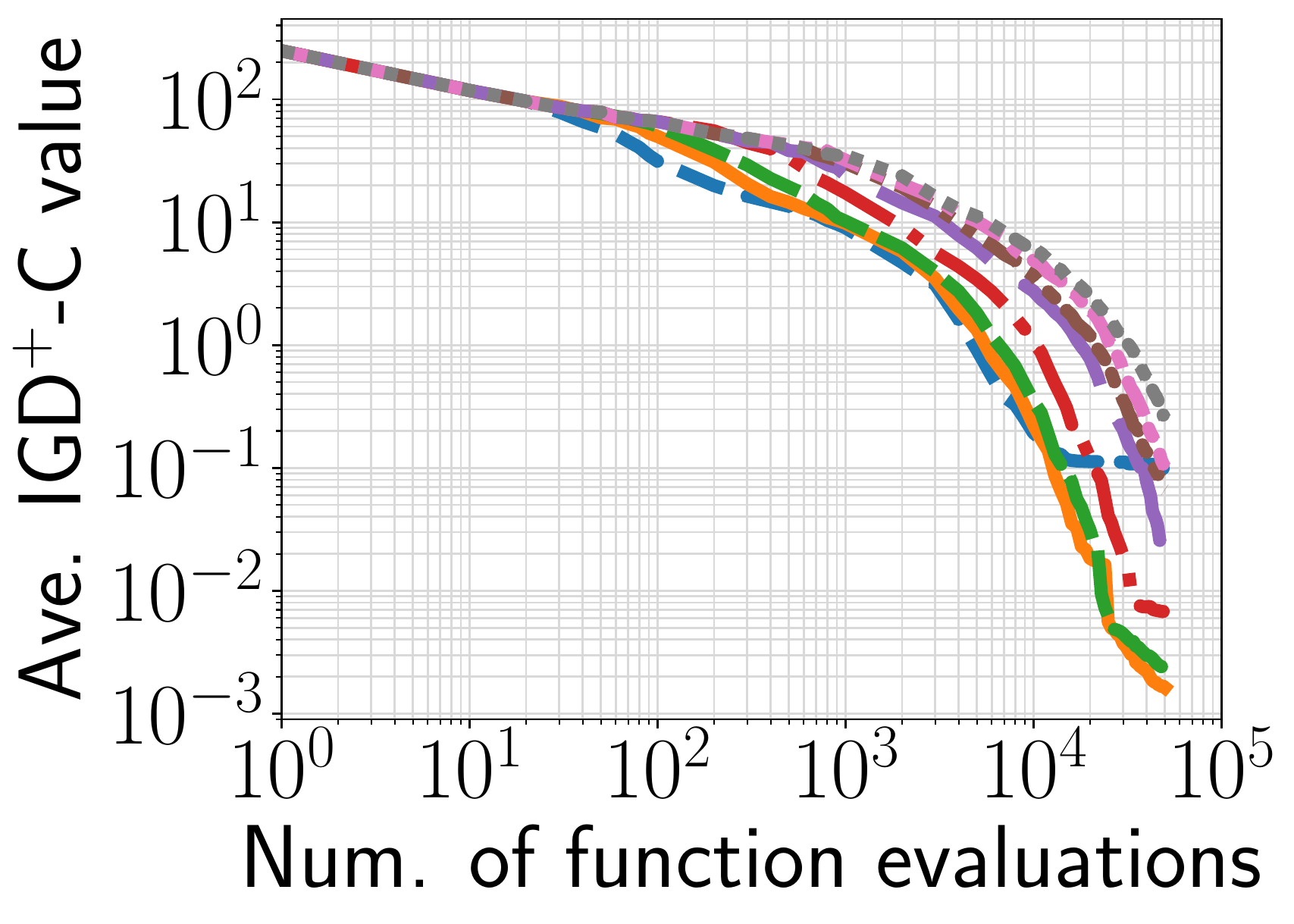}}
  \subfloat[DTLZ1 ($m=4$)]{\includegraphics[width=0.32\textwidth]{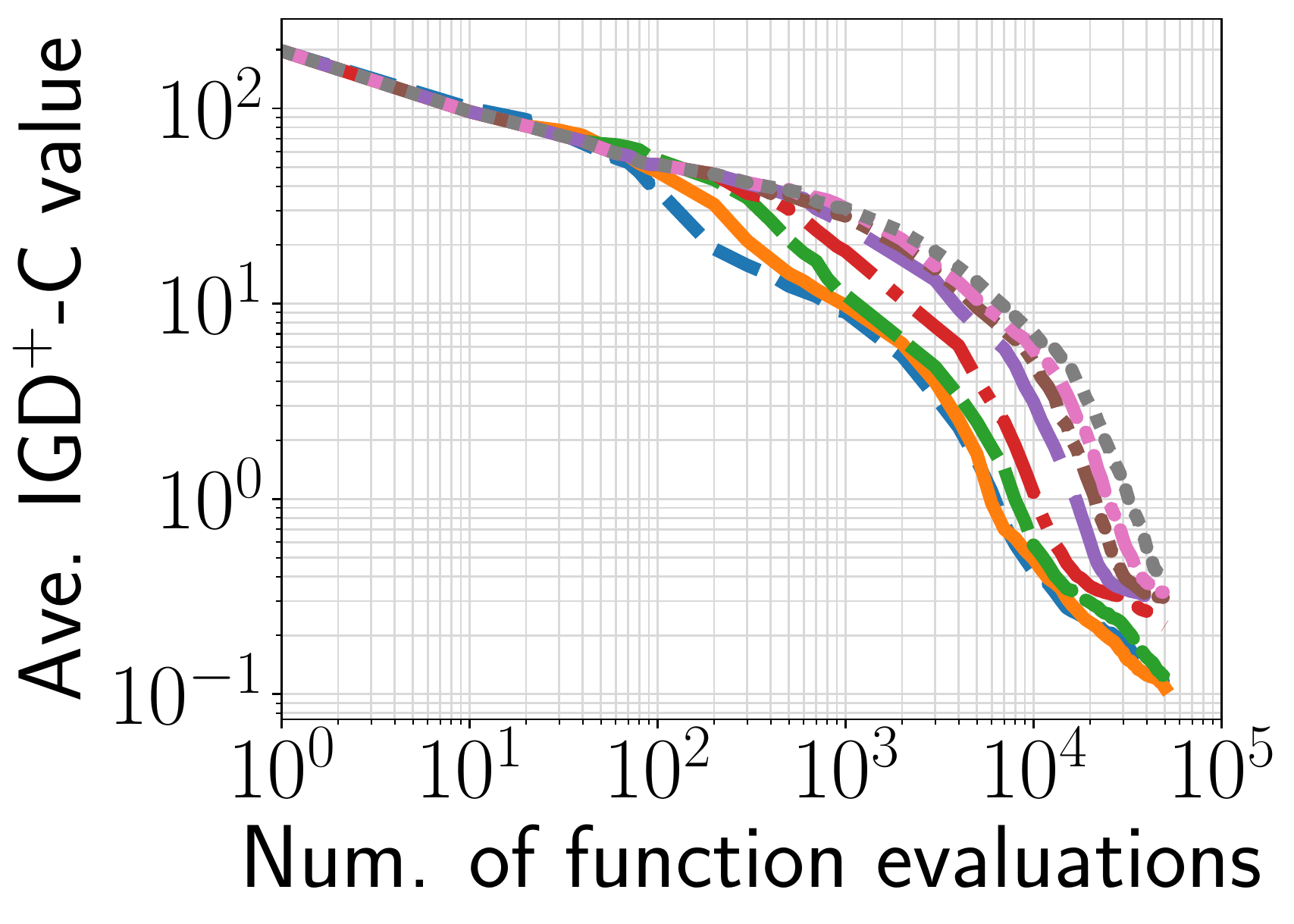}}
  \subfloat[DTLZ1 ($m=6$)]{\includegraphics[width=0.32\textwidth]{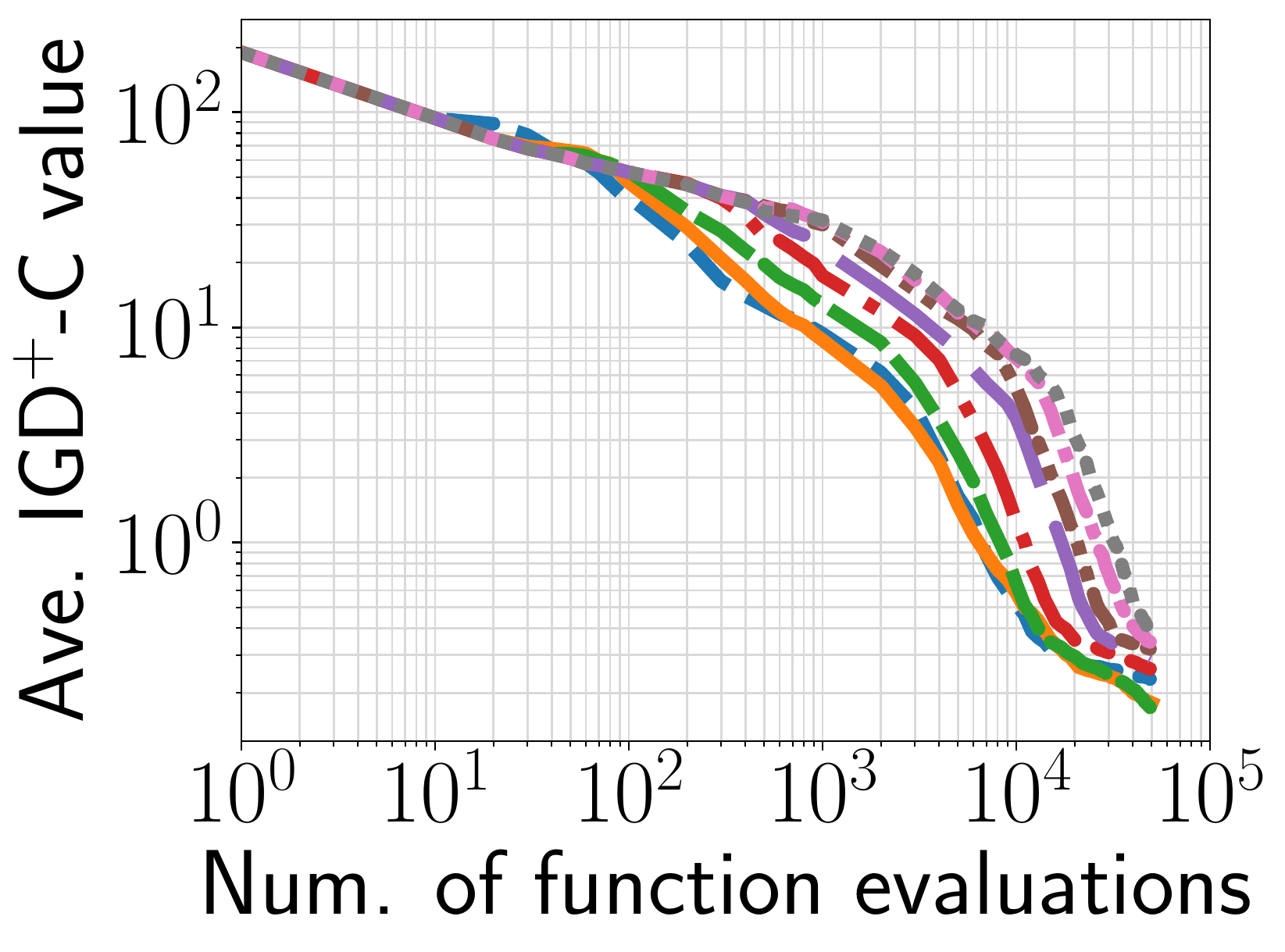}}
  \\
  \subfloat[DTLZ2 ($m=2$)]{\includegraphics[width=0.32\textwidth]{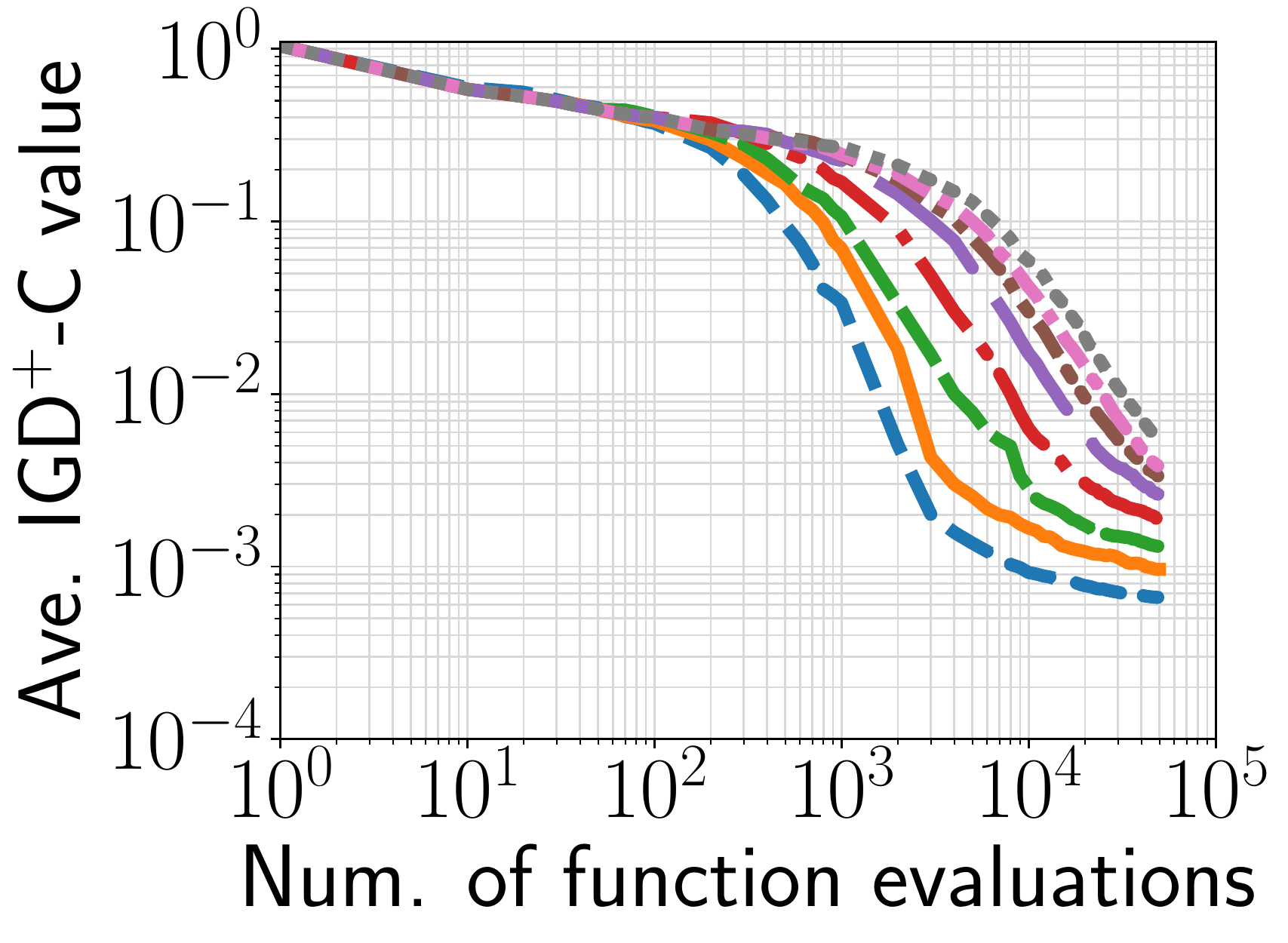}}
  \subfloat[DTLZ2 ($m=4$)]{\includegraphics[width=0.32\textwidth]{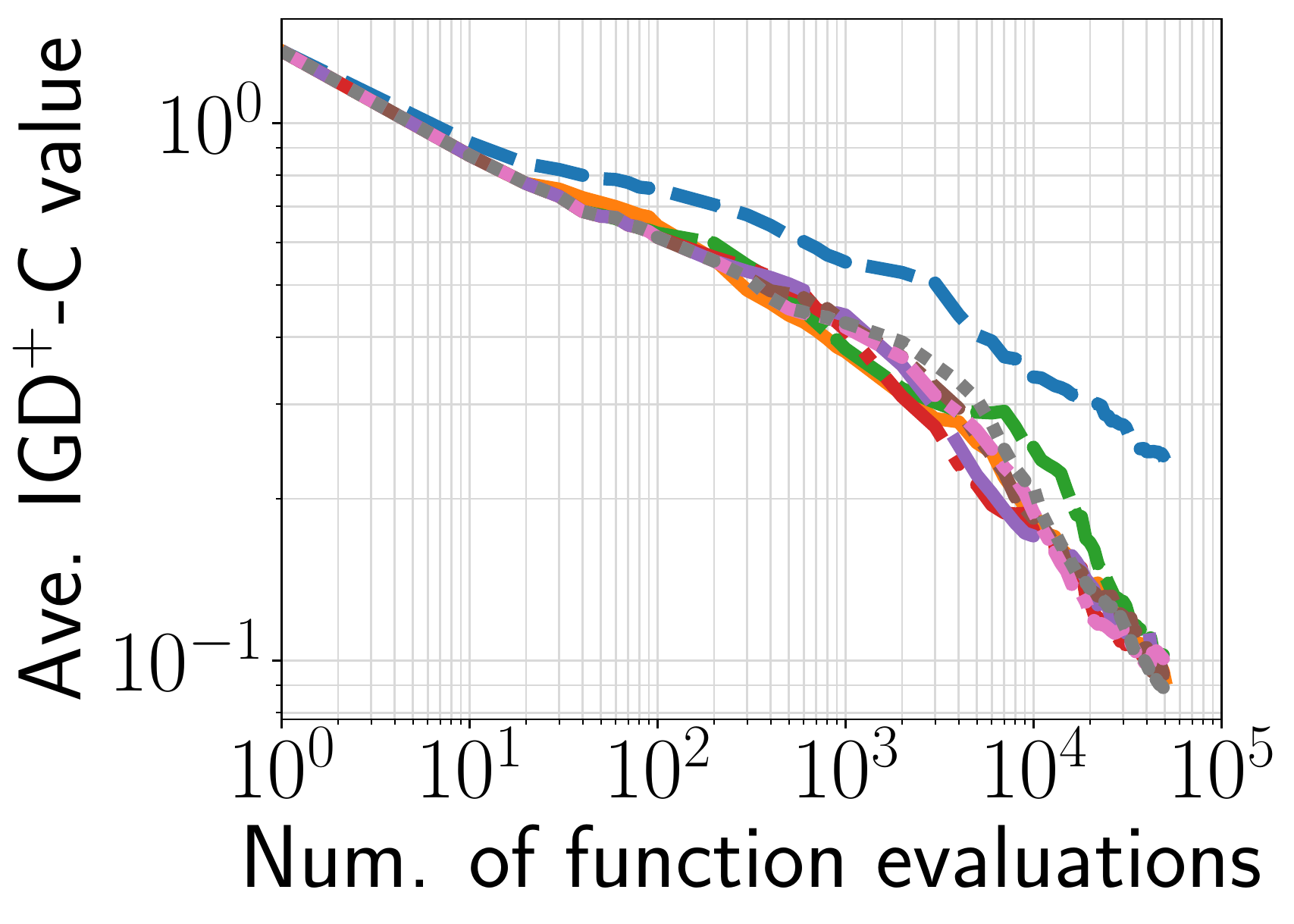}}
  \subfloat[DTLZ2 ($m=6$)]{\includegraphics[width=0.32\textwidth]{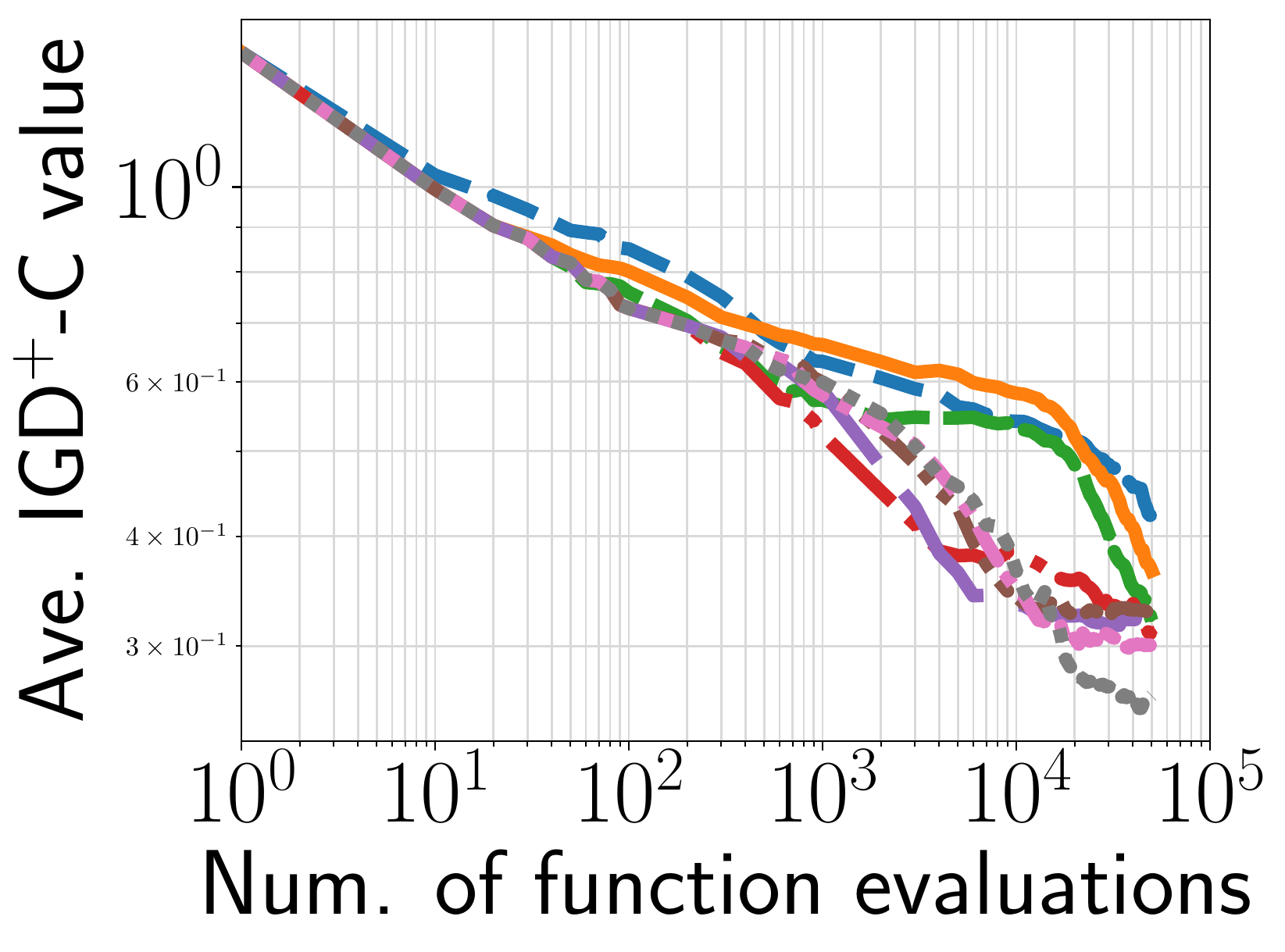}}
  \\
  \subfloat[DTLZ3 ($m=2$)]{\includegraphics[width=0.32\textwidth]{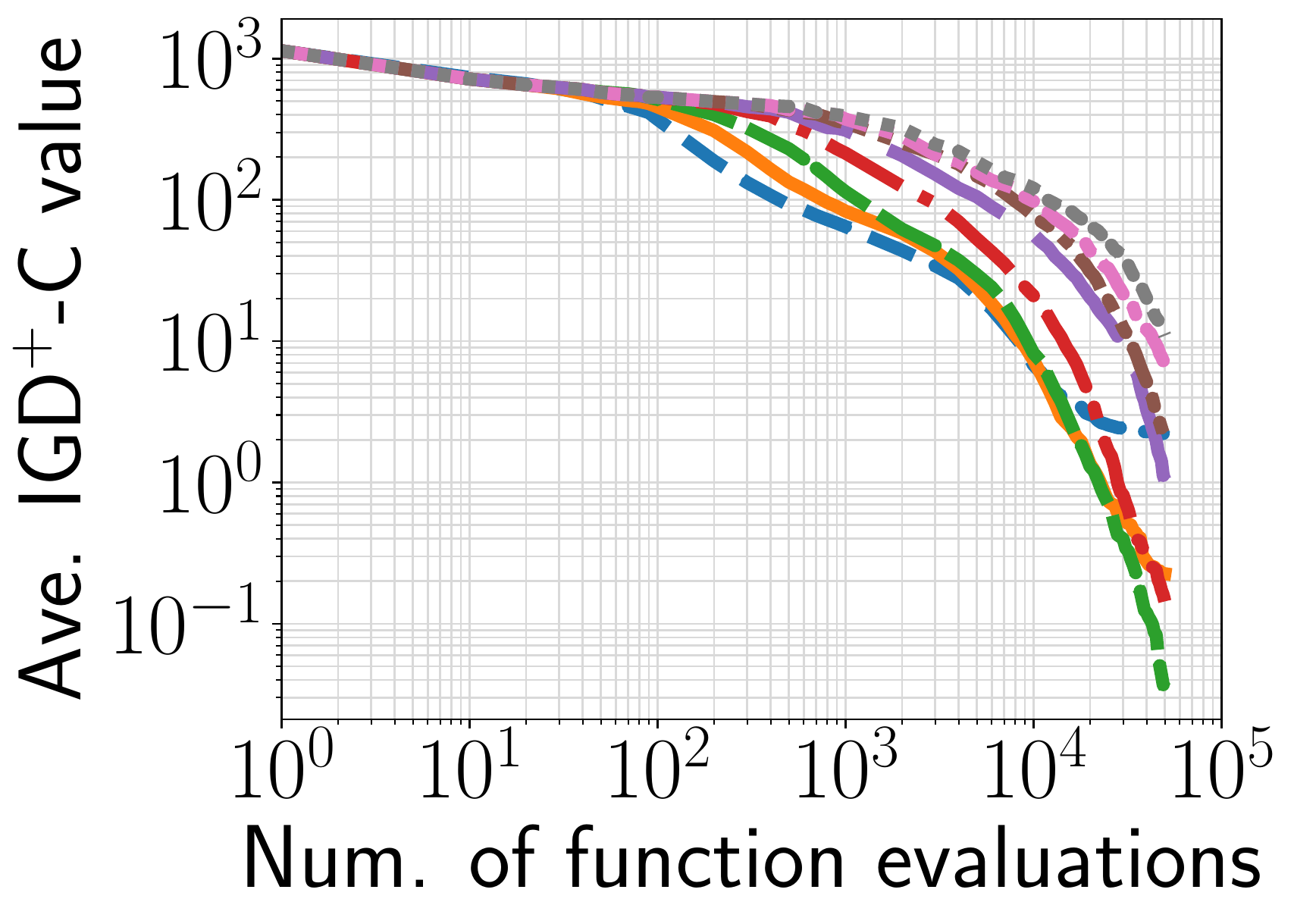}}
  \subfloat[DTLZ3 ($m=4$)]{\includegraphics[width=0.32\textwidth]{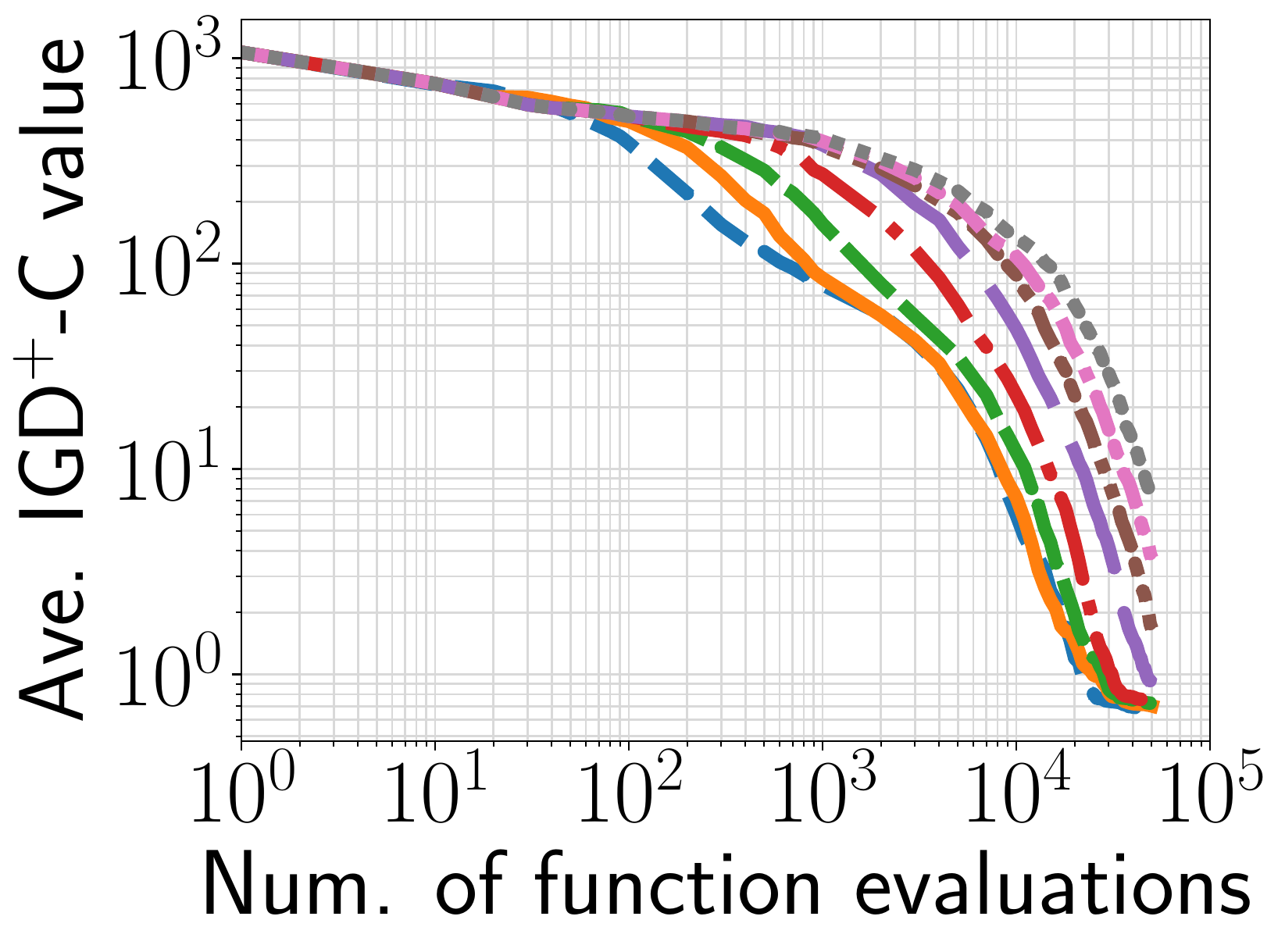}}
  \subfloat[DTLZ3 ($m=6$)]{\includegraphics[width=0.32\textwidth]{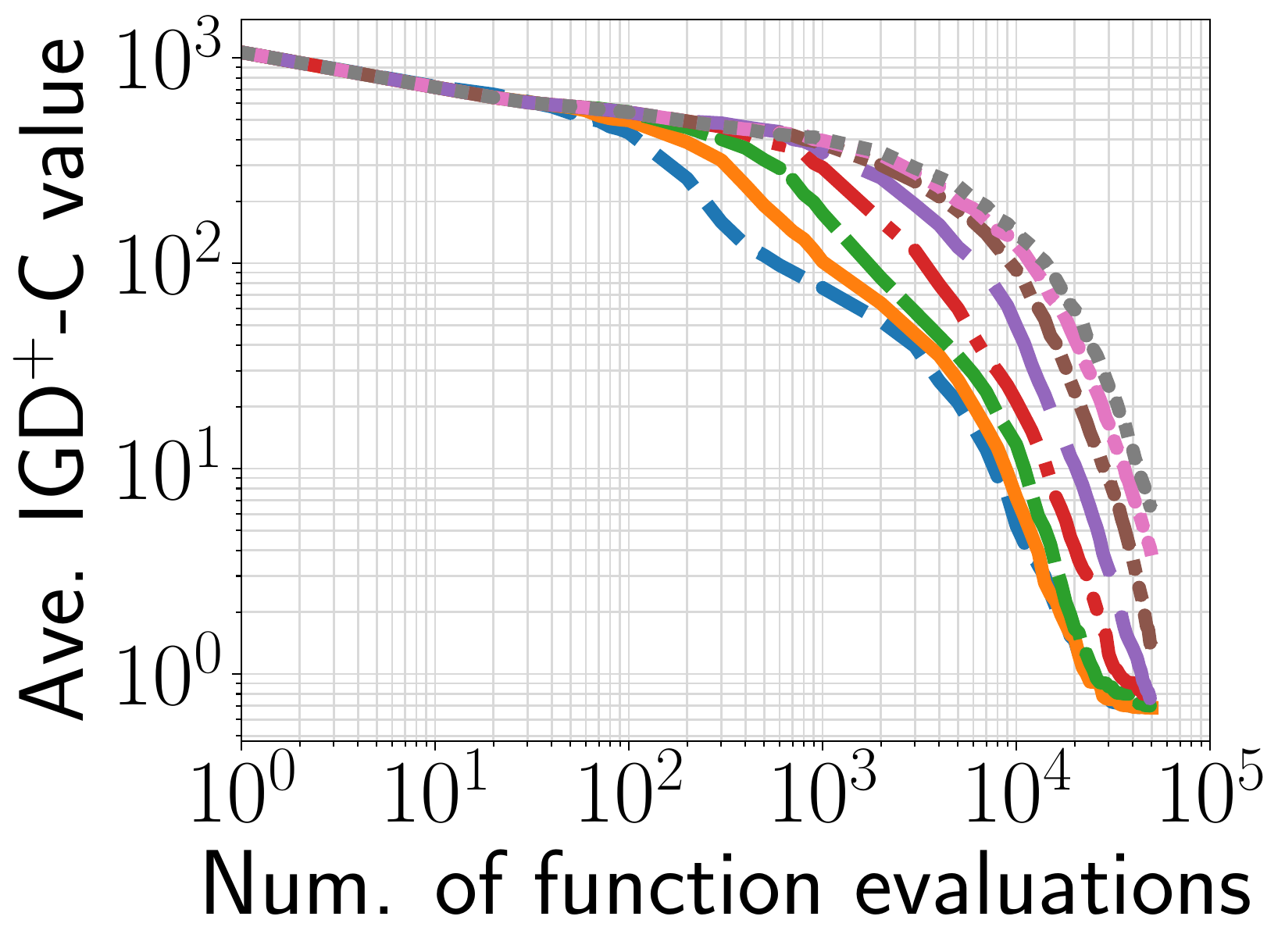}}
  \\
  \subfloat[DTLZ4 ($m=2$)]{\includegraphics[width=0.32\textwidth]{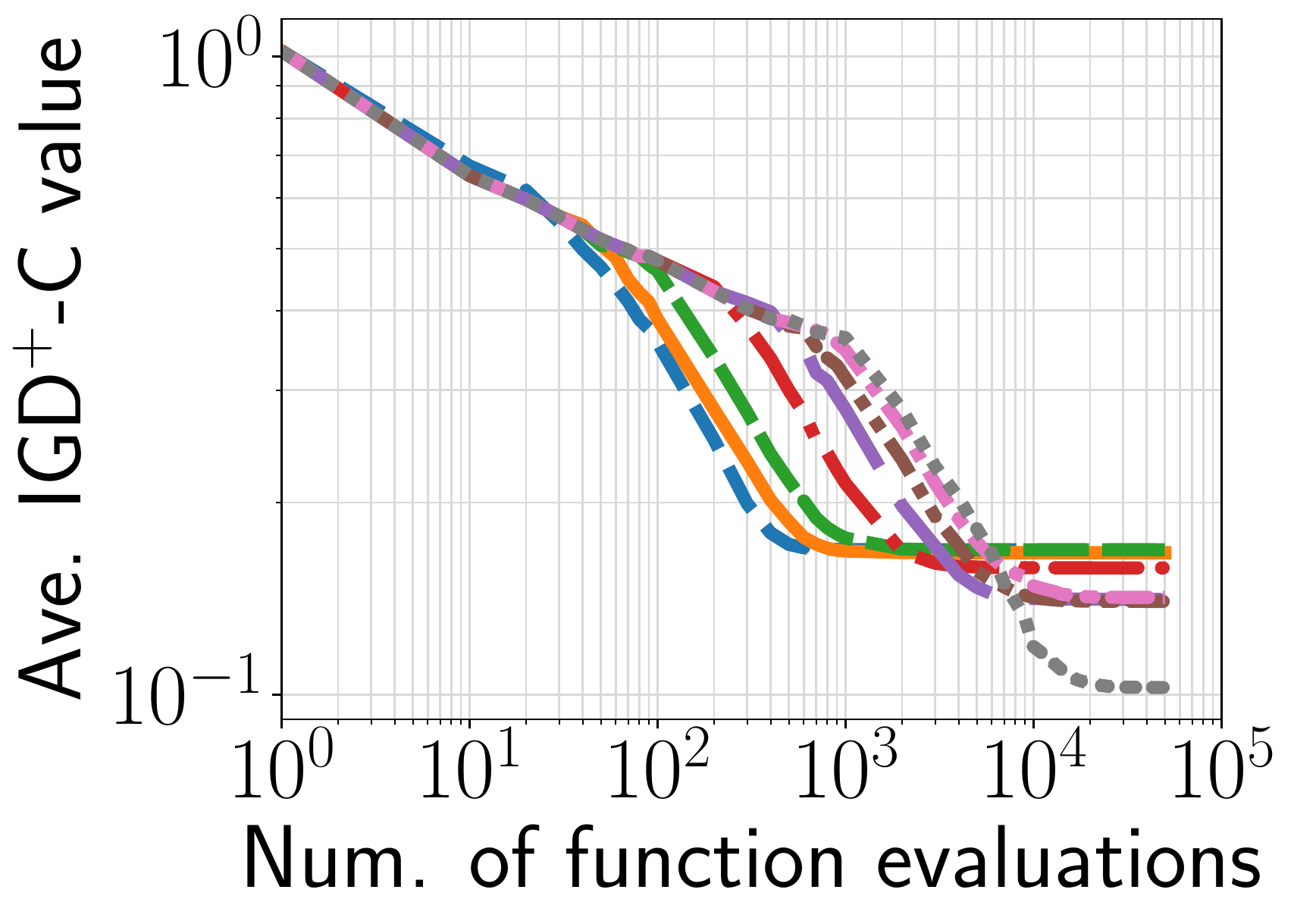}}
  \subfloat[DTLZ4 ($m=4$)]{\includegraphics[width=0.32\textwidth]{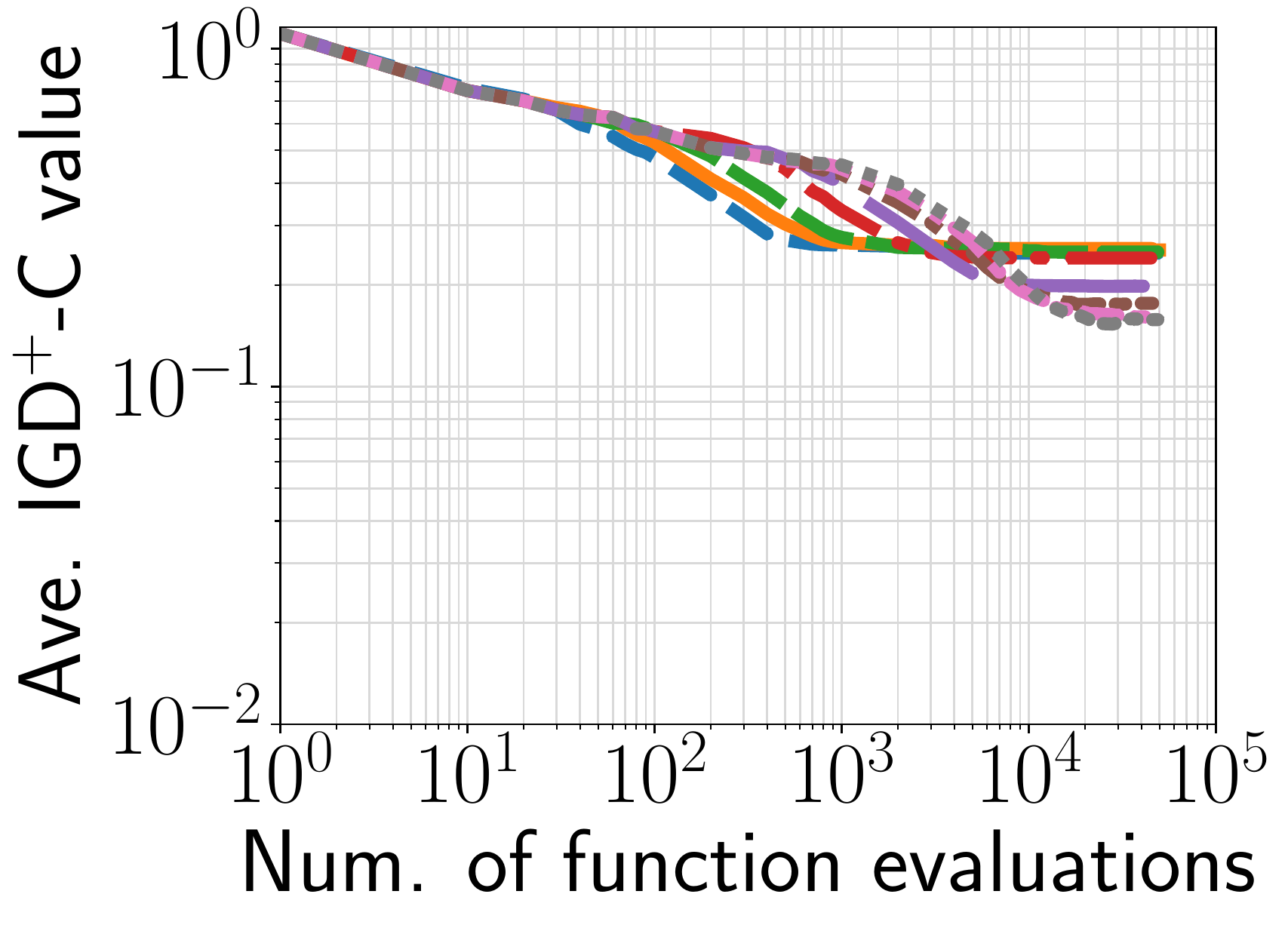}}
  \subfloat[DTLZ4 ($m=6$)]{\includegraphics[width=0.32\textwidth]{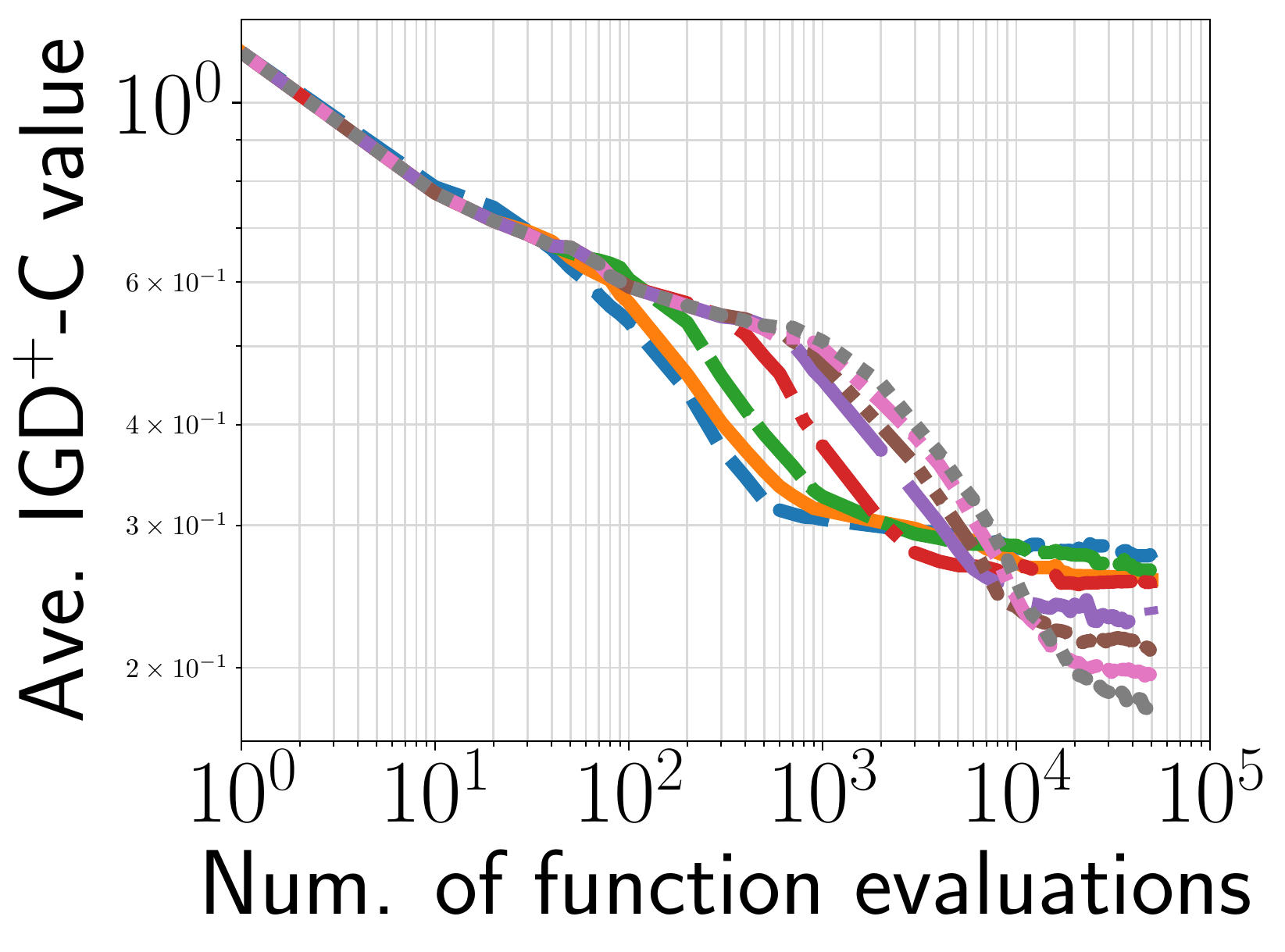}}
      \caption{Average IGD$^+$-C values of R-MEAD2 with different population sizes on the DTLZ1--DTLZ4 problems with $m \in \{2, 4, 6\}$.}
   \label{fig:sup_rmead2_dtlz}
\end{figure*}

\begin{figure*}[t]
  \centering
  \subfloat{\includegraphics[width=0.9\textwidth]{./figs/comp_mu/legend.pdf}}
  \\
  \subfloat[DTLZ1 ($m=2$)]{\includegraphics[width=0.32\textwidth]{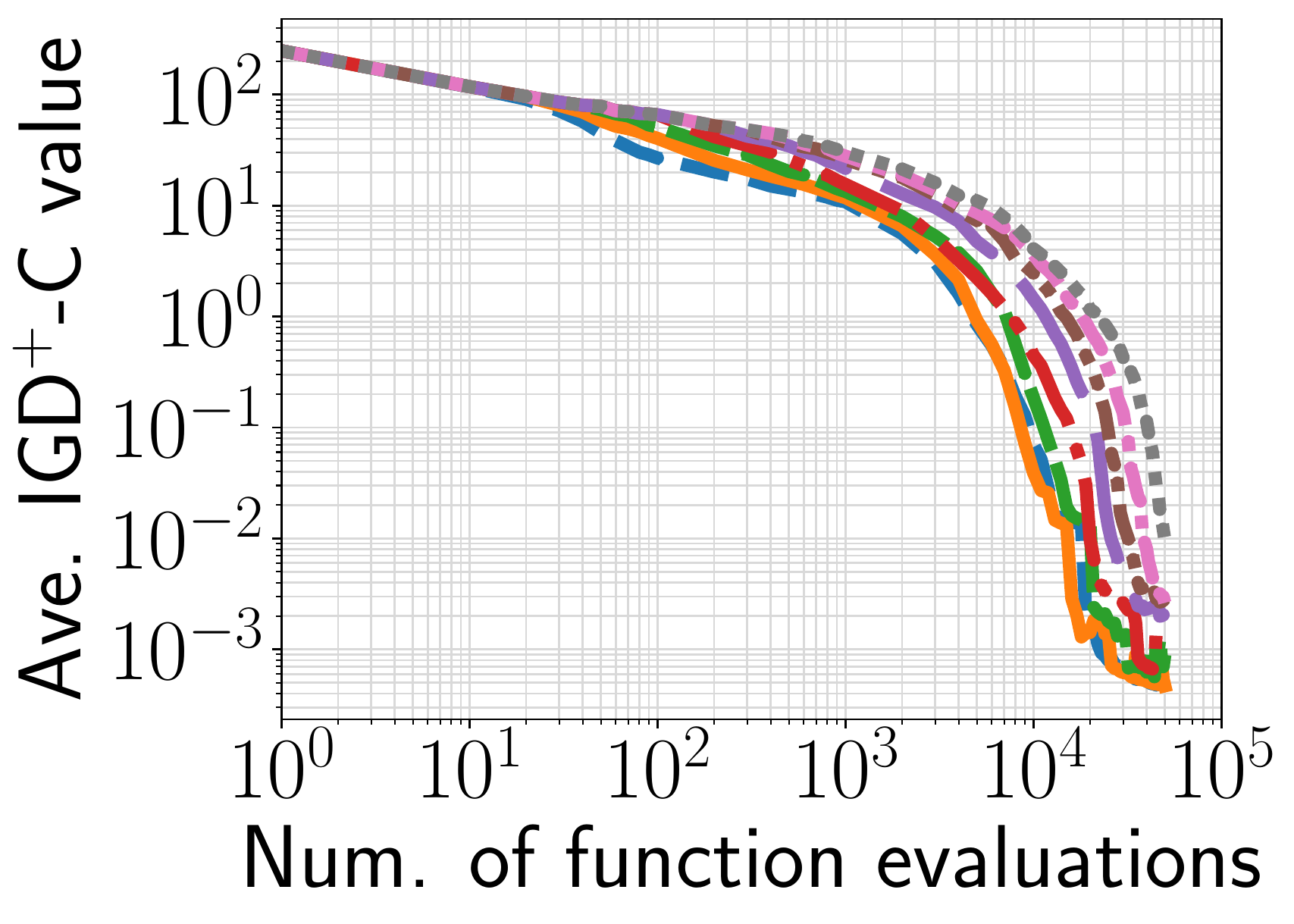}}
  \subfloat[DTLZ1 ($m=4$)]{\includegraphics[width=0.32\textwidth]{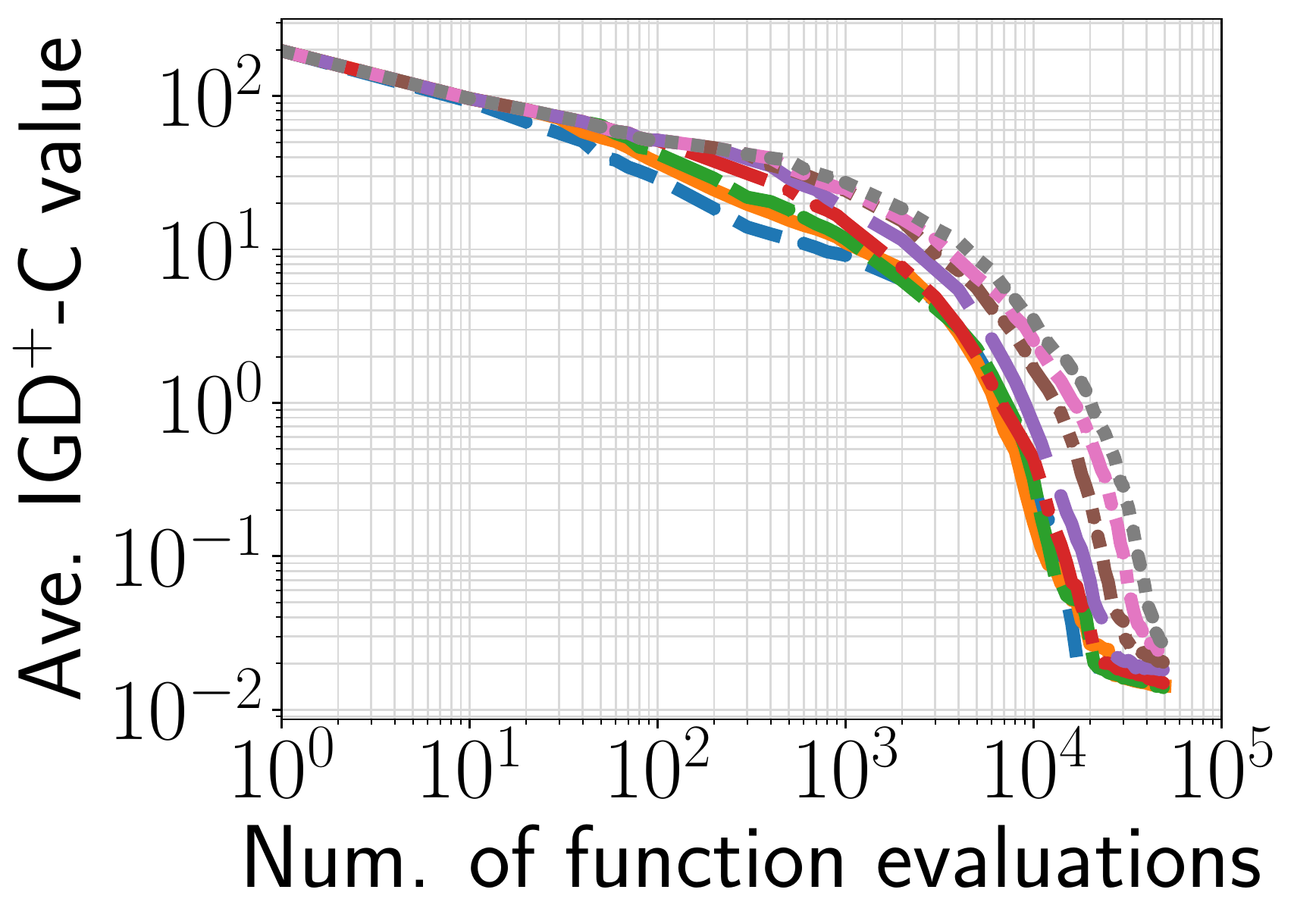}}
  \subfloat[DTLZ1 ($m=6$)]{\includegraphics[width=0.32\textwidth]{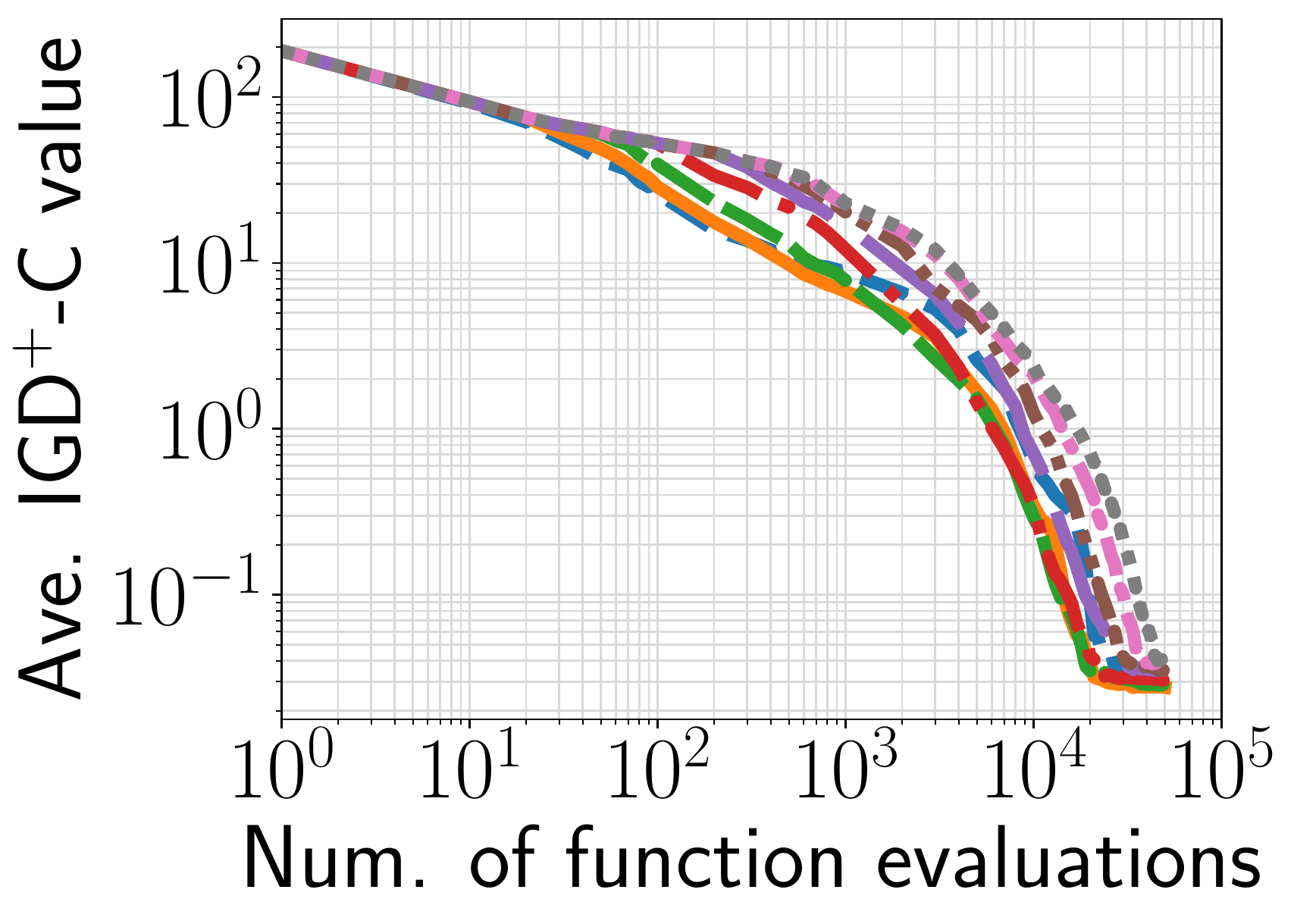}}
  \\
  \subfloat[DTLZ2 ($m=2$)]{\includegraphics[width=0.32\textwidth]{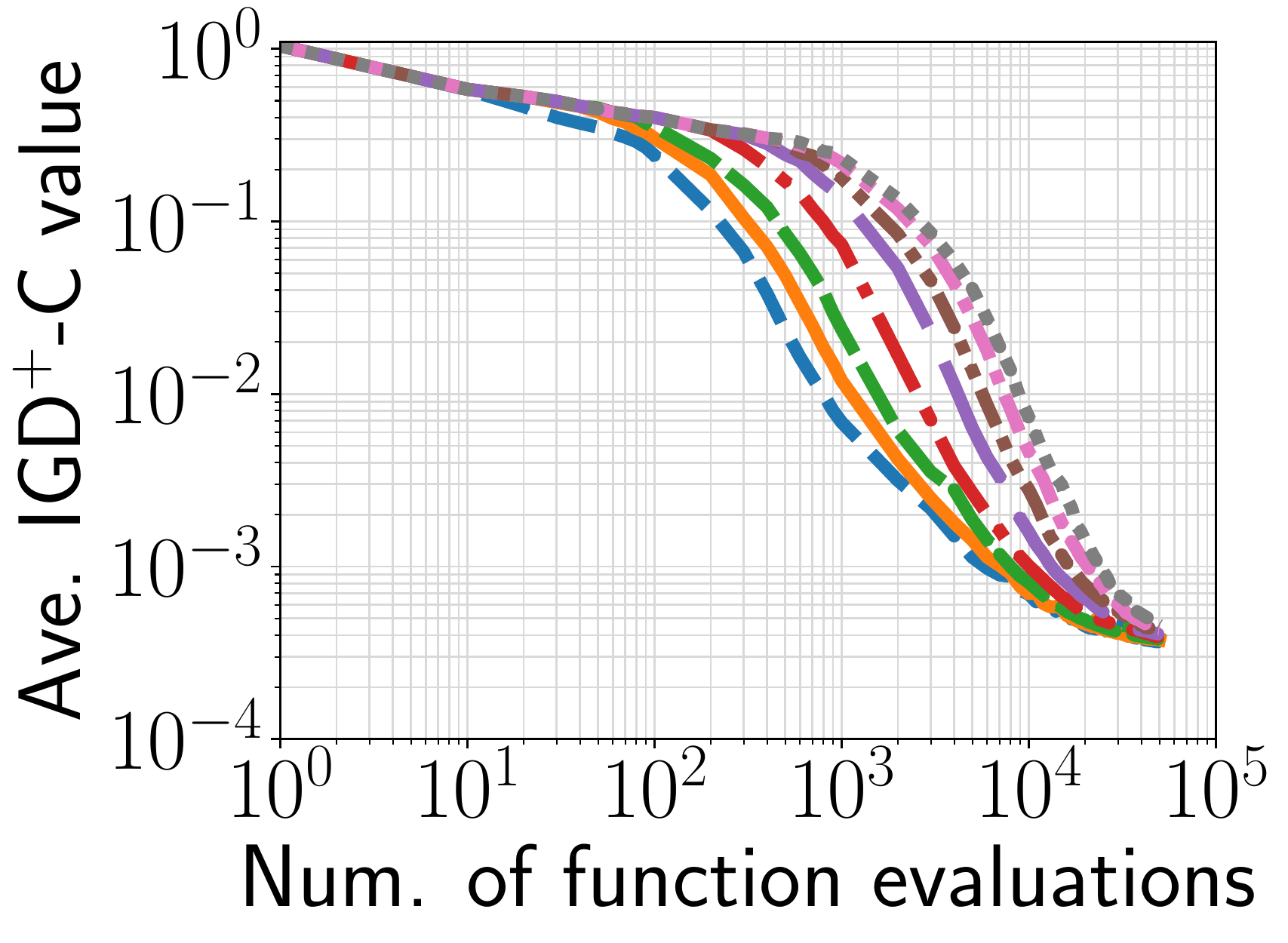}}
  \subfloat[DTLZ2 ($m=4$)]{\includegraphics[width=0.32\textwidth]{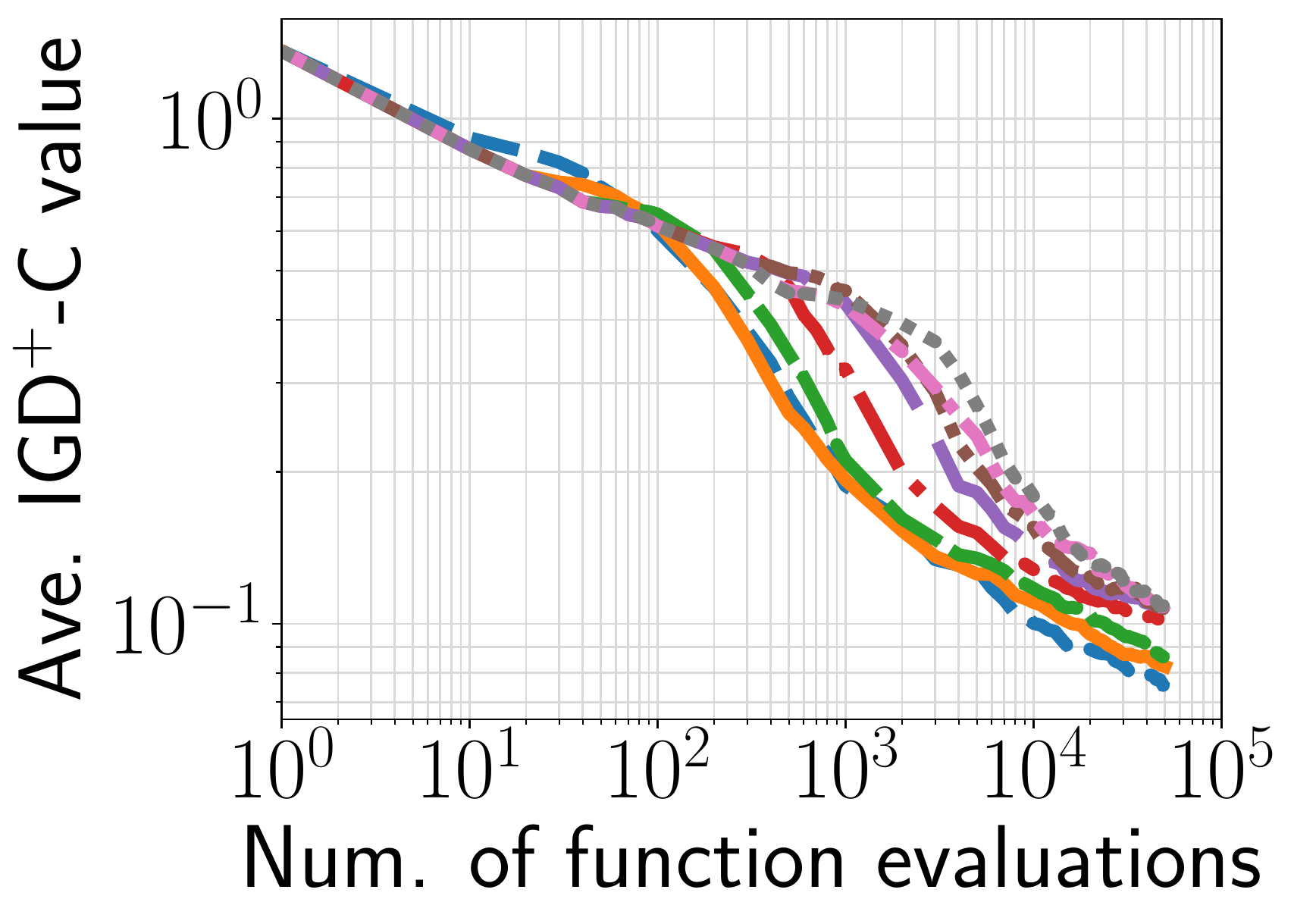}}
  \subfloat[DTLZ2 ($m=6$)]{\includegraphics[width=0.32\textwidth]{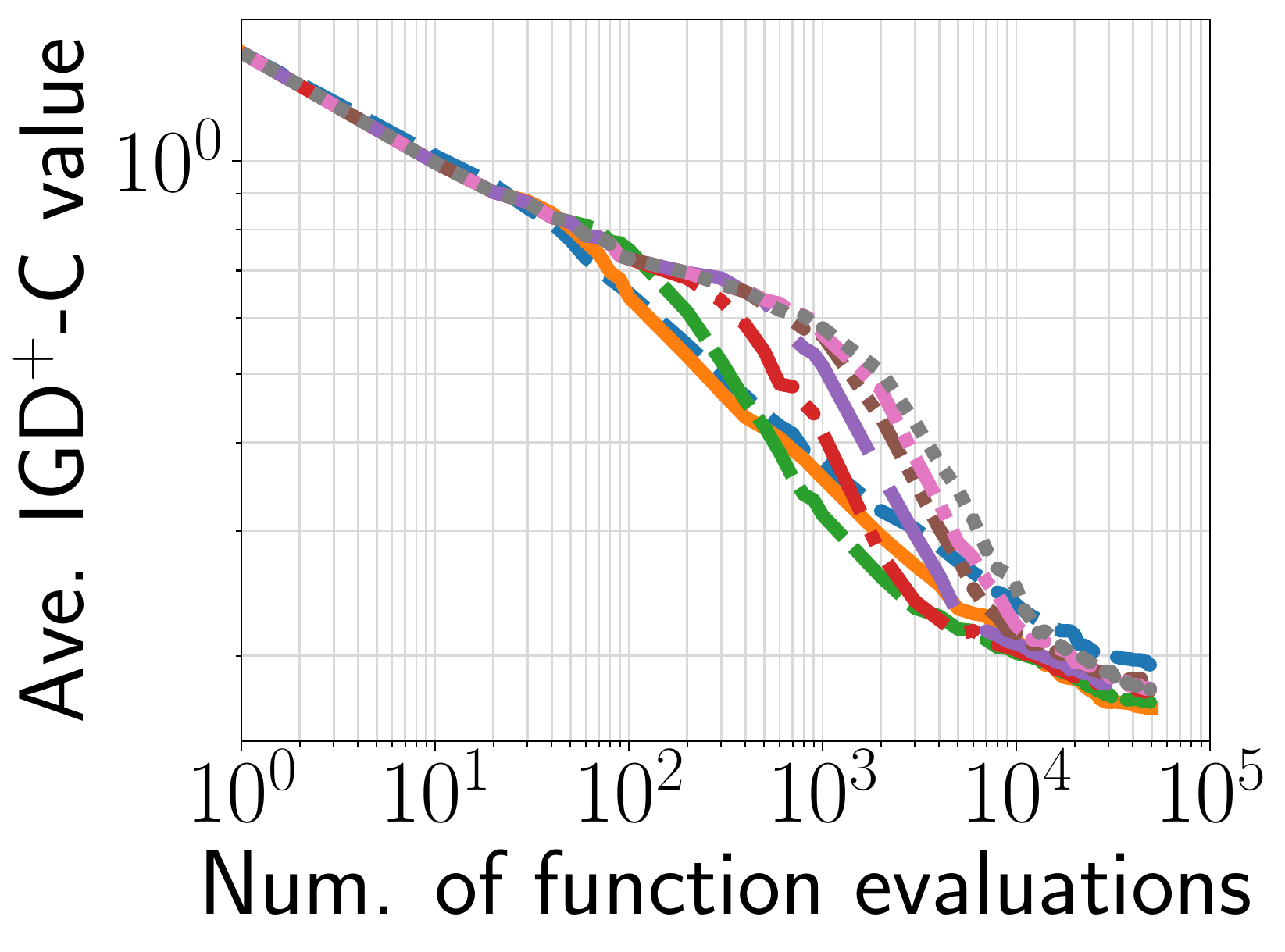}}
  \\
  \subfloat[DTLZ3 ($m=2$)]{\includegraphics[width=0.32\textwidth]{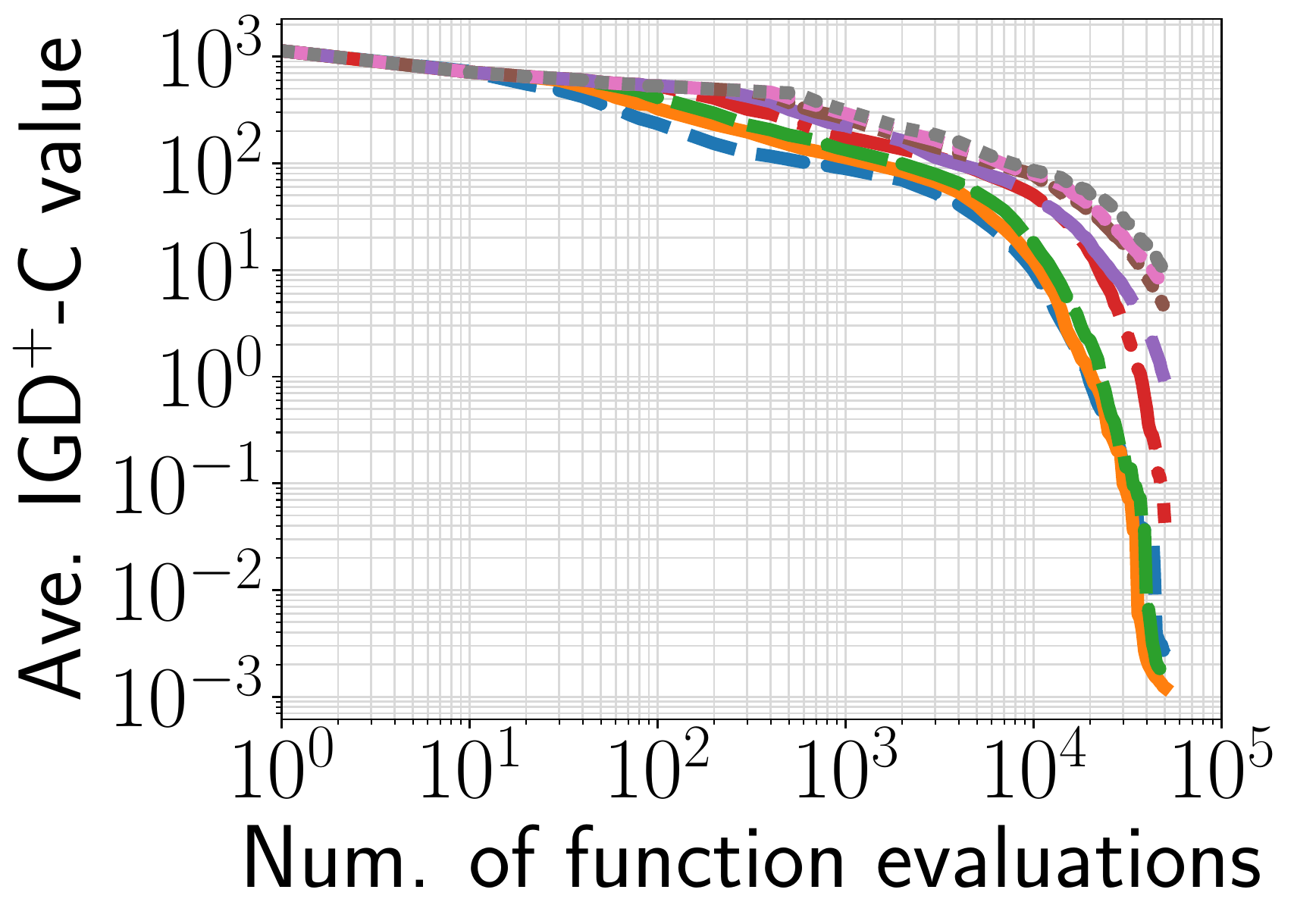}}
  \subfloat[DTLZ3 ($m=4$)]{\includegraphics[width=0.32\textwidth]{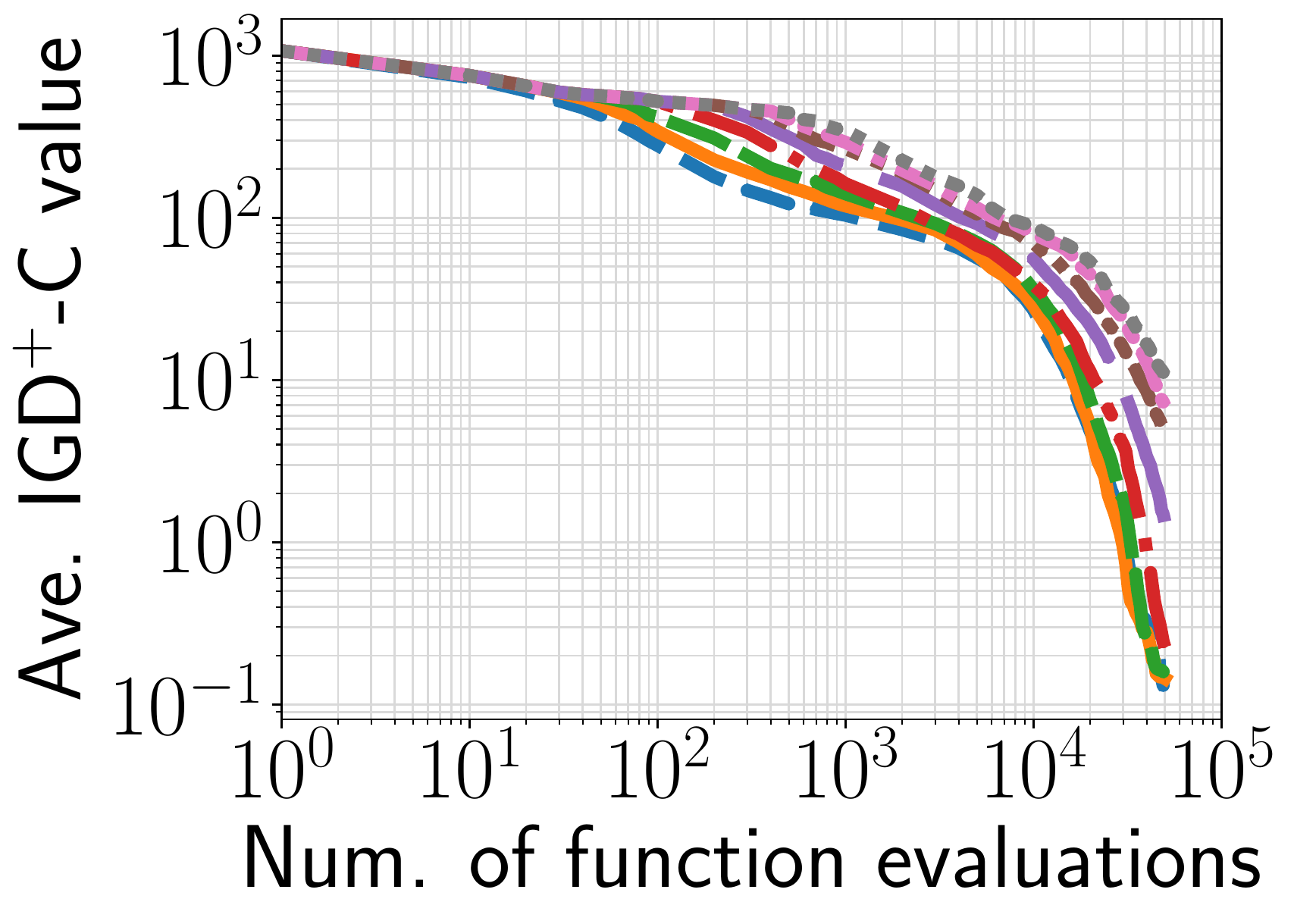}}
  \subfloat[DTLZ3 ($m=6$)]{\includegraphics[width=0.32\textwidth]{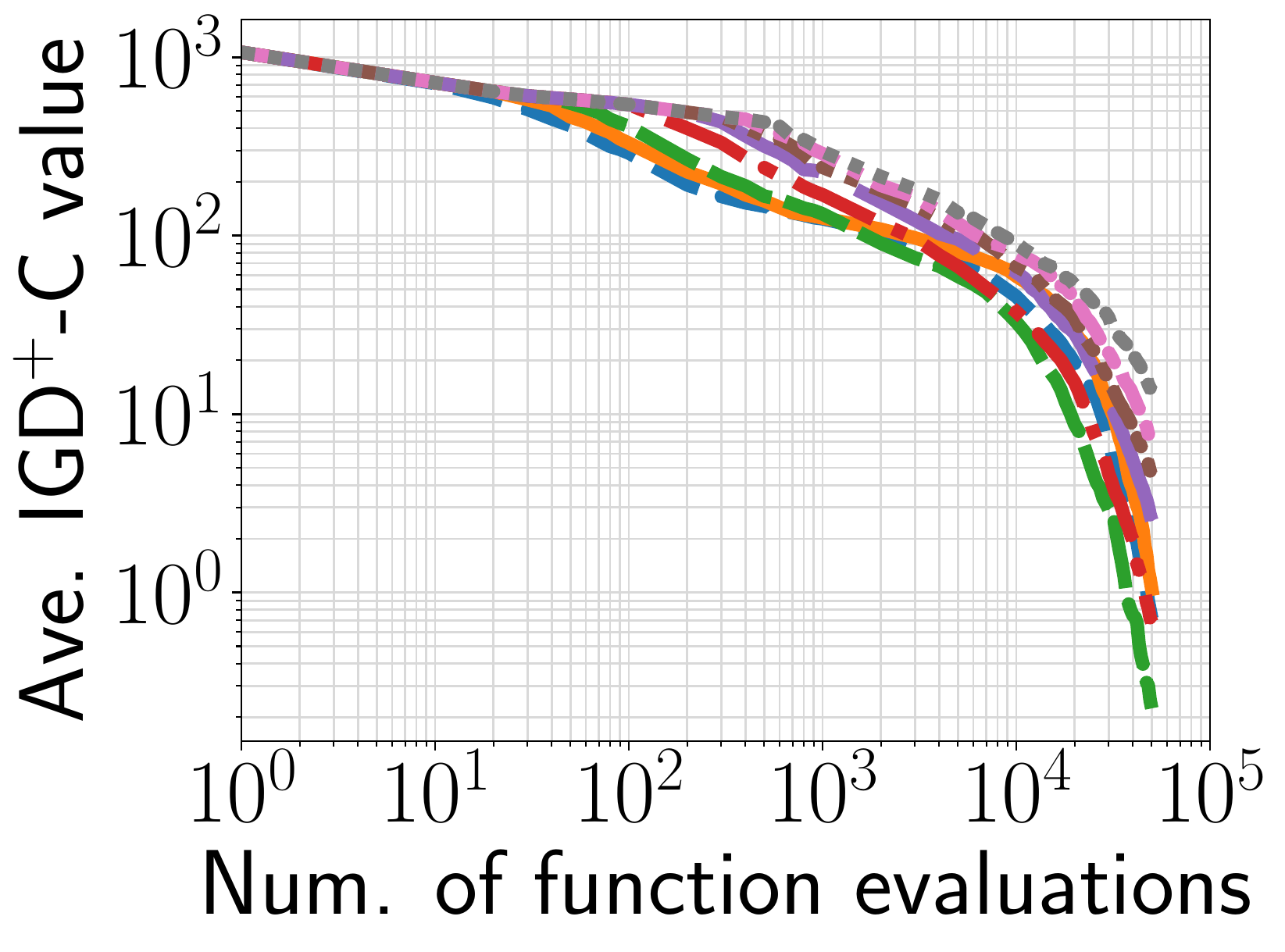}}
  \\
  \subfloat[DTLZ4 ($m=2$)]{\includegraphics[width=0.32\textwidth]{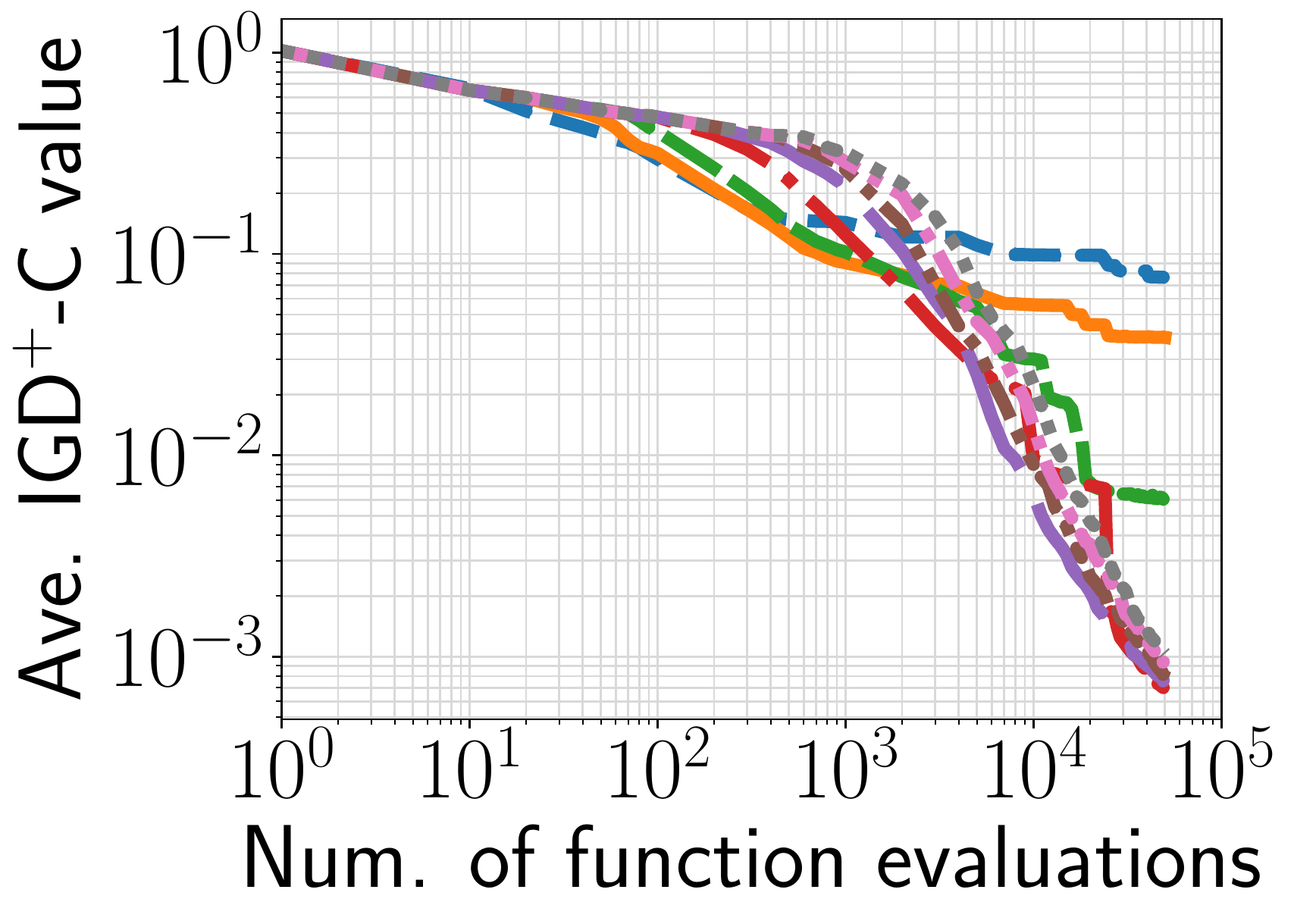}}
  \subfloat[DTLZ4 ($m=4$)]{\includegraphics[width=0.32\textwidth]{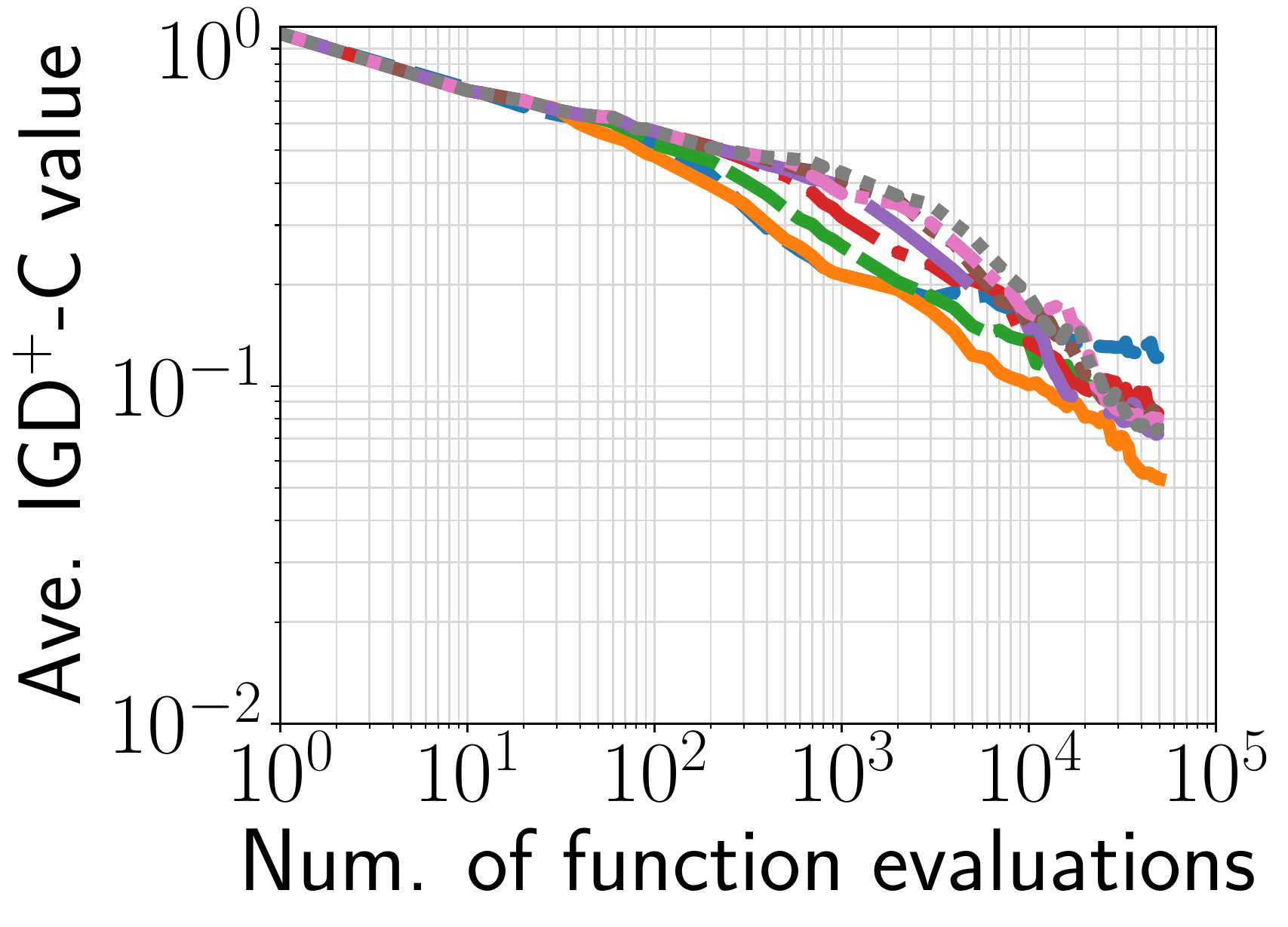}}
  \subfloat[DTLZ4 ($m=6$)]{\includegraphics[width=0.32\textwidth]{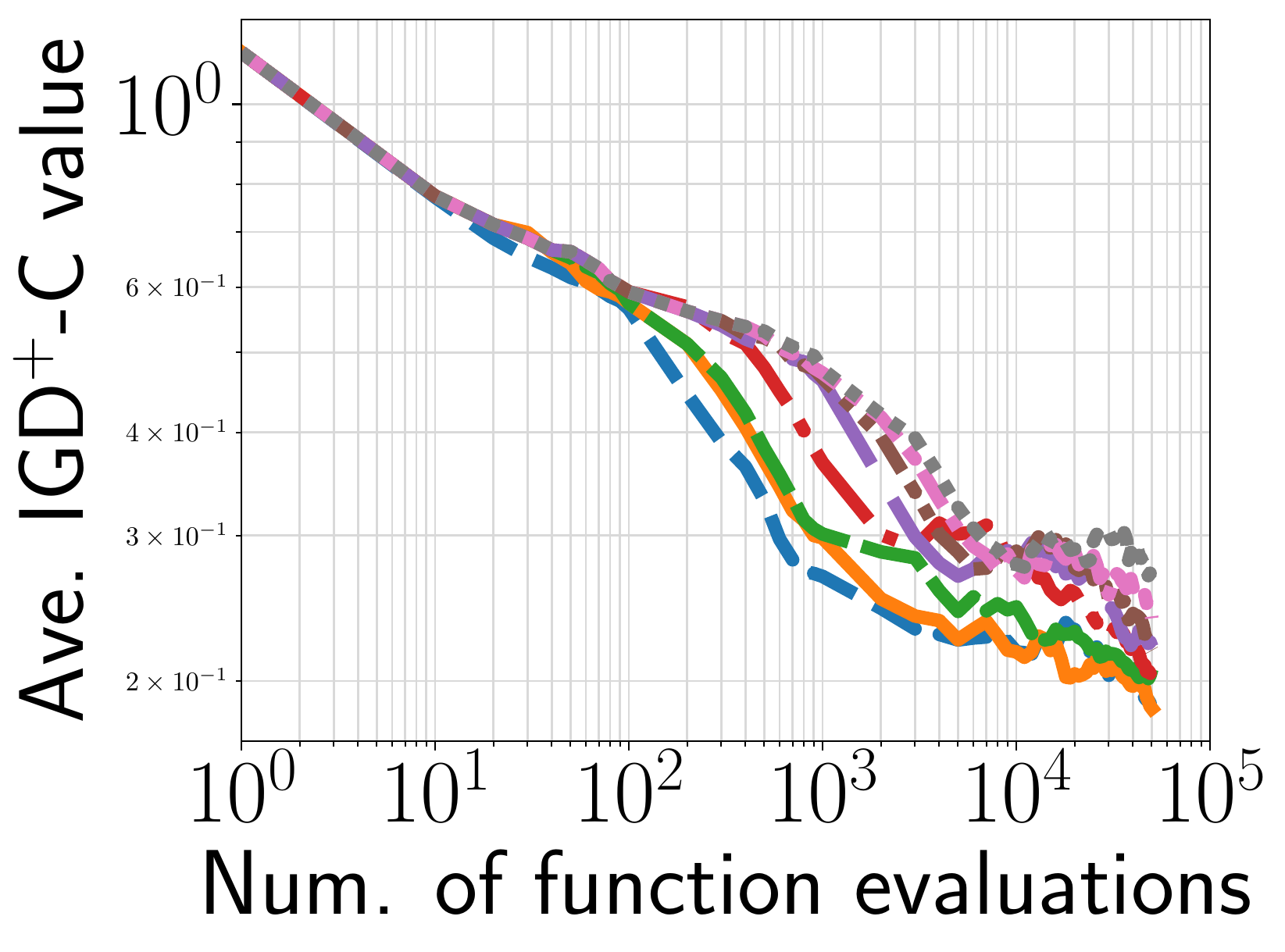}} 
      \caption{Average IGD$^+$-C values of MOEAD-NUMS with different population sizes on the DTLZ1--DTLZ4 problems with $m \in \{2, 4, 6\}$.}
   \label{fig:sup_nums_dtlz}
\end{figure*}

\begin{table*}[t]
  \renewcommand{\arraystretch}{0.84} 
\centering
  \caption{\small The best population size $\mu$ of r-NSGA-II on the DTLZ1--DTLZ4 problems with $r \in \{0.01, 0.05, 0.1, 0.2, 0.3\}$.}
  \label{tab:sup_effect_roirad_rnsga2}  
{\small
\subfloat[$m=2$]{
\begin{tabular}{lccccccc}
\toprule
$r$ & $1$K FEs & $5$K FEs  & $10$K FEs  & $30$K FEs  & $50$K FEs\\
\midrule
0.01 & 20 & 20 & 8 & 100 & 40\\
0.05 & 20 & 20 & 20 & 100 & 40\\
0.1 & 20 & 20 & 20 & 40 & 40\\
0.2 & 20 & 20 & 8 & 40 & 40\\
0.3 & 20 & 20 & 8 & 40 & 40\\
\bottomrule
\end{tabular}
}
\\
\subfloat[$m=4$]{
\begin{tabular}{lccccccc}
\toprule
$r$ & $1$K FEs & $5$K FEs  & $10$K FEs  & $30$K FEs  & $50$K FEs\\
\midrule
0.01 & 40 & 40 & 40 & 40 & 40\\
0.05 & 40 & 40 & 40 & 40 & 40\\
0.1 & 40 & 40 & 40 & 40 & 40\\
0.2 & 20 & 40 & 40 & 40 & 40\\
0.3 & 20 & 40 & 20 & 40 & 40\\
\bottomrule
\end{tabular}
}
\\
\subfloat[$m=6$]{
\begin{tabular}{lccccccc}
\toprule
$r$ & $1$K FEs & $5$K FEs  & $10$K FEs  & $30$K FEs  & $50$K FEs\\
\midrule
0.01 & 40 & 40 & 40 & 100 & 100\\
0.05 & 40 & 40 & 40 & 100 & 100\\
0.1 & 40 & 40 & 100 & 100 & 100\\
0.2 & 40 & 40 & 100 & 100 & 40\\
0.3 & 40 & 40 & 40 & 40 & 40\\
\bottomrule
\end{tabular}
}
}
\end{table*}

\begin{table*}[t]
  \renewcommand{\arraystretch}{0.84} 
\centering
  \caption{\small The best population size $\mu$ of g-NSGA-II on the DTLZ1--DTLZ4 problems with $r \in \{0.01, 0.05, 0.1, 0.2, 0.3\}$.}
  \label{tab:sup_effect_roirad_gnsga2}  
{\small
\subfloat[$m=2$]{
\begin{tabular}{lccccccc}
\toprule
$r$ & $1$K FEs & $5$K FEs  & $10$K FEs  & $30$K FEs  & $50$K FEs\\
\midrule
0.01 & 20 & 40 & 40 & 200 & 100\\
0.05 & 20 & 40 & 40 & 200 & 100\\
0.1 & 20 & 40 & 40 & 200 & 100\\
0.2 & 20 & 40 & 40 & 40 & 100\\
0.3 & 20 & 40 & 40 & 200 & 100\\
\bottomrule
\end{tabular}
}
\\
\subfloat[$m=4$]{
\begin{tabular}{lccccccc}
\toprule
$r$ & $1$K FEs & $5$K FEs  & $10$K FEs  & $30$K FEs  & $50$K FEs\\
\midrule
0.01 & 40 & 100 & 200 & 400 & 400\\
0.05 & 40 & 100 & 200 & 400 & 400\\
0.1 & 40 & 100 & 200 & 400 & 400\\
0.2 & 40 & 100 & 200 & 400 & 400\\
0.3 & 40 & 100 & 200 & 400 & 400\\
\bottomrule
\end{tabular}
}
\\
\subfloat[$m=6$]{
\begin{tabular}{lccccccc}
\toprule
$r$ & $1$K FEs & $5$K FEs  & $10$K FEs  & $30$K FEs  & $50$K FEs\\
\midrule
0.01 & 8 & 8 & 300 & 8 & 8\\
0.05 & 8 & 8 & 300 & 8 & 8\\
0.1 & 8 & 8 & 300 & 8 & 8\\
0.2 & 8 & 8 & 8 & 8 & 8\\
0.3 & 8 & 8 & 8 & 8 & 8\\
\bottomrule
\end{tabular}
}
}
\end{table*}

\begin{table*}[t]
  \renewcommand{\arraystretch}{0.84} 
\centering
  \caption{\small The best population size $\mu$ of PBEA on the DTLZ1--DTLZ4 problems with $r \in \{0.01, 0.05, 0.1, 0.2, 0.3\}$.}
  \label{tab:sup_effect_roirad_pbea}  
{\small
\subfloat[$m=2$]{
\begin{tabular}{lccccccc}
\toprule
$r$ & $1$K FEs & $5$K FEs  & $10$K FEs  & $30$K FEs  & $50$K FEs\\
\midrule
0.01 & 20 & 40 & 8 & 100 & 40\\
0.05 & 20 & 40 & 8 & 100 & 200\\
0.1 & 20 & 40 & 8 & 100 & 200\\
0.2 & 20 & 40 & 20 & 100 & 40\\
0.3 & 20 & 40 & 40 & 100 & 300\\
\bottomrule
\end{tabular}
}
\\
\subfloat[$m=4$]{
\begin{tabular}{lccccccc}
\toprule
$r$ & $1$K FEs & $5$K FEs  & $10$K FEs  & $30$K FEs  & $50$K FEs\\
\midrule
0.01 & 20 & 100 & 40 & 200 & 200\\
0.05 & 20 & 100 & 40 & 200 & 200\\
0.1 & 20 & 100 & 40 & 100 & 200\\
0.2 & 20 & 100 & 40 & 200 & 200\\
0.3 & 20 & 100 & 40 & 200 & 300\\
\bottomrule
\end{tabular}
}
\\
\subfloat[$m=6$]{
\begin{tabular}{lccccccc}
\toprule
$r$ & $1$K FEs & $5$K FEs  & $10$K FEs  & $30$K FEs  & $50$K FEs\\
\midrule
0.01 & 20 & 40 & 100 & 300 & 200\\
0.05 & 20 & 40 & 100 & 300 & 400\\
0.1 & 20 & 40 & 100 & 300 & 300\\
0.2 & 20 & 40 & 100 & 300 & 400\\
0.3 & 20 & 40 & 100 & 300 & 400\\
\bottomrule
\end{tabular}
}
}
\end{table*}

\begin{table*}[t]
  \renewcommand{\arraystretch}{0.84} 
\centering
  \caption{\small The best population size $\mu$ of R-MEAD2 on the DTLZ1--DTLZ4 problems with $r \in \{0.01, 0.05, 0.1, 0.2, 0.3\}$.}
  \label{tab:sup_effect_roirad_rmead2}  
{\small
\subfloat[$m=2$]{
\begin{tabular}{lccccccc}
\toprule
$r$ & $1$K FEs & $5$K FEs  & $10$K FEs  & $30$K FEs  & $50$K FEs\\
\midrule
0.01 & 8 & 8 & 8 & 20 & 20\\
0.05 & 8 & 8 & 8 & 20 & 20\\
0.1 & 8 & 8 & 8 & 20 & 20\\
0.2 & 8 & 8 & 8 & 20 & 20\\
0.3 & 8 & 8 & 8 & 20 & 20\\
\bottomrule
\end{tabular}
}
\\
\subfloat[$m=4$]{
\begin{tabular}{lccccccc}
\toprule
$r$ & $1$K FEs & $5$K FEs  & $10$K FEs  & $30$K FEs  & $50$K FEs\\
\midrule
0.01 & 20 & 100 & 20 & 20 & 20\\
0.05 & 20 & 100 & 20 & 20 & 20\\
0.1 & 20 & 100 & 100 & 100 & 20\\
0.2 & 20 & 100 & 100 & 100 & 20\\
0.3 & 20 & 100 & 200 & 100 & 100\\
\bottomrule
\end{tabular}
}
\\
\subfloat[$m=6$]{
\begin{tabular}{lccccccc}
\toprule
$r$ & $1$K FEs & $5$K FEs  & $10$K FEs  & $30$K FEs  & $50$K FEs\\
\midrule
0.01 & 8 & 100 & 200 & 20 & 40\\
0.05 & 8 & 100 & 200 & 20 & 40\\
0.1 & 8 & 100 & 200 & 20 & 40\\
0.2 & 8 & 100 & 200 & 20 & 40\\
0.3 & 8 & 100 & 300 & 20 & 20\\
\bottomrule
\end{tabular}
}
}
\end{table*}

\begin{table*}[t]
  \renewcommand{\arraystretch}{0.84} 
\centering
  \caption{\small The best population size $\mu$ of MOEA/D-NUMS on the DTLZ1--DTLZ4 problems with $r \in \{0.01, 0.05, 0.1, 0.2, 0.3\}$.}
  \label{tab:sup_effect_roirad_nums}  
{\small
\subfloat[$m=2$]{
\begin{tabular}{lccccccc}
\toprule
$r$ & $1$K FEs & $5$K FEs  & $10$K FEs  & $30$K FEs  & $50$K FEs\\
\midrule
0.01 & 8 & 8 & 8 & 100 & 20\\
0.05 & 8 & 8 & 8 & 20 & 40\\
0.1 & 8 & 8 & 8 & 20 & 20\\
0.2 & 8 & 20 & 20 & 20 & 20\\
0.3 & 8 & 8 & 20 & 8 & 20\\
\bottomrule
\end{tabular}
}
\\
\subfloat[$m=4$]{
\begin{tabular}{lccccccc}
\toprule
$r$ & $1$K FEs & $5$K FEs  & $10$K FEs  & $30$K FEs  & $50$K FEs\\
\midrule
0.01 & 8 & 20 & 20 & 20 & 20\\
0.05 & 8 & 20 & 20 & 20 & 20\\
0.1 & 8 & 20 & 20 & 20 & 20\\
0.2 & 8 & 20 & 20 & 20 & 20\\
0.3 & 8 & 20 & 20 & 20 & 20\\
\bottomrule
\end{tabular}
}
\\
\subfloat[$m=6$]{
\begin{tabular}{lccccccc}
\toprule
$r$ & $1$K FEs & $5$K FEs  & $10$K FEs  & $30$K FEs  & $50$K FEs\\
\midrule
0.01 & 20 & 40 & 40 & 20 & 20\\
0.05 & 20 & 40 & 40 & 20 & 20\\
0.1 & 20 & 40 & 40 & 20 & 20\\
0.2 & 8 & 40 & 40 & 20 & 20\\
0.3 & 8 & 40 & 40 & 20 & 20\\
\bottomrule
\end{tabular}
}
}
\end{table*}









\end{document}